\documentclass[runningheads]{llncs}

\usepackage{eccv}

\usepackage{eccvabbrv}

\usepackage{graphicx}
\usepackage{booktabs}
\usepackage{wrapfig}
\usepackage[accsupp]{axessibility}  

\usepackage{times}
\usepackage{epsfig}
\usepackage{amsmath}
\usepackage{amssymb}

\usepackage{adjustbox}
\usepackage[graphicx]{realboxes}

\usepackage{ntheorem}
\theoremseparator{:}
\newtheorem{hyp}{Hypothesis}
\usepackage{multirow}
\usepackage{caption}
\usepackage{subcaption}
\definecolor{cadmiumgreen}{rgb}{0.0, 0.42, 0.24}
\definecolor{custom}{cmyk}{0.1,0.48,0.49,0.2}
\definecolor{new}{rgb}{0.81,0.05,0.9}
\definecolor{BrickRed}{rgb}{0.81,0.1,0.1}
\definecolor{RoyalBlue}{rgb}{0.2,0.2,0.75}

\usepackage{hyperref}

\usepackage{orcidlink}

\usepackage[capitalize]{cleveref}
\crefname{section}{Sec.}{Secs.}
\Crefname{section}{Section}{Sections}
\Crefname{table}{Table}{Tables}
\crefname{table}{Tab.}{Tabs.}
\usepackage{xcolor}
\definecolor{cadmiumgreen}{rgb}{0.0, 0.42, 0.24}
\definecolor{custom}{cmyk}{0.1,0.48,0.49,0.2}
\definecolor{OliveGreen}{cmyk}{0.64,0,0.95,0.40}
\definecolor{new}{rgb}{0.81,0.05,0.9}
\definecolor{BrickRed}{rgb}{0.81,0.1,0.1}
\definecolor{RoyalBlue}{rgb}{0.2,0.2,0.75}

\begin{document}

\title{Improving Feature Stability during Upsampling -- Spectral Artifacts and the Importance of Spatial Context} 

\titlerunning{Improving Feature Stability during Upsampling}

\author{Shashank Agnihotri\inst{1}\orcidlink{0000-0001-6097-8551} \and
Julia Grabinski\inst{1,2,3}\orcidlink{0000-0002-8371-1734} \and
Margret Keuper\inst{1,4}\orcidlink{0000-0002-8437-7993}}

\authorrunning{S.~Agnihotri et al.}

\institute{Data and Web Science Group, University of Mannheim \and
Fraunhofer ITWM, Kaiserslautern \and
Institute for Machine Learning and Analytics (IMLA), Offenburg University \and
Max-Planck-Institute for Informatics, Saarland Informatics Campus \\
\email{\{shashank.agnihotri,julia.grabinski,keuper\}@uni-mannheim.de}}

\maketitle

\begin{abstract}
Pixel-wise predictions are required in a wide variety of tasks such as image restoration, image segmentation, or disparity estimation. Common models involve several stages of data resampling, in which the resolution of feature maps is first reduced to aggregate information and then increased to generate a high-resolution output. Previous works have shown that resampling operations are subject to artifacts such as aliasing. During downsampling, aliases have been shown to compromise the prediction stability of image classifiers. During upsampling, they have been leveraged to detect generated content. 
Yet, the effect of aliases during upsampling has not yet been discussed w.r.t.~the stability and robustness of pixel-wise predictions. While falling under the same term (\emph{aliasing}), the challenges for correct upsampling in neural networks differ significantly from those during downsampling: when downsampling, some high frequencies can not be correctly represented and have to be removed to avoid aliases. However, when upsampling for pixel-wise predictions, we actually require the model to restore such high frequencies that can not be encoded in lower resolutions. The application of findings from signal processing is therefore a necessary but not a sufficient condition to achieve the desirable output. In contrast, we find that the availability of large spatial context during upsampling allows to provide stable, high-quality pixel-wise predictions, even when fully learning all filter weights. 
\end{abstract}
\begin{figure}[t]
    \centering
\scalebox{0.92}{
\scriptsize
   \begin{tabular}{@{}c@{\hspace{0.15cm}}c@{}c@{\hspace{0.05cm}}c@{\hspace{0.1cm}}c@{\hspace{0.1cm}}c@{}}
    \rotatebox{90}{{\phantom{sub}Baseline} \cite{chen2022simple}} &  \rotatebox{90}{{\phantom{suB}Pixel Shuffle}} & &
\includegraphics[width=0.325\textwidth]{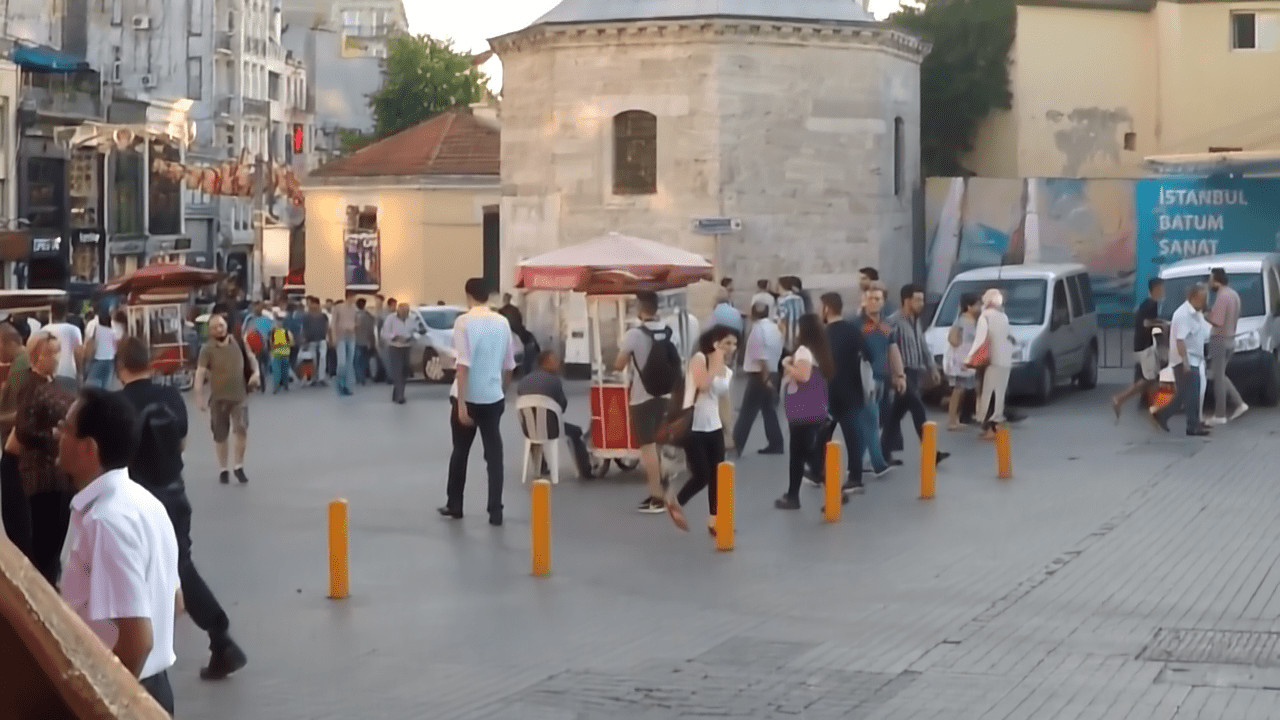} &
  \includegraphics[width=0.325\textwidth]{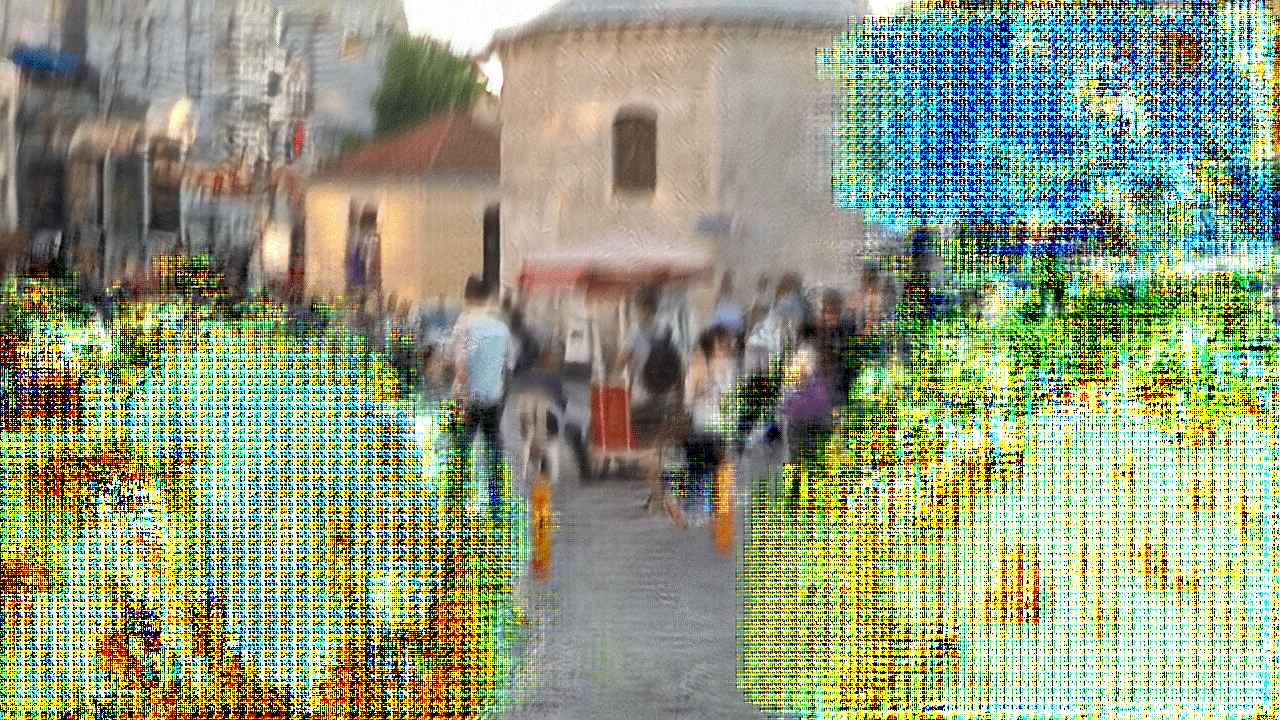} 
& \includegraphics[width=0.325\textwidth, height=2.24cm]{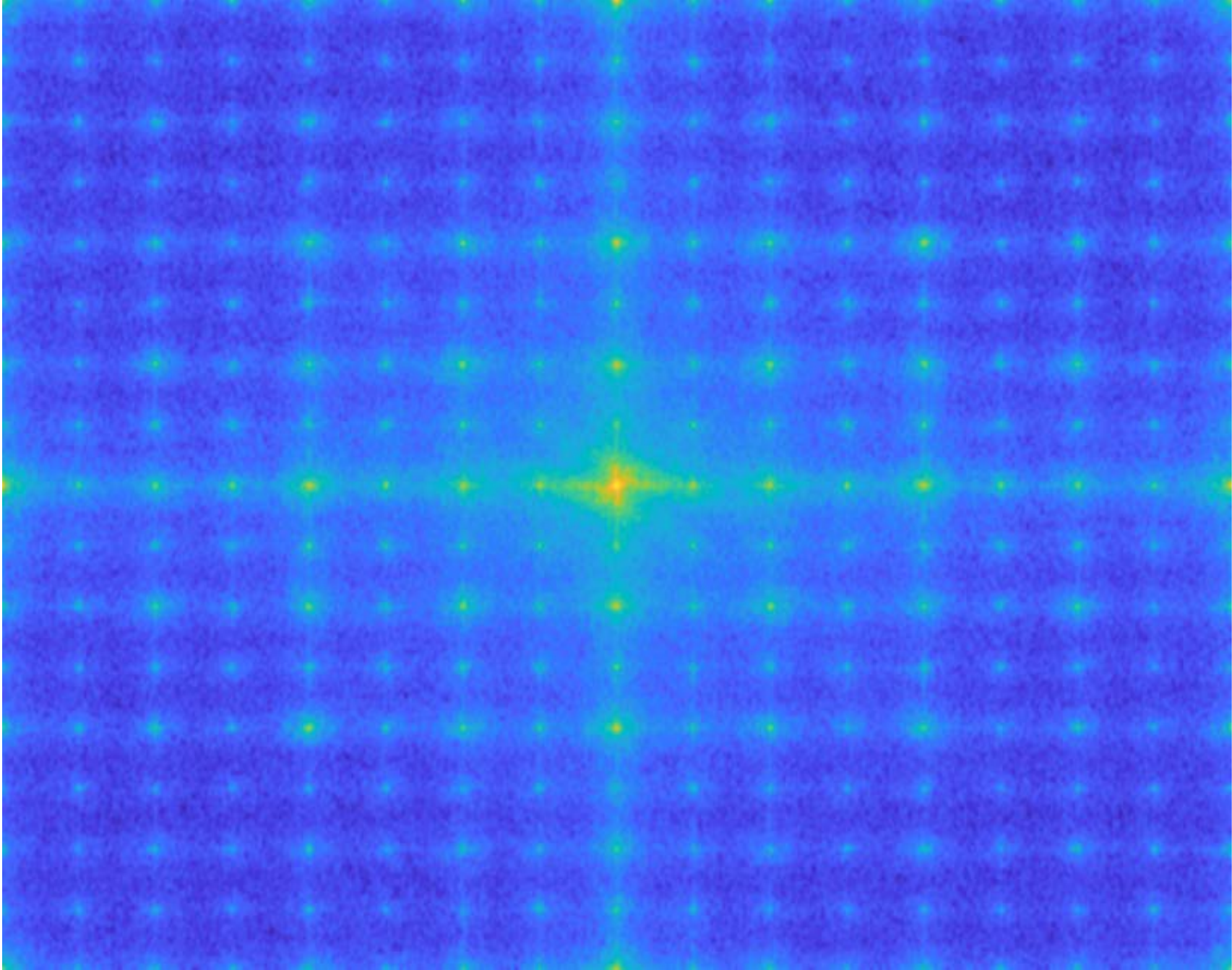} 
  \\
  
  \rotatebox{90}{{\phantom{su}}} &  \rotatebox{90}{{\phantom{suB}Transp.~conv.}} & &
\includegraphics[width=0.325\textwidth]{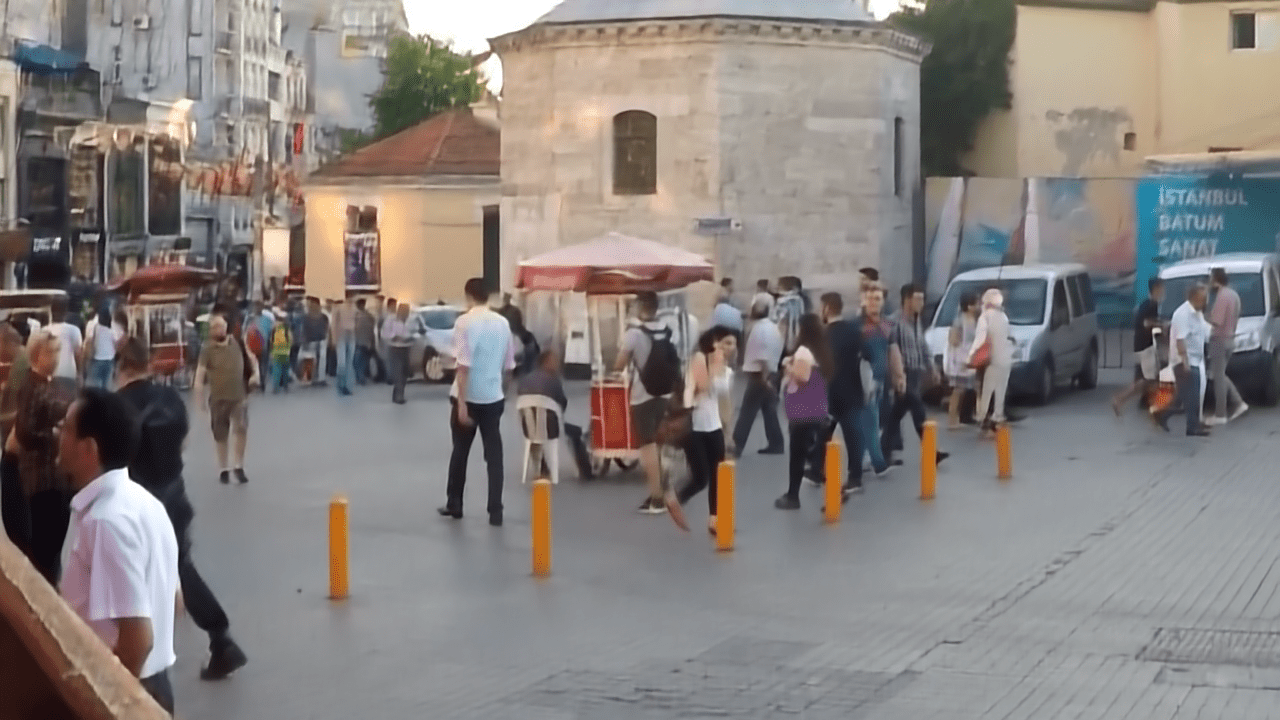} &
  \includegraphics[width=0.325\textwidth]{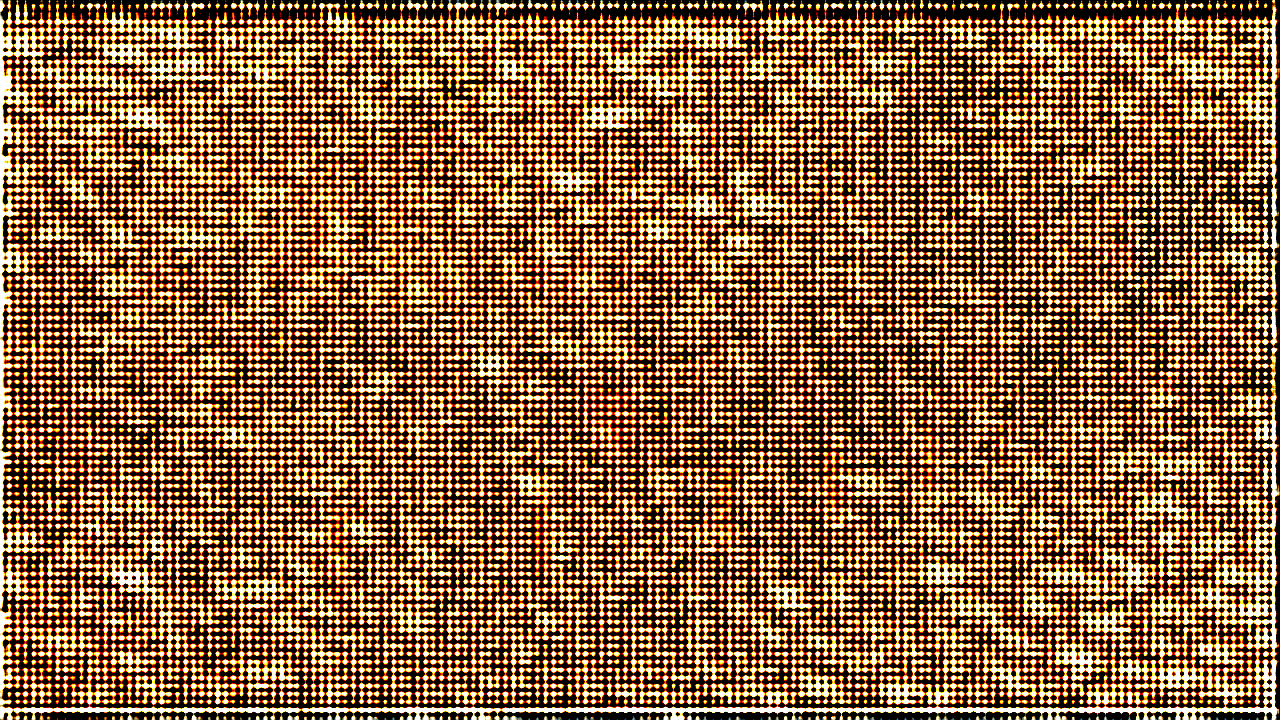} 
& \includegraphics[width=0.325\textwidth, height=2.24cm]{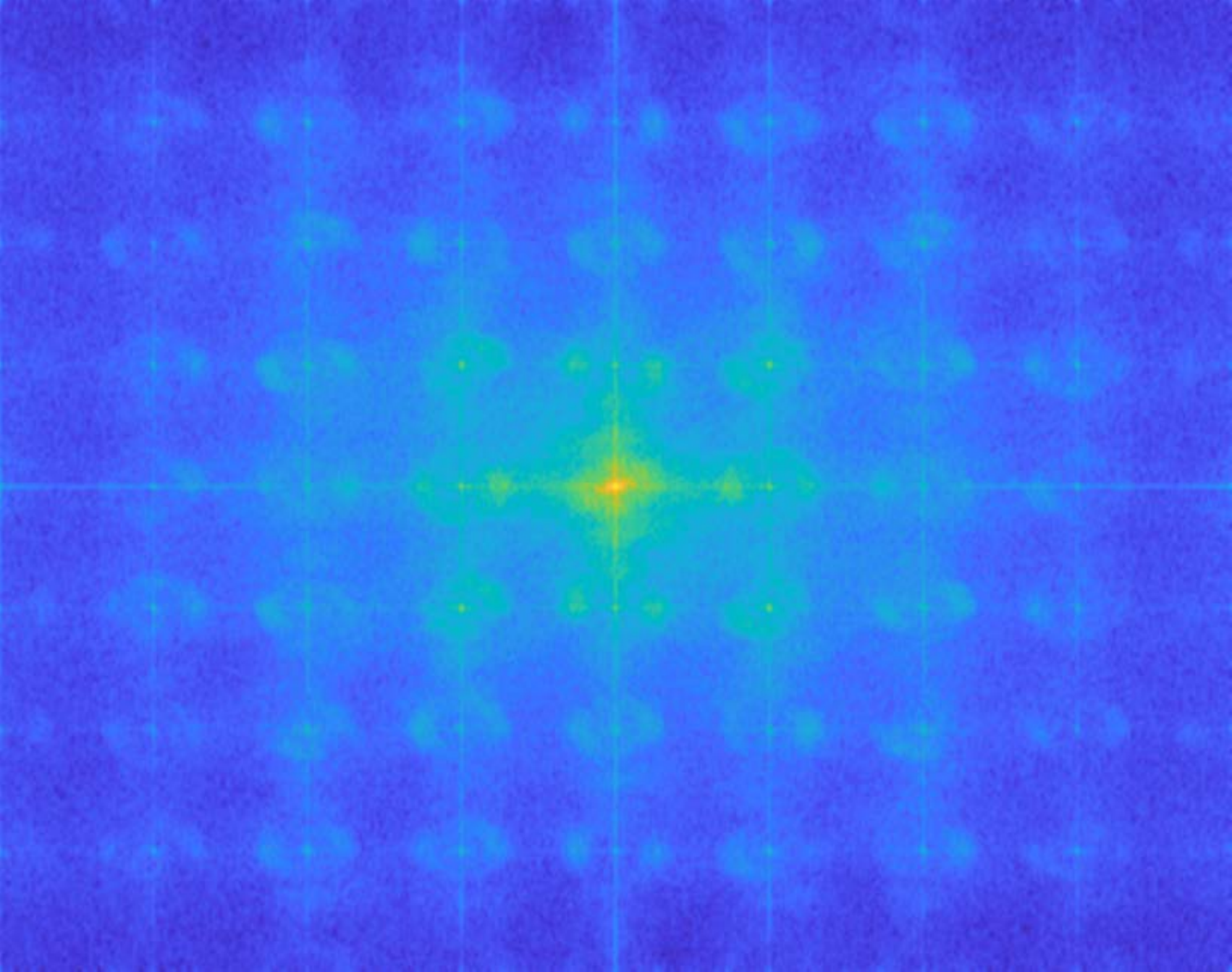} 
  \\
  
  \rotatebox{90}{{{\phantom{s}} Large Context}} & \rotatebox{90}{{Transp.~conv. (Ours)}}
  &  
  &
  \includegraphics[width=0.325\textwidth]{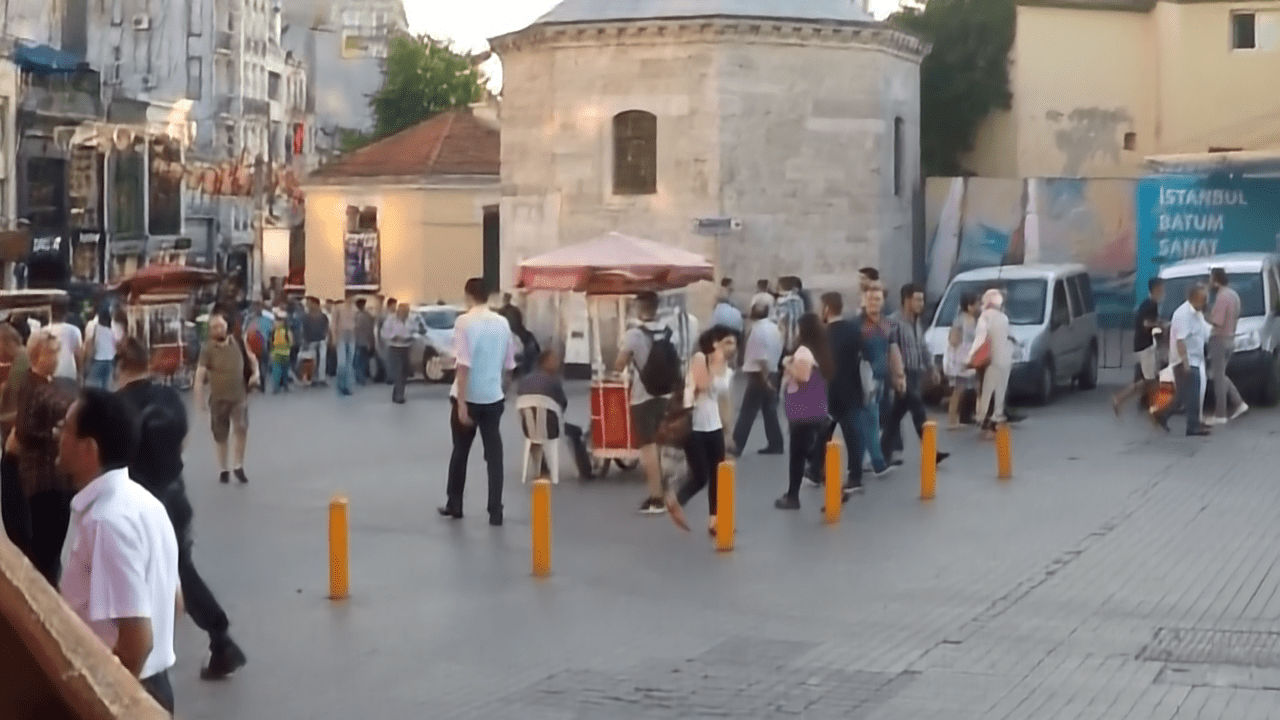} &
  \includegraphics[width=0.325\textwidth]{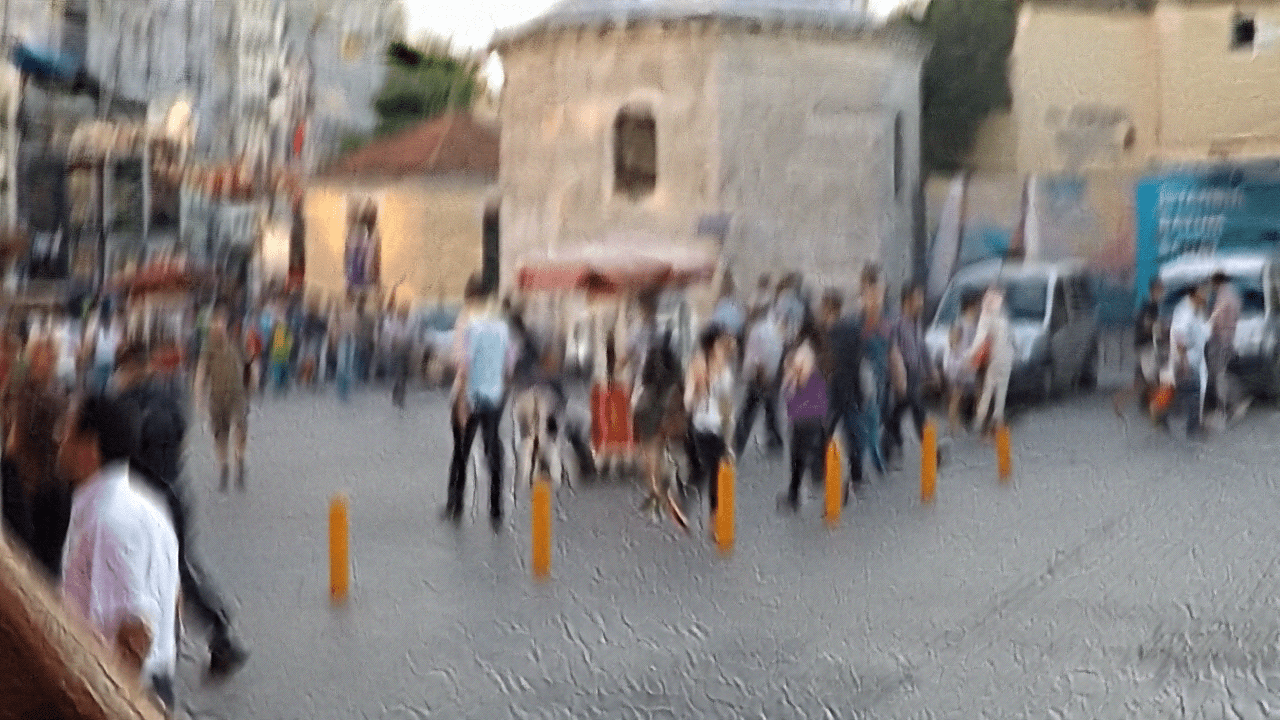} &
  \includegraphics[width=0.325\textwidth, height=2.24cm]{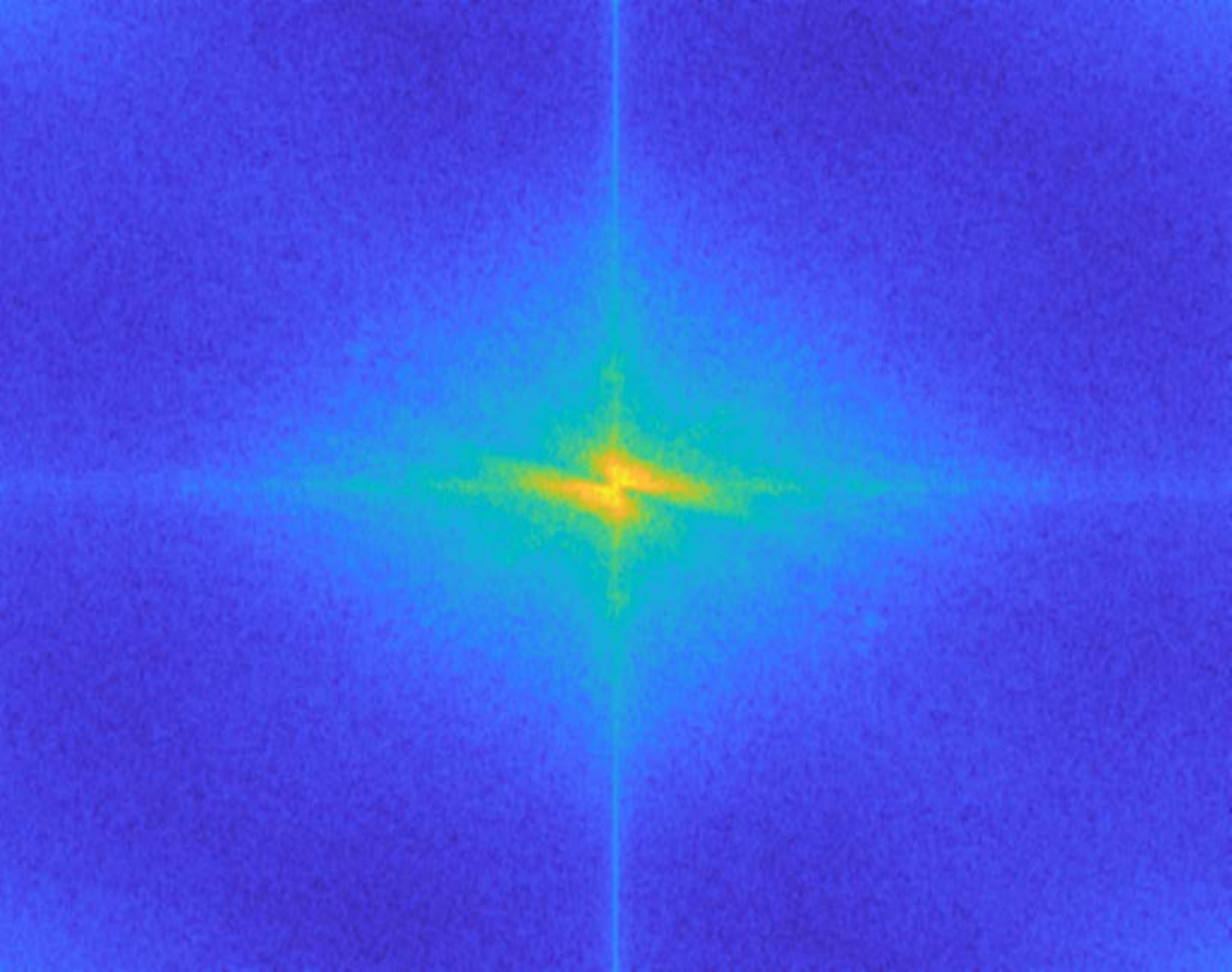} 
  \\

\multicolumn{3}{c}{} & Clean - within domain & Attacked & 2D Frequency Spectra \\

\end{tabular}
}
   \caption{Image restoration example using NAFNet~\cite{chen2022simple} variants on GoPro~\cite{gopro}. Upsampling techniques like Pixel Shuffle\cite{pixelshuffle} (first row) and transposed convolution\cite{dumoulin2016guide} using small learnable filters (2$\times$2 or 3$\times$3) (second row) are used by most prior art. Both lead to spectral artifacts for which the model needs to compensate. The clean (in-domain) restored images look appealing - while adversaries (here 5-step PGD~\cite{pgd} attack) can leverage aliases such that 
   artifacts become easily visible. 
   When observed in the frequency domain, they manifest as repeating peaks all over the spectra.
   Based on sampling theoretic considerations, we propose 
   Large Context Transposed Convolutions (7$\times$7 or larger) (bottom row). 
   They significantly increase the model's stability during upsampling, observable in the restored image under attack and the frequency spectrum.} 
    \label{fig:teaser}
    \vspace{-1em}
\end{figure}

\section{Introduction}





Most computer vision models addressing perceptual tasks such as image restoration~\cite{zamir2022restormer, chen2022simple}, semantic segmentation~\cite{unet, segnet, SegNeXt}, optical flow estimation~\cite{flownet, flownet2, raft} and disparity estimation~\cite{sttr, badki2020Bi3D, dttr} in realistic scenarios are required to behave in a stable way, at least under mild corruptions.
Interestingly, for the slightly simpler task of image classification, recent progress has shown that a model's robustness does not only depend on its training but also on its architecture \cite{maiya2021frequency,grabinski2022aliasing,grabinski2022AAAIw,grabinski2022frequencylowcut,grabinski2023fix,zhang2019making,zou2020delving,hossain2021antialiasing,jung2023neural,hoffmann2021towards}. 
Specifically, \emph{aliasing}, \ie spectral artifacts that emerge from na\"ive image resampling, have shown to compromise prediction stability, in particular in the context of classical convolutional models \cite{resnet, vgg, alexnet, vision_transformer2, small_kernel1,sommerhoff2023differentiable,grabinski2024as} which predominantly use small filter kernels in combination with severe data aggregation during downsampling \cite{maiya2021frequency,grabinski2022aliasing}. 
Principled cures usually refer to basic concepts from signal processing such as anti-aliasing by blurring before downsampling \cite{zhang2019making,grabinski2022frequencylowcut}. 
While this discussion on classifier (\ie encoder) networks is insightful, it does not provide a recipe to counteract aliases emerging during \emph{upsampling} for pixel-wise prediction tasks such as image restoration. 
Specifically, na\"{i}ve upsampling introduces artifacts in the feature representation, such as grid artifacts~\cite{checkerboard_odena2016deconvolution, aitken2017checkerboard} or ringing artifacts~\cite{ringing_artifacts}. 
As shown in \cref{fig:teaser}, these artifacts, an inherent property of inadequate upsampling (refer \cref{sec:fft}) are not always visible to the human eye, are accentuated under adversarial attack such that they can also be seen with a human eye. 
We leverage this effect in our analysis.
When observed in the frequency domain, these artifacts are apparent as multiple peaks, \ie \emph{aliases} of the original data.

While for downsampling, signal processing laws basically prescribe which part of the information can be retained at lower resolutions without aliases \cite{shannon}, ``correct'', alias-free upsampling can not restore the original high-resolution information. 
Thus, learning to upsample feature maps such that the feature stability is not harmed is of paramount importance. 
In this paper, we therefore first provide a synopsis of different aliases that emerge from different upsampling techniques. 
Based on this work, we propose a simple, transposed convolution-based upsampling block. We study our proposed operation in the context of various models, from image restoration \cite{zamir2022restormer,chen2022simple} over semantic segmentation~\cite{unet} to disparity estimation~\cite{sttr}.

Our main contributions can be summarized as follows:
\begin{itemize}
    \item Motivated by sampling theory \cite{shannon}, we study 
    upsampling in models for diverse pixel-wise prediction tasks. We find that the availability of large kernels in transposed convolutions helps the feature stability and significantly improves over standard, small kernel transposed convolutions as well as pixel shuffle~\cite{pixelshuffle}. 
    \item While large kernels are required to allow for reduced aliasing and to provide the necessary spatial context for increasing the resolution, additional small kernels can add details and remain useful. 
    \item We provide empirical evidence for our findings on diverse architectures (including vision transformer-based architectures) and downstream tasks such as image restoration, semantic segmentation, and depth estimation.
    \item We show empirically that our proposed upsampling operation complements other feature stability-increasing approaches like adversarial training.
\end{itemize}

\section{Related Work}
\label{sec:related}
In the following, we discuss recent challenges for neural networks regarding artifacts introduced by spatial sampling methods~\cite{checkerboard_odena2016deconvolution, aitken2017checkerboard,ringing_artifacts}. 
Further, we review related work on the most recent use of large kernels in CNNs. 
Finally, we provide an overview of adversarial attacks to gauge the quality of representations learned by a network. 
%
%
\paragraph{\textbf{Spectral Artifacts.}}
\label{subsec:related:artifacts}
Several prior works have studied the effect of downsampling operations on model robustness, \eg~\cite{agnihotri2024beware,karras2021alias,hossain2021antialiasing,grabinski2022aliasing,grabinski2022frequencylowcut,zhang2019making,zou2020delving}. Inspired by \cite{grabinski2022aliasing},  
\cite{grabinski2022frequencylowcut} propose an aliasing-free downsampling in the frequency domain which translates to an infinitely large blurring filter before downsampling in the spatial domain. 
Thus, for image classification, using large filter kernels has been shown to remove artifacts from downsampled representations and it leads to favorable robustness in all these cases \cite{karras2021alias,hossain2021antialiasing,grabinski2022aliasing}. However, all these works focus on improving the properties of encoder networks.

Models that use transposed convolutions in their decoders\footnote{For more details on Transposed Convolutions refer to \cite{dumoulin2016guide}.} are widely used for tasks like image generation~\cite{goodfellow2020generative, radford2015unsupervised} or segmentation~\cite{long2015fully, noh2015learning, unet, segnet}. 
However, 
in simple transposed convolutions, the convolution kernels overlap based on the chosen stride and kernel size. 
If the stride is smaller than the kernel size, this will cause overlaps in the operation, leading to uneven contributions to different pixels in the upsampled feature map and thus to grid-like artifacts~\cite{aitken2017checkerboard, checkerboard_odena2016deconvolution}. Further, image resampling can lead to aliases that become visible as ringing artifacts~\cite{shannon}. 
In the context of deepFake detection, image generation, and deblurring, several works analyzed~\cite{durall2020watch,Chandrasegaran_2021_CVPR,khayatkhoei2022spatial,ijcai2021p349,jung2021spectral,dosovitskiy2016generating,dosovitskiy2017learning} and improved upsampling techniques~\cite{karras2021alias, gal2021swagan,NIPS2014_1c1d4df5, blind_deconvolution} to reduce visual artifacts. 

Some architectures like PSPNet~\cite{pspnet}, PSANet~\cite{zhao2018psanet}, or PSMNet~\cite{psmnet} simply use bilinear interpolation operations for upsampling the feature representations.
While this reduces grid artifacts as bilinear interpolation smoothens out the feature maps, it also has major drawbacks as they sample incorrectly. These new artifacts are sometimes visible as overly smooth predictions, in particular, apparent in the PSPNet segmentation masks.
The segmentation masks over-smoothen around edges and often miss out on thin details (predictions showing these are included in the \cref{subsubsec:experiments:pspnet_interpolation}). 
This observation already shows why image encoding and decoding have to be considered separately when it comes to sampling artifacts. 
While during encoding, artifacts can be reduced by blurring, the main purpose of decoder networks is \emph{reducing} blur in many applications, to create fine-granular, pixel-wise accurate outputs, which our approach facilitates. 

\paragraph{\textbf{Large Kernels.}}
\label{subsec:related:kernels}
For image classification, \cite{convnext} showed that using large kernels like 7$\times$7 in the CNN convolution layer can outperform self-attention based vision transformers~\cite{vision_transformer1, liu2021Swin}. 
In \cite{SegNeXt,replk,slak,Peng_2017_CVPR,grabinski2024as}, the receptive field of the convolution operations was further expanded by using larger kernels, up to 31$\times$31 and 51$\times$51. These larger receptive fields provide more context to the \textbf{encoder}, leading to better performance on classification, segmentation, or object detection tasks. \cite{replk, slak} use a small kernel in parallel to capture the local context along with the global context. 
In contrast to these works, which are limited to exploring increased context only during encoding, we investigate if larger kernels can benefit upsampling when considering pixel-wise prediction tasks such as image restoration or segmentation. 
\vspace{-0.3cm}
\paragraph{\textbf{Adversarial Attacks.}}
\label{subsec:related:adversarial}
The purpose of adversarial attacks is to reveal neural networks' weaknesses~\cite{grabinski2022aliasing, intriguing, agnihotri2023cospgd, schmalfuss2022perturbationconstrained} by perturbing pixel values in the input image~\cite{fgsm, c_and_w, pgd}. 
These perturbations should lead to a false prediction even though the changes are hardly visible~\cite{fgsm,deepfool,intriguing}. 
Especially attacks that have access to the network's architecture and weights, so-called white-box attacks, are a common approach to analyzing weaknesses within the networks' structure~\cite{c_and_w, fgsm}. They employ the gradient of the network to optimize the perturbation, which is bounded within an $\epsilon$-ball of the original image, \ie $\epsilon$ defines the strength of the attack. 
Most adversarial attacks are proposed to attack classification networks like the one-step Fast Gradient Sign Method (FGSM)~\cite{fgsm} or the multi-step Projected Gradient Descent (PGD) attack~\cite{pgd}. 
However, they can be adapted to other tasks as \eg in~\cite{adv_segment, monocular_depth_adv, Mathew2020MonocularDE}. 
Furthermore, there are dedicated methods like SegPGD~\cite{segpgd} for attacks on semantic segmentation models or PCFA~\cite{schmalfuss2022perturbationconstrained} and~\cite{schmalfuss2023distracting, scheurer2023detection} for optical flow models and CosPGD~\cite{agnihotri2023cospgd} and others~\cite{schmalfuss2022attacking} for other pixel-wise prediction tasks. 
We evaluate the stability of upsampled features using adversarial attacks such as PGD and CosPGD for image restoration and FGSM and SegPGD for segmentation. 

\section{Spectral Upsampling Artifacts and How They Can Be Reduced}
\label{sec:fft}
 Following, we first theoretically review artifacts that are caused during upsampling from a signal processing aspect. 
 We start by describing the spectral artifacts~\cite{shannon} induced by the bed of nails interpolation, similar to the discussion in~\cite{durall2020watch}, and then extend the theoretical analysis to further upsampling schemes. Second, we derive from this analysis two hypotheses for the prediction stability of encoder-decoder networks, depending on their architecture. These hypotheses will motivate the remainder of the manuscript.

Consider, w.l.o.g., a one-dimensional signal $I$ and its discrete Fourier Transform $\mathcal{F}({I})$ with $k$ being the index of discrete frequencies 
\begin{equation}
\mathcal{F}({I})_k=\sum_{j=0}^{N-1}e^{-2\pi i\cdot\frac{jk}{N}}\cdot I_j, \quad \mathrm{for }\quad k=0,\dots,N-1.\nonumber
\end{equation}
During decoding, we need to upsample the spatial resolution of $I$ to get $I^{\mathrm{up}}$. For example for an upsampling factor of $2$ (often used in DNNs \cite{agnihotri2023unreasonable, chen2022simple,vision_transformer2,zamir2022restormer,dosovitskiy2016inverting}) 
we have for $\bar{k}=0,\dots,2N-1$
\begin{align}
\mathcal{F}(I)^{\mathrm{up}}_{\bar{k}}
=\sum_{j=0}^{2N-1}e^{-2\pi i\cdot\frac{j\bar{k}}{2\cdot N}}\cdot I^{\mathrm{up}}_j=\sum_{j=0}^{N-1}e^{-2\pi i\cdot\frac{2\cdot j\bar{k}}{2\cdot N}} I_j
+ \sum_{j=0}^{N-1}e^{-2\pi i\cdot\frac{(2j+1)\bar{k}}{2\cdot N}} \bar{I}_j,
\label{eq:theory2}
\end{align}
where $\bar{I}_j=0$ in \textbf{bed of nails interpolation}. 
Therefore, the second term in \eqref{eq:theory2} can be dropped and the first term resembles the original $\mathcal{F}(I)$. 
Equivalently, 
we can rewrite Eq.~\eqref{eq:theory2}, for $\bar{I}_j=0$, using a Dirac impulse comb as
\begin{align}
\eqref{eq:theory2} &=\sum_{j=0}^{2N-1}e^{-2\pi i\cdot\frac{j\bar{k}}{2\cdot N}}\cdot \sum_{t=-\infty}^{\infty}I^{\mathrm{up}}_j\cdot\delta(j-2t).\label{eq:theory3}
\end{align}
If we now apply the pointwise multiplication with the Dirac impulse comb as convolution in the Fourier domain (assuming periodicity)~\cite{forsyth2003computer}, it is
\begin{align}
\mathcal{F}(I)^{\mathrm{up}}_{\bar{k}}&=\frac{1}{2}\sum_{t=-\infty}^{\infty} \left(\sum_{j=-\infty}^{\infty}
e^{-2\pi i\cdot\frac{j\bar{k}}{2N}}I^{\mathrm{up}}_j\right)\left(\bar{k}-\frac{t}{2}\right)\\
&\overset{\eqref{eq:theory2}}{=}\frac{1}{2}\sum_{t=-\infty}^{\infty} \left(\sum_{j=-\infty}^{\infty}
e^{-2\pi i\cdot\frac{j\bar{k}}{N}}\cdot I_j\right)\left(\bar{k}-\frac{t}{2}\right) = \frac{1}{2}  \sum_{t=-\infty}^{\infty} \mathcal{F}(I)_{\bar{k}}\left(\bar{k}-\frac{t}{2}\right).\nonumber
\end{align}
We can see that such upsampling creates high-frequency replica of the signal at $\frac{t}{2}$ for $t$ in $-\infty, \dots, \infty$ in $\mathcal{F}({I})^{\mathrm{up}}$ and spatial frequencies apparent beyond array positions $\frac{N}{2}$ will be impacted by spectral artifacts if no appropriate countermeasures are taken.

\begin{figure*}[t]
	\begin{center}
	\includegraphics[width=0.95\linewidth]{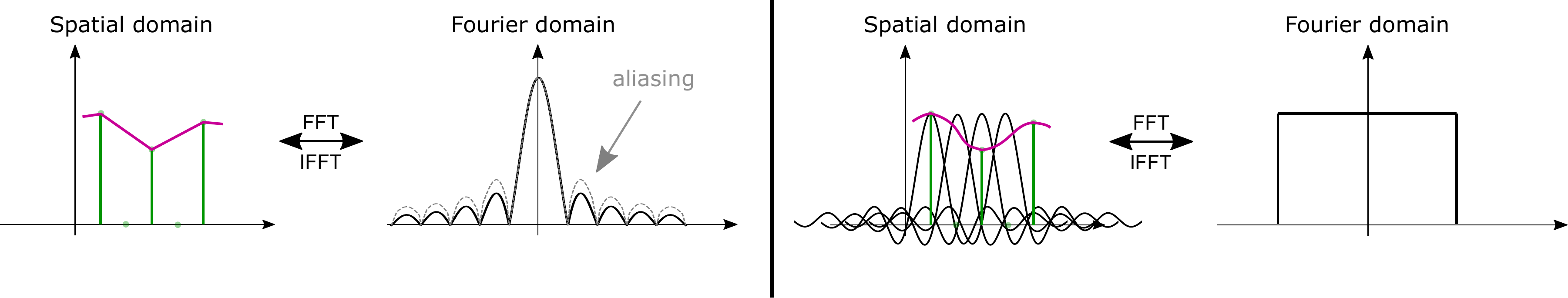}
	\caption[]{(Left) Linear interpolation (pink) of the samples (green) causes aliases. (Right) Optimal signal reconstruction (pink) is achieved by $\mathrm{sinc}$ interpolation. In practice our spatial context is limited and the interpolation function is discrete. Yet, increasing the kernel size enables the approximation of larger $\mathrm{sinc}$-like structures.
	}\label{fig:linear_interpoaltion}
	\end{center}
	\vspace{-2em}
\end{figure*}

\begin{figure}[t]
    \centering
\scalebox{0.75}{
   \begin{tabular}{@{}c@{\hspace{0.05cm}}c@{\hspace{0.05cm}}c@{\hspace{0.05cm}}c@{}}
   
\includegraphics[width=0.325\textwidth, height=2.24cm]{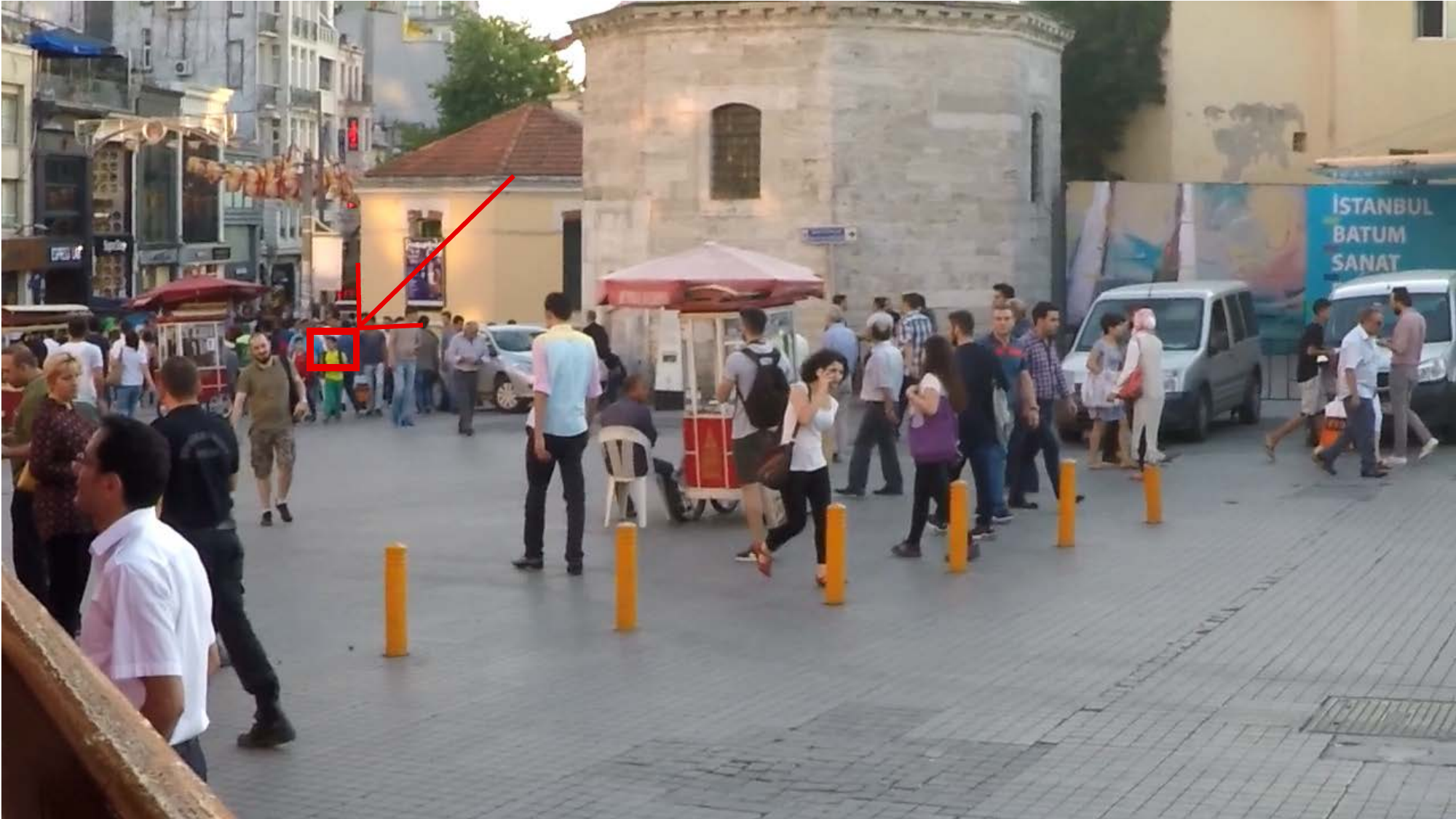}
& \includegraphics[width=0.325\textwidth, height=2.24cm]{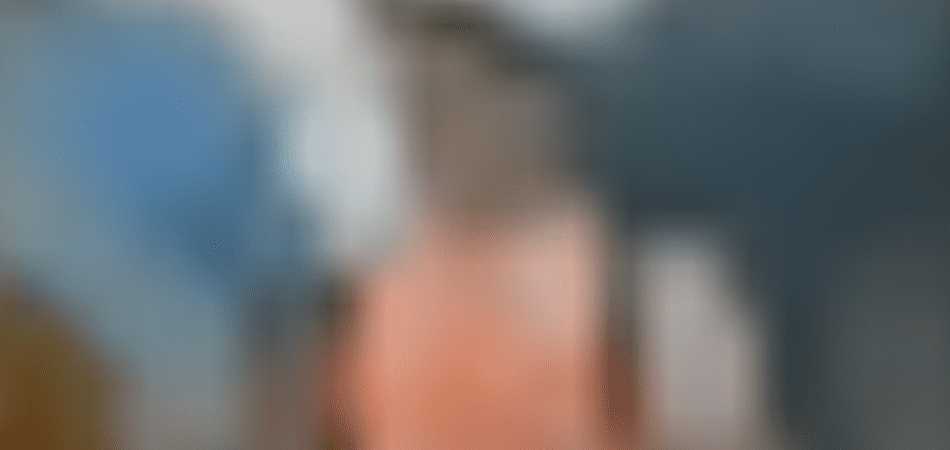}
& \includegraphics[width=0.325\textwidth, height=2.24cm]{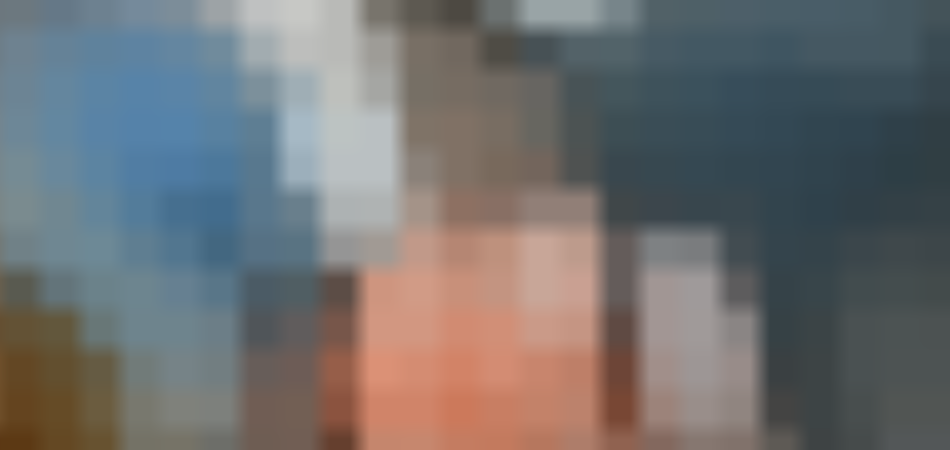}
& \includegraphics[width=0.325\textwidth, height=2.24cm]{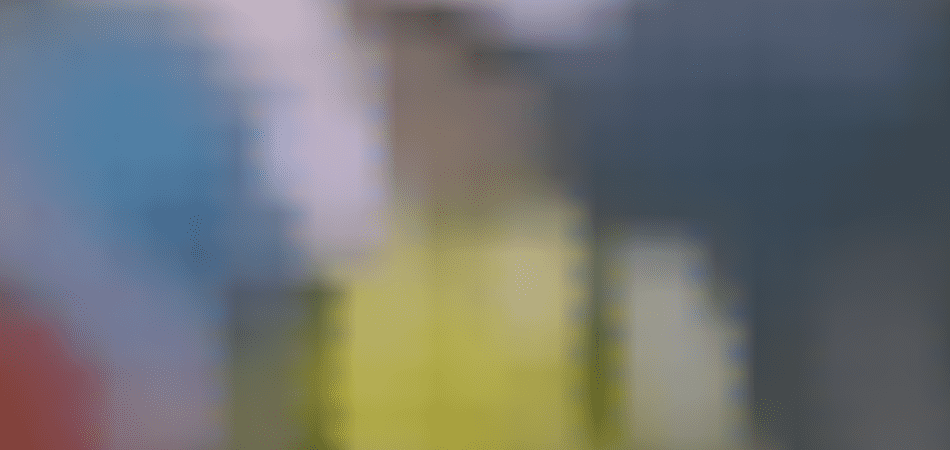}
  \\

  Artifact-free Ground Truth & Bicubic Interpolation & Nearest Neighbor Interpolation & Small{\tiny (3$\times$3)} Transposed Conv \\

  \includegraphics[width=0.325\textwidth, height=2.24cm]{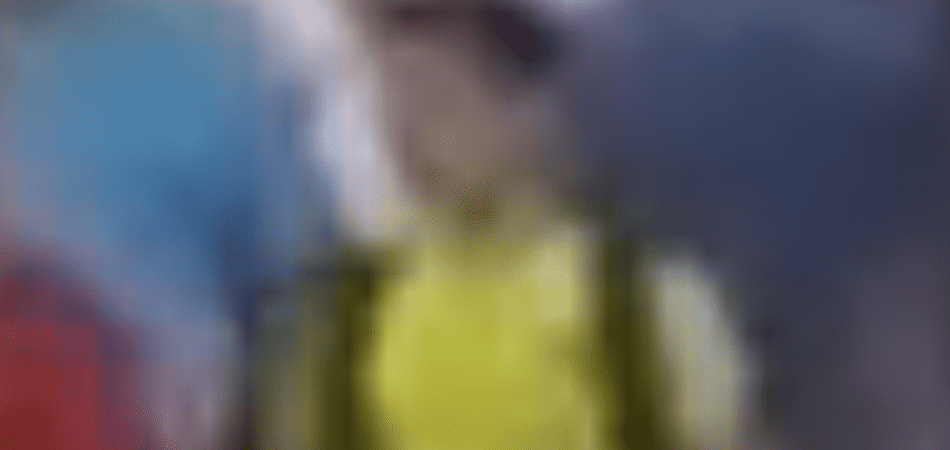}
& \includegraphics[width=0.325\textwidth, height=2.24cm]{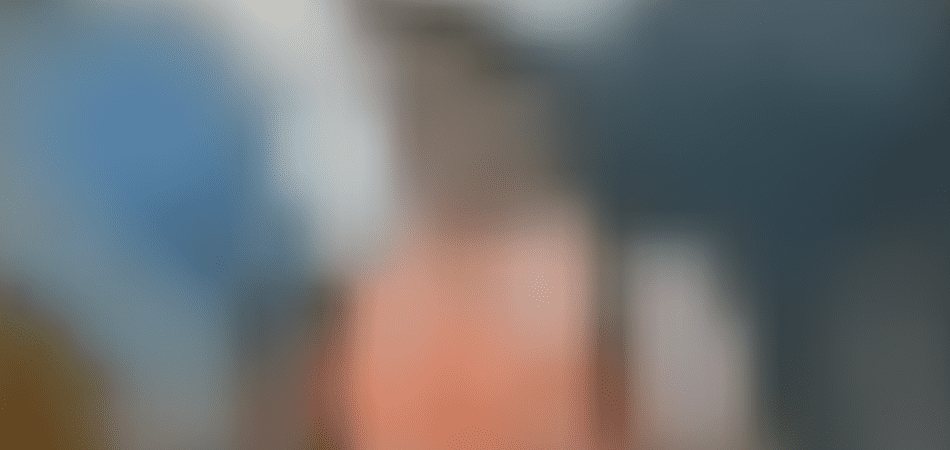}
& \includegraphics[width=0.325\textwidth, height=2.24cm]{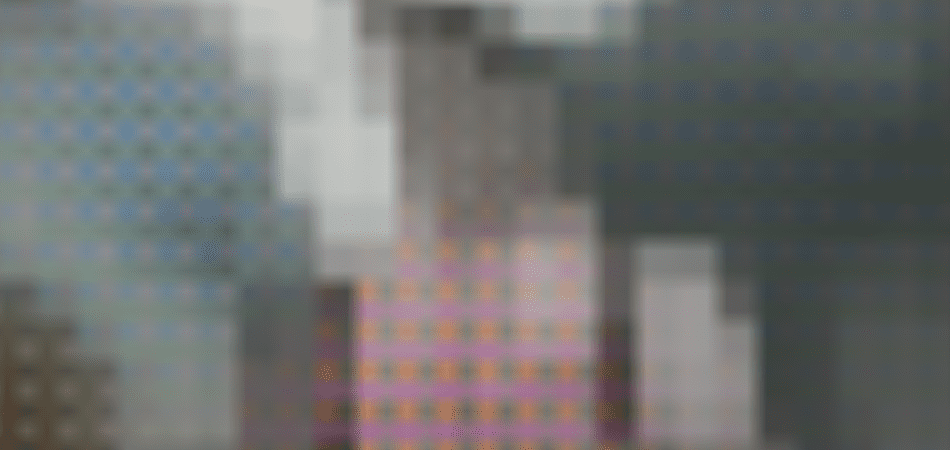}
& \includegraphics[width=0.325\textwidth, height=2.24cm]{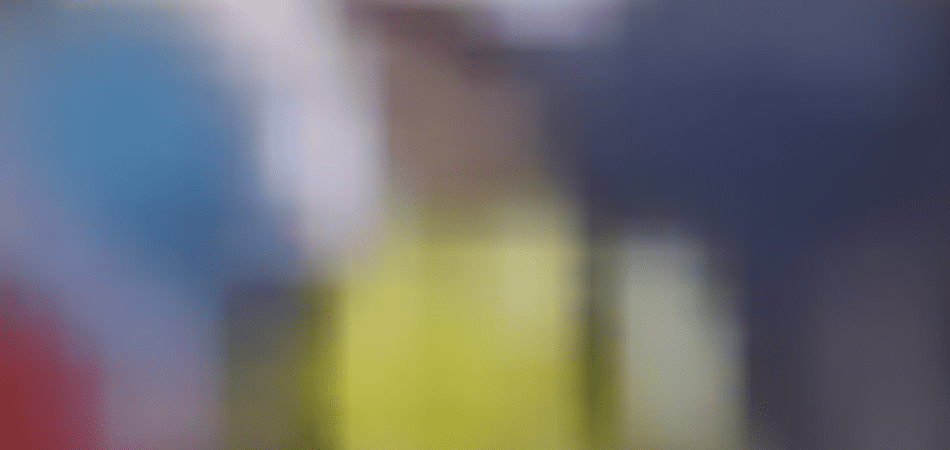}
  \\

  Zoomed-in Ground Truth & Bilinear Interpolation & Pixel Shuffle & Large{\tiny (7$\times$7+3$\times$3)} Transposed Conv \\

\end{tabular}
}
   \caption{An image from GoPro\cite{gopro} downsampled with 3$\times$3 MaxPooling and then upsampled
using various upsampling techniques. The resulting artifacts are compared on \textcolor{red}{zoomed-in red box regions} for better visibility. Bilinear interpolation causes over-smoothing. Bicubic interpolation causes overestimation along image boundaries while Pixel Shuffle and Nearest Neighbor cause strong grid artifacts along with discoloration. Small kernel transposed convolutions cause grid artifacts, however, on increasing kernel size we start getting better upsampling.} 
    \label{fig:upsampling_artifacts}
    \vspace{-1em}
\end{figure}

A standard countermeasure is interpolation of the inserted values with $\bar{I}_j=\frac{I_{j-1} + I_{j}}{2}$ for \textbf{linear interpolation} in Eq.~\eqref{eq:theory2}.
Linear interpolation (and in consequence bi-linear interpolation in 2D signals) corresponds to a convolution with a triangular impulse with width $2$, which can be represented as the convolution of two rectangle functions with width $1$. 
Accordingly, the Fourier response for frequency $\ell$, $\mathcal{F}_\ell$ of the triangular impulse is a squared sinc function ($\mathrm{sinc}^2(\ell)$) with $\mathrm{sinc}(\ell)=\frac{\mathrm{sin(\pi \ell)}}{\pi \ell}$. Since the output signal after interpolation is still discrete, i.e.~sampled with sampling rate $\frac{1}{2}$, a replica of the interpolation function, the $\mathrm{sinc}^2$ function, will appear with rate $2$ in the resulting spectrum (see also~\cref{fig:linear_interpoaltion}). 
The resulting interpolated signal is not optimal for several reasons. 
Most importantly, the spectrum of the interpolation function is not flat although the estimated values appear overly smooth (see \cref{fig:upsampling_artifacts}. ). 
This is arguably suboptimal for, for example, image restoration or segmentation tasks, where fine structural details are supposed to emerge in the upsampled data. 

Note that, in Eq.~\eqref{eq:theory2}, \textbf{pixel shuffle}~\cite{pixelshuffle} will set $\bar{I}_j$ to completely unrelated values of a different feature map channel, leading to a highly non-smooth signal with frequencies at the band limit. The resulting issues in the spectrum are similar to the ones caused by the bed of nails interpolation.
These spectral artifacts can be visually observed in \cref{fig:upsampling_artifacts}.

Therefore, in \textbf{transposed convolutions}, the interpolation function is not fixed to a predefined smoothing kernel but learned so that the resulting signal can represent fine details after the initial bed of nails interpolation and potentially learn to add fine details. One issue is that the learned convolution kernels may overlap based on the chosen stride and kernel size. 
If the stride is smaller than the kernel size, this will cause overlaps in the operation, leading to uneven contributions to different pixels in the upsampled feature map and thus to grid-like artifacts~\cite{aitken2017checkerboard, checkerboard_odena2016deconvolution}. 
Besides this rather technical aspect, transposed convolutions, if sufficiently large (thus also containing more context), could in principle learn to approximate correct upsampling functions. 
This can be understood when again looking at the Fourier representation. 
When interpolating, we want to increase the signal array size so that all the original information is preserved and the model can easily learn additional details. 
Such upsampling to preserve the information from the original low-resolution data is most easily achieved by transforming the signal to the Fourier domain, then padding the missing high-frequency parts with zeros and transforming the resulting array back to the spatial domain~\cite{PASPWEB2010}. 
In the Fourier domain, this padding operation can be understood as a point-wise multiplication of the desired full spectrum with a rectangle function with width $N$ (denoted $\mathrm{rect}_N$). 
Conversely, this operation corresponds to a convolution with $\mathcal{F}^{-1}(\mathrm{rect}_N) = \frac{1}{N}\mathrm{sinc}(xN)$ in the spatial domain. 
While the $sinc$ function drops off as $x$ increases, it never drops to zero. 
When applied for interpolation, its crests and the troughs cancel out the aliasing to a large extent as shown in \cref{fig:linear_interpoaltion}.
Thus, in order to allow the approximation of the optimal interpolation function, the kernel size in transposed convolutions has to be chosen as large as possible. 
This is, however, at odds with the ``learnability'' of suitable filter weights. 
Note that for pixel-wise predictions, models not only need to correctly interpolate, but they also need to ``fill in'' the missing details, which requires global as well as local context. 
Therefore, we expect a trade-off on the kernel size of transposed convolutions, where larger kernels improve the stability of the upsampled features and thus can reduce artifacts while the absolute prediction quality can suffer from very large learnable kernels. 
Sufficiently but not overly large kernels provide sufficient spatial context and are appropriate to allow for the model to learn when to blur and when to preserve/sharpen upsampled features. 
We illustrate this in \cref{fig:rebuttal_visualizing_weights} in \cref{subsec:appendix:image_restoration:kernel_weights}.

From this theoretical analysis of common upsampling methods, we derive the following hypotheses that we deem relevant for encoder-decoder architectures:
\begin{hyp}[H\ref{hyp:first}] \label{hyp:first}
\textbf{Large Context Transposed Convolutions (LCTC)} i.e. Large kernels in transposed convolution operations provide more context and reduce spectral artifacts and can therefore be leveraged by the network to facilitate better and more robust pixel-wise predictions.
\end{hyp}
\begin{hyp}[H\ref{hyp:second}, Null Hypothesis] \label{hyp:second}
To leverage prediction context and reduce spectral artifacts, it is crucial to increase the size of the \emph{transposed} convolution kernels (upsample using large filters). Increasing the size of normal (i.e.~non-upsampling) decoder convolutions does not have this effect.
\end{hyp}
In the following, we show the proposed, simple, and principled architecture changes that allow for studying the above hypotheses and improving robustness by improving feature stability.


\begin{figure*}[t]
	\begin{center}
	\includegraphics[width=\linewidth]{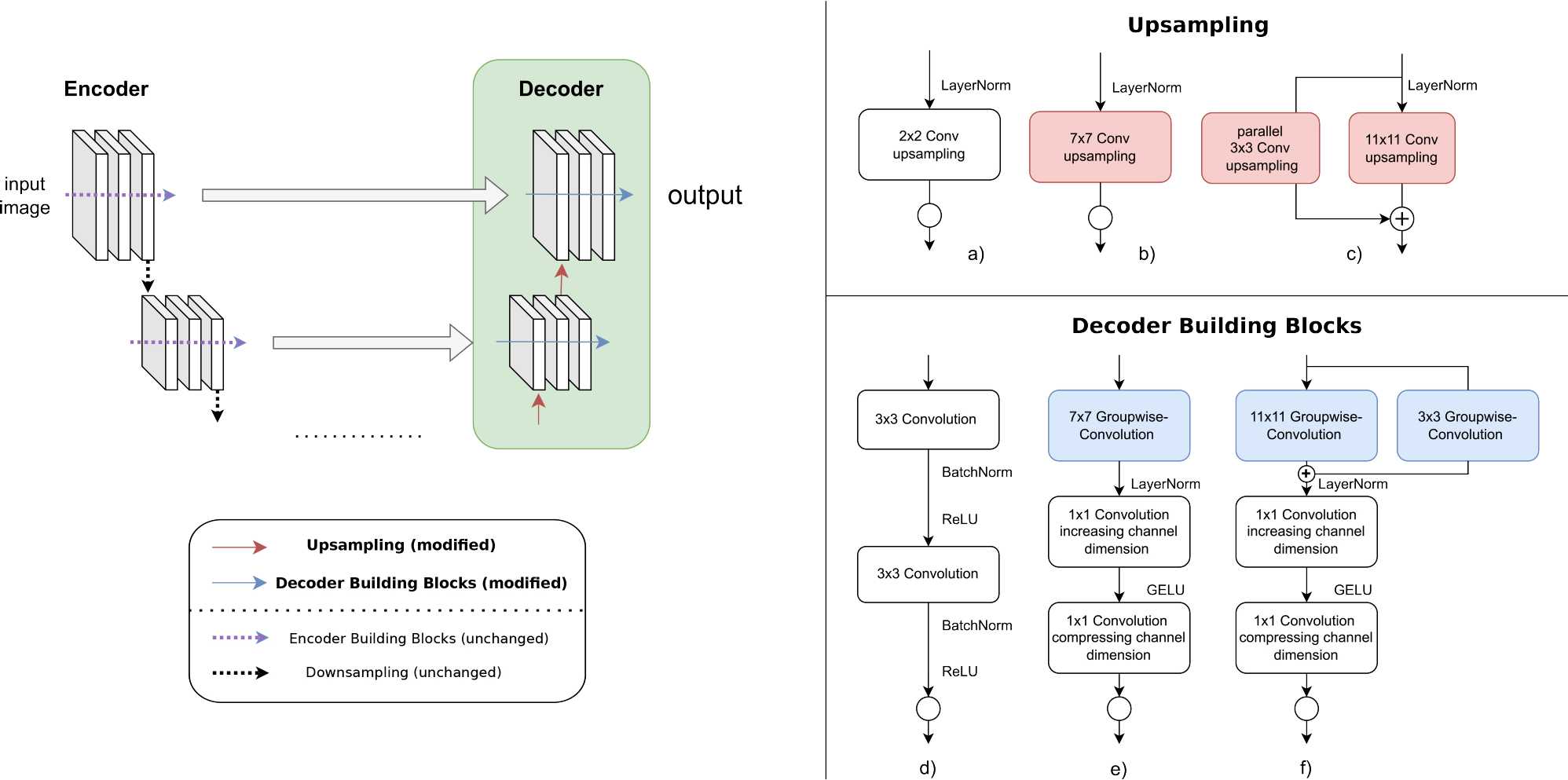}
	\caption[]{Abstract representation of an encoder-decoder architecture. 
	While for different tasks, the implementation of the model encoder varies~(including transformer-based encoders), our study focuses on the \emph{model decoder} (in green). The backbone for the decoder is commonly a ResNet-like structure for feature extraction \cite{unet,segnet}, additionally we also used a ConvNeXt-like~\cite{convnext} structure. We investigate variants of different upsampling operations (the operations along the {\color{BrickRed} red arrows} in the decoder) for fixed decoder blocks. We consider, as a probe for H\ref{hyp:first}, the baseline transposed deconvolution (a) in the top right), and for LCTC an increased convolution kernel size (b) in the top right), and an increased convolution kernel with a second path using a small convolution kernel (c) in the top right). 
To test whether the plain increase in parameters is responsible for improved results (zero hypotheses, H\ref{hyp:second}), we also ablate on the increase of convolution kernel size in the decoder block~(operations along the {\color{RoyalBlue} blue arrows} in the green block), as shown on the bottom right. 
 We consider 
 the common ResNet-like decoder building block structure (in d)) and two ConvNext-like structured backbones for the decoder building block 
 in e) and f), where f) has an additional small convolution applied in parallel, analog to c). 
	}\label{fig:unet_plus_block}
	\end{center}
	\vspace{-2em}
\end{figure*}

\section{Upsampling using Large Context Transposed Convolutions}
\label{sec:method}
Driven by the observations on upsampling artifacts, we investigate the advantage of larger kernel sizes during upsampling, for applications such as semantic segmentation or disparity estimation. 
Therefore, we keep the models' encoder part fixed and exclusively change operations in the architecture of the decoder part of the model. 
There, we have two design choices: {\color{BrickRed}Upsampling} -- The kernel size for the transposed convolution operations that learn upsampling, and {\color{RoyalBlue}Decoder Block} -- The kernel size in the convolution operations of blocks that learn to decode the features. 
Probing options for {\color{BrickRed}Upsampling} works towards proving H\ref{hyp:first} while a combination of both options proves H\ref{hyp:second}, \ie shows that a pure increase in the decoder parameters does not have the desired effect. 
This is considered in our ablation study in \cref{sec:ablation}.

\Cref{fig:unet_plus_block} summarizes the studied options for an abstract encoder-decoder architecture like \cite{unet}. The model decoder is depicted in the green box. 
Operations that we consider to be executed along the red upwards arrows ({\color{BrickRed}Upsampling} Operators) are detailed in the top right part of the figure (operations a) to c)). 
Operations that we consider to be executed along the blue sideways arrows ({\color{RoyalBlue}Decoder Building Blocks}) are depicted in the bottom right (operations d) to f)).
\paragraph{\textbf{Model Details. }}
Here, we provide details on the studied models. 
All implementation details are given in the Appendix \ref{appenxdix:subsec:exp:setup}. 

\noindent\textit{\textbf{Transposed Convolution Kernels for Upsampling. }}
The upsampling operation is typically performed with small kernels (2$\times$2 or 3$\times$3) in the transposed convolution operations~\cite{unet, SWav2, ddpm_baranchuk2021labelefficient}.
We aim to increase the spatial context during upsampling and to reduce grid artifacts. 
Thus we use \textbf{Large Context Transposed Convolutions (LCTC)}. 
We either use 7$\times$7 transposed convolutions or 11$\times$11 transposed convolutions with a parallel 3$\times$3 transposed convolution. 
Adding a parallel 3$\times$3 kernel is motivated by \cite{replk}, as large convolution kernels tend to lose local context, and thus adding a parallel small kernel helps to overcome this potential drawback~(see \cref{subsubsec:ablation:parallel_kernel}).

\noindent\textit{\textbf{Decoder Building Blocks. }}
\label{subsec:methods:decoder}
To verify that the measurable effects are due to the improved upsampling and not due to merely increasing the decoder capacity, we ablate on decoder convolution blocks similar to convolution blocks used in the ConvNeXt~\cite{convnext} basic block for encoding. 
While the standard ConvNeXt block uses a 7$\times$7 depth-wise convolution, we consider 7$\times$7 and 11$\times$11 group-wise convolutions, followed by layers present in a ConvNeXt basic block to analyze the importance of the receptive field within the block. 
\Cref{fig:unet_plus_block} (bottom right e) and f)) shows the structure of a ConvNeXt-style building block used in our work. 
First, a group-wise convolution is performed, followed by a LayerNorm~\cite{layernorm} and two 1$\times$1 convolutions which, similar to \cite{convnext}, creates an inverted bottleneck by first increasing the channel dimension and after a GELU~\cite{gelu} activation compressing the channel dimension again. 
We consider the ResNet-style building block (\Cref{fig:unet_plus_block}, d)), with 3$\times$3 convolution, yet without skip connection, as our baseline when studying this architectural design choice.

\begin{figure*}[t]
    \centering 
    \scriptsize
   \begin{tabular}{@{}c@{\hspace{0.1cm}}c@{\hspace{0.1cm}}c@{\hspace{0.1cm}}c@{\hspace{0.1cm}}c@{\hspace{0.1cm}}c@{\hspace{0.1cm}}c@{}}
   && PixelShuffle&$3\times 3$&$7\times 7 + 3\times 3$ (LCTC)&$11\times 11 + 3\times 3$ (LCTC)\\
  \rotatebox{90}{\phantom{su}Debluring}& \rotatebox{90}{\phantom{su}clean input}&\includegraphics[width=0.23\textwidth]{eccv_2024/figures/ablation/restoration/NAFNet_no_attack_GOPR0384_11_00-000002.png} &
 \includegraphics[width=0.23\textwidth]{eccv_2024/figures/ablation/restoration/NAFNet_3_0_no_attack_GOPR0384_11_00-000002.png}&
 \includegraphics[width=0.23\textwidth]{eccv_2024/figures/ablation/restoration/NAFNet_7_3_no_attack_GOPR0384_11_00-000002.png} &
 \includegraphics[width=0.23\textwidth]{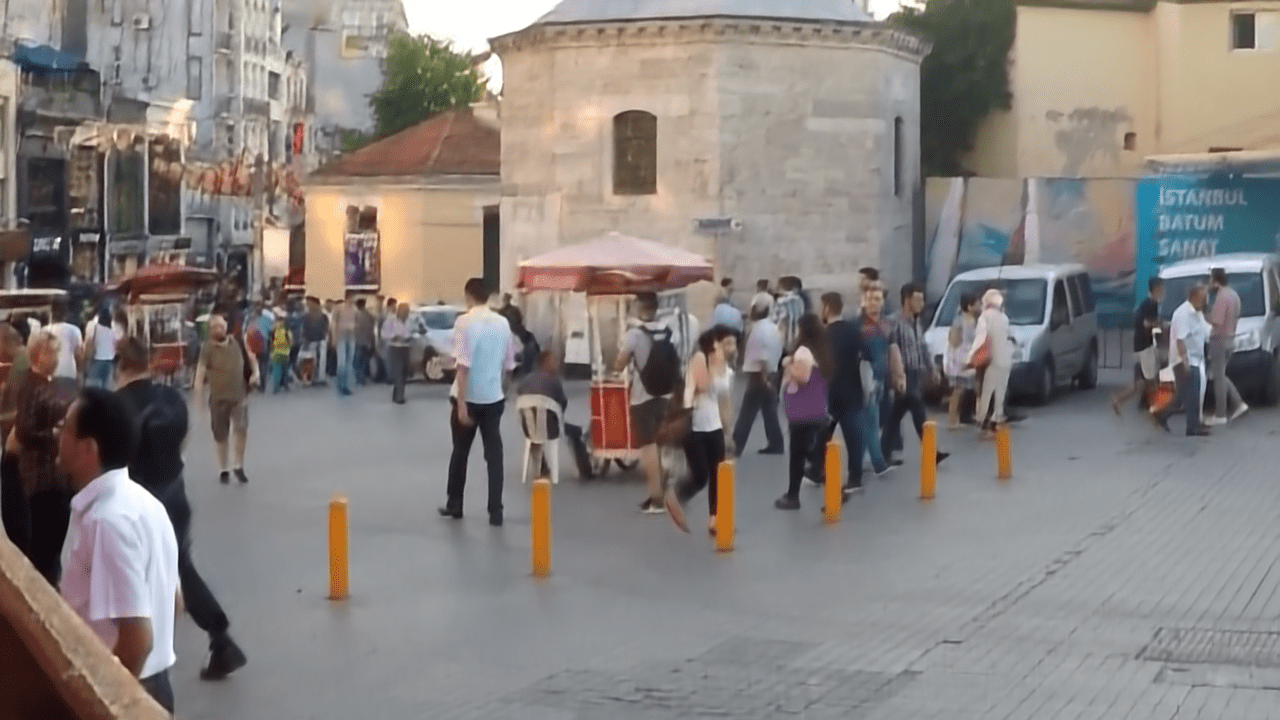}\\
   \rotatebox{90}{\phantom{su}Debluring}& \rotatebox{90}{attacked input}&\includegraphics[width=0.23\textwidth]{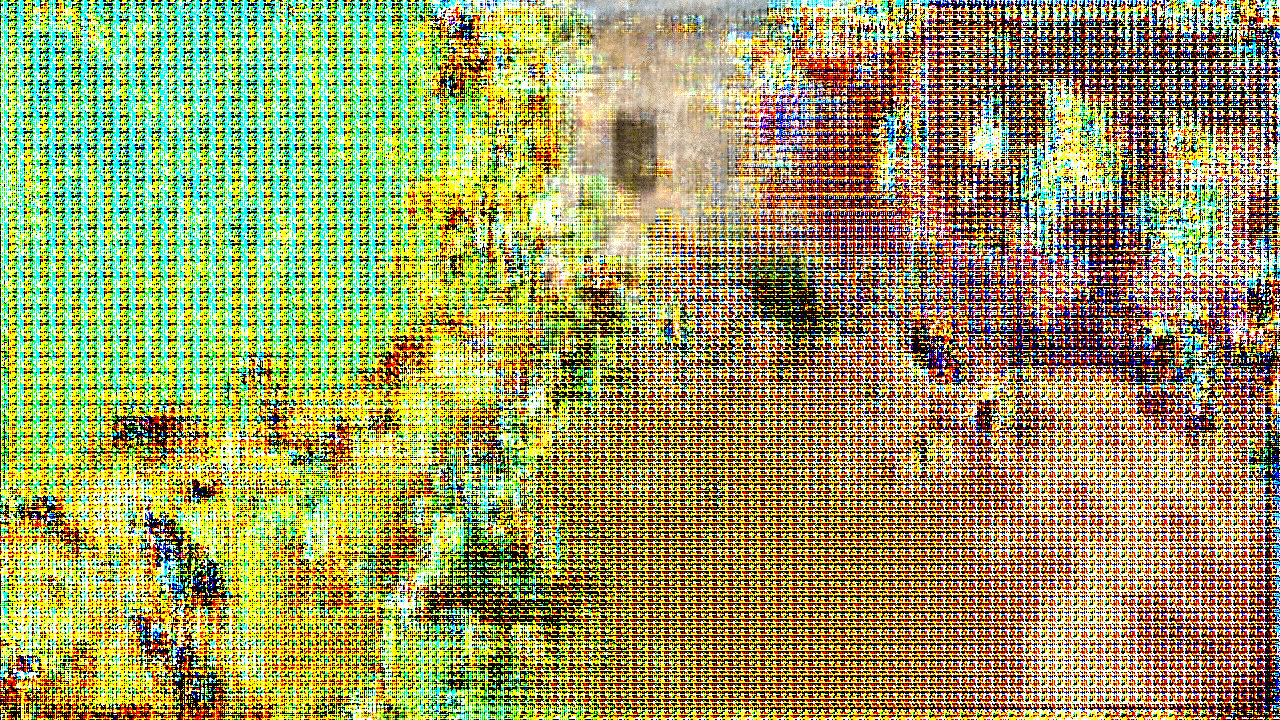} &
  \includegraphics[width=0.23\textwidth]{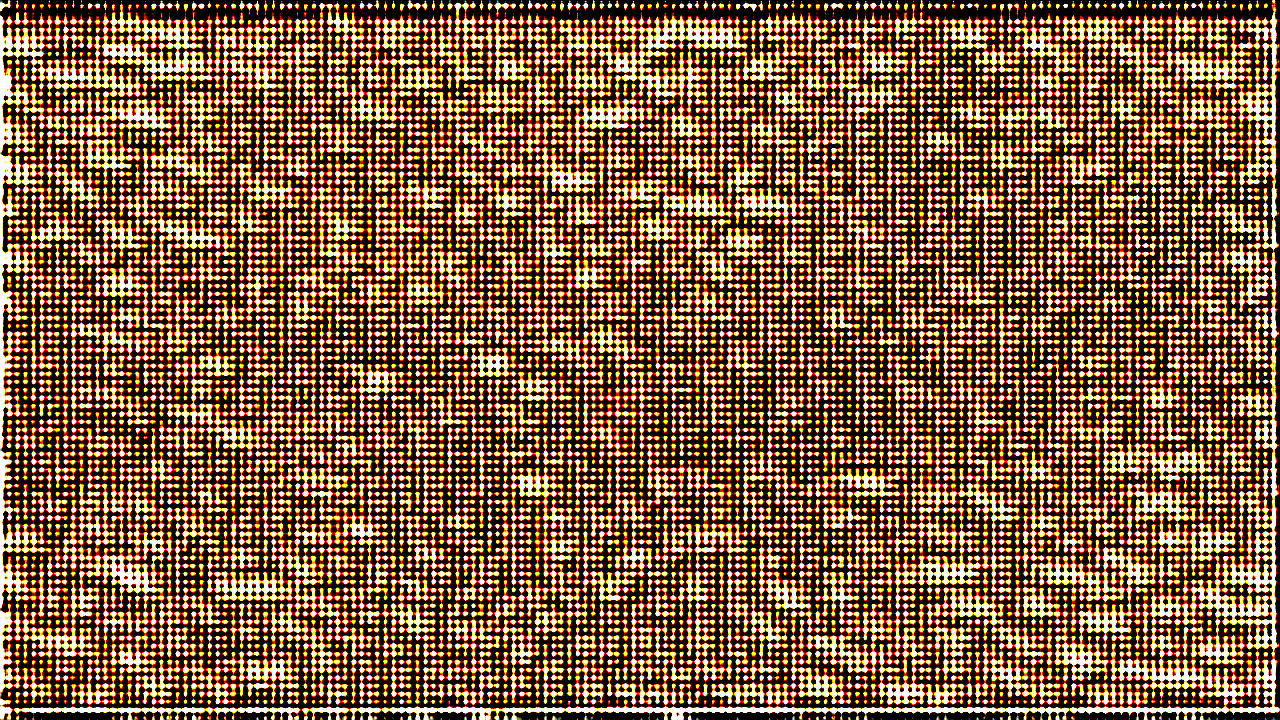} &
  \includegraphics[width=0.23\textwidth]{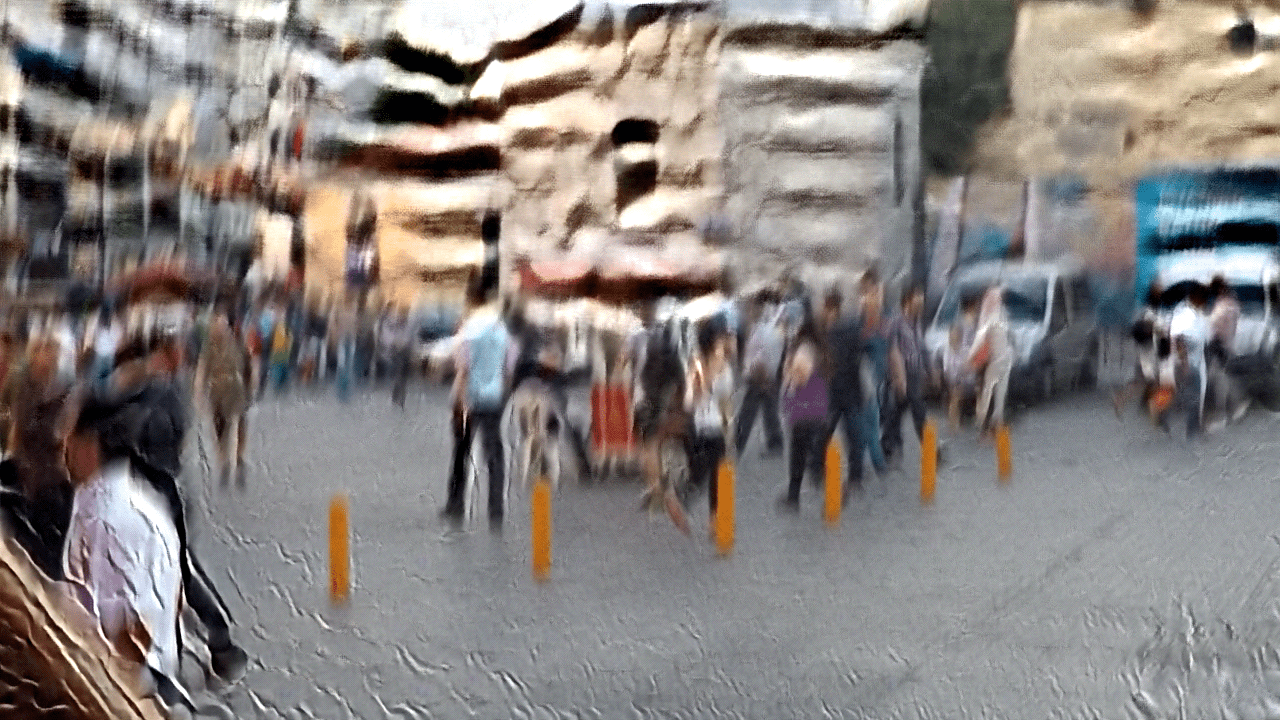} &
  \includegraphics[width=0.23\textwidth]{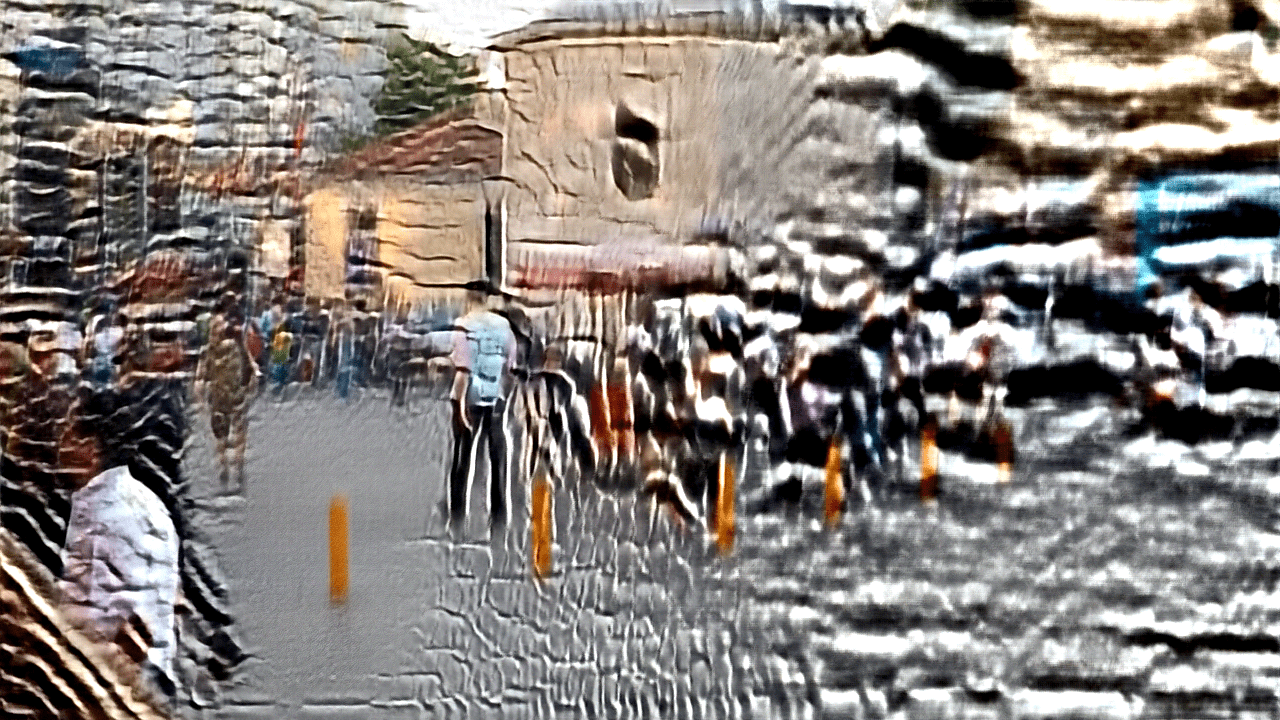} 
\end{tabular}

\caption{NAFNet, as proposed, uses Pixel Shuffle for upsampling. We modify only the upsampling operations to transposed convolution with kernel size (3$\times$3) and LCTC (Ours) for comparisons. We observe, for example, under a 10-step PGD attack with $\epsilon\approx\frac{8}{255}$ our proposed H\ref{hyp:first} gains validity. More examples for \cite{chen2022simple, zamir2022restormer} using different attacks and budgets are in \cref{subsec:appendix:image_restoration:visual_results}.}
\label{fig:reconstruction_pgd_attack_main}


\end{figure*}

\section{Experiments}
\label{sec:exp}
In the following, we evaluate the effect of the considered upsampling operators in several applications. 
We start by evaluating the effect on the upsampled feature stability of recent \textit{state-of-the-art} (SotA) image restoration models~\cite{zamir2022restormer,chen2022simple}, then provide results on semantic segmentation using more generic convolutional architectures that allow us to provide compulsory ablations. 
Last, we show that our results also extend to disparity estimation~\cite{sttr}.
We provide details on the used adversarial attacks, datasets, reported metrics, and other experimental details in \Cref{appenxdix:subsec:exp:setup}.



\begin{table*}[t]
\caption{Comparison of performances of different \textcolor{BrickRed}{upsampling} methods in \textit{SotA} Image Restoration Networks on the GoPro dataset. The architectures use Pixel Shuffle for Upsampling, we propose replacing the Pixel Shuffle with Large Context Transposed Convolutions (LCTC). We report additional results using adversarial training in \cref{tbl:ablation:image_restoration_adv_train}. Note, that some trade-off between clean performance and robustness is expected~\cite{zhang2019theoretically,tsipras2018robustness}. }
\label{tbl:ablation:image_restoration}
\centering
\scalebox{0.6}{
\begin{tabular}{@{}p{1.5cm} c@{\hspace{0.4cm}}|@{\hspace{0.2cm}} c @{\hspace{0.2cm}} c  @{\hspace{0.4cm}} |@{\hspace{0.2cm}} c @{\hspace{0.1cm}} c @{\hspace{0.2cm}} c @{\hspace{0.1cm}} c @{\hspace{0.2cm}} c @{\hspace{0.1cm}} c   @{\hspace{0.2cm}} | @{\hspace{0.2cm}} c @{\hspace{0.1cm}} c @{\hspace{0.2cm}} c @{\hspace{0.1cm}} c @{\hspace{0.2cm}} c @{\hspace{0.1cm}} c @{}}
\multirow{3}{1.5cm}{\textbf{Network}} & \multirow{3}{*}{\textbf{Upsampling Method}} & \multicolumn{2}{c}{\textbf{Test Accuracy}}   & \multicolumn{6}{c}{\textbf{CosPGD ($\epsilon\approx\frac{8}{255}$) attack iterations}} &  \multicolumn{6}{c}{\textbf{PGD ($\epsilon\approx\frac{8}{255}$) attack iterations}} \\
  &  & &  & \multicolumn{2}{c}{5} & \multicolumn{2}{c}{10} & \multicolumn{2}{c}{20}   & \multicolumn{2}{c}{5} & \multicolumn{2}{c}{10} & \multicolumn{2}{c}{20}                                  \\
& &  PSNR & SSIM &  PSNR & SSIM &  PSNR & SSIM &  PSNR & SSIM &  PSNR & SSIM &  PSNR & SSIM &  PSNR & SSIM \\
\midrule

\multirow{4}{*}{Restormer} & Pixel Shuffle & 31.99 & 0.9635 & 11.36 & 0.3236 & 9.05 & 0.2242 & 7.59 & 0.1548 & 11.41 & 0.3256 & 9.04 & 0.2234 & 7.58 & 0.1543 \\

& Transposed Conv 3$\times$3 & 9.68 & 0.095 &  8.24 & 0.0452 & 8.53 & 0.0628 & 8.44 & 0.0631 & 7.66 & 0.0464 &  7.72 & 0.0577 & 8.64 & 0.0527  \\

& LCTC: 7$\times$7 + 3 $\times$3 (Ours) & 29.51 & 0.9337 & 13.69 & 0.4186 & 11.53 & 0.3136 & 10.16 & 0.2484 & 13.69 & 0.4183 & 11.54 & 0.3137 & 10.16 & 0.2483 \\

& LCTC: 11$\times$11 + 3$\times$3 (Ours) & 29.44 & 0.9324 & \textbf{14.65} & \textbf{0.4251} &  \textbf{12.83} & \textbf{0.3438} & \textbf{11.48} & \textbf{0.29} & \textbf{14.65} & \textbf{0.4253} & \textbf{12.84} & \textbf{0.3445} & \textbf{11.48} & \textbf{0.2893} \\

\midrule

\multirow{4}{*}{NAFNet} & Pixel Shuffle &  32.87 & 0.9606 & 8.67 & 0.2264 & 6.68 & 0.1127  & 5.81 & 0.0617 & 10.27 & 0.3179  & 8.66 & 0.2282  &  5.95 & 0.0714\\

& Transposed Conv 3$\times$3 & 31.02 & 0.9422 & 6.15 & 0.0332 &  5.95 & 0.0258 & 5.87 & 0.0233 & 6.15 & 0.0332 & 5.95 & 0.0258 & 5.87 & 0.0234 \\

& LCTC: 7$\times$7 + 3 $\times$3 (Ours) & 31.12 & 0.9430 & \textbf{14.54} & \textbf{0.4827} & 11.05 & 0.3220 & 9.06 & 0.2213 & \textbf{14.53} & \textbf{0.4823} & 11.03 & 0.3201 & 9.08 & 0.2224 \\

& LCTC: 11$\times$11 + 3$\times$3 (Ours) & 30.77 & 0.9392 & 14.34 & 0.4492 & \textbf{11.41} & \textbf{0.3244} & \textbf{9.54} & \textbf{0.2411} & 14.34 & 0.45 & \textbf{11.4} & \textbf{0.3236} & \textbf{9.55} & \textbf{0.2398} \\


\end{tabular}
}
\vspace{-1em}
\end{table*}

In all cases, we observe that Large Context Transposed Convolutions (LCTC) improve the results of the respective pixel-wise prediction task in terms of 
stability under attack, showing that H\ref{hyp:first} holds. Further, our extensive ablation on image segmentation shows that increasing the convolution kernel in the decoder building blocks does not have this beneficial effect, providing experimental evidence for our hypothesis~H\ref{hyp:second} and confirming the impact of spectral artifacts on pixel-wise predictions.

\subsection{Image Restoration}
For image restoration, we consider the Vision Transformer-based Restormer~\cite{zamir2022restormer} and NAFNet~\cite{chen2022simple}.
Both originally use the Pixel Shuffle~\cite{pixelshuffle} for upsampling. 
Here, we compare the reconstructions from these proposed architectures to their variants using the proposed operators with large transposed convolution filters.
We use the same metrics as \cite{zamir2022restormer, chen2022simple},  Peak Signal-to-Noise Ratio (PSNR), and structural similarity index measure (SSIM)~\cite{ssim}. 
We perform our experiments on the GoPro~\cite{gopro} image deblurring dataset,  
following the experimental setup  in~\cite{agnihotri2023unreasonable}.
\paragraph{\textbf{Results on Image Restoration. }}
\label{subsec:experiments:restoration}
We first consider qualitative results on NAFNet~\cite{chen2022simple} in \Cref{fig:reconstruction_pgd_attack_main} and Restormer~\cite{zamir2022restormer} in \cref{fig:reconstruction_pgd_attack}, \cref{fig:reconstruction_cospgd_attack}(in \cref{subsec:appendix:image_restoration:visual_results}), where we see that the proposed upsampling operators allow for visually good results in image deblurring on clean data (similar to pixel shuffle). Yet, in contrast to pixel shuffle and the baseline small transposed convolution filters, the proposed Large Context Transposed Convolutions~(LCTC) significantly reduces artifacts that arise on attacked images (in this case, 10-step PGD with $\epsilon\approx\frac{8}{255}$). 
attacks with varying numbers of steps.

In \Cref{tbl:ablation:image_restoration}, we report the average PSNR and SSIM values of the reconstructed images from the GoPro test set. 
These results confirm that at filter size 3$\times$3, the performance of the transposed convolution variant of both the considered networks is significantly worse than the originally proposed Pixel Shuffle variant, justifying the community's extensive use of Pixel Shuffle. 
However, we observe on increasing context by increasing the kernel size to 7$\times$7 that the performance of the transposed convolution variants significantly improves, especially making the networks more stable when facing adversarial attacks.
This boost in performance is further accentuated by increasing the kernel size to 11$\times$11 (both with parallel small kernels).
These results provide evidence for Hypothesis~\ref{hyp:first}.

Note that the slightly reduced performance on clean images, seen in \Cref{tbl:ablation:image_restoration}, is expected to some degree: here, we only investigate sampling in the decoder, while pixel unshuffle is used in the encoder, potentially causing a mismatch.
Further, previous works have shown that there exists a trade-off between adversarial robustness and clean performance \cite{zhang2019theoretically,tsipras2018robustness}. However, we do not observe this trade-off for matching encoder-decoder architectures, \eg in semantic segmentation.
 
%
\begin{figure*}[htb]
    \centering 
    \scalebox{0.7}{
    \begin{tabular}{@{}c@{\hspace{0.1cm}}c@{\hspace{0.1cm}}c@{\hspace{0.1cm}}c@{\hspace{0.1cm}}c@{\hspace{0.1cm}}c@{\hspace{0.1cm}}c@{}}
    \rotatebox[origin=l]{90}{\small \centering \phantom{sub} Input} & \rotatebox[origin=l]{90}{\small \phantom{sub} Image}
    & \includegraphics[height=1.5cm, width=3cm]{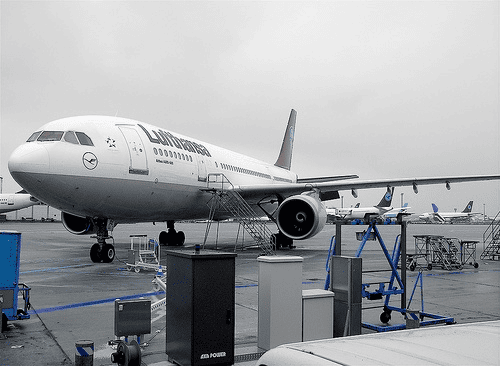}
    & \includegraphics[height=1.5cm, width=3cm]{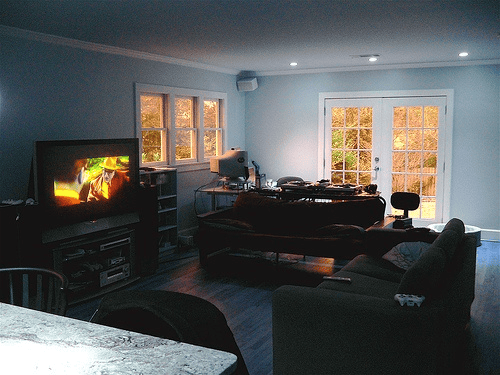}
    & \includegraphics[height=1.5cm, width=3cm]{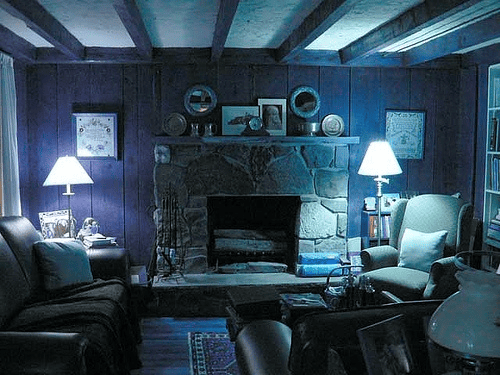}
    & \includegraphics[height=1.5cm, width=3cm]{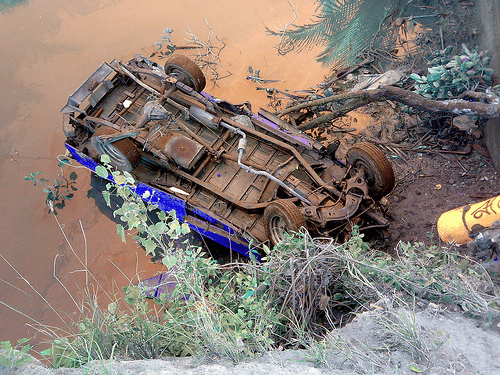} 
    & \includegraphics[height=1.5cm, width=3cm]{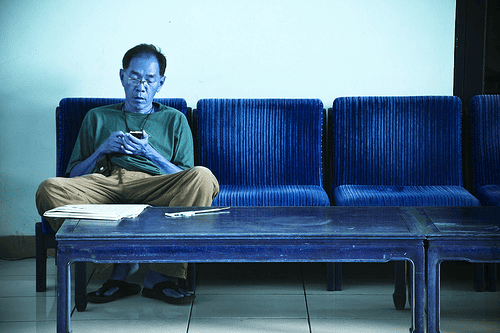}   \\
    \rotatebox[origin=l]{90}{\small Prediction} & \rotatebox[origin=l]{90}{\small Difference}
    & \includegraphics[height=1.5cm, width=3cm]{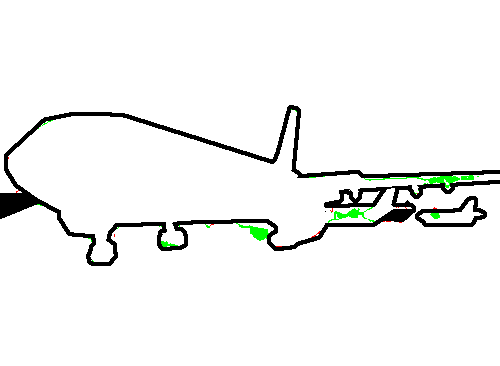}
    & \includegraphics[height=1.5cm, width=3cm]{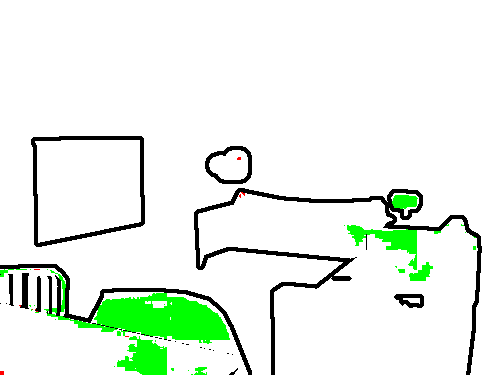}
    & \includegraphics[height=1.5cm, width=3cm]{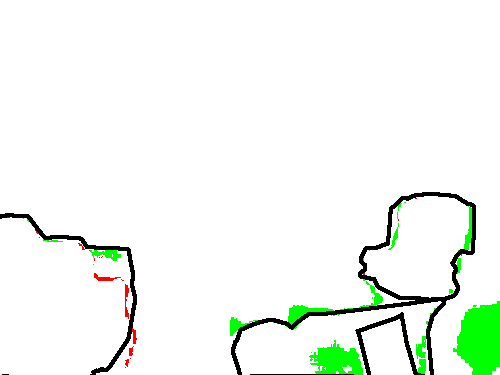}
    & \includegraphics[height=1.5cm, width=3cm]{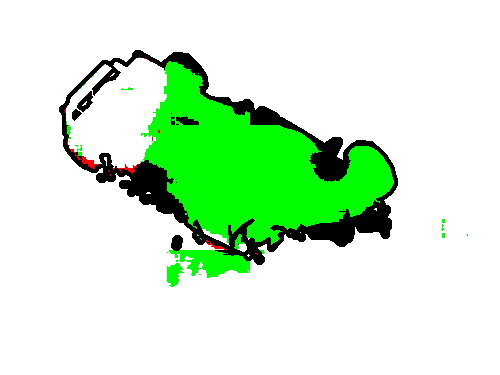}
    &  \includegraphics[height=1.5cm, width=3cm]{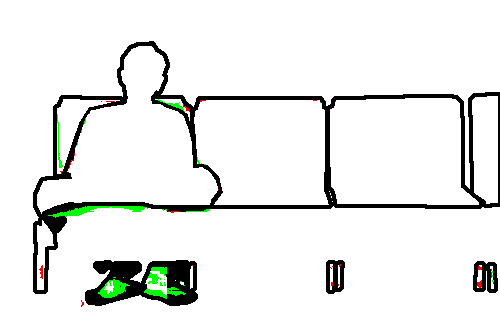}\\
\end{tabular}
}
\caption{A comparison of semantic segmentation mask predictions for the shown input images. 
The row labeled ``Prediction Difference'' shows the difference in predictions 
between the baseline model and the model with \textcolor{BrickRed}{Large Context Transposed Convolutions (11$\times$11+$3\times3$ kernels)}. On white pixels, both models agree. Red pixels indicate that the baseline model predicts correctly but our modified model predicts incorrectly. Green pixels indicate that our modified model predicts correctly but the baseline does not. The ground truth segmentation boundaries are drawn in black. Our modification improves the segmentation result along object boundaries, which can be attributed to spectral artifact removal, but also in more extended regions, where the context plays a more crucial role.}
\label{fig:comparison}
\vspace{-1em}
\end{figure*}
\subsection{Semantic Segmentation}
As baseline architecture for semantic segmentation, we consider a UNet-like architecture~\cite{unet} with encoder backbone layers from ConvNeXt~\cite{convnext} (see \cref{subsubsec:appendix:ablation:encoder} on the choice of encoder). 
This generic architecture facilitates providing a thorough ablation on all considered blocks in the decoder network. 
Our experiments are conducted on the PASCAL VOC 2012 dataset~\cite{pascal-voc-2012}. We report the mean Intersection over Union (mIoU) of the predicted and ground truth segmentation mask, the mean accuracy over all pixels (mAcc), and the mean accuracy over all classes (allAcc).
\paragraph{\textbf{Results on Semantic Segmentation. }}
\label{subsec:exp:semantic}

\begin{table*}[t]
\caption{Semantic Segmentation performance on the PASCAL VOC2012 validation set for UNet with ConvNeXt encoder, and the baseline UNet decoder (see \Cref{fig:unet_plus_block}) with differently sized \textcolor{BrickRed}{kernels in transposed convolution for feature map upscaling} while keeping rest of the architecture fixed. Additional results are provided in \cref{tbl:appendix:ablation:backbone} and \cref{tbl:appendix:ablation:segpgd_backbone} in \cref{subsubsec:appendix:exp:segment}.}
\label{tbl:exp:semantic:unet-convnext-adv}
\centering
\scalebox{.72}{
\begin{tabular}{@{}p{3.8cm}|ccc|ccc|ccc|ccc|ccc@{}}
 \multirow{3}{3cm}{\textbf{Transposed Convolution Kernels}} & \multicolumn{3}{c|}{\textbf{Clean}} & \multicolumn{6}{c|}{\textbf{FGSM attack epsilon}} & \multicolumn{6}{c}{\textbf{SegPGD ($\epsilon\approx\frac{8}{255}$) attack iterations}} \\
& \multicolumn{3}{c|}{\textbf{Test Accuracy}} & \multicolumn{3}{c}{$\frac{1}{255}$} & \multicolumn{3}{c|}{$\frac{8}{255}$}  &  \multicolumn{3}{c}{3} & \multicolumn{3}{c}{20} \\
 &  mIoU & mAcc & allAcc &   mIoU & mAcc & allAcc & mIoU & mAcc & allAcc & mIoU & mAcc & allAcc & mIoU & mAcc & allAcc\\
\midrule
 2$\times$2 (baseline)  &    78.34 & 86.89 & 95.15 &       53.54 & 70.96 & 86.08 &  47.02 & 65.41 & 82.78 &   23.06 & 46.51 & 45.30 & 5.54 & 18.79 & 23.72 \\
 
 LCTC: 7$\times$7 (Ours) &  78.92 & \textbf{88.06} & 95.23   &     56.02 & 74.13 & 86.45  &    49.24 & 68.89 & 82.87     &  26.53 & 53.05 & 61.16    &  \textbf{7.17} & 23.05 & \textbf{27.52} \\
 
 \textbf{LCTC: 11$\times$11 + 3$\times$3 (Ours)} &  \textbf{79.33} & 87.81 & \textbf{95.41} & \textbf{58.04} & \textbf{74.93} & \textbf{87.80}  & \textbf{51.25} & \textbf{69.31} & \textbf{84.64}  &    \textbf{27.49} & \textbf{53.08} & \textbf{64.13}  & 7.08 & \textbf{23.30} & 26.82 \\

\end{tabular}
}
\vspace{-1em}
\end{table*}

\begin{table}[t]
\caption{Adversarially trained models using FGSM ($\epsilon\approx\frac{8}{255}$) from \Cref{tbl:exp:semantic:unet-convnext-adv} tested against SegPGD adversarial attacks ($\epsilon\approx\frac{8}{255}$) on UNet with ConvNeXt encoder and decoder with different sized kernels in the \textcolor{BrickRed}{transposed convolution for upsampling}, while keeping rest of the architecture identical. See \cref{tbl:exp:semantic:unet-full-convnext-adv-training} in \cref{subsec:appendix:ablation:semseg:adv_training} for more evaluations including PGD training.}
\label{tbl:exp:semantic:unet-convnext-adv-training}
\centering
\scalebox{.8}{
\begin{tabular}{@{}p{4cm}|@{\hspace{0.2cm}}c@{\hspace{0.1cm}}c@{\hspace{0.1cm}}c@{\hspace{0.2cm}}|@{\hspace{0.2cm}}c@{\hspace{0.1cm}}c@{\hspace{0.1cm}}c@{\hspace{0.2cm}}|@{\hspace{0.2cm}}c@{\hspace{0.1cm}}c@{\hspace{0.1cm}}c@{}}
 \multirow{3}{3cm}{\textbf{Transposed Convolution Kernels}} & \multicolumn{3}{c@{\hspace{0.05cm}}}{\textbf{Clean}} & 
 \multicolumn{6}{c@{\hspace{0.05cm}}}{\textbf{SegPGD attack iterations}} \\
& \multicolumn{3}{c@{\hspace{0.05cm}}}{\textbf{Test Data}}  & 
\multicolumn{3}{c}{3} & \multicolumn{3}{c}{20} \\
 &   mIoU & mAcc & allAcc &   
 mIoU & mAcc & allAcc & mIoU & mAcc & allAcc\\
\midrule
 2$\times$2 & 78.57 & 86.68 & 95.23 & 26.59 & 48.99 & 67.71 & 7.6 & 24.06 & 31.37 \\
 LCTC: 7$\times$7 (Ours) & 78.41  & 86.22 & 95.20 & 28.11 & 53.39 & 66.30 & 8.36 & 28.54 & 28.13 \\
 \textbf{LCTC: 11$\times$11 + 3$\times$3 (Ours)} & \textbf{79.57} & \textbf{88.1} & \textbf{95.3} & \textbf{30.37} & \textbf{55.54} & \textbf{68.3} & \textbf{9.4} & \textbf{29.79} & \textbf{32.37}   \\

\end{tabular}
}
\end{table}

We first discuss the results for \textcolor{BrickRed}{different upsampling operations}.  
The remaining architecture is kept identical, with ResNet-style building blocks in the decoder, throughout these experiments.
The clean test accuracies are shown in \Cref{tbl:exp:semantic:unet-convnext-adv}. 
We see that as we increase the kernel size of the transposed convolution layers, there is a slight increase across all three evaluation metrics.
Moreover, \Cref{fig:comparison} visually demonstrates that, as we increase the size of the kernels in transposed convolution from 2$\times$2 (baseline) to 11$\times$11, the segmentations of the thin end and protrusions, for example, in the wing of the aircraft sample image are improving. 
The baseline model with small transposed convolution kernels could not predict these details. 
As hypothesized in H\ref{hyp:first}, we observe that increasing the context can reduce spectral artifacts caused when representation and images are upsampled using LCTC.
%
%
%
%
%

Further, in \Cref{tbl:exp:semantic:unet-convnext-adv}, we evaluate the performance of the segmentation models against FGSM~\cite{fgsm} and the multi-step attack SegPGD~\cite{segpgd} adversarial attacks for the indicated $\epsilon$ values. 
As expected, with the increasing intensity of the attack, the performance of all models drops.
Yet, even at high attack intensities, the larger kernels perform better than the small ones, and we see a trend of improvement in performance as we increase the kernel size, providing more evidence for Hypothesis~\ref{hyp:first}.

\vspace{-0.3cm}
\subsubsection{Ablation Study. }
\label{sec:ablation}
In the following, we first consider the effects of additional adversarial training, then ablate on the impact of other decoder building blocks and the filter size. Variations of the model encoder are ablated in the \cref{subsec:ablation:encoder_choice}, the impact of using small parallel kernels in addition to large kernels is ablated and discussed in \cref{subsubsec:ablation:parallel_kernel}, and competing upsampling techniques are ablated in \cref{subsec:appendix:ablation:upsampling}.

%
%
\vspace{-0.3cm}
\paragraph{\textbf{Adversarial Training. }}
In \Cref{tbl:exp:semantic:unet-convnext-adv-training}, we report results for FGSM adversarially trained models under SegPGD attack, with attacks as in \Cref{tbl:exp:semantic:unet-convnext-adv}. 
While the overall performance under attack is improved as expected, the trend of LCTC providing better results persists. 
More results for FGSM attack and SegPGD attacks with different numbers of iterations are given in \cref{tbl:appendix:ablation:backbone} and \cref{tbl:appendix:ablation:segpgd_backbone} in the Appendix. 
In \Cref{tbl:ablation:image_restoration_adv_train}, we additionally evaluate image restoration models under adversarial training.
\vspace{-0.3cm}
\paragraph{\textbf{Change in the decoder backbone architecture. }}
\label{subsec:ablation:backbone}
While all previous experiments focused on the upsampling using transposed convolutions in the decoder, we now evaluate the influence of the convolutional kernel size within the decoder which does not upsample (see \Cref{subsec:methods:decoder}).
For these experiments, we use a UNet-like architecture with a ConvNeXt backbone in the encoder and the PASCAL VOC 2012 dataset.

\begin{table*}[t]
\caption{Empirical evaluations for H\ref{hyp:second} using a UNet with ConvNeXt encoder. We observe that across different-sized kernels in \textcolor{BrickRed}{transposed convolution}, for a fixed kernel size, increasing the context in the \textcolor{RoyalBlue}{decoder building blocks} by using larger kernels causes performance deterioration. These observations for image decoding contrast the findings on image encoding by \cite{convnext,replk,slak}.}
\label{tbl:ablation:backbone_short}
\centering
\tiny
\scalebox{.93}{
\begin{tabular}{@{}p{2.3cm}|c@{}|c|cc|cc@{}}
\multirow{3}{2.3cm}{\textbf{Transposed Convolution Kernels}} &  \multirow{3}{2.5cm}{\textbf{Decoder Building Block Style}} & \multicolumn{1}{c}{\textbf{Test Accuracy}}        & \multicolumn{2}{c}{\textbf{FGSM attack epsilon}} & \multicolumn{2}{c}{\textbf{SegPGD} ($\epsilon\approx\frac{8}{255}$) \textbf{attack iterations}} \\
               &         & \multirow{2}{*}{mIoU  / mAcc / allAcc} &                $\frac{1}{255}$ & $\frac{8}{255}$                        &    3 & 20                           \\
                &        &                     &     mIoU / mAcc / allAcc & mIoU / mAcc / allAcc & mIoU / mAcc / allAcc  & mIoU / mAcc / allAcc  \\
\midrule
\multirow{3}{*}{2$\times$2} &   ResNet Style 3$\times$3   &   78.34 / 86.89 / 95.15   &    53.54 / 70.96 / 86.08  &     47.02 / 65.41 / 82.78   &     23.06 / 46.51 / 60.04   &     5.54 / 18.79 / 23.72   \\

&  ConvNeXt style 7$\times$7      & 77.17 / 86.86 / 94.81    &  49.98 / 72.22 / 83.93         &  42.04 / 64.86 / 79.08    &     17.94 / 44.81 / 47.96    &   3.20 / 14.73 / 9.81    \\

&  ConvNeXt style 11$\times$11 + 3$\times$3     &   77.17 / 86.86 / 94.81  &     47.34 / 67.72 / 83.34  &  37.91 / 57.79 / 78.21     &    13.97 / 35.82 / 45.68   &    2.21 / 10.75 / 5.29    \\

\midrule
\multirow{3}{*}{LCTC: 7$\times$7  (Ours)}  &  \textbf{ResNet Style 3$\times$3}    &   78.92 / \textbf{88.06} / 95.23  & 56.02 / 74.13 / 86.45    &   49.24 / 68.89 / 82.87   &  26.53 / 53.05 / 61.16     &    \textbf{7.17} / 23.05 / \textbf{27.52} \\

&  ConvNeXt style 7$\times$7   &  77.57 / 87.04 / 94.92    & 52.93 / 72.18 / 85.51 &  44.89 / 65.71 / 80.74     &   17.64 / 43.32 / 47.80     &      1.86 / 7.18 / 3.55  \\

&  ConvNeXt style 11$\times$11 + 3$\times$3   &   77.99 / 87.86 / 94.96   &   51.61 / 73.01 / 84.85   &  43.93 / 66.22 / 80.73    &  17.07 / 42.30 / 48.78  &  1.80 / 7.11 / 3.04   \\

\midrule
\multirow{3}{*}{\textbf{LCTC: 11$\times$11 +3$\times$3} (Ours)} &    \textbf{ResNet Style 3$\times$3}    &    \textbf{79.33} / 87.81 / \textbf{95.41}   & \textbf{58.04 / 74.93 / 87.80}   & \textbf{51.25 / 69.31 / 84.64}     &     \textbf{27.49 / 53.08 / 64.13}    & 7.08 / \textbf{23.30} / 26.82 \\

&   ConvNeXt style 7$\times$7   &   78.32 / 86.98 / 95.09   &  53.31 / 72.45 / 86.16   & 44.89 / 65.18 / 82.03     &   16.14 / 40.65 / 50.39    &    1.93 / 9.35 / 3.90   \\

&  ConvNeXt style 11$\times$11 + 3$\times$3    &   77.42 / 86.24 / 94.94   &  54.48 / 72.53 / 86.25    &   46.67 / 66.59 / 82.29     &   18.76 / 44.60 / 51.49     &    2.31 / 8.70 / 3.50    \\

\end{tabular}
}
\vspace{-1em}
\end{table*}
In \Cref{tbl:ablation:backbone_short} we observe, for a \textcolor{BrickRed}{fixed transposed convolution} kernel size, as \textcolor{RoyalBlue}{we increase the size of the convolution kernel in the decoder building blocks}, the performance of the model \emph{decreases}. This phenomenon extends to the performance of the architectures under adversarial attacks, showing that a mere increase in parameters in the model decoder does not have a positive effect on model performance or on its stability.
This proves the validity of hypothesis~H\ref{hyp:second}.
An explanation for this phenomenon could be that we only need to increase context during the \textcolor{BrickRed}{actual upsampling step}, increasing context in the consequent \textcolor{RoyalBlue}{decoder building blocks} has a negligible effect on the quality of representations learned.
However, the increase in the number of parameters makes the architecture more susceptible to adversarial attacks.

\vspace{-0.3cm}
\paragraph{\textbf{Ablation on filter size saturation. }}
\label{subsec:ablation:kernel_limit}
\begin{figure}[t]
	\begin{center}
	\includegraphics[width=0.7\linewidth]{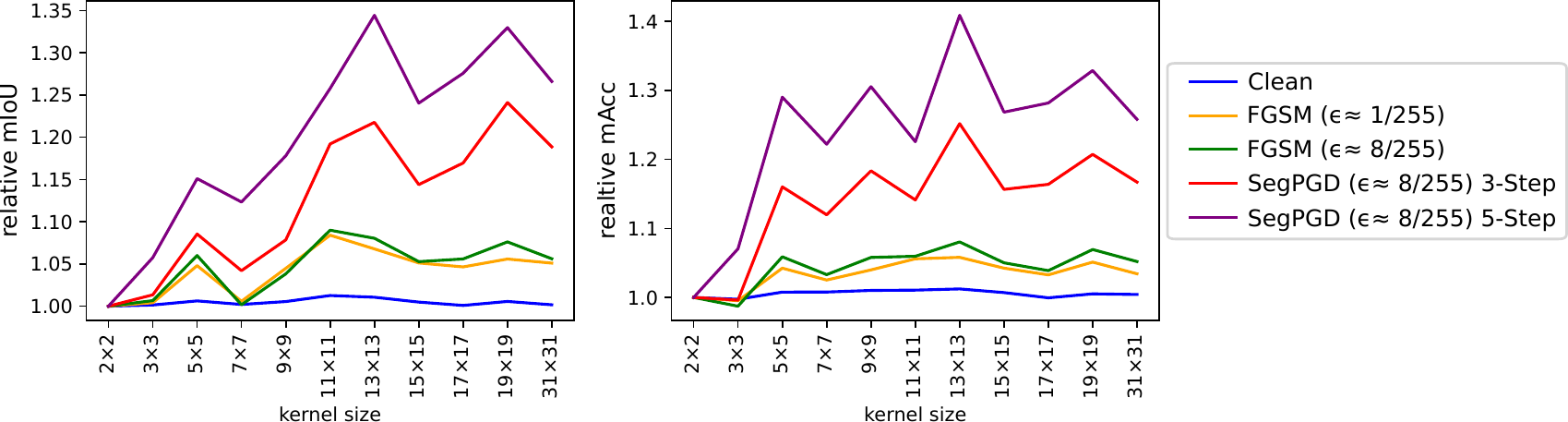}
	\caption[]{Performance comparison on PASCAL VOC2012  using UNet with ConvNeXt encoder for different LCTC sizes 
 from $2\times2$ (small) to $31\times31$ (LCTC) kernels. 
 All, besides the baseline with $2\times2$ and $3\times3$, have a parallel 3×3 kernel, as shown in \Cref{fig:unet_plus_block} (bottom left). For the \textcolor{RoyalBlue}{decoder building block backbone, a ResNet Style $3\times3$ style is used}. See \cref{tbl:ablation:kernel_limit} for the values.
	}\label{fig:transpose_conv_plots}
	\end{center}
	\vspace{-1em}
\end{figure}
After proving H\ref{hyp:first} one could argue that networks will consistently improve with increased \textcolor{BrickRed}{kernel size for Large Context Transposed Convolutions}. Hence, 
we test larger kernel sizes of 15$\times$15, 17$\times$17, 19$\times$19 and 31$\times$31 kernels. Yet, as seen in  \Cref{fig:transpose_conv_plots}, the effect of the kernel size appears to saturate: the performance after 13$\times$13 and the performance of 31$\times$31 kernels is not better than for 11$\times$11 kernels. Yet, they are significantly better than the baseline's performance.

\vspace{-0.3cm}
\paragraph{\textbf{Ablation on different Upsampling Methods. }}
\label{subsec:ablation:different_upsampling}


\begin{table*}[t]
\caption{Comparison of performances of different upsampling methods in the UNet-like architecture. All architectures use the baseline (ConvNeXt) encoder and 3$\times$3 convolution kernels in the decoder block. Please refer to \Cref{tbl:appendix:ablation:upsampling_full} in \Cref{subsec:appendix:ablation:upsampling} for more evaluations and discussion, including those with ConvNeXt style 7$\times$7+3$\times$3 Convolution kernels in the decoder blocks.}
\label{tbl:ablation:upsampling}
\centering
\scalebox{0.65}{
\begin{tabular}{@{}p{4.2cm} | @{\hspace{0.2cm}} c @{\hspace{0.1cm}} c @{\hspace{0.1cm}} c @{\hspace{0.2cm}}| @{\hspace{0.2cm}} c @{\hspace{0.1cm}} c @{\hspace{0.1cm}} c @{\hspace{0.2cm}} c @{\hspace{0.1cm}} c @{\hspace{0.1cm}} c @{\hspace{0.2cm}} | @{\hspace{0.2cm}} c @{\hspace{0.1cm}} c @{\hspace{0.1cm}} c @{\hspace{0.2cm}} c @{\hspace{0.1cm}} c @{\hspace{0.1cm}} c @{}}
\multirow{3}{5.2cm}{\textbf{Upsampling Method}} & \multicolumn{3}{c}{\textbf{Test Accuracy}}   & \multicolumn{6}{c}{\textbf{FGSM attack epsilon}} &  \multicolumn{6}{c}{\textbf{SegPGD} ($\epsilon\approx\frac{8}{255}$) \textbf{attack iterations}} \\
  &   &  & & \multicolumn{3}{c}{$\frac{1}{255}$}  & \multicolumn{3}{c}{$\frac{8}{255}$}  &  \multicolumn{3}{c}{5}   &  \multicolumn{3}{c}{20} \\
&   mIoU & mAcc & allAcc & mIoU & mAcc & allAcc  & mIoU & mAcc & allAcc & mIoU & mAcc & allAcc & mIoU & mAcc & allAcc \\
\midrule

Pixel Shuffle & 78.54 & 87.32 & 95.18 & 53.82 & 71.58 & 85.88 & 46.67 & 65.03 & 81.71 & 15.06 & 38.85 & 41.71 & 6.69 & 23.43 & 24.05 \\

 Nearest Neighbour Interpolation & 78.40 & \textbf{88.16} & 95.09 & 52.68 & 73.51 & 84.55 & 46.08 & 67.96 & 80.22 & 15.34 & \textbf{44.53} & 36.21 & \textbf{7.65} & \textbf{27.89} & 20.48   \\

  Transposed Convolution 2$\times$2 &   78.45 & 86.66 & 95.20 & 53.76 & 70.62 & 86.32 & 47.33 & 64.58 & 83.16 &  14.43 & 35.50 & 45.30 & 5.54 & 18.79 & 23.72 \\

 \textbf{LCTC: 11$\times$11+3$\times$3 (Ours)} & \textbf{79.33} & 87.81 & \textbf{95.41} & \textbf{58.04} & \textbf{74.93} & \textbf{87.80} & \textbf{51.25} & \textbf{69.31} & \textbf{84.64} & \textbf{18.15} & 43.51 & \textbf{49.36} & 7.08 & 23.30 & \textbf{26.82}  
\vspace{-2em}


\end{tabular}
}
\end{table*}

Following, we compare different upsampling techniques thus justifying our advocacy for using LCTC instead of other upsampling techniques like interpolation and pixel shuffle in the real world.

We report the comparison in \Cref{tbl:ablation:upsampling} and observe that both Pixel shuffle and Nearest Neighbor interpolation perform better than the usually used Transposed Convolution with a 2$\times$2 kernel size. 
However, as we increase the kernel size for Transposed Convolution to 11$\times$11 with a 3$\times$3 small kernel in parallel, we observe that LCTC is strictly outperforming Pixel Shuffle, on both clean unperturbed images and under adversarial attacks, across all metrics used. 
Large Context Transposed Convolutions are either outperforming or performing at par with Nearest Neighbor interpolation.
Thus we demonstrate the superior clean and adversarial performance of Large Context Transposed Convolutions operation over other commonly used techniques.

\subsection{Disparity Estimation}
\label{subsec:exp:disparity}
To show that the observations extend from image restorations and segmentation to other tasks, we conduct additional experiments for disparity estimation. 
We consider the STTR-light~\cite{sttr} architecture, built from STTR, which is a recent state-of-the-art vision-transformer based model for disparity estimation and occlusion detection. 
To implement the proposed modification, we alter the kernel sizes in the transposed convolution layers used for pixel-wise upsampling in the ``feature extractor'' module of the architecture from 3$\times$3 kernels to larger kernels. 
We conduct evaluations on FlyingThings3D~\cite{flyingthings_MIFDB16} and keep all other details as implemented in \cite{sttr}.
%

\begin{wraptable}{r}{6.9cm}
\vspace{-2.8em}
\caption{Comparison of performance of STTR-light architecture with different sized kernels in \textcolor{BrickRed}{transposed convolution for upsampling} the feature maps in the feature extractor (lower is better). The entire set of results is provided as \cref{tbl:appendix:exp:disparity:sttr} in \cref{subsubsec:appendix:exp:disparity}.}
\label{tbl:exp:disparity:sttr}
\centering
\scalebox{.65}{
\begin{tabular}{@{}p{4cm} @{\hspace{0.2cm}}| @{\hspace{0.2cm}}c@{\hspace{0.1cm}}c@{\hspace{0.2cm}}|@{\hspace{0.2cm}}c@{\hspace{0.1cm}}c@{}}
 \multirow{2}{3.5cm}{\textbf{Transposed Convolution Kernels}} & \multicolumn{2}{c}{\textbf{Test Accuracy}}    & \multicolumn{2}{c}{\textbf{3-Step PGD attack}}   \\
     & \multirow{1}{*}{epe\textcolor{green}{$\downarrow$}}   & \multirow{1}{*}{3px error\textcolor{green}{$\downarrow$}} & \multirow{1}{*}{epe\textcolor{green}{$\downarrow$}}    & \multirow{1}{*}{3px error\textcolor{green}{$\downarrow$}}\\
\midrule
STTR-light \cite{sttr} reported   &   0.5        &    1.54  & -   &  - \\
\midrule
   3$\times$3 \cite{sttr} reproduced  &  0.4927    &  1.54   &  4.05  &  18.5    \\
   \textbf{LCTC: 7$\times$7 + 3$\times$3 (Ours)} &   \textbf{0.4788} & \textbf{1.50}   & \textbf{4.02}   &   \textbf{18.3}   \\
\end{tabular}
}
\vspace{-2em}
\end{wraptable}

%
In \Cref{tbl:exp:disparity:sttr}, we report the improvements in performance due to our architecture modification of increasing the size of the \textcolor{BrickRed}{transposed convolution kernels used for upsampling}, from the 3$\times$3 in the baseline model to 7$\times$7 (LCTC). %
Similar to previous applications, the increased kernel sizes with parallel 3$\times$3 kernels further facilitate to stabilize the model when attacked, as evaluated here for 3 attack iterations using PGD with $\epsilon\approx\frac{8}{255}$ on the disparity loss.
Indicating that larger kernels in the transposed convolutions can better decode learned representations from the encoder regardless of the specific downstream task.
We provide visual results in \cref{subsubsec:appendix:exp:disparity}.
%

\section{Conclusion}
\label{sec:conclusion}
We provide conclusive reasoning and empirical evidence for our hypotheses on the importance of context during upsampling. While increasing the \textcolor{BrickRed}{size of convolutions during upsampling (LCTC)} increases prediction stability, increasing the size of those \textcolor{RoyalBlue}{convolution layers without upsampling} does not benefit the network. 
This indicates that observations made for increased context during encoding do not translate to decoding.
Further, we show that our simple LCTC can be directly incorporated into recent models, yielding better stability even in ViT-based architectures like Restormer, NAFNet, and STTR-light as well as in classical CNNs. 
Our observations are consistent across several architectures and downstream tasks.

\noindent\textbf{Limitations. }Current metrics for measuring performance do not completely account for spectral artifacts.
Spectral artifacts begin affecting these metrics only when they become pronounced such as under adversarial attacks, and here LCTC consistently performs better across tasks and architectures.
Ideally, we would want infinitely large kernels, however, with increasing kernel size and task complexity, training extremely large kernels can be challenging. Thus, in this work, while having ablated over kernels as large as 31$\times$31, we propose using kernels only as large as 7$\times$7 to 11$\times$11 for good practical trade-offs. Further improvements might be possible when jointly optimizing the encoder \emph{and} decoder. 
%
Moreover, there might exist other factors that contribute to the introduction and existence of spectral artifacts such as spatial bias.

\section*{Acknowledgements. }
Margret Keuper acknowledges funding by the DFG Research Unit 5336 - Learning to Sense.
The OMNI cluster of the University of Siegen was used for some of the initial computations.
Additionally, Shashank Agnihotri would like to thank Dr. Bin Zhao for his help in translating \cite{voc_instructions}.

%
%
\bibliographystyle{splncs04}
\bibliography{main}

\appendix
\hfill
\newpage

\onecolumn
{
    \centering
    \Large
    \textbf{Improving Feature Stability during Upsampling -- Spectral Artifacts and the Importance of Spatial Context} \\
    \vspace{0.5em}Supplementary Material \\
    \vspace{1.0em}
}
In the following, we present results and figures to support our statements in the main paper and provide additional information.
The following has been covered in the appendix:

\vspace{1em}
\begin{itemize}
\setlength\itemsep{3em}
    \item \Cref{appenxdix:subsec:exp:setup}: \textbf{Detailed experimental setup} for all downstream tasks.    
    \vspace{1em}
    \begin{itemize}
    \setlength\itemsep{1em}
        \item \Cref{subsubsec:exp:setup:restoration}: Image Restoration experimental setup
        \item \Cref{subsubsec:exp:setup:segment}: Semantic Segmentation experimental setup
        \item \Cref{subsubsec:exp:setup:disparity}: Disparity Estimation experimental setup
        \item \Cref{subsubsec:exp:setup:adversarial}: Detailed setup of Adversarial attacks for all downstream tasks.
        \item \Cref{subsubsec:exp:setup:adv_training}: Detailed setup of adversarial training for semantic segmentation and image restoration.
    \end{itemize}
    \item \Cref{subsec:appendix:experiments}: \textbf{Semantic Segmentation}: Additional Experiments and Ablations. In detail:
    \vspace{1em}
    \begin{itemize}
    \setlength\itemsep{1em}
    \item \Cref{subsubsec:appendix:exp:segment}: Detailed results from \cref{subsec:exp:semantic} and \cref{sec:ablation}.
    \item \Cref{subsubsec:appendix:limit_large}: Discussion on saturation of kernel size for upsampling.
    \item \Cref{subsubsec:appendix:ablation:encoder}: An ablation on the impact of the capacity of the encoder block for standard options such as ResNet or ConvNeXt blocks.
    \item \Cref{subsubsec:ablation:parallel_kernel}: Ablation about including or excluding a small parallel kernel during upsampling using transposed convolution.
    \item \Cref{subsubsec:experiments:pspnet_interpolation}: Short study on drawbacks of using interpolation for pixel-wise upsampling.
    \item \Cref{subsec:appendix:ablation:upsampling}: A comparison to different kinds of upsampling Operations on Segmentation Models. 
    \item  \Cref{subsec:appendix:ablation:semseg:adv_training}: A comparison of the performance of different sized kernels in the transposed convolution operations of UNet-like models adversarially trained using FGSM attack and 3-step PGD attack on 50\% of the mini-batches during training.
    \end{itemize}
\item \Cref{subsec:appendix:ablation:image_restoration}: \textbf{Image Restoration} : Additional Results:
    \vspace{1em}
    \begin{itemize}
    \setlength\itemsep{1em}
        \item \Cref{subsec:appendix:image_restoration:latency}: Here we report the number of parameters and latency study of LCTC.
        \item \Cref{subsec:appendix:image_restoration:adv_training}: Adversarial training evaluation for Restormer and NAFNet for Image deblurring task.
        \item \Cref{subsec:appendix:image_restoration:visual_results}: Qualitative results for image reconstruction models using Restormer and NAFNet and evaluated on clean data, PDG and CosPGD attack with varying numbers of attack iterations.
        \item \Cref{subsec:appendix:image_restoration:kernel_weights}: \textbf{Visualizing Kernel Weights}: Here we visualize kernel weights from a random channel for models from \Cref{fig:reconstruction_pgd_attack_main} to show the how different kernels handle uneven contributions of pixels that leads to spectral artifacts.
        \item \Cref{subsec:appendix:image_restoration:OOD_performance}: Out-Of-Distribution and Real World Generalization.
        
    \end{itemize}
 \item \Cref{sec:appendix:disparity}: \textbf{Disparity Estimation} : We provide additional results for \Cref{subsec:exp:disparity}: including performance against adversarial attacks.
    \vspace{1em}
    \begin{itemize}
        \item \Cref{subsubsec:appendix:exp:disparity} Additional discussion on the results and importance of a parallel 3$\times$3 kernel with large kernels for transposed convolution operation.
    \end{itemize}

    \item \Cref{sec:appendix:nomenclature}: \textbf{Nomenclature- What are ``Large Context Transposed Convolutions?''}: We discuss the nomenclature used in this work and describe what comprises a LCTC.

    \item \Cref{sec:appendix:teaser_extension}: \textbf{Additional visualizations of Upsampling Artifacts and their Frequency Spectra}: Here we extend \Cref{fig:teaser} with more examples showing failure of upsampling operations used in prior work and superiority of LCTC both in the spatial and frequency domain.

    \item \Cref{sec:appendix:limitations}: \textbf{Limitations}: Here we discuss the limitations of our work in detail.
\end{itemize}

\section{Experimental Setup}
\label{appenxdix:subsec:exp:setup}
All the experiments were done using NVIDIA V100 16GB GPUs or NVIDIA Tesla A100 40GB GPUs.
For image restoration, models were trained on 1 NVIDIA Tesla A100 40GB GPU.
For the semantic segmentation downstream task, UNet~\cite{unet} was trained using 1 GPU. 
For the disparity estimation task, STTR-light~\cite{sttr} was trained using 4 NVIDIA V100 GPUs in parallel.
%
\subsection{Image Restoration}
\label{subsubsec:exp:setup:restoration}

\textbf{Architectures. } We consider the recently proposed state-of-the-art transformer-based Image Restoration architectures Restormer~\cite{zamir2022restormer} and NAFNet~\cite{chen2022simple}.
Both architectures as proposed use Pixel Shuffle\cite{pixelshuffle} to upsample feature maps.
We use these as our baseline models.
We replace this pixel shuffle operation with a transposed convolution operation. 

\noindent\textbf{Dataset. } For the Image Restoration task, we focus on Image Deblurring. 
For this, we use the GoPro image deblurring dataset\cite{gopro}.
This dataset consists of 3214 real-world images with realistic blur and their corresponding ground truth (deblurred images) captured using a high-speed camera.
The dataset is split into 2103 training images and 1111 test images.

\noindent\textbf{Training Regime. } For Restormer we follow the same training regime of progressive training as that used by \cite{zamir2022restormer}.
Similarly, for NAFNet we use the same training regime as that used by \cite{chen2022simple}.

\noindent\textbf{Evaluation Metrics. } Following common practice\cite{agnihotri2023unreasonable, zamir2022restormer, chen2022simple}, We report the PSNR and SSIM scores of the reconstructed images w.r.t. to the ground truth images, averaged over all images.
PSNR stands for Peak Signal-to-Noise ratio, a higher PSNR indicates a better quality image or an image closer to the image to which it is being compared.
SSIM stands for Structural similarity\cite{ssim}.
A higher SSIM score corresponds to better higher similarity between the reconstruction and the ground-truth image.
\subsection{Semantic Segmentation}
\label{subsubsec:exp:setup:segment}
Here we describe the experimental setup for the segmentation task, the architectures considered, the dataset considered and the training regime.\newline
\textbf{Architectures. }We considered UNet~\cite{unet} with encoder layers from ConvNeXt~\cite{convnext}. 
For the decoder, the baseline comparison is done with 2$\times$2 kernels in the transposed convolution layers and the commonly used ResNet~\cite{resnet} BasicBlock style layers for the convolution layers in the decoder building blocks. 
In our experiments, we used larger sized kernels, e.g. 7$\times$7 and 11$\times$11 in the transposed convolution while keeping the rest of the architecture, including the convolution blocks in the decoder identical to \cref{subsec:exp:semantic}. 
When using kernels larger than 7$\times$7 for transposed convolution we follow the work of \cite{replk, slak} and additionally include a parallel 3$\times$3 kernel to keep the local context. 
Usage of this parallel kernel is denoted by ``+3$\times$3"
Further, we analyze the behavior of a different block of convolution layers in the decoder, as explained in \cref{subsec:methods:decoder} and replace the ResNet-style layers with ConvNeXt-style layers in \cref{subsec:ablation:backbone}.\newline
\textbf{Dataset. }We considered the PASCAL VOC 2012 dataset~\cite{pascal-voc-2012} for the semantic segmentation task. 
We follow the implementation of~\cite{semseg2019, pspnet, zhao2018psanet} and augment the training examples with semantic contours from \cite{SBD_BharathICCV2011} as instructed by \cite{voc_instructions}. \newline
\textbf{Training Regime. }We follow a similar training regime as \cite{semseg2019, pspnet}, and train for 50 epochs, with an AdamW optimizer~\cite{adamw} and the learning rate was scheduled using Cosine-Annealing~\cite{cosineannealing}. 
In the implementation of \cite{pspnet}, the authors slide over the images using a window of size 473$\times$473, however for computation reasons and for symmetry we use a window of size 256$\times$256. We use a starting learning rate of $10^{-4}$ and a weight decay of $5\times10^{-2}$.
\newline\textbf{Evaluation Metrics. }We report the mean Intersection over Union (mIoU) of the predicted and the ground truth segmentation mask, the mean accuracy over all pixels (mAcc) and the mean accuracy over all classes (allAcc). 

%
%
%
%

%
\subsection{Disparity Estimation}
\label{subsubsec:exp:setup:disparity}
Following, we describe the experimental setup for disparity estimation and occlusion detection tasks.\newline
\textbf{Architectures. }We consider the STTR-light~\cite{sttr} architecture for our work. 
To analyze the influence of implementing larger kernels in transposed convolution as described in \Cref{sec:method} we alter the kernel sizes in the transposed convolution layers used for pixel-wise upsampling in the ``feature extractor" module of the architecture. 
We consider the STTR-light architecture as proposed by \cite{sttr} with 3$\times$3 kernels in the transposed convolution layers as our baseline.\newline
\textbf{Dataset. }Similar to \cite{sttr} we train and test our models on FlyingThings3D dataset~\cite{flyingthings_MIFDB16}.\newline
\textbf{Training Regime. }We follow the training regime as implemented in \cite{sttr}.\newline
\textbf{Evaluation Metrics. }We report the end-point-error (epe) and the 3-pixel error (3px) for the disparity estimation w.r.t. the ground truth.
%
%
%
\subsection{Adversarial Attacks}
\label{subsubsec:exp:setup:adversarial}
We consider the commonly used \cite{segpgd, adv_segment, monocular_depth_adv, Mathew2020MonocularDE} FGSM attack \cite{fgsm} and a new segmentation-specific SegPGD attack \cite{segpgd} for testing the robustness of the models against adversarial attacks.
For the \textbf{semantic segmentation downstream task}, each crop of the input was perturbed with FGSM and SegPGD, while for the disparity estimation downstream task, each of the left and right inputs were perturbed using FGSM.\newline
For FGSM, we test our model against epsilons $ \epsilon \in \{\frac{1}{255}, \frac{8}{255}\}$.
Where, we follow common practice and use $\frac{1}{255}\approx$0.004 and $\frac{8}{255}\approx$0.03 .

For SegPGD we follow the testing parameters as originally proposed in \cite{segpgd}, with $\epsilon\approx\frac{8}{255}$, $\alpha$=0.01 and number of iterations $\in \{3, 5, 10, 20, 40, 100\}$. We use the same scheduling for loss balancing term $\mathbf{\lambda}$ as suggested by the authors.
We use SegPGD for the semantic segmentation task as it is a stronger attack specifically designed for segmentation. Thus providing more accurate insights into the models' performance and giving a better evaluation of the architectural design choices made. 

For the \textbf{Image Restoration task}, we follow the evaluation method of \cite{agnihotri2023unreasonable}, and evaluate against CosPGD\cite{agnihotri2023cospgd} and PGD\cite{pgd} adversarial attacks. 
For both attacks, we use $\epsilon\approx\frac{8}{255}$, $\alpha$=0.01 and test for number of attack iterations $\in \{5, 10, 20\}$.

For the \textbf{Depth Estimation task}, we use the PGD attack with $\epsilon\approx\frac{8}{255}$, $\alpha$=0.01 and test for number of attack iterations $\in \{5, 10, 20\}$.

\subsection{Adversarial Training}
\label{subsubsec:exp:setup:adv_training}
Following, we describe the adversarial training setup employed in this work for adversarially training models for semantic segmentation and image restoration.

\noindent\paragraph{Semantic Segmentation. }We follow the commonly used\cite{segpgd} procedure and split the batch into two 50\%-50\% mini-batches. 
One mini-batch is used to generate adversarial examples using FGSM attack with $\epsilon\approx\frac{8}{255}$ and PGD attack with 3 attack iterations and with $\epsilon\approx\frac{8}{255}$ and $\alpha$=0.01 during training.

\noindent\paragraph{Image Restoration. }We follow the training procedure used by \cite{agnihotri2023unreasonable}. 
We split each training batch into two equal 50\%-50\% mini-batches.
We use one of the mini-batches to generate adversarial samples using FGSM attack with $\epsilon\approx\frac{8}{255}$. 

\subsection{Frequency spectrum analysis}
\label{subsubsec:exp:setup:frequency}
To analyze the images in the frequency domain, we use the Fast Fourier Transform\cite{fft} (FFT) $X_c=FFT(x_c)$ for all channels $c$ of feature maps $x$ and aggregate a 2D representation 
over frequencies $w$. We compute the mean over $C$ channels of the FFT of the difference between the prediction and the ground truth.
\begin{equation}
\label{eqn:appendix:freq_calculation}
    \text{2D Frequency Spectra} = \frac{1}{C}\sum_{c\in C}FFT(x^{pred}_c - x^{gt}_c)
\end{equation}
Here, $x^{pred}$ are the predictions from the model, $x^{gt}$ is the ground truth, and in \cref{fig:teaser} and \cref{fig:teaser_extension_appendix} $C$=3 for the RGB channels.
For better visualization, we plot the $\mathrm{log}$ of the magnitude of the Discrete Fourier Transform.

Next, we describe, from the literature, the process of performing a Discrete Fourier Transform.

\textbf{Fast Fourier Transform (FFT)}\cite{fft}. The discrete Fourier transform has been used in this work to convert the images from the spatial domain to the frequency domain. 
\begin{quote}
    ``~DFT is a linear operator (i.e. a matrix) that maps the data points in $f$ to the frequency domain $\hat{f}$~''\cite{brunton_kutz_2019}
\end{quote}
Equation 2.26 in \cite{brunton_kutz_2019} shows the formula to perform DFT is:
\begin{equation}
\label{equation_from_brunton_kutz_2019}
    \hat{f}_k = \sum_{j=0}^{n-1} f_{j}\epsilon^{-i2\pi jk/n}
\end{equation}
where $\hat{f}_k$ from each sample n contains the amplitude and phase (of the sine and cosine components) information at frequency $k$.
These are integer multiples of $\epsilon^{-2\pi j/n}$, the fundamental frequency, short-handed as $\omega_{n}$\cite{brunton_kutz_2019}.
Equation 2.29 in \cite{brunton_kutz_2019} shows the Discrete Fourier transform matrix (in terms of $\omega_n$) that when multiplied by the samples in f, converts the information in those samples to frequency domain (a basis transformation).
FFT is an algorithm by \cite{fft} to perform Discrete Fourier transform in an efficient manner.
In \cref{eqn:appendix:freq_calculation}, we use these frequencies $w$ (referred to as $k$ in \cref{equation_from_brunton_kutz_2019}) from sample $x_c$ obtained using an FFT(~) function that uses the FFT algorithm.
%
%
%
%
%
\section{Additional Experiments and Ablation}
\label{subsec:appendix:experiments}
Here we provide detailed results from \cref{sec:exp} and \cref{sec:ablation} and additional results as mentioned in the main paper.
%
%
\subsection{Semantic Segmentation}
\label{subsubsec:appendix:exp:segment}

\begin{figure*}[htbp]
	\begin{center}
	\includegraphics[width=0.98\linewidth]{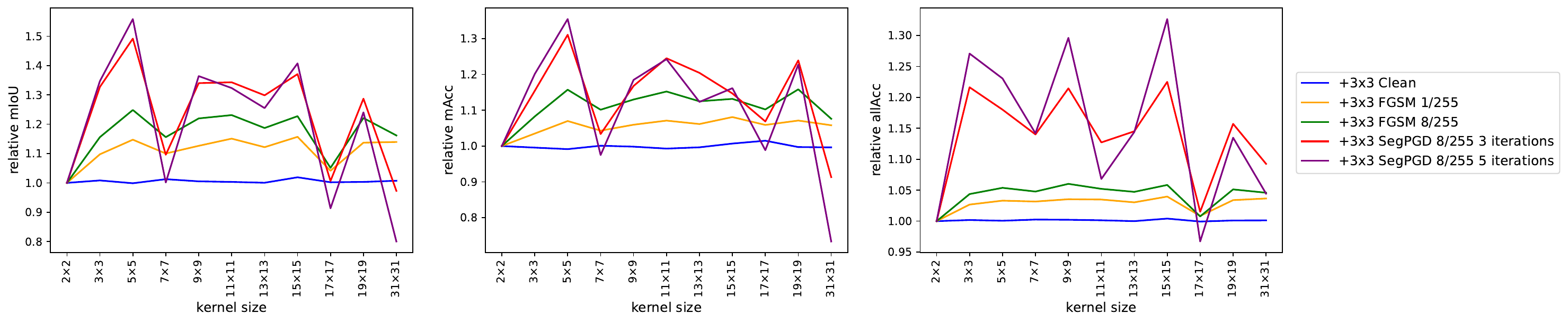}
	\caption[]{Comparison of performance of different sizes of transpose convolutions from standard sizes like $2\times2$ as well as very large $31\times31$ kernels with \textcolor{RoyalBlue}{ConvNeXt style 11$\times$11 + 3$\times$3 style in the decoder building blocks}. All have a parallel 3×3 kernel, as shown in \Cref{fig:unet_plus_block} (bottom left). 
	}\label{fig:transpose_conv_plots_largestconvnext}
	\end{center}
	\vspace{-1em}
\end{figure*}

\Cref{tbl:appendix:ablation:backbone} and \Cref{tbl:appendix:ablation:segpgd_backbone} provide all the results of empirical performance (across the considered upsampling blocks) on clean inputs images and input images perturbed by varying intensities of FGSM and SegPGD attacks respectively.


\begin{table*}[ht]
\caption{ Complete comparison of performances against FGSM attack, of UNet with ConvNeXt encoder and decoder with architectures along with different sized kernels in transposed convolution and different convolution blocks in the decoder for upscaling the feature maps.}
\label{tbl:appendix:ablation:backbone}
\centering

\scalebox{0.4}{
\begin{tabular}{p{4cm}cccc|ccc|ccc}
\toprule
\multirow{3}{4cm}{\textbf{Transposed Convolution Kernels}} &  \multirow{3}{*}{\textbf{Backbone Style}} & \multicolumn{3}{c}{\textbf{Test Accuracy}}   & \multicolumn{6}{c}{\textbf{FGSM attack epsilon}} \\
               &         & \multirow{2}{*}{mIoU} & \multirow{2}{*}{mAcc} & \multirow{2}{*}{allAcc} &   \multicolumn{3}{c}{$\epsilon$=$\frac{1}{255}$} & \multicolumn{3}{c}{$\epsilon$=$\frac{8}{255}$}                                  \\
                &        &     &  &  & mIoU & mAcc & allAcc & mIoU & mAcc & allAcc\\
\midrule
\multirow{3}{*}{2$\times$2} &   ResNet Style 3$\times$3  &  78.34 & 86.89 & 95.15 &  53.54 & 70.96 & 86.08 & 47.02 & 65.41 & 82.78  \\

&  ConvNeXt style 7$\times$7     & 77.17 & 86.86 & 94.81 & 77.42 & 86.24 & 94.94 & 42.04 & 64.86 & 79.08 \\

&  ConvNeXt style 7$\times$7 + 3$\times$3    & 77.24 & 86.03 & 94.84 & 51.09 & 70.53 & 85.29 & 43.52 & 63.74 & 81.18 \\

&  ConvNeXt style 11$\times$11    & 77.68 & 86.42 & 94.97  & 50.73 & 69.78 & 84.88 & 42.33 & 61.80 & 80.36 \\

&  ConvNeXt style 11$\times$11 + 3$\times$3    & 77.17 & 86.86 & 94.81  & 47.34 & 67.72 & 83.34 & 37.91 & 57.79 & 78.21 \\

\midrule
\multirow{3}{*}{3$\times$3} &   ResNet Style 3$\times$3  &  78.45 & 86.66 & 95.20 & 53.76 & 70.62 & 86.32 & 47.33 & 64.58 & 83.16 \\

&  ConvNeXt style 7$\times$7     & 77.70 & 86.89 & 94.99 & 52.30 & 71.56 & 85.73 & 44.80 & 65.38 & 81.99 \\

&  ConvNeXt style 7$\times$7 + 3$\times$3  &  77.33 & 87.53 & 94.79 & 50.90 & 72.77 & 83.78 & 44.40 & 67.08 & 79.11 \\

&  ConvNeXt style 11$\times$11    & 77.86 & 86.75 & 94.99 & 51.30 & 70.39 & 85.33 & 42.78 & 62.76 & 81.08  \\

&  ConvNeXt style 11$\times$11 + 3$\times$3    & 77.81 & 86.48 & 94.98 & 51.95 & 70.08 & 85.57 & 43.82 & 62.56 & 81.63 \\

\midrule

\multirow{3}{*}{5$\times$5 (Ours)} &   ResNet Style 3$\times$3  &  79.19 & 87.62 & 95.36 & 55.57 & 73.51 & 86.65 & 48.96 & 67.97 & 83.41  \\

&  ConvNeXt style 7$\times$7     & 76.94 & 86.92 & 94.75 & 51.32 & 72.37 & 84.96 & 44.19 & 66.56 & 81.13  \\

&  ConvNeXt style 7$\times$7 + 3$\times$3 & 78.52 & 87.39 & 95.13 & 54.4 & 72.48 & 86.29 & 46.33 & 65.65 & 82.0  \\

&  ConvNeXt style 11$\times$11    & 77.83 & 86.99 & 94.91 & 53.76 & 72.8 & 85.96 & 45.32 & 65.82 & 81.82  \\

&  ConvNeXt style 11$\times$11 + 3$\times$3    & 77.92 & 86.92 & 95.02 & 48.67 & 68.11 & 83.96 & 38.88 & 58.13 & 78.96 \\

\midrule

\multirow{3}{*}{5$\times$5 + 3$\times$3 (Ours)} &   ResNet Style 3$\times$3  & 78.83 & 87.56 & 95.28 & 56.11 & 73.97 & 86.91 & 49.84 & 69.26 & 83.44  \\

&  ConvNeXt style 7$\times$7     & 78.11 & 86.90 & 95.01 & 53.17 & 71.55 & 86.0& 45.98 & 66.05 & 82.18  \\

&  ConvNeXt style 7$\times$7 + 3$\times$3  & 78.73 & 87.81 & 95.24 & 53.86 & 73.12 & 85.86 & 45.93 & 66.83 & 81.51 \\

&  ConvNeXt style 11$\times$11    & 77.83 & 86.57 & 95.07 & 52.12 & 70.29 & 85.79 & 44.05 & 63.11 & 81.63 \\

&  ConvNeXt style 11$\times$11 + 3$\times$3    & 77.07 & 86.11 & 94.87 & 54.31 & 72.45 & 86.1 & 47.33 & 66.88 & 82.42 \\

\midrule

\multirow{3}{*}{LCTC: 7$\times$7 (Ours)} &   ResNet Style 3$\times$3  & 78.92 & 88.06 & 95.23 & 56.02 & 74.13 & 86.45 & 49.24 & 68.89 & 82.87 \\

&  ConvNeXt style 7$\times$7     & 77.57 & 87.04 & 94.92  & 52.93 & 72.18 & 85.51 & 44.89 & 65.71 & 80.74 \\

&  ConvNeXt style 7$\times$7 + 3$\times$3     & 77.88 & 87.0& 95.05 & 51.63 & 70.74 & 85.37 & 43.15 & 62.74 & 80.83 \\

&  ConvNeXt style 11$\times$11    & 77.9 & 87.35 & 94.94 & 53.47 & 72.61 & 85.79 & 45.49 & 67.04 & 81.36 \\

&  ConvNeXt style 11$\times$11 + 3$\times$3 & 77.99 & 87.86 & 94.96 & 51.61 & 73.01 & 84.85 & 43.93 & 66.22 & 80.73 \\
\midrule

\multirow{3}{*}{LCTC: 7$\times$7 + 3$\times$3 (Ours)} &   ResNet Style 3$\times$3  & 78.5 & 87.57 & 95.13 & 53.85 &  72.75 & 85.87 & 47.1	& 67.57 & 82.04 \\

&  ConvNeXt style 7$\times$7     & 78.09 & 87.14 & 95.04 & 52.42 & 71.88 & 85.59 & 43.43 & 65.39 & 80.88 \\

&  ConvNeXt style 7$\times$7 + 3$\times$3 & 78.37 & 88.11 & 95.07 & 52.15 & 72.31 & 84.95 & 42.77 & 63.69 & 79.78 \\

&  ConvNeXt style 11$\times$11    & 77.71 & 87.22 & 94.97 & 52.47 & 73.22 & 85.55 & 44.07 & 65.84  & 81.31 \\

&  ConvNeXt style 11$\times$11 + 3$\times$3 & 78.14 & 86.94 & 95.05 & 52.08 & 70.63 & 85.98 & 43.82 & 63.65 & 81.95 \\

\midrule

\multirow{3}{*}{LCTC: 9$\times$9 (Ours)} &   ResNet Style 3$\times$3  &  78.36 & 86.88 & 95.18 & 55.62 & 72.62 & 86.9 & 49.5 & 67.03 & 83.9  \\

&  ConvNeXt style 7$\times$7     & 77.17 & 86.74 & 94.84 & 52.76 & 72.31 & 85.56 & 44.23 & 64.98 & 81.39 \\

&  ConvNeXt style 7$\times$7 + 3$\times$3 & 77.93 & 86.97 & 95.04 & 51.01 & 70.59 & 84.87 & 41.93 & 61.63 & 80.18  \\

&  ConvNeXt style 11$\times$11    & 77.80 & 86.80 & 94.99 & 52.42 & 72.22 & 85.39 & 44.14 & 65.56 & 81.16 \\

&  ConvNeXt style 11$\times$11 + 3$\times$3    & 78.25 & 86.71 & 95.07 & 54.59 & 72.04 & 86.48 & 46.88 & 65.56 & 82.73 \\

\midrule

\multirow{3}{*}{LCTC: 9$\times$9  + 3$\times$3 (Ours)} &   ResNet Style 3$\times$3  & 78.77 & 87.77 & 95.24 & 55.94 & 73.79 & 86.67 & 48.82 & 69.2 & 82.76  \\

&  ConvNeXt style 7$\times$7     & 77.79 & 86.65 & 94.92 & 52.6 & 70.51 & 85.75 & 43.3 & 62.16 & 80.89 \\

&  ConvNeXt style 7$\times$7 + 3$\times$3 & 77.96 & 87.24 & 94.98 & 51.21 & 70.01 & 85.24 & 41.75 & 61.16 & 80.64 \\

&  ConvNeXt style 11$\times$11    & 77.92 & 86.82 & 95.03 & 52.71 & 71.17 & 86.02 & 44.33 & 63.26 & 82.2  \\

&  ConvNeXt style 11$\times$11 + 3$\times$3  & 77.57 & 86.71 & 95.02 & 53.32 & 71.75 & 86.29 & 46.24 & 65.3 & 82.92 \\

\midrule

\multirow{3}{*}{LCTC: 11$\times$11 (Ours)} &   ResNet Style 3$\times$3  & 79.11 & 87.06 & 95.36 & 56.18 & 72.11 & 87.27 & 49.51 & 66.15 & 84.12 \\

&  ConvNeXt style 7$\times$7 & 77.87 & 86.98 & 95.06 & 54.32 & 72.59 & 86.42 & 47.14 & 67.05 & 82.71 \\

&  ConvNeXt style 7$\times$7 + 3$\times$3 & 78.34 & 87.06 & 95.07 & 51.93 & 71.19 & 85.54  & 41.77
 & 62.31 & 80.8 \\

&  ConvNeXt style 11$\times$11 & 77.42 & 86.68 & 94.94 & 53.11 & 71.43 & 86.03 & 44.55 & 63.45 & 81.75 \\

&  ConvNeXt style 11$\times$11 + 3$\times$3  & 77.75 & 86.83 & 95.01 & 52.88 & 71.47 & 85.93 & 43.55 & 62.75 & 81.4 \\

\midrule

\multirow{3}{*}{LCTC: 11$\times$11 + 3$\times$3 (Ours)} &   ResNet Style 3$\times$3  & 79.33 & 87.81 & 95.41 & 58.04 & 74.93 & 87.8 & 51.25 & 69.31 & 84.64 \\

&  ConvNeXt style 7$\times$7  & 78.32 & 86.98 & 95.09 & 53.31 & 72.45 & 86.16 & 44.89 & 65.18 & 82.03 \\

&  ConvNeXt style 7$\times$7 + 3$\times$3 & 78.64 & 86.78 & 95.17 & 54.32 & 71.27 & 86.63 & 45.48 & 63.62 & 82.32 \\

&  ConvNeXt style 11$\times$11    & 77.15 & 85.93 & 94.87 & 51.19 & 69.72 & 85.45 & 42.02 & 61.09 & 81.1 \\

&  ConvNeXt style 11$\times$11 + 3$\times$3 & 77.42 & 86.24 & 94.94 & 54.48 & 72.53 & 86.25 & 46.67 & 66.59 & 82.29 \\

\midrule

\multirow{3}{*}{LCTC: 13$\times$13 (Ours)} &   ResNet Style 3$\times$3  & 79.41 & 88.18 & 95.36 & 56.89 & 74.71 & 87.36 & 51.06 & 70.39 & 84.48   \\

&  ConvNeXt style 7$\times$7     & 77.99 & 87.11 & 95.06 & 54.96 & 73.32 & 86.69 & 47.39 & 67.2 & 82.73 \\

&  ConvNeXt style 7$\times$7 + 3$\times$3  & 78.44 & 87.22 & 95.13 & 54.21 & 72.18 & 86.34 & 47.27 & 65.72 & 82.95 \\

&  ConvNeXt style 11$\times$11    &  77.57 & 85.99 & 95.00 & 53.51 & 70.31 & 86.67 & 45.63 & 63.59 & 83.11 \\

&  ConvNeXt style 11$\times$11 + 3$\times$3  & 77.40 & 86.53 & 94.89  & 53.16 & 71.62 & 86.12 & 45.09 & 64.23 & 82.39 \\
\midrule

\multirow{3}{*}{LCTC: 13$\times$13  + 3$\times$3 (Ours)} &   ResNet Style 3$\times$3  & 79.17 & 87.96 & 95.38 & 57.17 & 75.08 & 87.44 & 50.8 & 70.67 & 84.06  \\

&  ConvNeXt style 7$\times$7     & 78.05 & 86.73 & 95.02 & 53.41 & 71.62 & 86.12 & 45.07 & 65.04 & 81.76  \\

&  ConvNeXt style 7$\times$7 + 3$\times$3 & 77.76 & 86.14 & 95.06 & 54.09 & 72.11 & 86.29 & 45.69 & 65.15 & 82.2  \\

&  ConvNeXt style 11$\times$11    & 77.81 & 87.43 & 95.01 & 51.71 & 71.77 & 85.25 & 41.97 & 62.61 & 80.66 \\

&  ConvNeXt style 11$\times$11 + 3$\times$3  & 77.20 & 86.55 & 94.81 & 53.1 & 71.88 & 85.87 & 45.0& 65.01 & 81.91 \\

\midrule

\multirow{3}{*}{LCTC: 15$\times$15 (Ours)} &   ResNet Style 3$\times$3  &  79.17 & 87.68 & 95.28 & 58.08 & 73.56 & 87.58 & 51.11 & 67.94 & 84.36  \\

&  ConvNeXt style 7$\times$7     & 78.34 & 87.14 & 95.03 & 53.86 & 72.77 & 86.11 & 45.12 & 65.22 & 81.65  \\

&  ConvNeXt style 7$\times$7 + 3$\times$3  &  77.39 & 86.40 & 94.95 & 51.2 & 69.42 & 85.27 & 42.65 & 60.88 & 81.24 \\

&  ConvNeXt style 11$\times$11    & 77.14 & 86.36 & 94.82 & 50.14 & 69.32 & 84.49 & 40.97 & 60.11 & 79.81 \\

&  ConvNeXt style 11$\times$11 + 3$\times$3    & 77.67 & 86.78 & 94.90 & 54.44 & 72.74 & 86.54 & 46.37 & 66.24 & 82.29  \\

\midrule

\multirow{3}{*}{LCTC: 15$\times$15  + 3$\times$3 (Ours)} &   ResNet Style 3$\times$3  & 78.72 & 87.50 & 95.25 & 56.28 & 73.97 & 87.15 & 49.5 & 68.69 & 83.53  \\

&  ConvNeXt style 7$\times$7 &  77.56 & 87.01 & 94.93  & 53.28 & 72.15 & 85.78 & 45.51 & 64.84 & 81.57 \\

&  ConvNeXt style 7$\times$7 + 3$\times$3 & 77.09 & 86.27 & 94.76 & 52.25 & 70.01 & 85.41 & 44.01 & 62.49 & 81.16  \\

&  ConvNeXt style 11$\times$11    & 77.40 & 86.39 & 94.92 & 53.59 & 71.49 & 86.21 & 45.48 & 64.37 & 82.28 \\

&  ConvNeXt style 11$\times$11 + 3$\times$3  &  78.64 & 87.46 & 95.20  & 54.77 & 73.2 & 86.65 & 46.53 & 65.4 & 82.78 \\

\midrule

\multirow{3}{*}{LCTC: 17$\times$17 (Ours)} &   ResNet Style 3$\times$3  & 79.22 & 87.77 & 95.37 & 56.5 & 73.3 & 87.27 & 50.1 & 68.23 & 84.11  \\

&  ConvNeXt style 7$\times$7  & 77.36 & 87.64 & 94.89  & 54.06 & 73.88 & 85.84 & 47.25 & 68.3 & 82.19 \\

&  ConvNeXt style 7$\times$7 + 3$\times$3 & 78.03 & 87.56 & 95.01 & 52.75 & 72.0& 85.65 & 44.32 & 64.16 & 81.54  \\

&  ConvNeXt style 11$\times$11    & 77.82 & 87.40 & 94.92 & 51.43 & 70.57 & 85.22 & 42.53 & 62.68 & 80.79 \\

&  ConvNeXt style 11$\times$11 + 3$\times$3 & 77.74 & 86.69 & 94.99 & 51.31 & 69.71 & 85.53 & 41.58 & 60.43 & 80.83 \\

\midrule

\multirow{3}{*}{LCTC: 17$\times$17  + 3$\times$3 (Ours)} &   ResNet Style 3$\times$3  &  78.41 & 86.84 & 95.26 & 56.03 & 73.28 & 87.16 & 49.65 & 67.95 & 83.74  \\

&  ConvNeXt style 7$\times$7   & 78.14 & 86.99 & 94.98 & 53.44 & 72.34 & 86.01 & 45.02 & 65.35 & 81.85 \\

&  ConvNeXt style 7$\times$7 + 3$\times$3 & 78.62 & 87.64 & 95.14 & 55.54 & 73.87 & 86.85 & 47.86 & 67.22 & 83.18 \\

&  ConvNeXt style 11$\times$11    & 77.59 & 87.73 & 94.84 & 52.84 & 74.14 & 84.63 & 44.1 & 67.34 & 79.57  \\

&  ConvNeXt style 11$\times$11 + 3$\times$3 &  77.33 & 88.15 & 94.75  & 49.29 & 71.71 & 84.04 & 39.85 & 63.7 & 78.81 \\

\midrule

\multirow{3}{*}{LCTC: 19$\times$19 (Ours)} &   ResNet Style 3$\times$3  &  78.54 & 87.64 & 95.12 & 56.63 & 74.09 & 87.25 & 50.02 & 68.73 & 83.99 \\

&  ConvNeXt style 7$\times$7 &  78.74 & 87.66 & 95.15  & 56.28 & 73.79 & 87.11 & 49.44 & 68.74 & 83.84 \\

&  ConvNeXt style 7$\times$7 + 3$\times$3 & 77.05 & 86.33 & 94.89 & 54.47 & 72.38 & 86.78 & 45.63 & 64.94 & 82.81  \\

&  ConvNeXt style 11$\times$11  &  77.66 & 86.61 & 95.00 & 51.58 & 71.51 & 84.83 & 42.48 & 63.44 & 79.58  \\

&  ConvNeXt style 11$\times$11 + 3$\times$3  & 77.61 & 86.59 & 94.93  & 50.34 & 69.39 & 84.54 & 41.82 & 61.29 & 79.75 \\
\midrule

\multirow{3}{*}{LCTC: 19$\times$19 + 3$\times$3 (Ours)} &   ResNet Style 3$\times$3  & 78.78 & 87.34 & 95.28 & 56.53 & 74.59 & 86.97 & 50.6 & 69.95 & 83.98  \\

&  ConvNeXt style 7$\times$7  &  77.44 & 86.70 & 94.91  & 54.05 & 72.52 & 86.09 & 45.52 & 65.29 & 81.52 \\

&  ConvNeXt style 7$\times$7 + 3$\times$3 & 78.14 & 87.14 & 95.02 & 55.82 & 74.54 & 86.96 & 48.97 & 69.98 & 83.3  \\

&  ConvNeXt style 11$\times$11    & 78.03 & 86.64 & 95.08 & 53.5 & 71.21 & 86.26 & 45.79 & 64.16 & 82.42  \\

&  ConvNeXt style 11$\times$11 + 3$\times$3  & 77.42 & 86.61 & 94.91 & 53.83 & 72.54 & 86.17 & 46.29 & 66.94 & 82.22 \\

\midrule

\multirow{3}{*}{LCTC: 31$\times$31 (Ours)} &   ResNet Style 3$\times$3  & 78.69 & 86.98 & 95.30 & 56.61 & 73.22 & 87.08 & 49.49 & 66.69 & 83.68  \\

&  ConvNeXt style 7$\times$7  &  77.54 & 87.30 & 94.84  & 52.36 & 72.27 & 85.14 & 43.56 & 65.14 & 8 .\\

&  ConvNeXt style 7$\times$7 + 3$\times$3 & 76.96 & 86.38 & 94.77 & 53.59 & 72.14 & 86.05 & 45.22 & 65.22 & 81.84  \\

&  ConvNeXt style 11$\times$11  &  76.84 & 86.72 & 94.71 & 50.74 & 70.53 & 84.61 & 41.62 & 61.96 & 79.96 \\

&  ConvNeXt style 11$\times$11 + 3$\times$3  &  76.77 & 85.60 & 94.71 & 51.42 & 69.17 & 85.2 & 42.12 & 60.32 & 80.77 \\

\midrule

\multirow{3}{*}{LCTC: 31$\times$31 + 3$\times$3 (Ours)} &   ResNet Style 3$\times$3  & 78.47 & 87.26 & 95.16 & 56.27 & 73.39 & 87.22 & 49.66 & 68.81 & 83.92  \\

&  ConvNeXt style 7$\times$7  &  77.43 & 86.56 & 94.93  & 53.45 & 72.74 & 86.17 & 45.84 & 66.41 & 82.16 \\

&  ConvNeXt style 7$\times$7 + 3$\times$3 & 78.43 & 87.07 & 95.17 & 56.72 & 73.65 & 87.6 & 49.56 & 68.15 & 84.22  \\

&  ConvNeXt style 11$\times$11    & 78.00 & 87.04 & 94.94 & 50.66 & 70.23 & 84.83 & 40.71 & 61.31 & 79.94 \\

&  ConvNeXt style 11$\times$11 + 3$\times$3 &  77.73 & 86.54 & 94.93  & 53.94 & 71.65 & 86.39 & 44.04 & 62.19 & 81.8 \\

\bottomrule

\end{tabular}
}
\end{table*}


\begin{table*}[ht]
\caption{Comparison of performances against SegPGD attack, of UNet with ConvNeXt encoder and decoder with architectures along with different sized kernels in transposed convolution and different convolution blocks in the decoder for upscaling the feature maps.}
\label{tbl:appendix:ablation:segpgd_backbone}
\centering
\scalebox{0.4}{
\begin{tabular}{@{}p{4cm}c @{\hspace{0.1cm}} c @{\hspace{0.1cm}}c @{\hspace{0.1cm}}c @{\hspace{0.5cm}}c @{\hspace{0.1cm}}c @{\hspace{0.1cm}}c @{\hspace{0.5cm}}c @{\hspace{0.1cm}}c @{\hspace{0.1cm}}c @{\hspace{0.5cm}}c @{\hspace{0.1cm}}c @{\hspace{0.1cm}}c @{\hspace{0.5cm}}c @{\hspace{0.1cm}}c @{\hspace{0.1cm}}c @{\hspace{0.5cm}}c @{\hspace{0.1cm}}c @{\hspace{0.1cm}}c@{}}
\toprule
\multirow{3}{4cm}{\textbf{Transposed Convolution Kernels}} &  \multirow{3}{*}{\textbf{Backbone Style}} &  \multicolumn{18}{c}{\textbf{SegPGD attack iterations}} \\
     &          &    \multicolumn{3}{c}{3}     & \multicolumn{3}{c}{5} & \multicolumn{3}{c}{10} & \multicolumn{3}{c}{20} &   \multicolumn{3}{c}{40} & \multicolumn{3}{c}{100}                                  \\
    &  & mIoU & mAcc & allAcc & mIoU & mAcc & allAcc & mIoU & mAcc & allAcc & mIoU & mAcc & allAcc & mIoU & mAcc & allAcc & mIoU & mAcc & allAcc \\
\midrule
\multirow{3}{*}{2$\times$2} &   ResNet Style 3$\times$3  & 23.06 & 46.51 & 60.04 & 14.43 & 35.50 & 45.30 & 08.12 & 24.67 & 29.88 & 05.54 & 18.79 & 23.72 & 04.39 & 14.98 & 23.70 & 03.50 & 11.61 & 27.93\\

&  ConvNeXt style 7$\times$7     & 17.94 & 	0.4481	 & 47.96 & 10.64	 & 33.63	 & 30.64  & 05.47	 & 21.74	 & 15.8 & 03.2	 & 14.73	 & 09.81 & 02.04 & 	0.1047 & 	0.0641 & 01.35	 & 07.57	 & 04.3 \\

&  ConvNeXt style 7$\times$7 + 3$\times$3    & 17.59	 & 42.55 & 	0.5168 & 09.88	 & 30.41 & 	0.3233 & 04.75	 & 16.83 & 	0.1431 & 02.65	 & 09.46 & 	0.0668 & 01.68	 & 05.64 & 	0.034 & 01.0& 	0.0316	 & 01.94  \\

&  ConvNeXt style 11$\times$11    & 16.39 & 	0.4013 & 	0.485  &  09.37 & 28.66 & 29.63 & 03.97 & 14.16 & 11.41 & 01.56 & 06.11 & 03.56 & 00.59 & 02.61 & 01.31  & 00.23 & 00.99 & 00.51 \\

&  ConvNeXt style 11$\times$11 + 3$\times$3    & 13.97 & 35.82 & 45.68 & 07.61 & 25.07 & 28.33 & 03.4 & 14.38 & 12.04 & 02.21 & 10.75 & 05.29  &  01.57 & 08.02 & 03.01 & 01.07 & 05.75 & 01.85  \\

\midrule

\multirow{3}{*}{3$\times$3} & ResNet Style 3$\times$3  & 23.37 & 46.33 & 60.78 & 15.26 & 38.0& 46.51 & 09.26 & 29.64 & 31.9 & 06.78 & 24.18 & 26.95 & 05.71 & 20.39 & 28.69 & 05.02 & 16.11 & 33.12 \\

&  ConvNeXt style 7$\times$7   & 18.48 & 43.81 & 54.97 & 09.51 & 29.92 & 34.86 & 03.63 & 15.1 & 13.03 & 01.64 & 08.23 & 04.51 & 01.0& 05.12 & 02.13 & 00.59 & 02.84 & 00.89   \\

&  ConvNeXt style 7$\times$7 + 3$\times$3  & 19.08 & 46.97 & 47.74 & 11.15 & 34.6 & 29.9 & 05.96 & 22.62 & 15.67 & 03.61 & 15.04 & 09.33 & 02.17 & 09.18 & 05.86 & 01.29 & 06.02 & 03.55 \\

&  ConvNeXt style 11$\times$11  & 16.2 & 39.11 & 50.93 & 09.52 & 29.32 & 32.61 & 04.93 & 20.31 & 14.82 & 02.86 & 13.94 & 06.46 & 02.05 & 10.94 & 03.58 & 01.4 & 08.23 & 02.21 \\

&  ConvNeXt style 11$\times$11 + 3$\times$3   & 18.54 & 41.34 & 55.56 & 10.25 & 30.11 & 36.0& 04.8 & 19.25 & 13.94 & 02.41 & 11.87 & 04.56 & 01.59 & 07.78 & 02.11 & 01.11 & 04.21 & 01.09  \\

\midrule

\multirow{3}{*}{5$\times$5 (Ours)} &   ResNet Style 3$\times$3  & 24.23 & 51.8 & 57.82 & 16.16 & 42.98 & 43.29 & 10.11 & 32.79 & 30.3 & 07.32 & 25.16 & 27.42 & 06.02 & 19.04 & 31.25 & 05.16 & 14.03 & 37.36 \\

&  ConvNeXt style 7$\times$7   & 17.59 & 43.57 & 51.41 & 09.9 & 30.84 & 33.14 & 04.74 & 18.3 & 14.55 & 02.23 & 09.47 & 05.21 & 01.47 & 06.03 & 02.32 & 00.97 & 03.64 & 01.28   \\

&  ConvNeXt style 7$\times$7 + 3$\times$3  & 18.7 & 43.18 & 52.74 & 10.56 & 31.41 & 33.32 & 04.87 & 18.5 & 14.78 & 02.49 & 10.84 & 05.59 & 01.39 & 05.61 & 02.69 & 00.91 & 03.38 & 01.46  \\

&  ConvNeXt style 11$\times$11   & 18.96 & 44.79 & 53.09 & 09.85 & 29.77 & 32.88 & 03.89 & 14.94 & 12.6 & 01.94 & 08.29 & 04.58 & 01.03 & 04.72 & 01.86 & 00.48 & 02.63 & 00.75  \\

&  ConvNeXt style 11$\times$11 + 3$\times$3   & 13.38 & 33.61 & 45.0& 06.84 & 20.94 & 25.99 & 02.51 & 08.85 & 08.5 & 01.18 & 04.39 & 02.62 & 00.71 & 02.48 & 01.07 & 00.48 & 01.52 & 00.53  \\

\midrule

\multirow{3}{*}{5$\times$5 + 3$\times$3 (Ours)} &   ResNet Style 3$\times$3  & 25.03 & 53.96 & 58.89 & 16.61 & 45.8 & 42.18 & 10.79 & 37.16 & 27.34 & 08.0& 29.62 & 21.71 & 06.16 & 21.69 & 22.25 & 04.83 & 13.87 & 28.97 \\

&  ConvNeXt style 7$\times$7   & 17.65 & 44.79 & 48.41 & 09.79 & 31.78 & 28.51 & 04.62 & 18.37 & 11.12 & 02.58 & 10.89 & 04.61 & 01.52 & 06.59 & 02.3 & 01.0& 04.04 & 01.33   \\

&  ConvNeXt style 7$\times$7 + 3$\times$3  & 18.31 & 42.75 & 49.26 & 09.89 & 28.58 & 30.02 & 03.78 & 12.49 & 11.08 & 01.34 & 04.76 & 03.54 & 00.48 & 02.13 & 01.45 & 00.19 & 00.88 & 00.76 \\

&  ConvNeXt style 11$\times$11   & 17.87 & 40.62 & 52.77 & 09.74 & 27.94 & 34.21 & 04.65 & 14.98 & 14.34 & 02.0& 05.95 & 04.77 & 01.07 & 02.81 & 01.71 & 00.32 & 00.98 & 00.63  \\

&  ConvNeXt style 11$\times$11 + 3$\times$3   & 20.84 & 46.95 & 53.91 & 11.86 & 33.96 & 34.86 & 05.65 & 19.8 & 16.66 & 02.83 & 10.73 & 08.2 & 01.59 & 06.21 & 04.68 & 01.11 & 04.2 & 02.62  \\

\midrule

\multirow{3}{*}{LCTC: 7$\times$7 (Ours)} &   ResNet Style 3$\times$3 & 26.53 & 53.05 & 61.16 & 17.75 & 43.31 & 46.99 & 10.26 & 30.92 & 32.62 & 07.17 & 23.05 & 27.52  & 05.69 & 17.24 & 29.48  & 04.37 & 11.29 & 35.16 \\

&  ConvNeXt style 7$\times$7   & 17.64 & 43.32 & 47.8  & 09.95 & 30.43 & 28.02  & 04.21 & 15.08 & 10.07  & 01.86 & 07.18 & 03.55 & 00.99 & 03.52 & 01.42  & 00.68 & 01.89 & 00.74 \\

&  ConvNeXt style 7$\times$7 + 3$\times$3     & 16.64 & 40.11 & 50.56  & 09.75 & 29.72 & 32.23 & 04.95 & 19.4 & 14.47 & 02.87 & 13.23 & 06.4 & 02.06 & 09.55 & 03.45 & 01.59 & 07.24 & 02.04 \\

&  ConvNeXt style 11$\times$11    & 17.37 & 45.07 & 47.32 &  08.86 & 30.03 & 26.48 &  03.47 & 14.22 & 07.94  & 01.53 & 06.55 & 02.45 & 00.93 & 03.9 & 01.2 & 00.61 & 02.3 & 00.64\\

&  ConvNeXt style 11$\times$11 + 3$\times$3    & 17.07 & 42.3 & 48.78  & 09.31 & 28.04 & 28.88 & 03.82 & 13.79 & 09.54 & 01.8 & 07.11 & 03.04 &  01.03 & 04.3 & 01.45 & 00.53 & 02.61 & 00.77 \\
\midrule


\multirow{3}{*}{LCTC: 7$\times$7 + 3$\times$3 (Ours)} & ResNet Style 3$\times$3  & 24.03 & 52.08 & 57.43 & 16.21 & 43.38 & 43.01 & 09.99 & 32.77 & 30.22  & 07.38 & 26.16 & 26.11 & 06.31 & 22.42 & 28.32 & 05.35 & 17.41 & 33.09  \\

&  ConvNeXt style 7$\times$7  & 16.19 & 43.4 & 48.59 & 09.02 & 32.38 & 29.17  &  04.23 & 19.63 & 10.47 & 02.46 & 12.18 & 03.99 &  01.53 & 06.85 & 01.97 & 00.91 & 03.94 & 01.1 \\

&  ConvNeXt style 7$\times$7 + 3$\times$3  & 16.04 & 39.67 & 48.16 & 08.94 & 27.45 & 30.33 & 03.81 & 14.69 & 12.79 & 01.91 & 09.17 & 04.63  & 01.2 & 06.16 & 01.95 & 00.84 & 03.96 & 00.95 \\

&  ConvNeXt style 11$\times$11  & 18.08 & 46.24 & 50.64 & 10.18 & 33.17 & 31.35 & 04.49 & 18.33 & 12.04 & 02.01 & 07.98 & 04.55 & 01.04 & 03.91 & 02.17 & 00.45 & 01.7 & 01.2 \\

&  ConvNeXt style 11$\times$11 + 3$\times$3    & 15.31 & 37.02 & 52.08 & 07.62 & 24.44 & 32.35 & 03.3 & 15.04 & 12.35 & 01.92 & 10.24 & 05.12 &  01.32 & 07.37 & 02.63 & 00.91 & 05.07 & 01.39 \\

\midrule

\multirow{3}{*}{LCTC: 9$\times$9 (Ours)} &   ResNet Style 3$\times$3  & 25.26 & 50.75 & 60.85 & 16.88 & 41.02 & 47.16 & 09.44 & 28.03 & 33.87 & 06.23 & 20.76 & 28.91 & 04.71 & 16.45 & 29.14 & 03.69 & 12.63 & 31.93 \\

&  ConvNeXt style 7$\times$7   & 18.11 & 44.53 & 50.69 & 10.46 & 31.69 & 32.26 & 04.92 & 18.52 & 14.48 & 02.86 & 12.13 & 06.36 & 02.1 & 09.3 & 03.51 & 01.5 & 06.59 & 01.9   \\

&  ConvNeXt style 7$\times$7 + 3$\times$3  & 16.2 & 39.55 & 50.82 & 09.0& 28.53 & 33.31 & 04.07 & 17.03 & 15.6 & 02.14 & 10.13 & 07.12 & 01.38 & 05.91 & 03.74 & 00.71 & 02.56 & 01.82 \\

&  ConvNeXt style 11$\times$11   & 17.02 & 43.01 & 48.45 & 08.92 & 28.35 & 28.13 & 03.64 & 14.36 & 10.06 & 01.17 & 06.28 & 03.11 & 00.55 & 04.04 & 01.35 & 00.32 & 02.76 & 00.77  \\

&  ConvNeXt style 11$\times$11 + 3$\times$3   & 19.34 & 43.6 & 54.41 & 10.71 & 31.22 & 33.98 & 04.6 & 15.76 & 12.75 & 01.98 & 07.78 & 04.04 & 00.95 & 03.95 & 01.69 & 00.51 & 01.96 & 00.78  \\

\midrule

\multirow{3}{*}{LCTC: 9$\times$9 + 3$\times$3 (Ours)} &   ResNet Style 3$\times$3  & 24.87 & 55.04 & 57.35 & 17.0& 46.34 & 42.08 & 10.88 & 36.04 & 28.55 & 07.91 & 28.17 & 22.86 & 06.02 & 21.13 & 22.87 & 04.63 & 14.45 & 27.39 \\

&  ConvNeXt style 7$\times$7   & 16.56 & 36.5 & 53.58 & 08.74 & 23.95 & 35.67 & 04.01 & 13.92 & 16.64 & 02.13 & 08.87 & 06.34 & 01.38 & 06.37 & 02.27 & 01.01 & 04.8 & 01.11   \\

&  ConvNeXt style 7$\times$7 + 3$\times$3  & 16.03 & 36.92 & 51.5 & 08.8 & 25.53 & 33.15 & 03.64 & 13.95 & 12.25 & 01.61 & 06.02 & 04.08 & 00.83 & 02.72 & 01.83 & 00.37 & 01.13 & 00.87 \\

&  ConvNeXt style 11$\times$11   & 16.42 & 39.19 & 51.71 & 08.32 & 26.64 & 31.11 & 03.66 & 15.7 & 11.61 & 01.94 & 10.11 & 04.4 & 01.19 & 06.75 & 02.23 & 00.83 & 04.83 & 01.36  \\

&  ConvNeXt style 11$\times$11 + 3$\times$3   & 18.72 & 41.83 & 55.48 & 10.38 & 29.7 & 36.72 & 04.74 & 18.16 & 17.44 & 02.49 & 11.2 & 07.18 & 01.69 & 08.21 & 03.56 & 01.24 & 06.07 & 01.93  \\

\midrule

\multirow{3}{*}{LCTC: 11$\times$11 (Ours)} &   ResNet Style 3$\times$3   & 26.02 & 48.81 & 63.76 & 16.8 & 39.62 & 49.72 & 09.62 & 29.4 & 34.22 & 06.85 & 24.07 & 27.66 & 05.63 & 20.38 & 26.45 & 04.56 & 15.64 & 28.86 \\

& ConvNeXt style 7$\times$7 & 19.04 & 45.39 & 52.63 & 10.17 & 32.3 & 32.46 & 04.58 & 20.16 & 13.36 & 02.44 & 13.63 & 05.33 & 01.74 & 10.13 & 03.04 & 01.21 & 07.07 & 01.7 \\

&  ConvNeXt style 7$\times$7 + 3$\times$3     & 16.08 & 39.09 & 53.1 & 08.86 & 28.27 & 35.06 & 03.94 & 16.77 & 15.75 & 02.25 & 11.87 & 06.31  & 01.32 & 07.98 & 02.72 & 00.82 & 05.14 & 01.28 \\

&  ConvNeXt style 11$\times$11 & 18.09 & 40.72 & 53.7 & 09.93 & 29.6 & 34.68 & 04.55 & 18.22 & 14.17 & 02.21 & 10.51 & 05.2 & 01.38 & 06.35 & 02.34 & 00.96 & 03.84 & 01.28 \\

&  ConvNeXt style 11$\times$11 + 3$\times$3    & 15.29 & 37.2 & 50.71 & 07.6 & 25.19 & 30.65 & 03.17 & 15.06 & 09.58 & 01.78 & 10.21 & 03.07  & 01.3 & 07.74 & 01.39 & 01.0& 05.6 & 00.88 \\

\midrule

\multirow{3}{*}{LCTC: 11$\times$11 + 3$\times$3 (Ours)} &   ResNet Style 3$\times$3  & 27.49 & 53.08 & 64.13 & 18.15 & 43.51 & 49.36 & 10.29 & 31.12 & 33.17 & 07.08 & 23.3 & 26.82 & 05.14 & 16.14 & 27.32 & 03.77 & 09.6 & 31.61 \\

&  ConvNeXt style 7$\times$7  & 16.14 & 40.65 & 50.39 & 08.08 & 27.2 & 31.4 & 03.34 & 15.36 & 12.29 & 01.93 & 09.35 & 03.9 & 01.36 & 05.77 & 01.76 & 00.92 & 03.51 & 00.83 \\

&  ConvNeXt style 7$\times$7 + 3$\times$3 & 17.7 & 39.71 & 54.64 & 09.71 & 26.92 & 35.8 & 04.32 & 13.93 & 15.8 & 02.37 & 08.49 & 06.7  & 01.59 & 05.85 & 03.43 &  01.09 & 03.87 & 01.83 \\ 

&  ConvNeXt style 11$\times$11    & 14.62  & 34.73 & 49.37 & 07.26 & 22.21 & 29.37 & 02.76 & 12.24 & 10.69 & 01.23 & 07.06 & 04.16 & 00.71 & 04.71 & 01.96 & 00.63 & 03.65 & 00.96 \\

&  ConvNeXt style 11$\times$11 + 3$\times$3  & 18.76 & 44.6 & 51.49 & 10.07 & 31.15 & 30.26 & 04.4 & 17.02 & 10.56 & 02.31 & 08.7 & 03.5  & 01.34 & 04.85 & 01.66 & 00.73 & 02.56 & 00.81 \\

\midrule

\multirow{3}{*}{LCTC: 13$\times$13 (Ours)} &   ResNet Style 3$\times$3  & 28.51 & 57.18 & 63.94 & 19.71 & 48.99 & 50.08 & 11.99 & 37.69 & 33.26 & 08.31 & 28.29 & 26.23 & 06.17 & 21.38 & 25.65 & 04.83 & 15.34 & 29.52 \\

&  ConvNeXt style 7$\times$7   & 20.9 & 46.62 & 55.13 & 12.32 & 34.21 & 35.91 & 06.14 & 21.39 & 16.39 & 03.15 & 13.44 & 07.51 & 02.16 & 10.21 & 04.3 & 01.41 & 06.61 & 02.54   \\

&  ConvNeXt style 7$\times$7 + 3$\times$3  & 20.13 & 42.92 & 57.7 & 11.38 & 29.96 & 39.57 & 04.85 & 15.81 & 19.37 & 02.54 & 09.48 & 09.47 & 01.65 & 06.45 & 05.61 & 00.86 & 03.83 & 03.0\\

&  ConvNeXt style 11$\times$11   & 18.65 & 39.48 & 56.4 & 10.02 & 27.46 & 38.02 & 04.69 & 17.27 & 19.03 & 02.47 & 11.35 & 08.76 & 01.39 & 07.95 & 04.12 & 00.9 & 06.02 & 02.11  \\

&  ConvNeXt style 11$\times$11 + 3$\times$3   & 18.95 & 42.88 & 55.82 & 10.68 & 31.21 & 35.69 & 04.92 & 18.29 & 12.63 & 02.35 & 09.29 & 03.78 & 01.26 & 05.02 & 01.6 & 00.79 & 02.56 & 00.72  \\

\midrule

\multirow{3}{*}{LCTC: 13$\times$13 + 3$\times$3 (Ours)} &   ResNet Style 3$\times$3 & 28.08 & 58.22 & 63.4 & 19.4 & 50.01 & 48.89 & 12.04 & 39.2 & 32.11 & 08.77 & 31.09 & 24.9 & 06.46 & 22.51 & 23.98 & 04.34 & 13.59 & 28.41  \\

&  ConvNeXt style 7$\times$7   & 18.42 & 43.52 & 51.26 & 10.23 & 30.56 & 30.5 & 04.37 & 16.41 & 11.29 & 02.08 & 09.09 & 04.39 & 01.35 & 06.65 & 02.49 & 00.86 & 04.28 & 01.37   \\

&  ConvNeXt style 7$\times$7 + 3$\times$3  & 16.7 & 41.09 & 50.56 & 08.54 & 26.94 & 30.39 & 03.31 & 13.44 & 11.22 & 01.53 & 06.72 & 04.13 & 01.02 & 04.01 & 02.01 & 00.56 & 02.07 & 01.03  \\

&  ConvNeXt style 11$\times$11   & 14.1 & 36.4 & 47.79 & 07.01 & 23.32 & 27.54 & 02.87 & 12.05 & 10.26 & 01.51 & 06.85 & 04.56 & 01.13 & 05.21 & 03.2 & 01.13 & 05.21 & 03.2  \\

&  ConvNeXt style 11$\times$11 + 3$\times$3   & 18.14 & 43.14 & 52.31 & 09.55 & 28.16 & 32.41 & 03.54 & 12.57 & 11.22 & 01.52 & 06.19 & 03.35 & 00.97 & 03.66 & 01.52 & 00.65 & 02.11 & 00.84  \\

\midrule

\multirow{3}{*}{LCTC: 15$\times$15 (Ours)} &   ResNet Style 3$\times$3  & 29.41 & 51.54 & 66.7 & 19.96 & 41.26 & 55.14& 11.51 & 29.26 & 41.04 & 07.17 & 20.8 & 31.7 & 05.13 & 15.79 & 28.53 & 03.9 & 11.59 & 29.37 \\

&  ConvNeXt style 7$\times$7   & 18.62 & 44.42 & 51.51 & 10.55 & 32.47 & 32.54 & 04.69 & 18.66 & 12.4 & 02.64 & 11.81 & 04.44 & 01.67 & 07.93 & 02.02 & 01.29 & 05.26 & 01.21  \\

&  ConvNeXt style 7$\times$7 + 3$\times$3  & 17.63 & 37.55 & 55.52 & 09.13 & 23.28 & 37.06 & 03.46 & 09.05 & 15.8 & 01.41 & 03.5 & 06.96 & 00.77 & 01.62 & 04.02 & 00.51 & 01.04 & 02.48 \\

&  ConvNeXt style 11$\times$11   & 15.24 & 36.62 & 49.45 & 08.05 & 24.89 & 31.93 & 03.68 & 14.62 & 13.68 & 02.03 & 07.68 & 05.66 & 01.26 & 04.48 & 03.25 & 00.61 & 02.51 & 01.87  \\

&  ConvNeXt style 11$\times$11 + 3$\times$3   & 19.01 & 44.78 & 52.74 & 10.35 & 31.98 & 32.35 & 04.38 & 19.15 & 12.42 & 02.31 & 12.1 & 04.84 & 01.53 & 07.72 & 02.24 & 01.06 & 04.86 & 01.17  \\

\midrule

\multirow{3}{*}{LCTC: 15$\times$15 + 3$\times$3 (Ours)} &   ResNet Style 3$\times$3  & 26.38 & 53.79 & 61.59 & 17.9 & 45.03 & 47.36 & 10.79 & 33.9 & 31.75 & 07.14 & 25.02 & 25.01 & 05.43 & 18.97 & 25.39 & 04.18 & 13.25 & 30.47 \\

&  ConvNeXt style 7$\times$7   & 19.81 & 42.18 & 53.73 & 11.14 & 28.25 & 37.03 & 04.72 & 13.45 & 17.41 & 01.92 & 05.27 & 07.64 & 01.06 & 03.07 & 04.37 & 00.7 & 01.97 & 02.61    \\

&  ConvNeXt style 7$\times$7 + 3$\times$3  & 17.53 & 39.4 & 54.39 & 09.51 & 26.67 & 35.52 & 03.73 & 12.39 & 13.27 & 00.93 & 04.2 & 03.22 & 00.44 & 02.14 & 01.01 & 00.16 & 00.94 & 00.36 \\

&  ConvNeXt style 11$\times$11   & 16.69 & 39.29 & 52.18 & 08.78 & 27.55 & 32.8 & 03.72 & 17.08 & 12.37 & 02.06 & 11.89 & 03.94 & 01.38 & 08.59 & 02.0& 00.99 & 06.41 & 01.24  \\

&  ConvNeXt style 11$\times$11 + 3$\times$3   & 19.15 & 41.08 & 55.96 & 10.71 & 29.12 & 37.58 & 05.28 & 19.35 & 18.54 & 02.88 & 13.69 & 08.43 & 02.0& 11.2 & 04.47 & 01.41 & 09.0& 02.39  \\

\midrule

\multirow{3}{*}{LCTC: 17$\times$17 (Ours)} &   ResNet Style 3$\times$3  & 27.74 & 53.24 & 64.48 & 18.51 & 43.47 & 51.02 & 10.72 & 32.21 & 36.08 & 07.43 & 25.5 & 28.78 & 05.85 & 20.69 & 28.85 & 04.93 & 16.94 & 32.03  \\

&  ConvNeXt style 7$\times$7   & 19.82 & 46.01 & 54.48 & 11.09 & 32.71 & 36.79 & 05.35 & 19.1 & 17.49 & 02.67 & 10.25 & 07.1 & 01.87 & 07.1 & 02.98 & 01.25 & 04.64 & 01.33   \\

&  ConvNeXt style 7$\times$7 + 3$\times$3  & 16.96 & 38.94 & 54.19 & 09.23 & 26.92 & 36.22 & 04.47 & 16.8 & 16.9 & 02.45 & 11.16 & 07.09 & 01.61 & 08.12 & 03.53 & 01.03 & 05.47 & 01.95 \\

&  ConvNeXt style 11$\times$11   & 13.72 & 34.03 & 48.09 & 06.57 & 20.94 & 26.93 & 02.32 & 09.22 & 07.84 & 01.08 & 04.61 & 02.28 & 00.63 & 02.55 & 01.0& 00.35 & 01.33 & 00.47  \\

&  ConvNeXt style 11$\times$11 + 3$\times$3   & 14.47 & 33.55 & 52.16 & 07.33 & 20.91 & 32.91 & 02.82 & 09.75 & 13.11 & 01.28 & 04.95 & 04.96 & 00.79 & 03.01 & 02.52 & 00.47 & 01.55 & 01.32  \\

\midrule

\multirow{3}{*}{LCTC: 17$\times$17 + 3$\times$3 (Ours)} &   ResNet Style 3$\times$3  & 26.97 & 54.13 & 62.04 & 18.41 & 45.5 & 47.66 & 11.05 & 34.55 & 32.01 & 07.43 & 25.65 & 24.78 & 05.07 & 17.38 & 24.12 & 03.51 & 10.48 & 27.4 \\

&  ConvNeXt style 7$\times$7   & 17.96 & 41.81 & 54.93 & 09.08 & 27.73 & 35.7 & 03.85 & 15.38 & 15.51 & 01.95 & 09.2 & 05.76 & 01.17 & 05.74 & 02.35 & 00.86 & 03.99 & 01.25   \\

&  ConvNeXt style 7$\times$7 + 3$\times$3  & 19.86 & 42.55 & 56.89 & 10.29 & 28.26 & 37.83 & 04.43 & 15.77 & 16.52 & 01.93 & 08.45 & 06.36 & 01.0& 05.27 & 02.61 & 00.68 & 03.7 & 01.31  \\

&  ConvNeXt style 11$\times$11   & 16.84 & 44.91 & 45.35 & 09.41 & 31.25 & 26.04 & 03.8 & 14.79 & 09.08 & 01.4 & 05.56 & 02.83 & 00.47 & 02.05 & 01.03 & 00.17 & 01.1 & 00.54  \\

&  ConvNeXt style 11$\times$11 + 3$\times$3   & 14.06 & 38.28 & 46.37 & 06.95 & 24.78 & 27.4 & 02.92 & 14.65 & 10.36 & 01.55 & 09.22 & 03.9 & 00.96 & 06.21 & 02.04 & 00.68 & 04.61 & 01.29  \\

\midrule

\multirow{3}{*}{LCTC: 19$\times$19 (Ours)} &   ResNet Style 3$\times$3  & 27.64 & 52.62 & 64.53 & 18.46 & 42.79 & 51.19 & 10.49 & 30.27 & 36.37 & 06.92 & 22.02 & 28.21 & 05.17 & 17.07 & 26.09 & 03.99 & 12.16 & 27.92 \\

&  ConvNeXt style 7$\times$7    & 20.28 & 46.96 & 56.75 & 10.06 & 30.29 & 36.2 & 03.5 & 13.07 & 14.56 & 01.51 & 06.13 & 05.64 & 00.72 & 03.24 & 02.4 & 00.64 & 02.2 & 01.4  \\

&  ConvNeXt style 7$\times$7 + 3$\times$3  & 18.14 & 39.86 & 56.98 & 09.34 & 26.13 & 37.16 & 03.5 & 13.58 & 14.24 & 01.52 & 07.79 & 04.4 & 00.83 & 05.76 & 01.83 & 00.54 & 03.82 & 00.93 \\

&  ConvNeXt style 11$\times$11   & 15.85 & 40.55 & 46.13 & 08.33 & 28.21 & 25.47 & 03.27 & 15.94 & 07.52 & 01.8 & 10.44 & 02.95 & 01.3 & 07.65 & 01.81 & 00.97 & 05.34 & 01.22  \\

&  ConvNeXt style 11$\times$11 + 3$\times$3   & 16.17 & 37.68 & 50.29 & 08.47 & 25.22 & 31.86 & 03.84 & 14.74 & 13.88 & 01.93 & 08.84 & 06.27 & 01.2 & 05.41 & 03.28 & 00.79 & 03.68 & 02.02  \\

\midrule

\multirow{3}{*}{LCTC: 19$\times$19 + 3$\times$3 (Ours)} &   ResNet Style 3$\times$3 & 28.62 & 56.15 & 63.93 & 19.19 & 47.17 & 48.9 & 10.96 & 34.41 & 31.54 & 07.15 & 25.67 & 23.6 & 05.25 & 19.32 & 21.84 & 04.24 & 15.12 & 24.17  \\

&  ConvNeXt style 7$\times$7   & 17.45 & 40.44 & 51.24 & 09.13 & 27.29 & 31.41 & 03.56 & 13.76 & 10.77 & 01.61 & 06.83 & 03.35 & 00.94 & 04.0& 01.37 & 00.56 & 02.22 & 00.54   \\

&  ConvNeXt style 7$\times$7 + 3$\times$3  & 20.9 & 48.61 & 54.54 & 11.59 & 35.05 & 33.3 & 04.6 & 18.94 & 11.57 & 02.05 & 10.9 & 03.51 & 01.42 & 07.83 & 01.63 & 00.9 & 05.28 & 00.88 \\

&  ConvNeXt style 11$\times$11   & 19.01 & 41.41 & 55.17 & 10.56 & 28.9 & 37.54 & 05.23 & 19.35 & 17.13 & 02.92 & 12.99 & 07.27 & 02.05 & 09.7 & 03.82 & 01.42 & 07.14 & 02.18  \\

&  ConvNeXt style 11$\times$11 + 3$\times$3   & 17.98 & 44.39 & 52.86 & 09.44 & 30.77 & 32.15 & 03.36 & 14.64 & 09.76 & 01.14 & 05.22 & 02.06 & 00.48 & 02.49 & 00.6 & 00.17 & 01.16 & 00.26  \\

\midrule

\multirow{3}{*}{LCTC: 31$\times$31 (Ours)} &   ResNet Style 3$\times$3  & 26.44 & 50.1 & 63.1 & 17.8 & 39.96 & 50.81 & 10.44 & 29.22 & 36.6 & 06.67 & 21.09 & 28.13 & 04.91 & 16.07 & 23.92 & 03.65 & 10.93 & 23.02 \\

&  ConvNeXt style 7$\times$7   & 17.75 & 41.69 & 51.94 & 09.26 & 27.63 & 32.32 & 03.52 & 12.48 & 11.73 & 01.37 & 04.95 & 04.18 & 00.62 & 02.57 & 01.93 & 00.37 & 01.68 & 01.02  \\

&  ConvNeXt style 7$\times$7 + 3$\times$3  & 16.53 & 40.82 & 50.9 & 08.0& 25.9 & 30.59 & 02.79 & 11.15 & 10.24 & 01.24 & 05.19 & 03.11 & 00.56 & 02.42 & 01.12 & 00.34 & 01.4 & 00.53 \\

&  ConvNeXt style 11$\times$11   & 13.08 & 31.95 & 45.87 & 05.85 & 17.71 & 25.83 & 02.06 & 07.35 & 08.65 & 00.88 & 03.15 & 02.74 & 00.39 & 01.55 & 01.05 & 00.26 & 01.15 & 00.55  \\

&  ConvNeXt style 11$\times$11 + 3$\times$3   & 15.42 & 35.92 & 51.53 & 07.44 & 21.84 & 31.72 & 02.43 & 09.33 & 10.18 & 00.85 & 03.85 & 02.59 & 00.41 & 01.79 & 01.04 & 00.22 & 00.99 & 00.56   \\

\midrule

\multirow{3}{*}{LCTC: 31$\times$31 + 3$\times$3 (Ours)} & ResNet Style 3$\times$3 & 27.41 & 54.28 & 64.15 & 18.27 & 44.66 & 49.97 & 11.02 & 33.64 & 34.65 & 07.24 & 25.06 & 26.54 & 05.39 & 18.81 & 22.82 & 04.3 & 14.03 & 22.46  \\

&  ConvNeXt style 7$\times$7   & 18.76 & 40.98 & 55.63 & 10.33 & 28.32 & 38.72 & 04.95 & 18.11 & 19.82 & 02.74 & 12.53 & 08.41 & 01.69 & 08.6 & 03.73 & 01.03 & 05.94 & 01.75   \\

&  ConvNeXt style 7$\times$7 + 3$\times$3  & 20.55 & 44.15 & 58.99 & 10.65 & 30.05 & 40.0 & 04.69 & 17.49 & 18.07 & 02.65 & 11.74 & 07.33 & 01.6 & 08.05 & 03.37 & 01.09 & 05.88 & 01.79 \\

&  ConvNeXt style 11$\times$11   & 14.47 & 36.3 & 49.11 & 07.12 & 22.87 & 29.65 & 02.57 & 11.14 & 10.67 & 01.24 & 06.52 & 03.69 & 00.9 & 05.06 & 01.68 & 00.63 & 03.65 & 00.96  \\

&  ConvNeXt style 11$\times$11 + 3$\times$3   & 13.59 & 32.71 & 49.91 & 06.09 & 18.39 & 29.59 & 01.96 & 06.7 & 08.76 & 00.79 & 02.53 & 02.11 & 00.4 & 01.47 & 00.89 & 00.12 & 00.65 & 00.45 \\

\bottomrule

\end{tabular}
}
\end{table*}

%
%
\subsubsection{Limit of large kernels for Upsampling}
\label{subsubsec:appendix:limit_large}
\begin{table*}[t]
\caption{Comparison of performance of Large Context Transposed Convolutions (LCTC) with very large: 31$\times$31 kernels in transposed convolution to large~(7$\times$7 to 17$\times$17) kernels.
All have a parallel 3$\times$3 kernel, as shown in \Cref{fig:unet_plus_block} (bottom left).
Here we observe the saturation of performance for very large kernels for upsampling. 
This comparison is for the same encoder (ConvNeXt) and same ResNet-like building blocks in the decoder~(our baseline).
The complete table is provided in \Cref{subsubsec:appendix:exp:segment}.}
\label{tbl:ablation:kernel_limit}
\centering
\scalebox{.71}{
\begin{tabular}{p{3cm}ccc|ccc|ccc|ccc}
\toprule
 \multirow{3}{3cm}{\textbf{Transposed Convolution Kernels}} & \multicolumn{3}{c}{\textbf{Test Accuracy}} & \multicolumn{6}{c}{\textbf{FGSM attack epsilon}} & \multicolumn{3}{c}{\textbf{SegPGD attack iterations}} \\
& \multirow{2}{*}{mIoU} & \multirow{2}{*}{mAcc} & \multirow{2}{*}{allAcc} & \multicolumn{3}{c}{$\frac{1}{255}$} & \multicolumn{3}{c}{$\frac{8}{255}$}  &  \multicolumn{3}{c}{20}  \\
 &   &  &  &  mIoU & mAcc & allAcc & mIoU & mAcc & allAcc & mIoU & mAcc & allAcc\\
\midrule
 7$\times$7 & 78.50 & 87.57 & 95.13 & 53.85 & 72.75 & 85.87  & 47.10 & 67.57 & 82.04  & 7.38 & \textbf{26.16} & 26.11 \\
 \textbf{11$\times$11} &  \textbf{79.33} & \textbf{87.81} & \textbf{95.41}  & \textbf{58.04} & \textbf{74.93} & \textbf{87.80}  &  \textbf{51.25} & 69.31 & \textbf{84.64} & 7.08 & 23.30 & \textbf{26.82} \\
  15$\times$15 & 78.72 & 87.50 & 95.25   & 56.28 & 73.97 & 87.15 & 49.50 & 68.69 & 83.53  & 7.14 & 25.02 & 25.01 \\
 17$\times$17 & 78.41 & 86.84 & 95.26  & 56.03 & 73.28 & 87.16 & 49.65 & 67.95 & 83.74  & \textbf{7.43} & 25.65 & 24.78 \\
 19$\times$19 & 78.78 & 87.34 & 95.28 & 56.53 & 74.59 & 86.97 & 50.60 & \textbf{69.95} & 83.98 & 7.15 & 25.67 & 23.60 \\
 31$\times$31 &  78.47 & 87.26 & 95.16  & 56.27 & 73.39 & 87.22 & 49.66 & 68.81 & 83.92 & 7.24 & 25.06 & 26.54 \\
\bottomrule

\end{tabular}
}
\end{table*}
As discussed in \cref{subsec:ablation:kernel_limit}, the performance of large kernels begins to saturate at a point.
We report results from \Cref{fig:transpose_conv_plots} in tabular form in \Cref{tbl:ablation:kernel_limit}.
In \Cref{tbl:ablation:kernel_limit}, we find that 13$\times$13 appears to be the saturation point for this setting and 31$\times$31 kernels are beyond this saturation point.
While 31$\times$31 performs worse or on-par with 17$\times$17, it still performs significantly better than the baseline of 2$\times$2.
In \Cref{subsec:ablation:kernel_limit} we explain the kernel size limit and that larger kernels are difficult to train.
\textbf{We also find that these results further strengthen our Hypothesis~\ref{hyp:second}.}
For ease of understanding, we visualize the trends from \Cref{tbl:ablation:kernel_limit} in \Cref{fig:transpose_conv_plots_largestconvnext}.
\begin{table*}[t]
\caption{Adversarially trained models using FGSM and PGD from \Cref{tbl:exp:semantic:unet-convnext-adv} tested against adversarial attacks on UNet with ConvNeXt encoder and decoder with different sized kernels in the transposed convolution for upscaling, while keeping rest of the architecture identical.}
\label{tbl:exp:semantic:unet-full-convnext-adv-training}
\centering
\scalebox{.76}{
\begin{tabular}{p{3cm}ccc|ccc|ccc|ccc|ccc}
\toprule
 \multirow{3}{3cm}{\textbf{Transposed Convolution Kernels}} & \multicolumn{3}{c|}{\textbf{Clean Test Accuracy}} & \multicolumn{6}{c|}{\textbf{FGSM attack epsilon}} & \multicolumn{6}{c}{\textbf{SegPGD attack iterations}} \\
& & & & \multicolumn{3}{c}{$\frac{1}{255}$} & \multicolumn{3}{c|}{$\frac{8}{255}$}  &  \multicolumn{3}{c}{3} & \multicolumn{3}{c}{20} \\
 &   mIoU & mAcc & allAcc &   mIoU & mAcc & allAcc & mIoU & mAcc & allAcc & mIoU & mAcc & allAcc & mIoU & mAcc & allAcc\\

\midrule
\multicolumn{16}{c}{\textbf{FGSM training}} \\
\midrule

 2$\times$2 (baseline) &78.57 & 86.68 & 95.23  & 54.28 & 70.80 & 86.91 & 52.45 & 68.38 & \textbf{86.26}  &  26.59 & 48.99 & 67.71 & 7.6 & 24.06 & 31.37 \\
 
 LCTC: 7$\times$7 (Ours) & 78.41  & 86.22 & 95.20& 56.87 & 72.92 & 87.70 &  51.31 & 68.4 & 85.17 & 28.11 & 53.39 & 66.30 & 8.36 & 28.54 & 28.13 \\
 
 \textbf{LCTC: 11$\times$11 + 3$\times$3 (Ours)} & \textbf{79.57} & \textbf{88.1} & \textbf{95.3}& \textbf{57.90} & \textbf{74.64} & 87.61 &  52.15 & \textbf{70.23} & 84.96 & \textbf{30.37} & \textbf{55.54} & \textbf{68.3} & \textbf{9.4} & \textbf{29.79} & \textbf{32.37}   \\

\midrule
\multicolumn{16}{c}{\textbf{PGD training with 3 attack iterations}} \\
\midrule

 2$\times$2 (baseline) & 75.33 & 84.66 & 94.39 & 53.87 & 72.17 & 86.58 & 58.57 & 73.93 & 89.01 & 29.38 & 57.82 & 66.67 & 9.39 & 33.15 & 28.11 \\
 
 LCTC: 7$\times$7 (Ours) & 75.79 & 84.89 & 94.38 & 54.82 & 72.31 & \textbf{86.80} & 61.29 & 74.33 & 89.96 & 31.12 & 58.36 & 68.58 & 10.24 & \textbf{33.99} & 31.14 \\
 
 \textbf{LCTC: 11$\times$11 + 3$\times$3 (Ours)} & \textbf{75.90} & \textbf{86.60} & 94.30 & \textbf{56.27} & \textbf{75.66} & 86.68 & \textbf{63.02} & \textbf{76.17} & \textbf{90.42} & \textbf{33.50} & 58.34 & \textbf{71.50} & \textbf{10.77} & 32.23 & \textbf{37.36} \\
\bottomrule

\end{tabular}
}
\end{table*}

%
%
%
%
%
%
%
%
%
\subsection{Choice of encoder}
\label{subsubsec:appendix:ablation:encoder}
\begin{table*}[ht]
\caption{Comparison of performances of different encoders in the UNet-like architecture. All architectures here have the baseline 2$\times$2 transposed convolution kernel for upsampling followed by 3$\times$3 convolution kernels in the decoder blocks. For more results please refer to \Cref{tbl:appendix:ablation:choice_of_encoder}.}
\label{tbl:ablation:choice_of_encoder}
\centering
\scalebox{0.75}{
\begin{tabular}{@{}p{2.5cm} c @{\hspace{0.1cm}} c @{\hspace{0.1cm}} c | @{\hspace{0.5cm}} c @{\hspace{0.1cm}} c @{\hspace{0.1cm}} c @{\hspace{0.5cm}} c @{\hspace{0.1cm}} c @{\hspace{0.1cm}} c | @{\hspace{0.5cm}} c @{\hspace{0.2cm}} c @{\hspace{0.2cm}} c @{}}
\toprule
\multirow{3}{2.5cm}{\textbf{Encoder}} & \multicolumn{3}{c}{\textbf{Test Accuracy}}   & \multicolumn{6}{c}{\textbf{FGSM attack epsilon}} &  \multicolumn{3}{c}{\textbf{SegPGD attack iterations}} \\
    & &  & & \multicolumn{3}{c}{$\frac{1}{255}$}  & \multicolumn{3}{c}{$\frac{8}{255}$}  & \multicolumn{3}{c}{20}  \\
 &  mIoU & mAcc & allAcc & mIoU & mAcc & allAcc  & mIoU & mAcc & allAcc & mIoU & mAcc & allAcc \\
\midrule
ResNet50 &  67.69 & 79.04 & 92.80 & 36.78 & 58.41 & 78.16 & 32.60 & 52.63 & 74.56 & 4.98 & 19.28 & 21.07 \\
ConvNeXt tiny & 78.45 & 86.66 & 95.20 & 53.76 & 70.62 & 86.32 & 47.33 & 64.58 & 83.16 &  5.54 & 18.79 & 23.72 \\
SLaK tiny & 78.82 & 87.01 & 95.17 & 55.22 & 71.72 & 86.97 & 48.69 & 66.45 & 83.57 & 8.45 & 25.42 & 32.37 \\

\bottomrule

\end{tabular}
}
\end{table*}


\begin{table*}[ht]
\caption{Comparison of performances of different encoders in the UNet-like architecture. All architectures here have the baseline 2$\times$2 transposed convolution kernel followed by 3$\times$3 convolution kernels in the decoder block.}
\label{tbl:appendix:ablation:choice_of_encoder}
\centering
\scalebox{0.4}{
\begin{tabular}{@{}p{2.5cm} c @{\hspace{0.1cm}} c @{\hspace{0.1cm}} c | @{\hspace{0.5cm}} c @{\hspace{0.1cm}} c @{\hspace{0.1cm}} c @{\hspace{0.5cm}} c @{\hspace{0.1cm}} c @{\hspace{0.1cm}} c | @{\hspace{0.5cm}} c @{\hspace{0.1cm}} c @{\hspace{0.1cm}} c @{\hspace{0.5cm}} c @{\hspace{0.1cm}} c @{\hspace{0.1cm}} c @{\hspace{0.5cm}} c @{\hspace{0.1cm}} c @{\hspace{0.1cm}} c @{\hspace{0.5cm}} c @{\hspace{0.1cm}} c @{\hspace{0.1cm}} c @{\hspace{0.5cm}} c @{\hspace{0.1cm}} c @{\hspace{0.1cm}} c @{\hspace{0.5cm}} c @{\hspace{0.1cm}} c @{\hspace{0.1cm}} c @{}}
\toprule
\multirow{3}{2.5cm}{\textbf{Encoder}} & \multicolumn{3}{c}{\textbf{Test Accuracy}}   & \multicolumn{6}{c}{\textbf{FGSM attack epsilon}} &  \multicolumn{18}{c}{\textbf{SegPGD attack iterations}} \\
    & &  & & \multicolumn{3}{c}{$\frac{1}{255}$}  & \multicolumn{3}{c}{$\frac{8}{255}$}    &    \multicolumn{3}{c}{3}     & \multicolumn{3}{c}{5} & \multicolumn{3}{c}{10} & \multicolumn{3}{c}{20} &   \multicolumn{3}{c}{40} & \multicolumn{3}{c}{100}                                  \\
 &  mIoU & mAcc & allAcc & mIoU & mAcc & allAcc  & mIoU & mAcc & allAcc & mIoU & mAcc & allAcc & mIoU & mAcc & allAcc & mIoU & mAcc & allAcc & mIoU & mAcc & allAcc & mIoU & mAcc & allAcc & mIoU & mAcc & allAcc \\
\midrule
ResNet50 &  67.69 & 79.04 & 92.80 & 36.78 & 58.41 & 78.16 & 32.60 & 52.63 & 74.56 & 16.18 & 37.46 & 50.04 & 11.32 & 30.59 & 38.98 & 7.21 & 23.76 & 27.58 & 4.98 & 19.28 & 21.07 & 3.95 & 16.49 & 18.35 & 3.09 & 13.87 & 15.87 \\
\midrule
ConvNeXt tiny & 78.45 & 86.66 & 95.20 & 53.76 & 70.62 & 86.32 & 47.33 & 64.58 & 83.16 & 23.06 & 46.51 & 60.04 & 14.43 & 35.50 & 45.30 & 8.12 & 24.67 & 29.88 & 5.54 & 18.79 & 23.72 & 4.39 & 14.98 & 23.70 & 3.50 & 11.61 & 27.93 \\
\midrule
SLaK tiny & 78.82 & 87.01 & 95.17 & 55.22 & 71.72 & 86.97 & 48.69 & 66.45 & 83.57 & 26.71 & 50.92 & 64.04 & 19.28 & 43.51 & 52.88 & 12.24 & 33.65 & 39.78 & 8.45 & 25.42 & 32.37 & 6.22 & 19.58 & 29.06 & - & - & - \\

\bottomrule

\end{tabular}
}
\end{table*}

\label{subsec:ablation:encoder_choice}
Following we aim to understand the importance of the encoder and its influence on the quality of representations later decoded during the upsampling.
Consequently, we justify our choice of using ConvNeXt tiny encoder for the majority of our studies.

In \Cref{tbl:ablation:choice_of_encoder} we compare different encoders: ResNet50, ConvNeXt tiny, and SLaK~\cite{slak} while fixing the decoder to the baseline implementation.
All encoders are pre-trained on the ImageNet-1k training dataset.

We observe that using ConvNeXt tiny and SLaK as the encoder backbone gives us significantly better performance than using ResNet50 as the encoder.
This observation holds true for both clean and adversarially perturbed samples. 
We additionally observe that SLaK gives us marginally better performance than ConvNeXt.
As shown by \cite{slak}, SLaK is a significantly better encoder than ConvNeXt tiny as it provides significantly more context than ConvNeXt by using kernel sizes up to 51$\times$51 in the convolution layers during encoding. 
This proves that better encoding can be harnessed during decoding which can lead to better upsampling.

However, in this work, we used the ConvNeXt tiny encoder since the SLaK encoder takes significantly longer to train for only a marginal gain in performance.
We report the performance results in \Cref{tbl:appendix:ablation:choice_of_encoder}. 
We observe that given our computation budget and the wall-clock time limit of 24 hours, we are unable to even compute the performance of the model with the SLaK encoder at 100 attack iterations.

\subsection{Ablation over small parallel kernel}
\label{subsubsec:ablation:parallel_kernel}
Following we ablate over the use of a small (3$\times$3) kernel in parallel to a large ($\geq$7$\times$7) kernel for Large Context Transposed Convolutions.
This concept is inspired by \cite{replk, slak} who use a small kernel in parallel with the large kernels to preserve local context when downsampling.
Similar behavior is observed while upsampling.
\Cref{tbl:appendix:ablation:backbone} compares the usage of this small parallel kernel.
We observe, that while not using the small kernel results in marginal better performance on clean images (for a fixed backbone style), it lacks context and thus performs poorly (when compared to using a small parallel kernel) against adversarial attacks.

This is further highlighted in\cref{tbl:appendix:ablation:segpgd_backbone} when the performance is compared against strong adversarial attacks.
Moreover, we observe that from medium-sized kernels i.e.~, the upsampling seems to lose local context, and adding a kernel in parallel helps the model in getting this additional context.
This effect can also be observed in the adversarial performances of the respective models.

%
%



\subsection{Drawbacks of interpolation}
\label{subsubsec:experiments:pspnet_interpolation}
\begin{figure*}[h]
     \centering
     \begin{subfigure}[t]{0.3\textwidth}
         \centering
         \includegraphics[width=\textwidth]{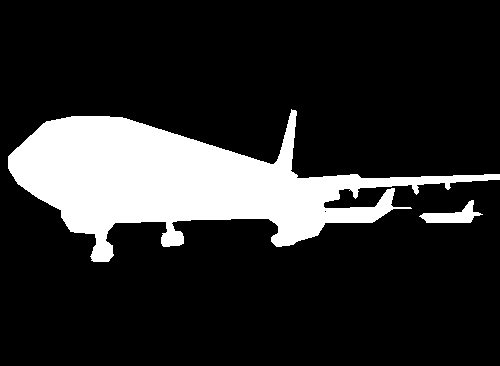}
         \caption{Ground truth segmentation mask of the third image in the test set.\newline\newline}
         \label{fig:appendix:exp:interpolation:aeroplane:gt}
     \end{subfigure}
     \hfill
     \begin{subfigure}[t]{0.3\textwidth}
         \centering
         \includegraphics[width=\textwidth]{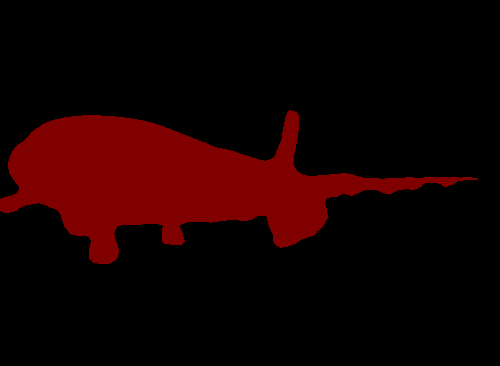}
         \caption{Prediction from PSPNet with ResNet 50 backbone as implemented by the authors.\newline\newline}
         \label{fig:appendix:exp:interpolation:aeroplane:pspnet_pred}
     \end{subfigure}
     \hfill
     \begin{subfigure}[t]{0.3\textwidth}
         \centering
         \includegraphics[width=\textwidth]{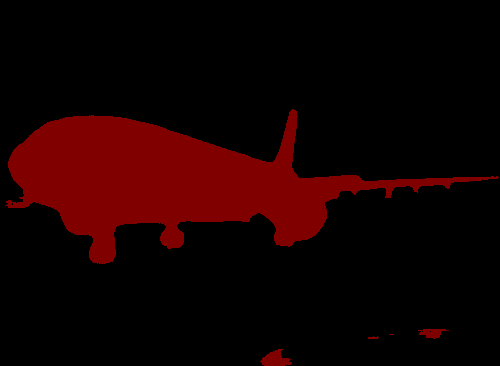}
         \caption{Prediction when using LCTC (11$\times$11 + 3$\times$3) and 3$\times$3 convolution kernels in the decoder building blocks of UNet.}
         \label{fig:appendix:exp:interpolation:aeroplane:trans11_decoder_conv_3_pred}
     \end{subfigure}
        \caption{A comparison of differences in the sharpness of final predictions due to different upsampling techniques.
        \cref{fig:appendix:exp:interpolation:aeroplane:gt} is the ground truth segmentation mask with sharp and thin edges in the rear fin and wing with protrusions in the wing of the aircraft. 
        We observe that PSPNet with a ResNet50 backbone as implemented by \cite{pspnet} is not able to accurately predict the thin edges and the protrusions, and is simply smoothening them out.
        This is due to the interpolation operation used in upsampling.
        However in comparison, as shown in \cref{fig:appendix:exp:interpolation:aeroplane:trans11_decoder_conv_3_pred}, when a transposed convolution operation is used for pixel-wise upsampling, the thin edges are sharper and protrusions are more accurately predicted.}
        \label{fig:appendix:experiments:interpolation:predictions}
\end{figure*}

As discussed in \Cref{sec:fft}, architecture designs that use interpolation for pixel-wise upsampling suffer with over-smoothening of feature maps. 
This can be seen in the final predictions, as shown in \cref{fig:appendix:exp:interpolation:aeroplane:pspnet_pred} compared to the ground truth segmentation mask in \cref{fig:appendix:exp:interpolation:aeroplane:gt} and prediction from a model with 11$\times$11 + 3$\times$3 transposed convolution kernel in \cref{fig:appendix:exp:interpolation:aeroplane:trans11_decoder_conv_3_pred}.

In their work, \cite{segpgd} showed that PSPNet has considerably lower performance against adversarial attacks, similar to the analysis made in \Cref{subsec:ablation:backbone}.
This is explained by H\ref{hyp:second}.
\subsection{Different Upsampling Methods}
\label{subsec:appendix:ablation:upsampling}
Following we compare different upsampling techniques thus justifying our advocacy for using Transposed Convolution instead of other upsampling techniques like interpolation and pixel shuffle.

We report the comparison in \Cref{tbl:ablation:upsampling} and observe that both Pixel shuffle and Nearest Neighbor interpolation perform better than the usually used Transposed Convolution with a 2$\times$2 kernel size. 
However, as we increase the kernel size for Transposed Convolution to 11$\times$11 with a 3$\times$3 small kernel in parallel, we observe that Large Context Transposed Convolutions are strictly outperforming pixel shuffle, on both clean unperturbed images and under adversarial attacks, across all metrics used. 
Transposed Convolution with a large kernel is either outperforming or performing at par with Nearest Neighbor interpolation as well.
Thus we demonstrate the superior clean and adversarial performance of large kernel-sized Transposed Convolution operation over other commonly used upsampling techniques.


\begin{table*}[ht]
\caption{Comparison of performances of different upsampling methods in the UNet-like architecture. All architectures here have the baseline i.e. ConvNeXt encoder and a ResNet style 3$\times$3 or ConvNext style 7
$\times$7+3$\times$3 convolution kernels in the decoder block.}
\label{tbl:appendix:ablation:upsampling_full}
\centering
\scalebox{0.34}{
\begin{tabular}{@{}p{3.5cm} c @{\hspace{0.1cm}} c @{\hspace{0.1cm}} c @{\hspace{0.1cm}} c | @{\hspace{0.5cm}} c @{\hspace{0.1cm}} c @{\hspace{0.1cm}} c @{\hspace{0.5cm}} c @{\hspace{0.1cm}} c @{\hspace{0.1cm}} c | @{\hspace{0.5cm}} c @{\hspace{0.1cm}} c @{\hspace{0.1cm}} c @{\hspace{0.5cm}} c @{\hspace{0.1cm}} c @{\hspace{0.1cm}} c @{\hspace{0.5cm}} c @{\hspace{0.1cm}} c @{\hspace{0.1cm}} c @{\hspace{0.5cm}} c @{\hspace{0.1cm}} c @{\hspace{0.1cm}} c @{\hspace{0.5cm}} c @{\hspace{0.1cm}} c @{\hspace{0.1cm}} c @{\hspace{0.5cm}} c @{\hspace{0.1cm}} c @{\hspace{0.1cm}} c @{}}
\toprule
\multirow{3}{3.5cm}{\textbf{Upsampling Method}} & \multirow{3}{4cm}{\textbf{Convolution Kernel in Decoder blocks}} & \multicolumn{3}{c}{\textbf{Test Accuracy}}   & \multicolumn{6}{c}{\textbf{FGSM attack epsilon}} &  \multicolumn{18}{c}{\textbf{SegPGD attack iterations}} \\
  &  & &  & & \multicolumn{3}{c}{$\frac{1}{255}$}  & \multicolumn{3}{c}{$\frac{8}{255}$}    &    \multicolumn{3}{c}{3}     & \multicolumn{3}{c}{5} & \multicolumn{3}{c}{10} & \multicolumn{3}{c}{20} &   \multicolumn{3}{c}{40} & \multicolumn{3}{c}{100}                                  \\
& &  mIoU & mAcc & allAcc & mIoU & mAcc & allAcc  & mIoU & mAcc & allAcc & mIoU & mAcc & allAcc & mIoU & mAcc & allAcc & mIoU & mAcc & allAcc & mIoU & mAcc & allAcc & mIoU & mAcc & allAcc & mIoU & mAcc & allAcc \\

\midrule
\multirow{2}{3.5cm}{Pixel Shuffle} & ResNet Style 3$\times$3 & 78.54 & 87.32 & 95.18 & 53.82 & 71.58 & 85.88 & 46.67 & 65.03 & 81.71 & 23.08 & 48.18 & 56.54 & 15.06 & 38.85 & 41.71 & 9.17 & 29.43 & 28.17 & 6.69 & 23.43 & 24.05 & 5.69 & 19.61 & 25.71 & 4.80 & 15.53 & 32.10 \\
& ConvNeXt Style 7$\times$7+3$\times$3 & 77.10 & 85.90 & 94.88 & 51.78 & 69.68 & 85.44 & 43.80 & 62.24 & 81.06 & 17.52 & 40.16 & 50.31 & 9.43 & 27.37 & 30.37 & 3.53 & 12.25 & 10.93 & 1.41 & 5.42 & 3.74 & 0.78 & 3.04 & 1.55 & 0.52 & 1.96 & 0.93 \\

\midrule

\multirow{2}{3.5cm}{\small Nearest Neighbour Interpolation} & ResNet Style 3$\times$3 & 78.40 & 88.16 & 95.09 & 52.68 & 73.51 & 84.55 & 46.08 & 67.96 & 80.22 & 22.82 & 53.16 & 51.75 &  15.34 & 44.53 & 36.21 & 10.02 & 34.83 & 23.84 & 7.65 & 27.89 & 20.48 & 6.43 & 23.23 & 21.48 & 5.40 & 17.34 & 28.05  \\
& ConvNeXt Style 7$\times$7+3$\times$3 & 77.86 & 86.92 & 94.97 & 50.71 & 71.21 & 84.45 & 41.97 & 64.92 & 78.89 & 15.77 & 44.36 & 42.09 & 8.56 & 30.25 & 23.74 & 2.96 & 12.56 & 7.19 & 1.27 & 5.70 & 2.10 & 0.52 & 2.08 & 0.75 & 0.17 & 0.85 & 0.35 \\
\midrule

\multirow{2}{3.5cm}{\small Transposed Convolution 2$\times$2} & ResNet Style 3$\times$3 &   78.45 & 86.66 & 95.20 & 53.76 & 70.62 & 86.32 & 47.33 & 64.58 & 83.16 & 23.06 & 46.51 & 60.04 & 14.43 & 35.50 & 45.30 & 8.12 & 24.67 & 29.88 & 5.54 & 18.79 & 23.72 & 4.39 & 14.98 & 23.70 & 3.50 & 11.61 & 27.93 \\
& ConvNeXt Style 7$\times$7+3$\times$3 &  77.24 & 86.03 & 94.84 & 51.09 & 70.53 & 85.29 & 43.52 & 63.74 & 81.18 & 17.59	 & 42.55 & 	51.68 & 9.88	 & 30.41 & 	32.33 & 4.75 & 16.83 & 	14.31 & 2.65	 & 9.46 & 	6.68 & 1.68	 & 5.64 & 	3.4 & 1.0 & 	3.16	 & 1.94 \\
\midrule
\multirow{2}{3.5cm}{\small \textbf{LCTC: 11$\times$11+3$\times$3 (Ours)}} & ResNet Style 3$\times$3 & 79.33 & 87.81 & 95.41 & 58.04 & 74.93 & 87.8 & 51.25 & 69.31 & 84.64 & 27.49 & 53.08 & 64.13 & 18.15 & 43.51 & 49.36 & 10.29 & 31.12 & 33.17 & 7.08 & 23.3 & 26.82 & 5.14 & 16.14 & 27.32 & 3.77 & 9.6 & 31.61 \\
& ConvNeXt Style 7$\times$7+3$\times$3 & 78.64 & 86.78 & 95.17 & 54.32 & 71.27 & 86.63 & 45.48 & 63.62 & 82.32  & 17.7 & 39.71 & 54.64 & 9.71 & 26.92 & 35.8 & 4.32 & 13.93 & 15.8 & 2.37 & 8.49 & 6.7  & 1.59 & 5.85 & 3.43 &  1.09 & 3.87 & 1.83 \\

\bottomrule

\end{tabular}
}
\end{table*}

There might be speculation if other downsampling techniques can utilize larger convolution kernels in the \textcolor{RoyalBlue}{decoder building blocks} better than transposed convolution. 
Thus, we additionally experiment using a ConvNeXt-like 7$\times$7+3$\times$3 kernel in the Convolution operations in the decoder building blocks that follow the upsampling operation. We report these results in \Cref{tbl:appendix:ablation:upsampling_full} and observe that similar to transposed convolution, other upsampling methods also do not benefit from an increase in the kernel size in the decoder building blocks. 

\subsection{Adversarial Training}
\label{subsec:appendix:ablation:semseg:adv_training}
Following, we present the results from adversarial training for semantic segmentation.
In \Cref{tbl:exp:semantic:unet-full-convnext-adv-training}, we report the performance of different transposed convolution kernel-sized adversarially trained UNet on clean input and adversarially perturbed inputs.
The observed performance improvement when increasing the transposed convolution kernel size during normal training also extends to adversarial training.

\section{Additional Results on Image Restoration}
\label{subsec:appendix:ablation:image_restoration}
Following we provide additional results for the Image deblurring tasks, like the performance of models after adversarial training and some visual results of the deblurring for a better understanding of the impact of increased spatial context against different adversarial attack methods and strengths.

\subsection{Latency Study}
\label{subsec:appendix:image_restoration:latency}
As PixelShuffle, when downsampling with a factor of 2, reduces the channel dims by a factor of 4, works \cite{chen2022simple,zamir2022restormer} use a 1$\times$1 convolution layer before the PixelShuffle to increase the number of channels by a factor of 4. 
This added complexity is not needed for Transposed Convolution. 
Thus, in \Cref{tab:latency_study} we report the number of parameters in the models from \Cref{fig:reconstruction_pgd_attack_main} and report latencies (mean over 1000 runs) of the upsampling operations, and show that these are comparable. 
In practice, these differences are negligible as other unchanged operations are more costly.
\begin{table}[t]
    \centering
    %
    \caption{Comparing latency and number of parameters for models from \Cref{fig:reconstruction_pgd_attack_main}.}
    \scalebox{1.0}{
    
    \begin{tabular}{@{}c@{\hspace{0.2cm}}c@{\hspace{0.2cm}}c@{}}
        \toprule
        Upsampling Method & Latency (ms) & No. of Params \\
        \midrule
        Pixel Shuffle & 0.26  & 17.11 M \\
        Trans.~Conv.~3$\times$3 & 0.27 & 16.43 M \\
        LCTC 11$\times$11+3$\times$3 & 0.38 & 16.54 M  \\
    \bottomrule
    \end{tabular}    
    }
    \label{tab:latency_study}
\end{table}

\subsection{Adversarial Training}
\label{subsec:appendix:image_restoration:adv_training}
\begin{table*}[ht]
\caption{Comparison of performances of adversarially trained \textit{SotA} Image Restoration Networks. The considered architectures use Pixel Shuffle for Upsampling, we propose replacing the Pixel Shuffle with Transposed Convolution operations using the large filter. Testing for image deblurring on GoPro dataset.}
\label{tbl:ablation:image_restoration_adv_train}
\centering
\scalebox{0.75}{
\begin{tabular}{@{}p{2.3cm} c@{\hspace{0.1cm}} c @{\hspace{0.1cm}} c @{\hspace{0.5cm}} | c @{\hspace{0.1cm}} c @{\hspace{0.1cm}} c @{\hspace{0.1cm}} c @{\hspace{0.1cm}} c @{\hspace{0.1cm}} c @{}}
\toprule
\multirow{3}{1.5cm}{\textbf{Network}} & \multirow{3}{5cm}{\textbf{Upsampling Method}} & \multicolumn{2}{c}{\textbf{Test Accuracy}}   & \multicolumn{6}{c}{\textbf{PGD attack iterations}} \\
  &  & &  & \multicolumn{2}{c}{5} & \multicolumn{2}{c}{10} & \multicolumn{2}{c}{20}  \\
& &  PSNR & SSIM &  PSNR & SSIM &  PSNR & SSIM &  PSNR & SSIM \\
\midrule

\multirow{4}{*}{NAFNet + ADV} & Pixel Shuffle &
29.91 & 0.9291 &  15.76 & 0.5228  & 13.91 & 0.4445  & 12.73 & 0.3859 \\

& Transposed Conv 3$\times$3 &  31.26 & 0.9448 & 15.89 & \textbf{0.5390} & 13.43 &\textbf{0.4627} & 11.62 &\textbf{0.4098} \\

& LCTC: 7$\times$7 + 3 $\times$3 (Ours) & 31.21 & 0.9446& \textbf{16.46} & 0.5061 & \textbf{14.55} & 0.4211 & \textbf{13.31} & 0.3688 \\

& LCTC:  11$\times$11 + 3$\times$3 (Ours) & 30.70 & 0.9390  & 13.68 & 0.4857 & 11.91 & 0.4085 & 10.92 & 0.3604 \\

\bottomrule

\end{tabular}
}
\end{table*}

In \Cref{tbl:ablation:image_restoration_adv_train} we provide additional results for adversarially training image restoration network NAFNet using FGSM attack on 50\% of the training minibatch of the GoPro dataset each iteration.
The state-of-the-art Image Restoration models are significantly larger w.r.t. the number of parameters, compared to the models considered for semantic segmentation.
Thus, they are significantly more difficult to train adversarially. 
They require more training iterations.
Due to the limited computing budget, we have only trained them for the same iterations as clean (non-adversarial) training iterations.
We already observe the advantages of using a larger kernel for transposed convolution over pixel-shuffle in these experiments.

\subsection{Visual Results}
\label{subsec:appendix:image_restoration:visual_results}
\begin{figure*}[htb]
    \centering 
   \begin{tabular}{@{}c@{\hspace{0.3cm}}c@{\hspace{0.3cm}}c@{\hspace{0.1cm}}c@{\hspace{0.1cm}}c@{\hspace{0.1cm}}c@{\hspace{0.1cm}}c@{}}
    \multicolumn{3}{c}{MODEL} & NO ATTACK & 5 iterations & 10 iterations & 20 iterations\\
  \rotatebox{90}{\textbf{Restormer}} & \rotatebox{90}{\tiny with Pixel Shuffle} & &
  \includegraphics[width=0.23\textwidth]{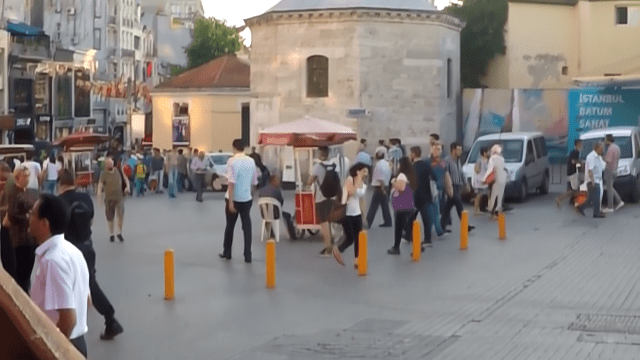} &
  \includegraphics[width=0.23\textwidth]{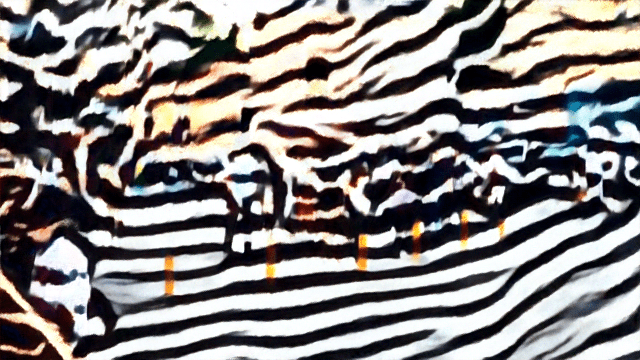} &
  \includegraphics[width=0.23\textwidth]{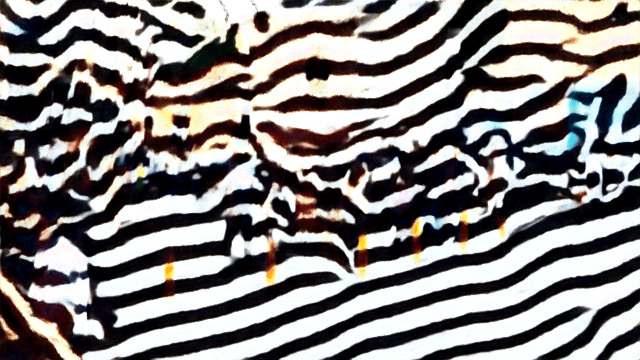} &
  \includegraphics[width=0.23\textwidth]{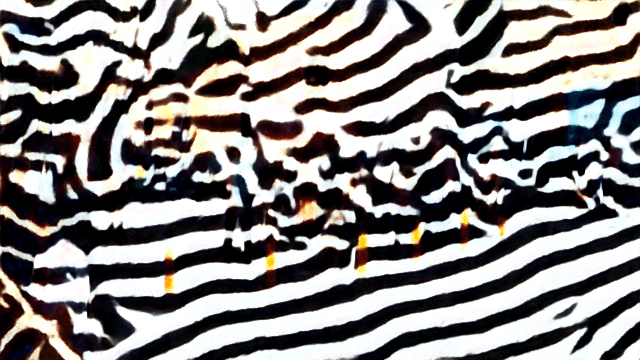}\\
  \rotatebox{90}{\textbf{Restormer}} & \rotatebox{90}{\tiny with Transposed Conv} & \rotatebox{90}{\phantom{su}~~~3$\times$3} &
  \includegraphics[width=0.23\textwidth]{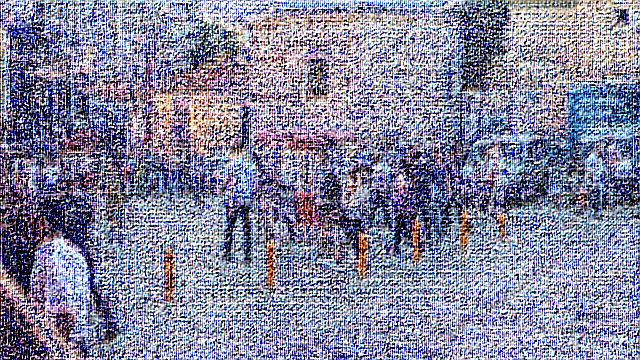} &
  \includegraphics[width=0.23\textwidth]{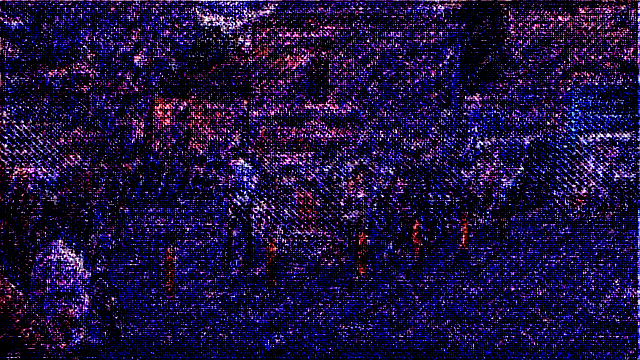} &
  \includegraphics[width=0.23\textwidth]{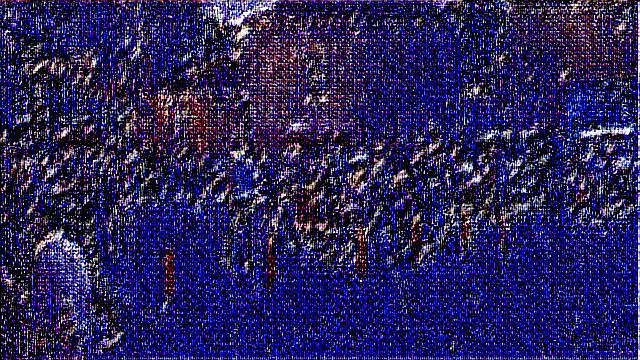} &
  \includegraphics[width=0.23\textwidth]{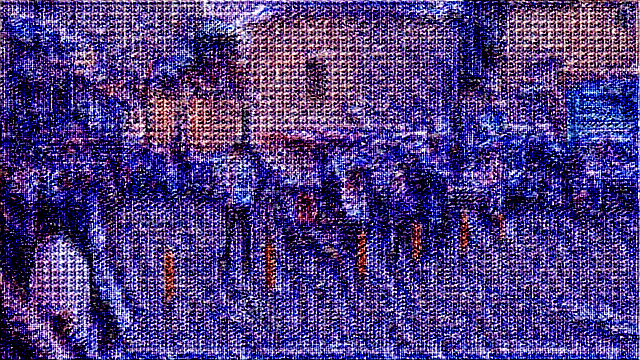}\\
  \rotatebox{90}{\textbf{Restormer}} & \rotatebox{90}{\tiny \phantom{substi}LCTC with }& \rotatebox{90}{~~7$\times$7 + 3$\times$3} &
  \includegraphics[width=0.23\textwidth]{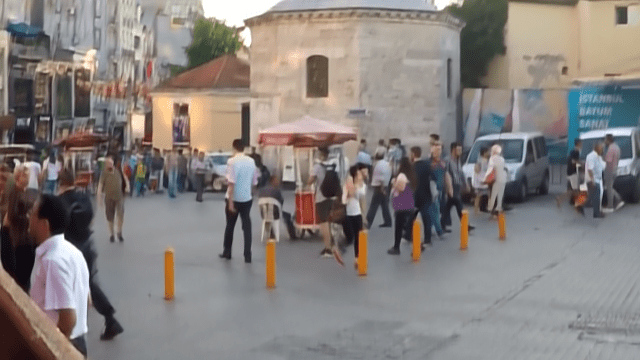} &
  \includegraphics[width=0.23\textwidth]{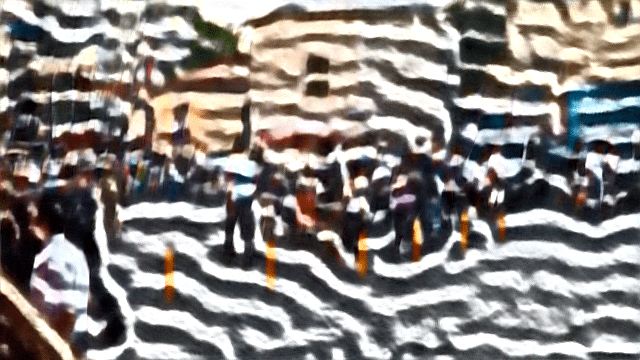} &
  \includegraphics[width=0.23\textwidth]{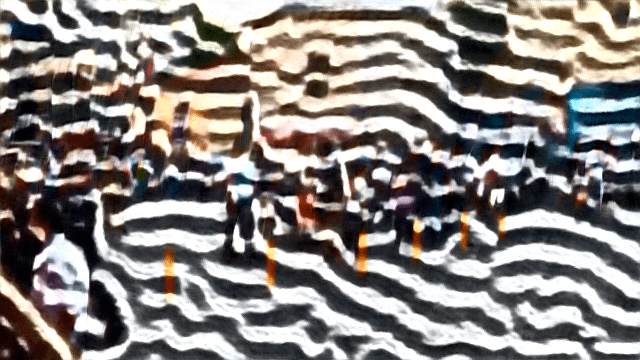} &
  \includegraphics[width=0.23\textwidth]{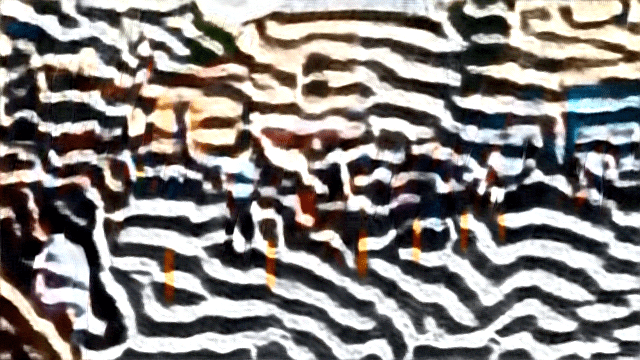}\\
  \rotatebox{90}{\textbf{Restormer}} & \rotatebox{90}{\tiny \phantom{substi}LCTC with }& \rotatebox{90}{11$\times$11 + 3$\times$3} &
  \includegraphics[width=0.23\textwidth]{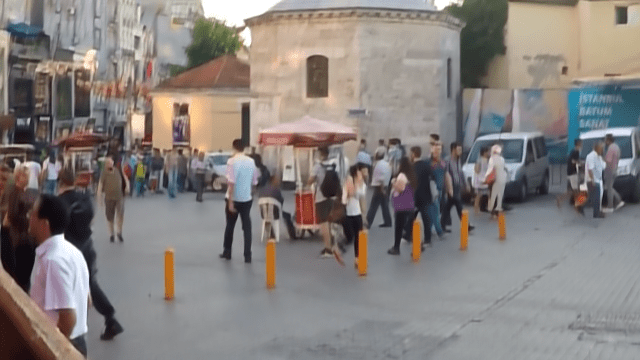} &
  \includegraphics[width=0.23\textwidth]{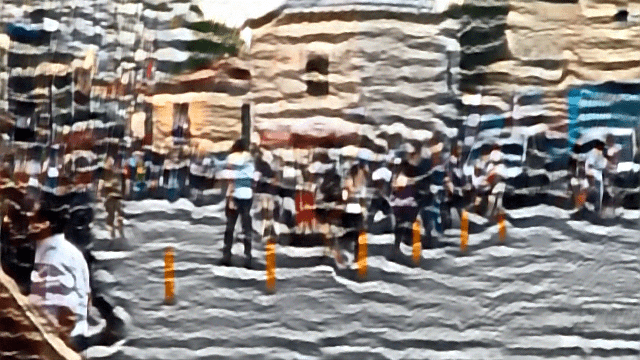} &
  \includegraphics[width=0.23\textwidth]{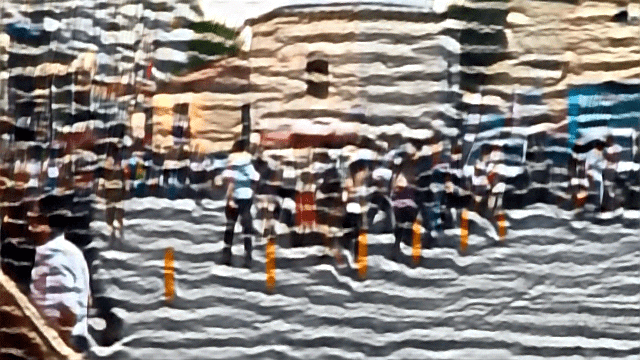} &
  \includegraphics[width=0.23\textwidth]{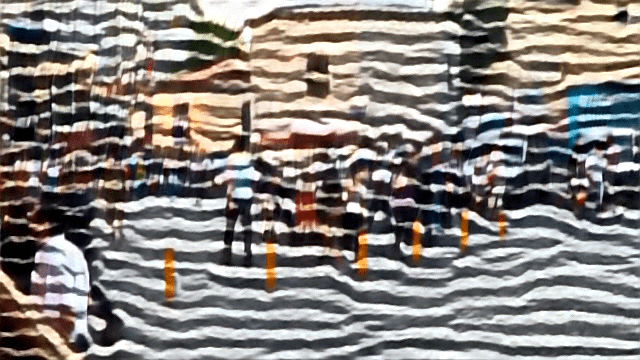}\\
    \rotatebox{90}{\textbf{NAFNet}} & \rotatebox{90}{\tiny with Pixel Shuffle} & &
  \includegraphics[width=0.23\textwidth]{eccv_2024/figures/ablation/restoration/NAFNet_no_attack_GOPR0384_11_00-000002.png} &
  \includegraphics[width=0.23\textwidth]{eccv_2024/figures/ablation/restoration/NAFNet_pgd_5_GOPR0384_11_00-000002.png} 
  &
  \includegraphics[width=0.23\textwidth]{eccv_2024/figures/ablation/restoration/NAFNet_pgd_10_GOPR0384_11_00-000002.png} 
  &
  \includegraphics[width=0.23\textwidth]{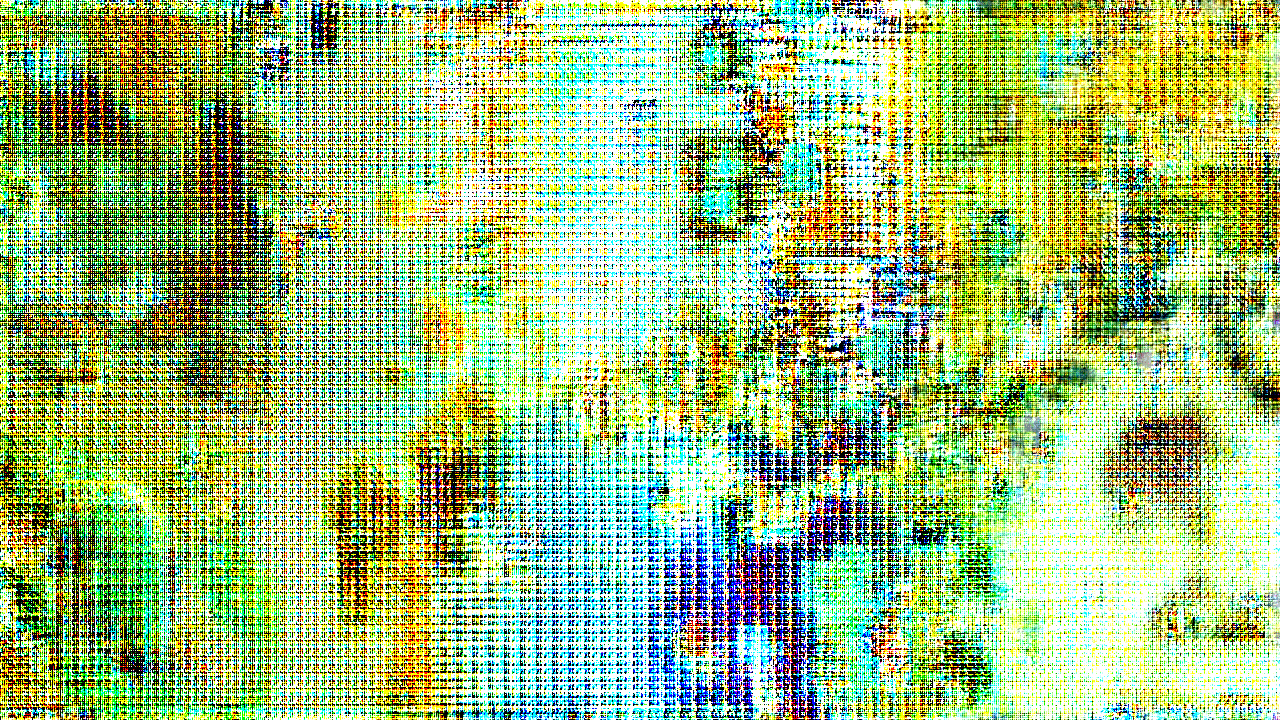}
  \\
  \rotatebox{90}{\textbf{NAFNet}} & \rotatebox{90}{\tiny with Transposed Conv} & \rotatebox{90}{\phantom{su}~~~3$\times$3} &
  \includegraphics[width=0.23\textwidth]{eccv_2024/figures/ablation/restoration/NAFNet_3_0_no_attack_GOPR0384_11_00-000002.png} &
  \includegraphics[width=0.23\textwidth]{eccv_2024/figures/ablation/restoration/NAFNet_3_0_pgd_5_GOPR0384_11_00-000002.png} &
  \includegraphics[width=0.23\textwidth]{eccv_2024/figures/ablation/restoration/NAFNet_3_0_pgd_10_GOPR0384_11_00-000002.png} &
  \includegraphics[width=0.23\textwidth]{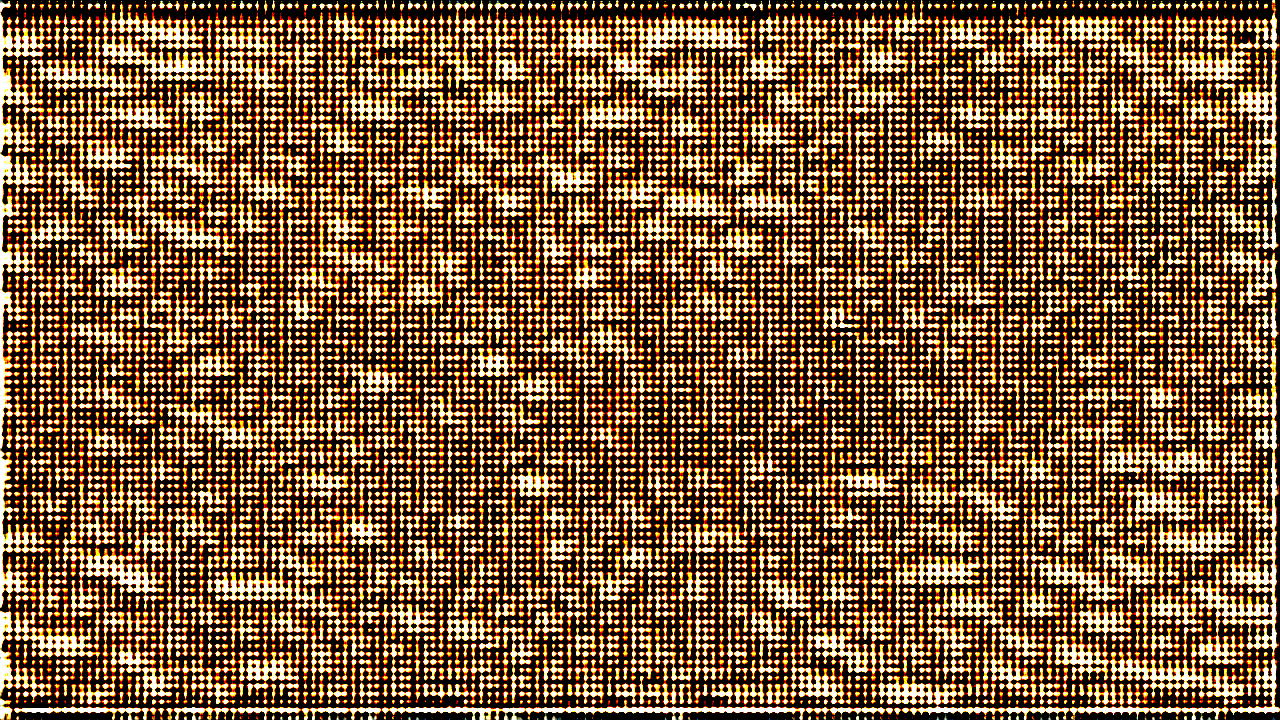}\\
  \rotatebox{90}{\textbf{NAFNet}} & \rotatebox{90}{\tiny \phantom{substi}LCTC with }& \rotatebox{90}{~~7$\times$7 + 3$\times$3} &
  \includegraphics[width=0.23\textwidth]{eccv_2024/figures/ablation/restoration/NAFNet_7_3_no_attack_GOPR0384_11_00-000002.png} &
  \includegraphics[width=0.23\textwidth]{eccv_2024/figures/ablation/restoration/NAFNet_7_3_pgd_5_GOPR0384_11_00-000002.png} &
  \includegraphics[width=0.23\textwidth]{eccv_2024/figures/ablation/restoration/NAFNet_7_3_pgd_10_GOPR0384_11_00-000002.png} &
  \includegraphics[width=0.23\textwidth]{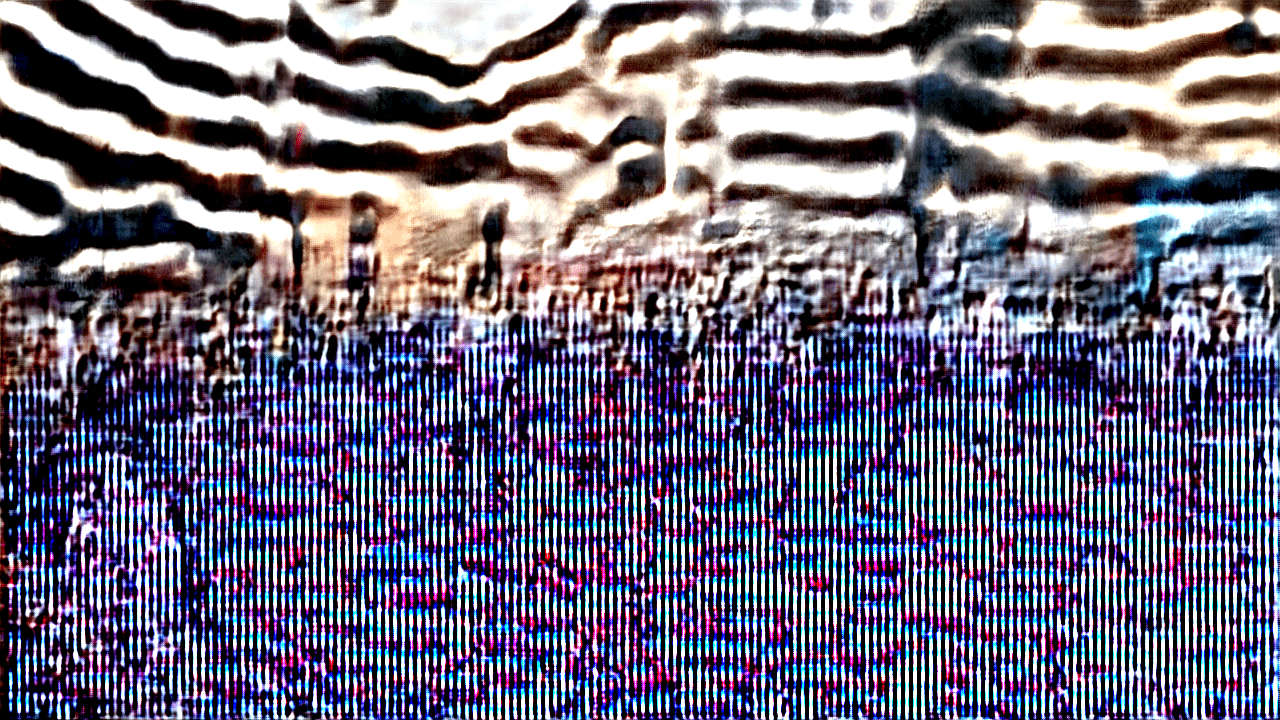}\\
  \rotatebox{90}{\textbf{NAFNet}} & \rotatebox{90}{\tiny \phantom{substi}LCTC with }& \rotatebox{90}{11$\times$11 + 3$\times$3} &
  \includegraphics[width=0.23\textwidth]{eccv_2024/figures/ablation/restoration/NAFNet_11_3_no_attack_GOPR0384_11_00-000002.png} &
  \includegraphics[width=0.23\textwidth]{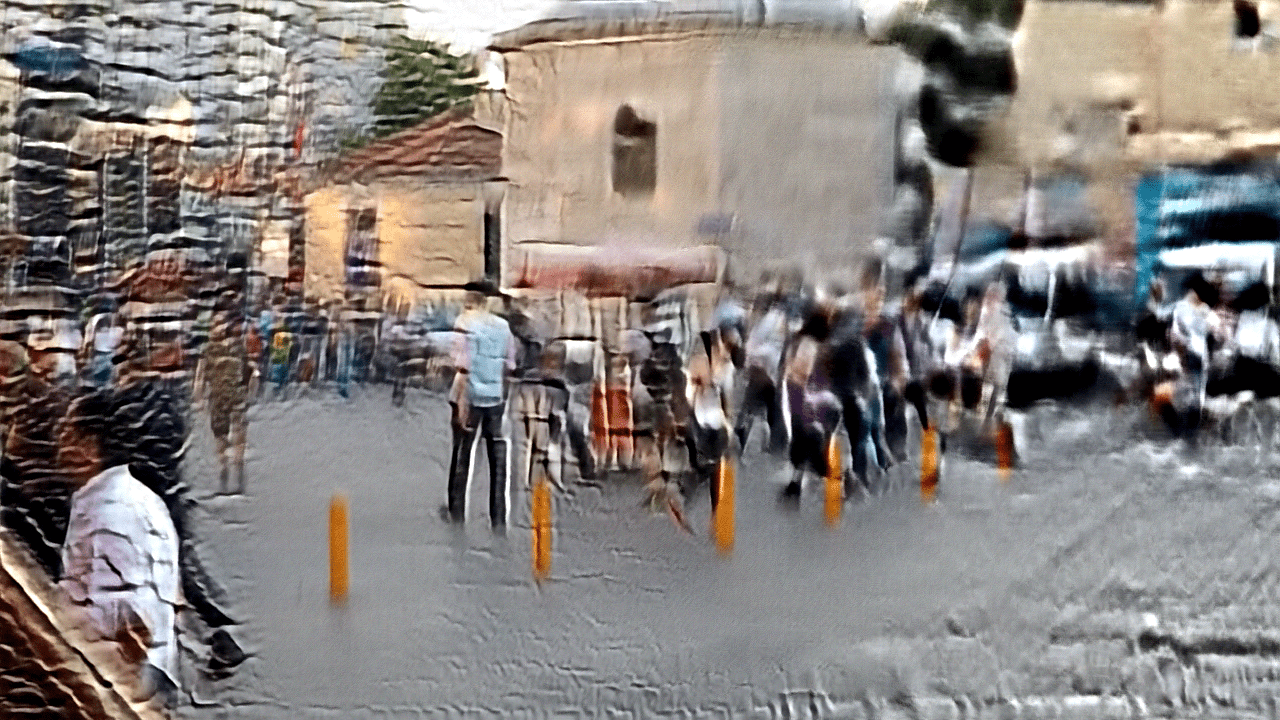} &
  \includegraphics[width=0.23\textwidth]{eccv_2024/figures/ablation/restoration/NAFNet_11_3_pgd_10_GOPR0384_11_00-000002.png} &
  \includegraphics[width=0.23\textwidth]{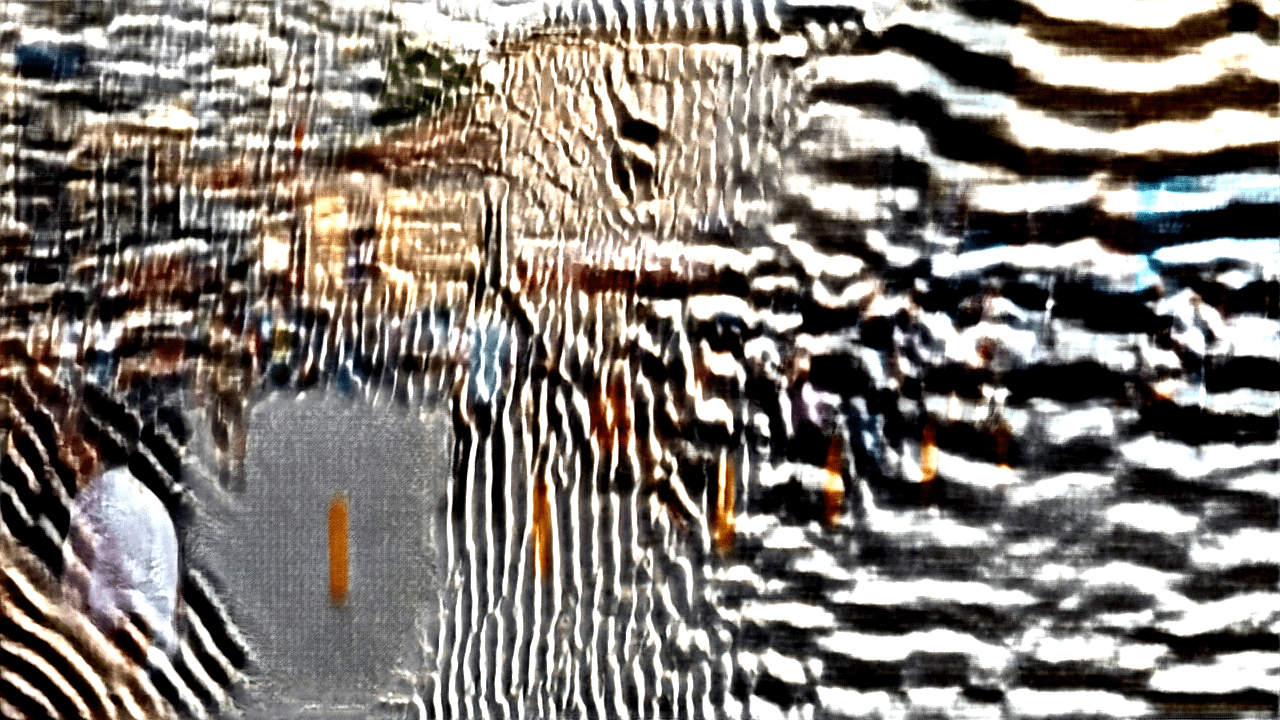}\\
\end{tabular}
\caption{Comparing images reconstructed by all models after \textbf{PGD attack} on variants of Upsampling.}
\label{fig:reconstruction_pgd_attack}
\end{figure*}

\begin{figure*}[htb]
    \centering 
\scalebox{0.92}{
   \begin{tabular}{@{}c@{\hspace{0.3cm}}c@{\hspace{0.3cm}}c@{\hspace{0.1cm}}c@{\hspace{0.1cm}}c@{\hspace{0.1cm}}c@{\hspace{0.1cm}}c@{}}
    \multicolumn{3}{c}{MODEL} & NO ATTACK & 5 iterations & 10 iterations & 20 iterations\\
  
  \rotatebox{90}{\textbf{Restormer}} & \rotatebox{90}{\tiny with Pixel Shuffle} & &
  \includegraphics[width=0.23\textwidth]{eccv_2024/figures/ablation/restoration/Restormer_no_attack_GOPR0384_11_00-000002.png} &
  \includegraphics[width=0.23\textwidth]{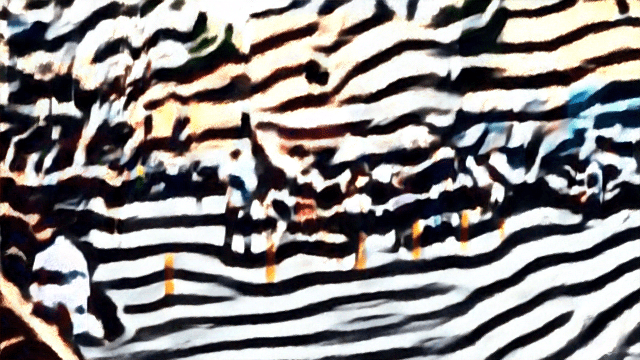} &
  \includegraphics[width=0.23\textwidth]{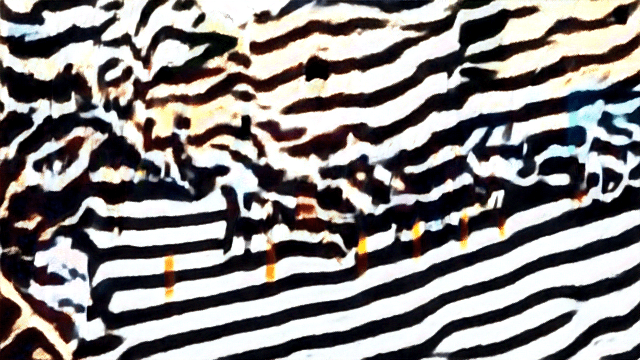} &
  \includegraphics[width=0.23\textwidth]{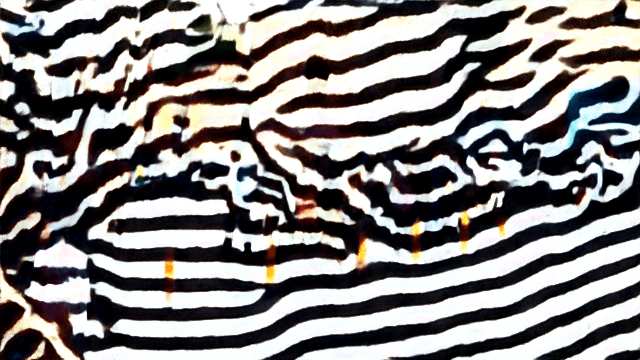}\\

  \rotatebox{90}{\textbf{Restormer}} & \rotatebox{90}{\tiny with Transposed Conv} & \rotatebox{90}{\phantom{su}~~~3$\times$3} &
  \includegraphics[width=0.23\textwidth]{eccv_2024/figures/ablation/restoration/Restormer_3_0_no_attack_GOPR0384_11_00-000002.png} &
  \includegraphics[width=0.23\textwidth]{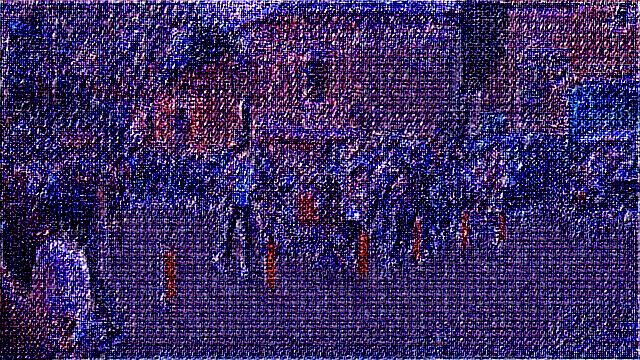} &
  \includegraphics[width=0.23\textwidth]{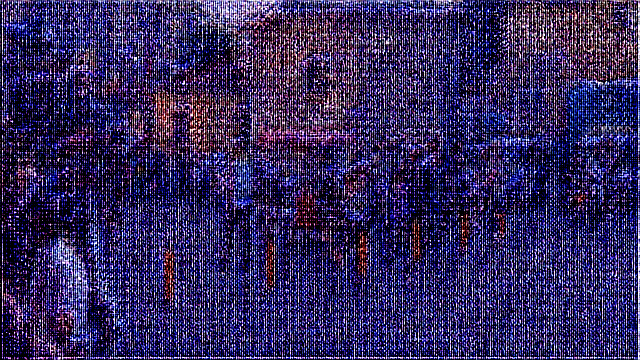} &
  \includegraphics[width=0.23\textwidth]{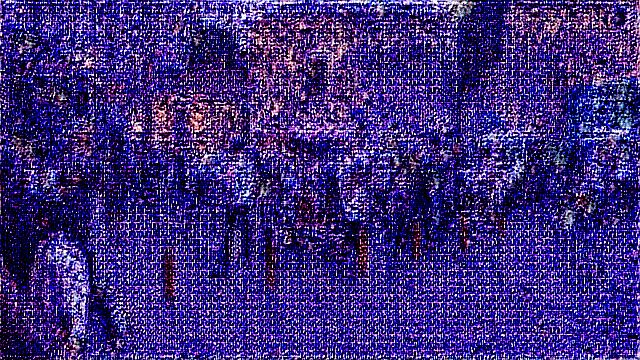}\\

  \rotatebox{90}{\textbf{Restormer}} & \rotatebox{90}{\tiny \phantom{substi}LCTC with }& \rotatebox{90}{~~7$\times$7 + 3$\times$3} &
  \includegraphics[width=0.23\textwidth]{eccv_2024/figures/ablation/restoration/Restormer_7_3_no_attack_GOPR0384_11_00-000002.png} &
  \includegraphics[width=0.23\textwidth]{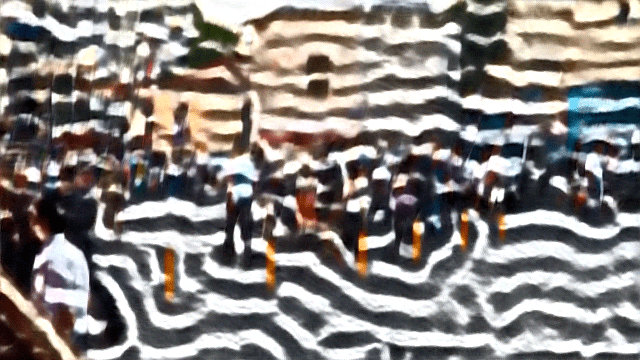} &
  \includegraphics[width=0.23\textwidth]{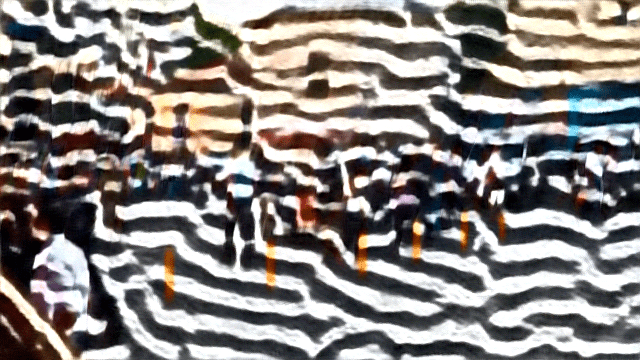} &
  \includegraphics[width=0.23\textwidth]{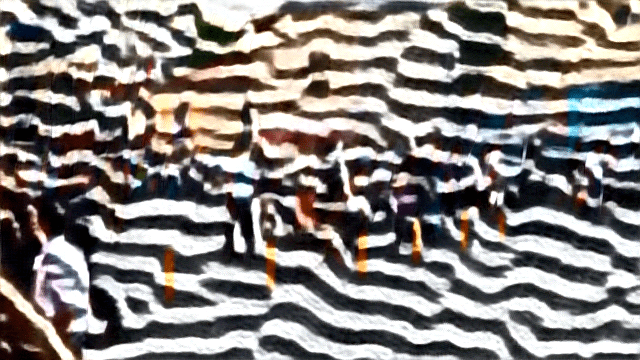}\\

  \rotatebox{90}{\textbf{Restormer}} & \rotatebox{90}{\tiny \phantom{substi}LCTC with }& \rotatebox{90}{11$\times$11 + 3$\times$3} &
  \includegraphics[width=0.23\textwidth]{eccv_2024/figures/ablation/restoration/Restormer_11_3_no_attack_GOPR0384_11_00-000002.png} &
  \includegraphics[width=0.23\textwidth]{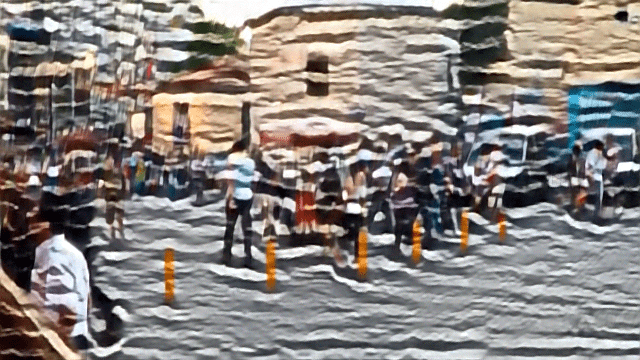} &
  \includegraphics[width=0.23\textwidth]{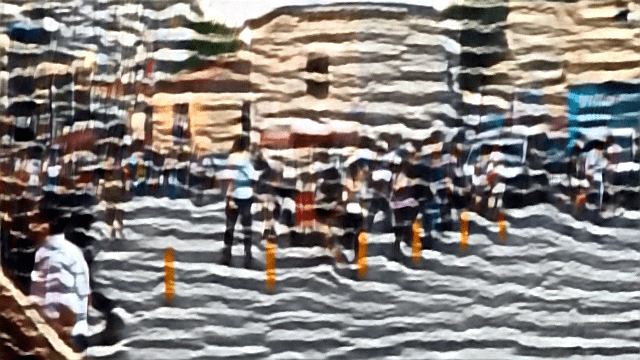} &
  \includegraphics[width=0.23\textwidth]{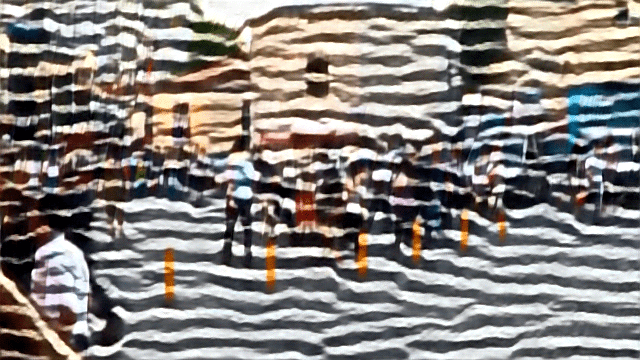}\\

    \rotatebox{90}{\textbf{NAFNet}} & \rotatebox{90}{\tiny with Pixel Shuffle} & &
  \includegraphics[width=0.23\textwidth]{eccv_2024/figures/ablation/restoration/NAFNet_no_attack_GOPR0384_11_00-000002.png} &
  \includegraphics[width=0.23\textwidth]{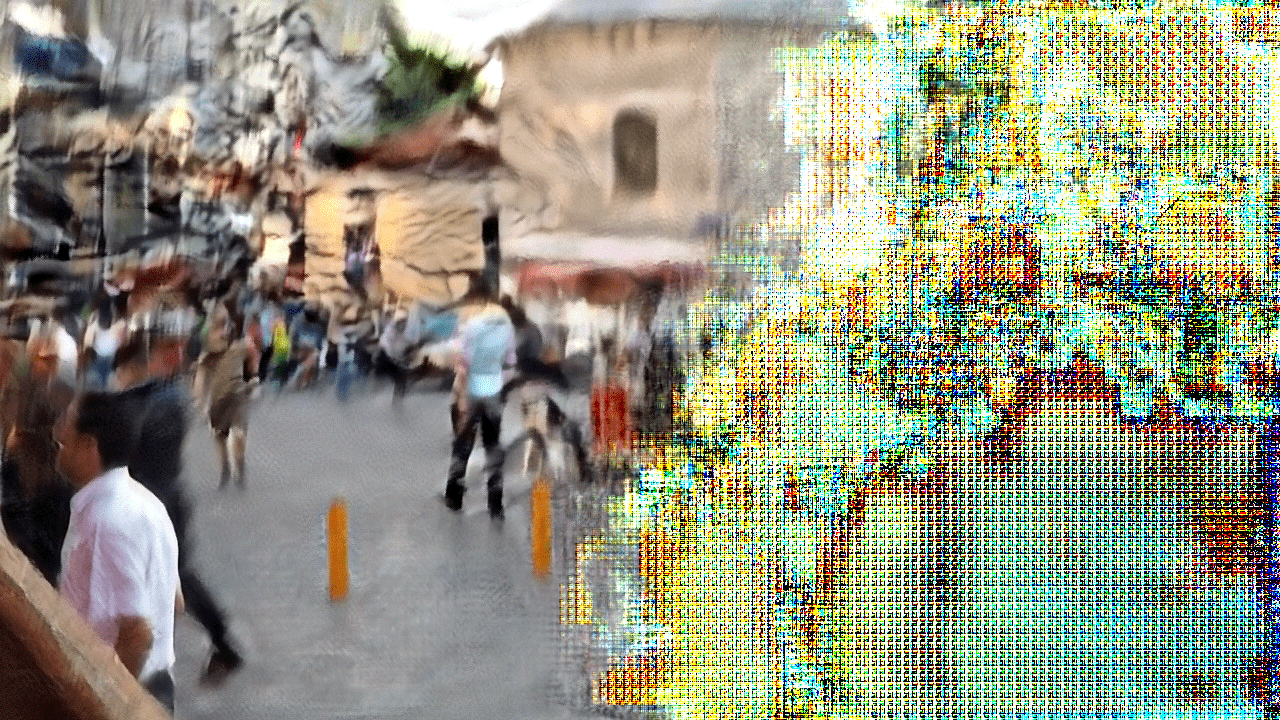} &
  \includegraphics[width=0.23\textwidth]{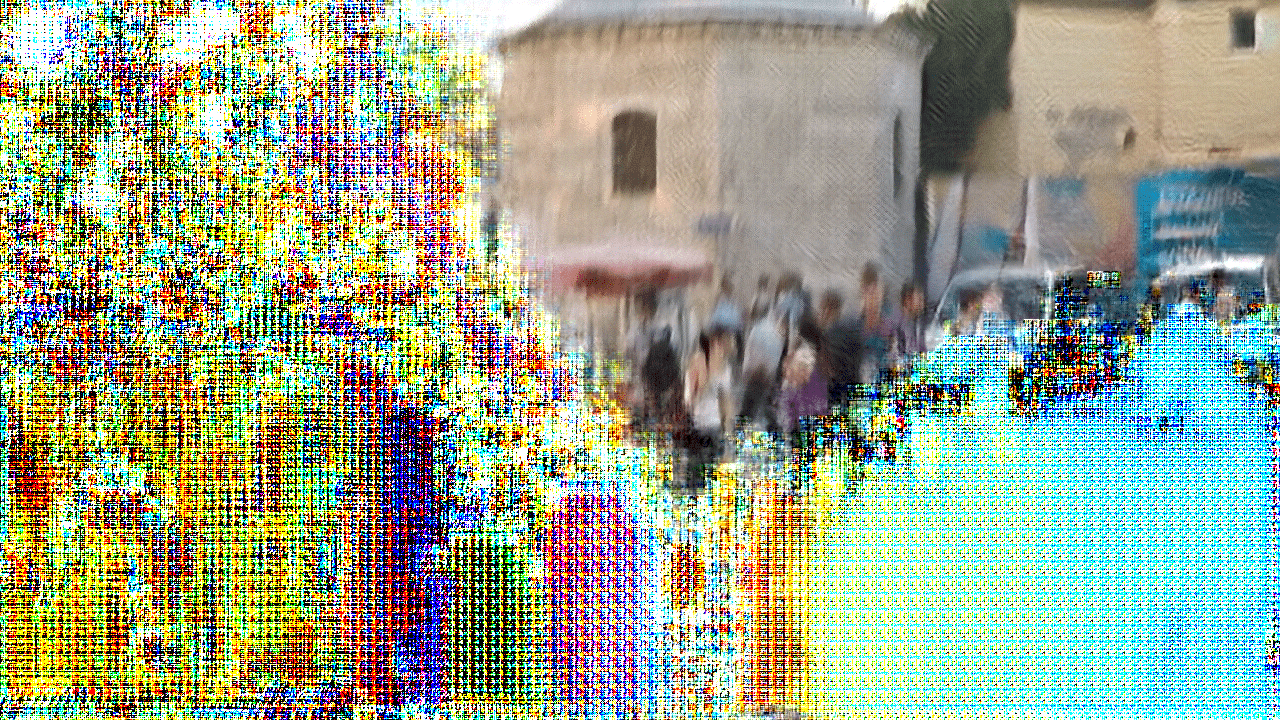} &
  \includegraphics[width=0.23\textwidth]{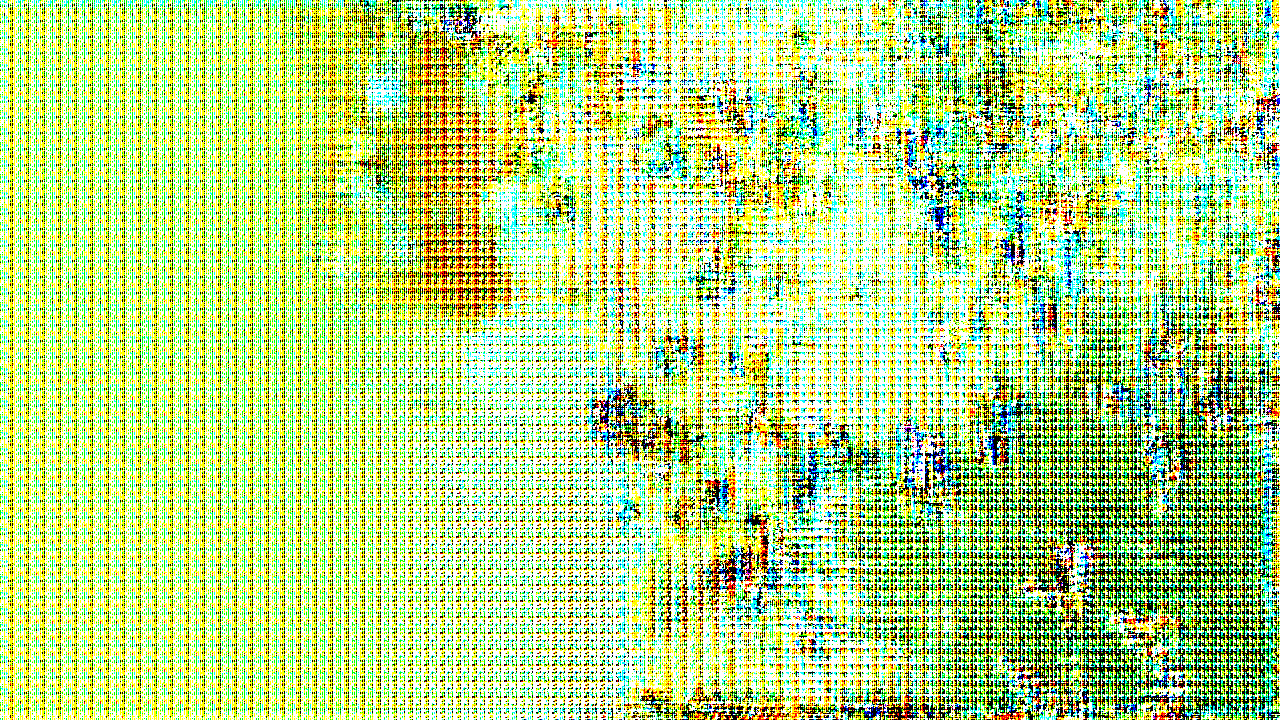}\\

  \rotatebox{90}{\textbf{NAFNet}} & \rotatebox{90}{\tiny with Transposed Conv} & \rotatebox{90}{\phantom{su}~~~3$\times$3} &
  \includegraphics[width=0.23\textwidth]{eccv_2024/figures/ablation/restoration/NAFNet_3_0_no_attack_GOPR0384_11_00-000002.png} &
  \includegraphics[width=0.23\textwidth]{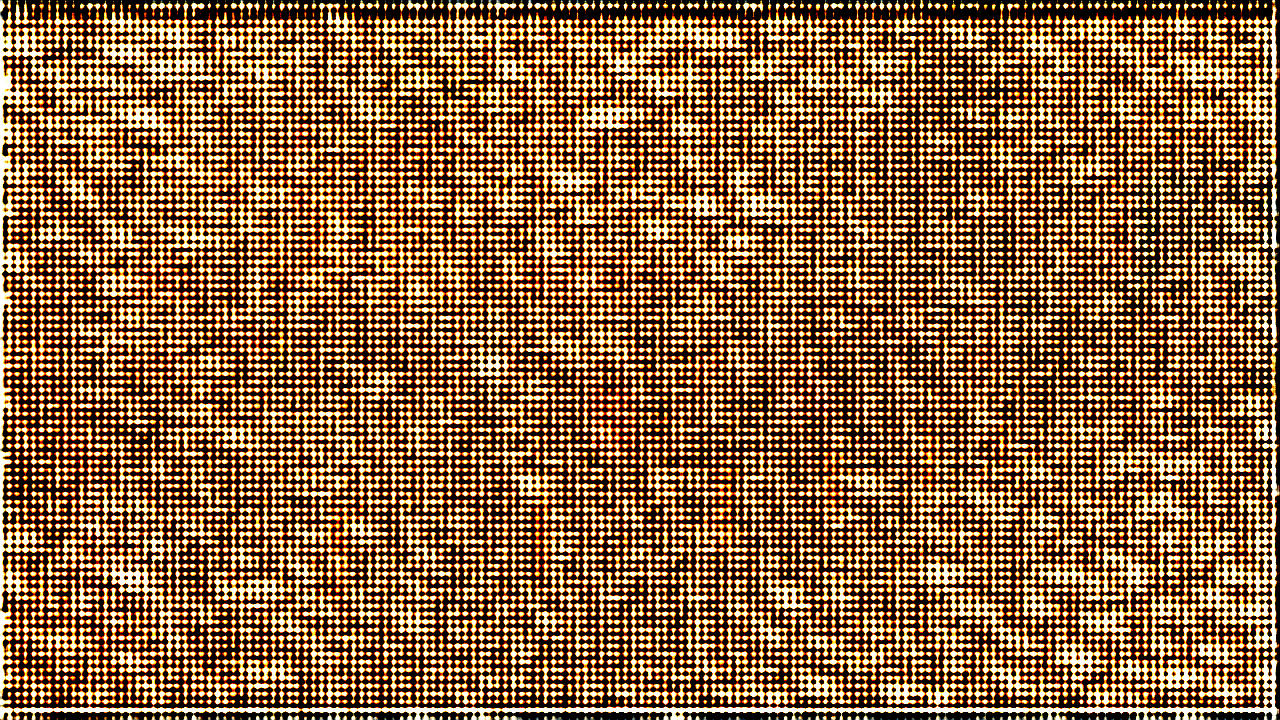} &
  \includegraphics[width=0.23\textwidth]{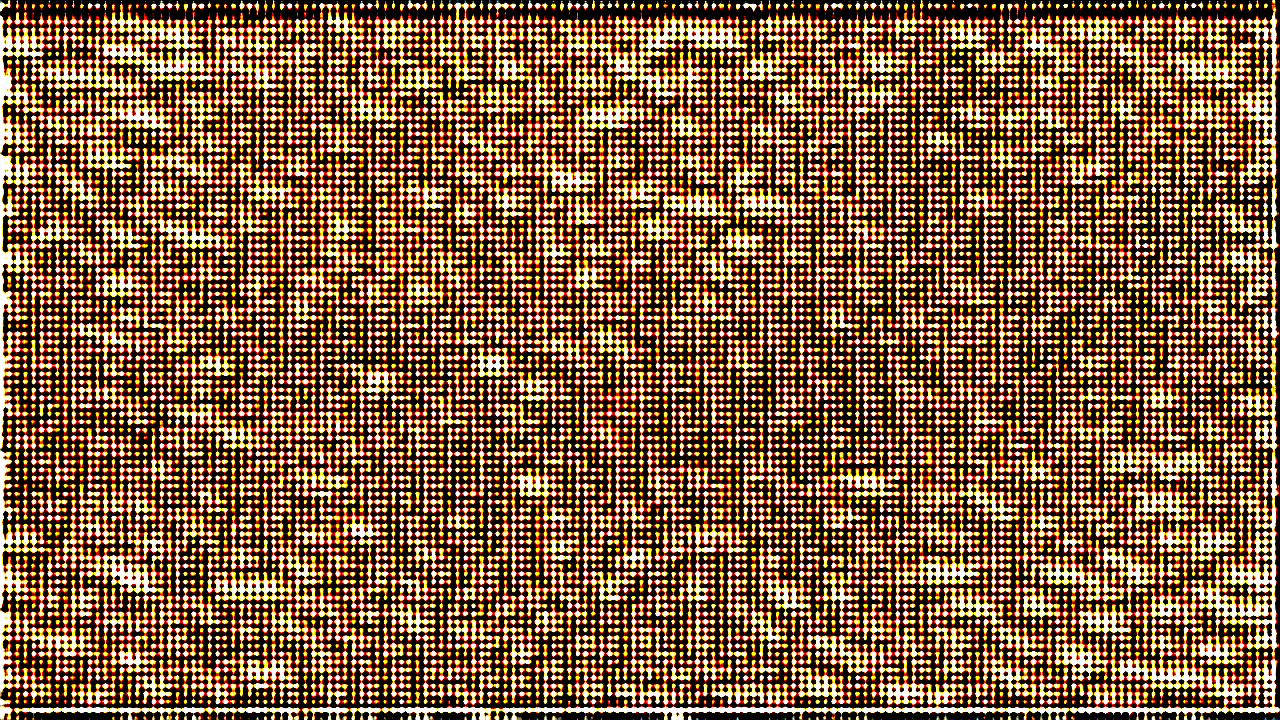} &
  \includegraphics[width=0.23\textwidth]{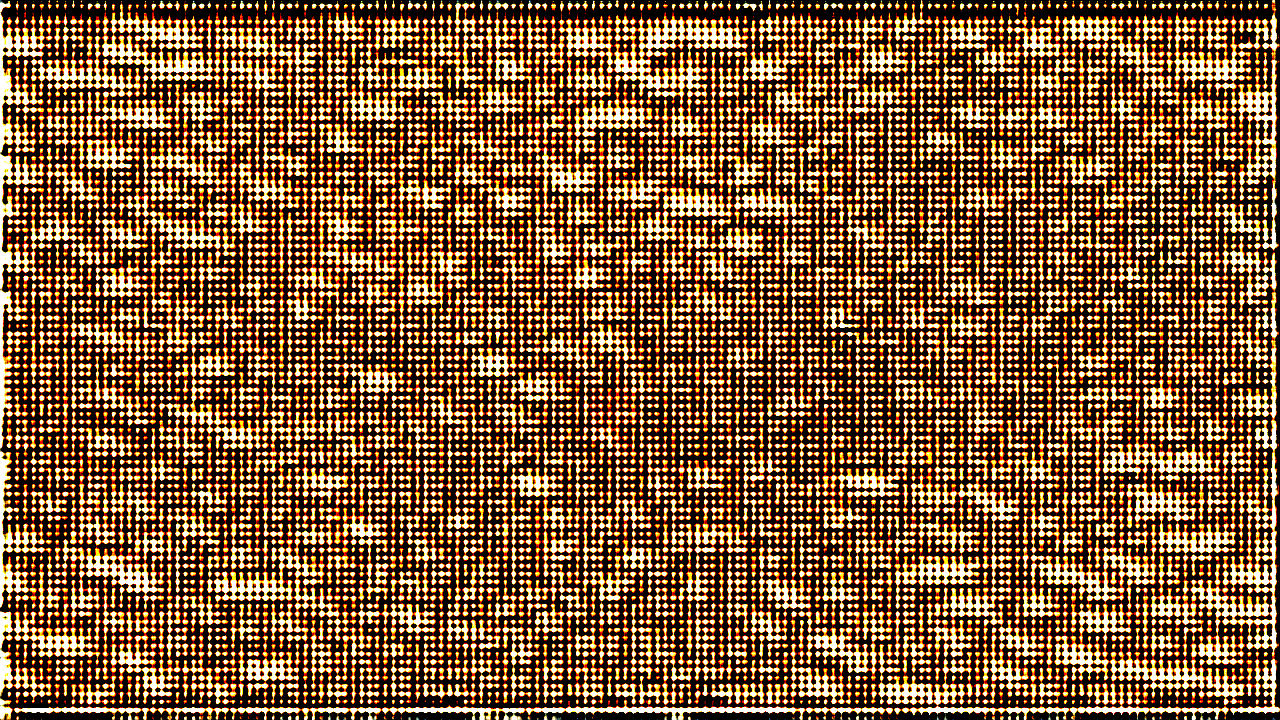}\\

  \rotatebox{90}{\textbf{NAFNet}} & \rotatebox{90}{\tiny \phantom{substi}LCTC with }& \rotatebox{90}{~~7$\times$7 + 3$\times$3} &
  \includegraphics[width=0.23\textwidth]{eccv_2024/figures/ablation/restoration/NAFNet_7_3_no_attack_GOPR0384_11_00-000002.png} &
  \includegraphics[width=0.23\textwidth]{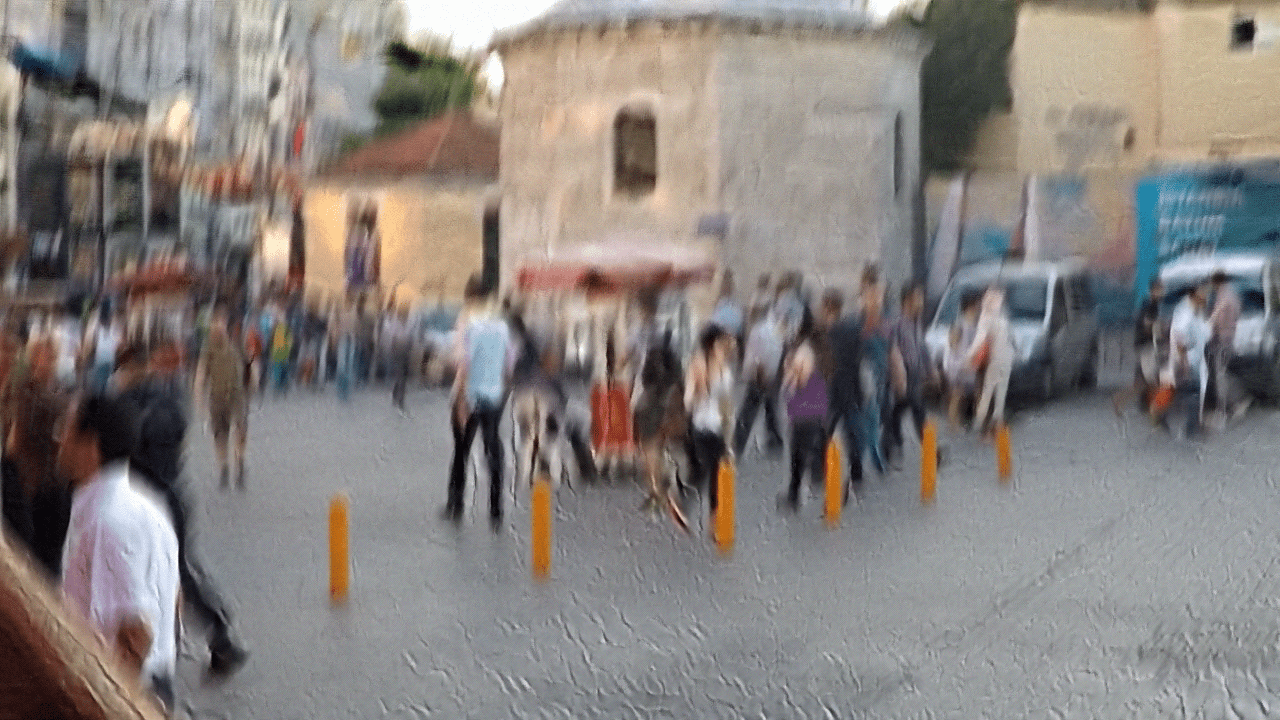} &
  \includegraphics[width=0.23\textwidth]{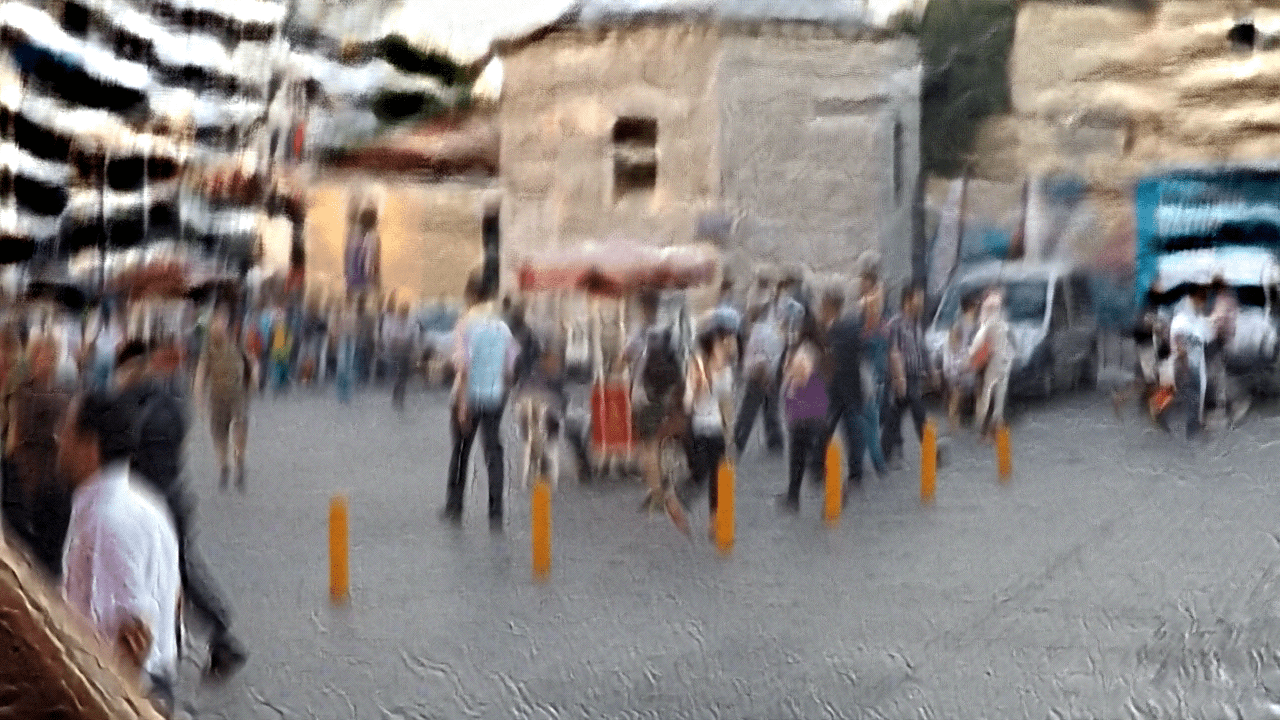} &
  \includegraphics[width=0.23\textwidth]{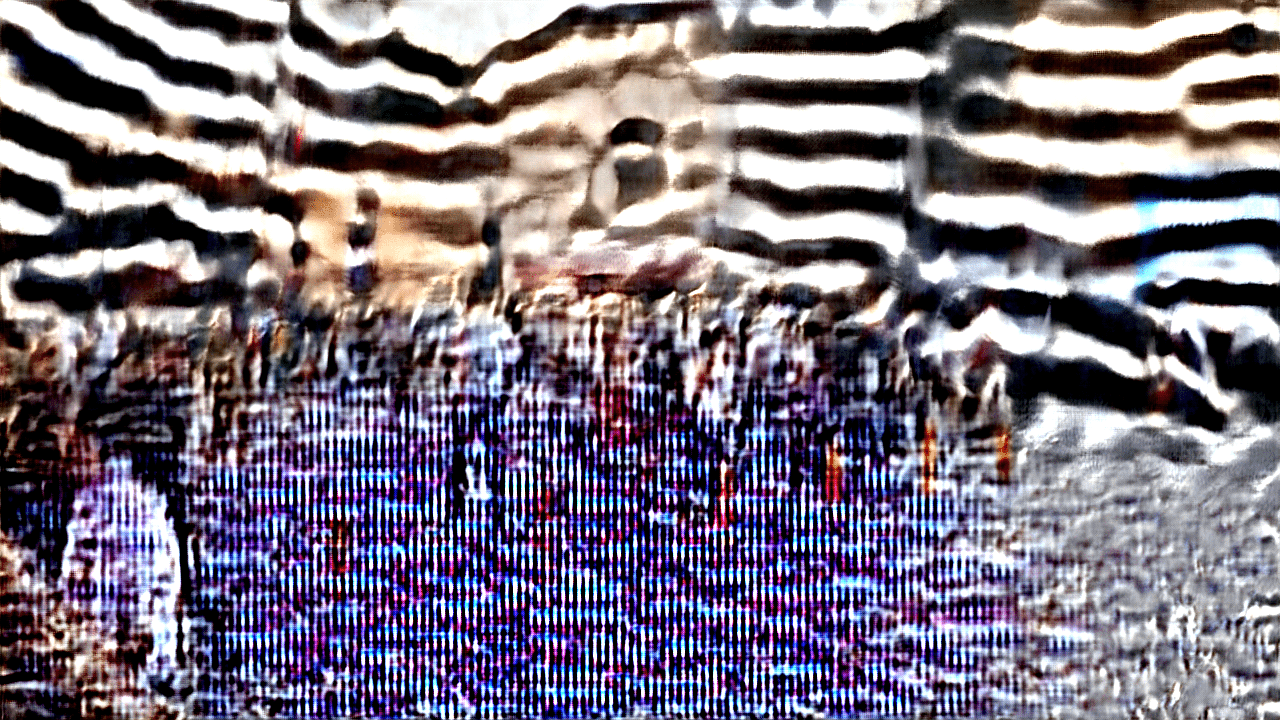}\\

  \rotatebox{90}{\textbf{NAFNet}} & \rotatebox{90}{\tiny \phantom{substi}LCTC with }& \rotatebox{90}{11$\times$11 + 3$\times$3} &
  \includegraphics[width=0.23\textwidth]{eccv_2024/figures/ablation/restoration/NAFNet_11_3_no_attack_GOPR0384_11_00-000002.png} &
  \includegraphics[width=0.23\textwidth]{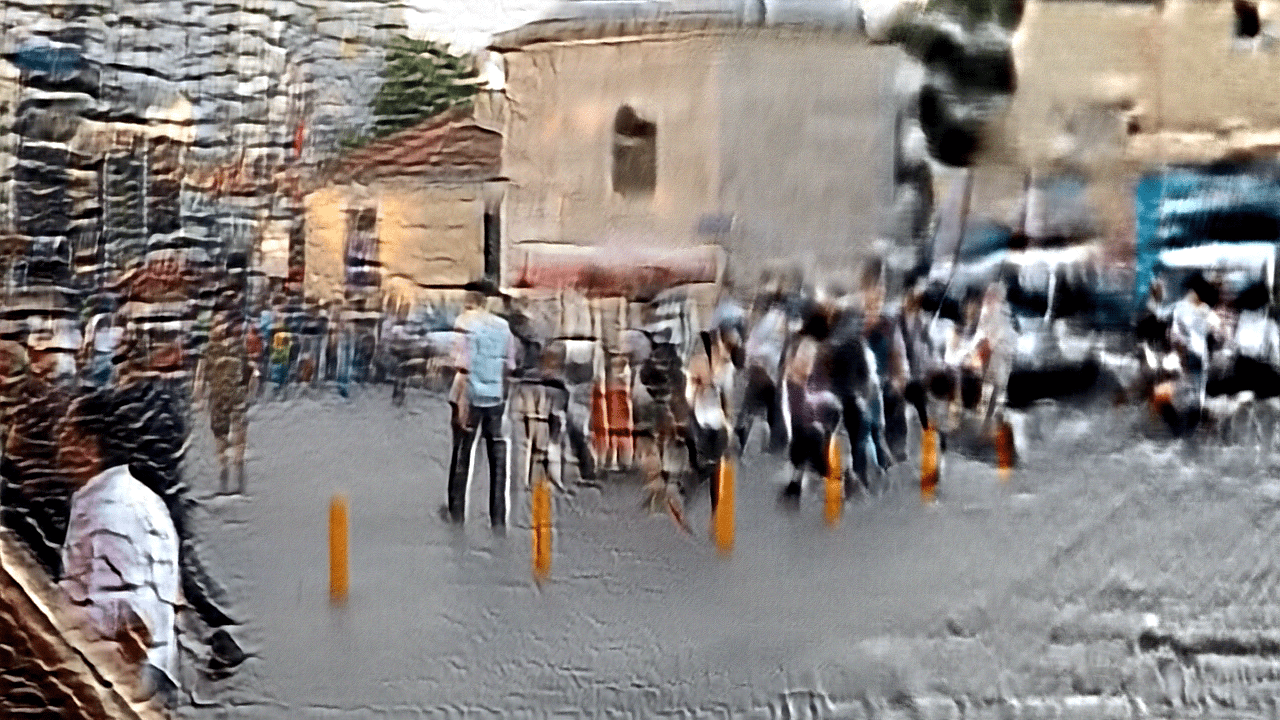} &
  \includegraphics[width=0.23\textwidth]{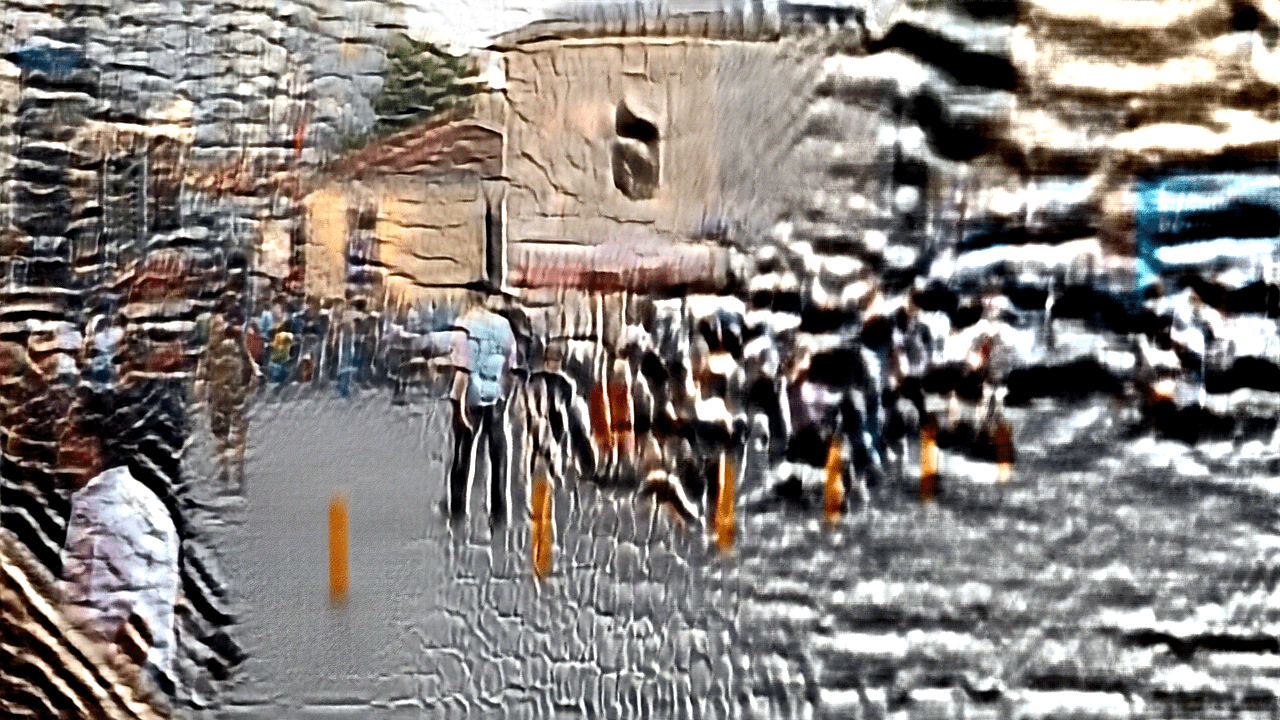} &
  \includegraphics[width=0.23\textwidth]{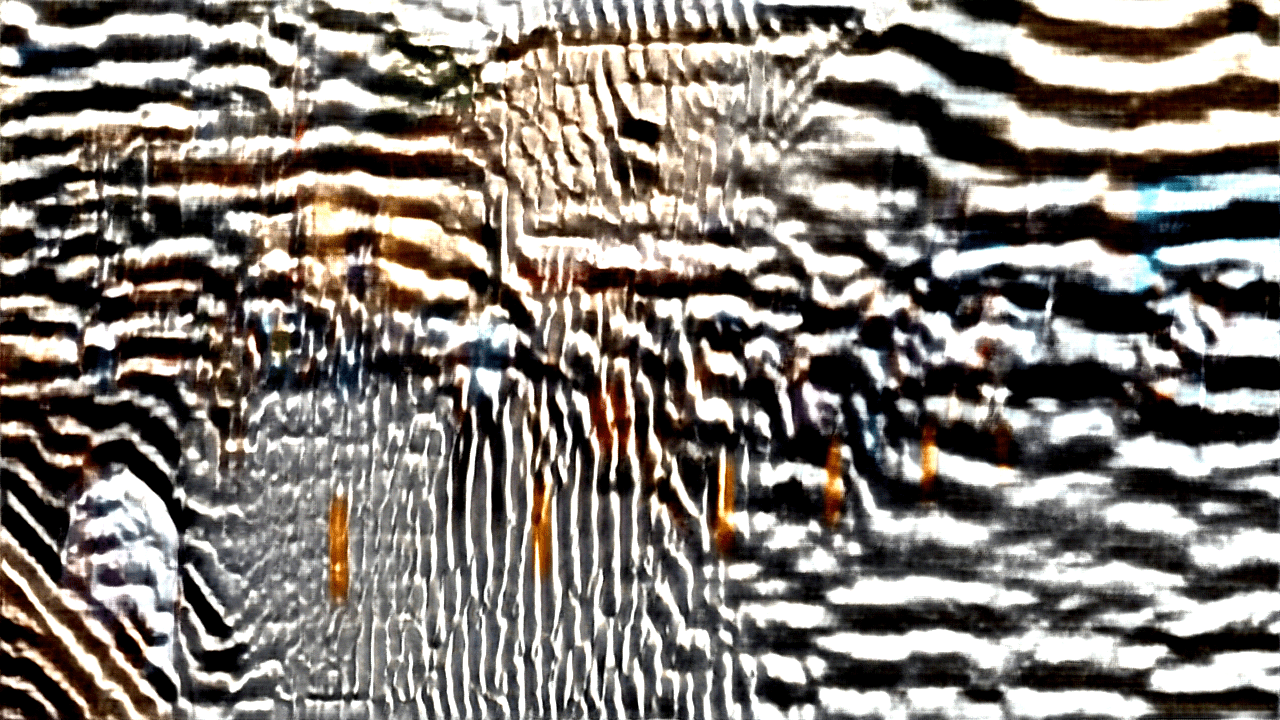}\\

\end{tabular}
}
\caption{Comparing images reconstructed by the considered variants of the SotA models after CosPGD attack \cite{agnihotri2023cospgd}. We observe that the originally proposed Restormer and NAFNet architectures that use Pixel Shuffle for upsampling perform considerably well under no adversarial attack but even a small perturbation of $\epsilon$=$\frac{8}{255}$ causes ringing and other spectral artifacts to occur in the deblurred images to the extent that the images are unrecognizable. However, on replacing the Pixel Shuffle operation in these architectures with a Transposed Convolution operation with a large kernel (11$\times$11+3$\times$3), we observe a significant reduction in the spectral artifacts in the images restored under adversarial attack while the image restored under no attack are very comparable to those restored by the original architectures.}
\label{fig:reconstruction_cospgd_attack}
\end{figure*}

\Cref{fig:reconstruction_pgd_attack} shows reconstruction under PGD attack for Restormer~\cite{zamir2022restormer} and NAFNet~\cite{chen2022simple}.
\Cref{fig:reconstruction_cospgd_attack} shows reconstruction under CosPGD attack for Restormer~\cite{zamir2022restormer} and NAFNet~\cite{chen2022simple}.

\subsection{Visualizing Kernel Weights}
\label{subsec:appendix:image_restoration:kernel_weights}
\begin{figure}[h]
    \vspace{-2em}
    \centering
    \includegraphics[width=\columnwidth]{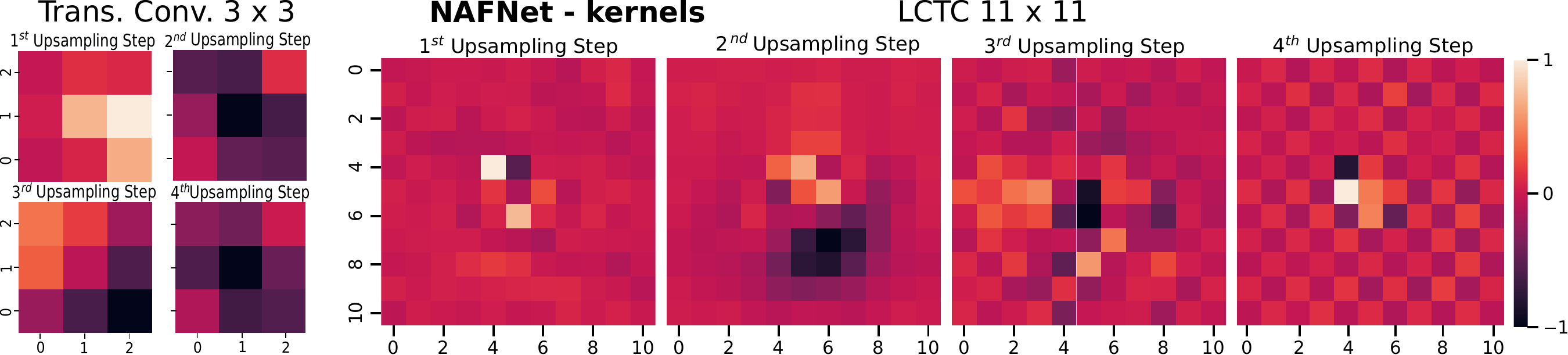}
    \vspace{-1em}
    \caption{Normalized kernel weights from a random channel each for the models from \Cref{fig:reconstruction_pgd_attack_main}.}
    \label{fig:rebuttal_visualizing_weights}
    \vspace{-1em}
\end{figure}

An increase in kernel size leads to an increase in context and since the context is increased, the effect of uneven contributions of pixels is negated leading to reduced spectral artifacts.
This can be seen in \Cref{fig:rebuttal_visualizing_weights}.
Here we observe that the weights for 3×3 are high at the edges, causing the described grid effect, whereas for 11×11 kernels there is a smooth fading towards the border of kernels, negating this effect.

\subsection{Real World and Out-Of-Distribution (ODD) Generalization}
\label{subsec:appendix:image_restoration:OOD_performance}
\begin{table}[t]
    \centering

    \caption{Performance of different upsampling methods in NAFNet in real-world ODD setting by applying 2D common corruption\cite{commoncorruptions} (severity=3) on GoPro dataset. We use all common corruptions from \cite{commoncorruptions} GitHub repository except weather conditions (ideally these should happen before the motion blurring) and blurring (since the images are already motion blurred). Here ``Mean'' is performance over all the considered corruptions:}
    \scalebox{1.0}{
    \begin{tabular}{@{}c@{\hspace{0.2cm}}c@{\hspace{0.2cm}}c@{\hspace{0.4cm}}|c@{\hspace{0.2cm}}c@{\hspace{0.4cm}}|c@{\hspace{0.2cm}}c@{}}
    \toprule
    \multirow{3}{*}{\textbf{Common Corruption}} & \multicolumn{6}{c}{\textbf{Upsampling Method}} \\
    & \multicolumn{2}{c}{Pixel Shuffle} & \multicolumn{2}{c}{Trans.~Conv.~3$\times$3} & \multicolumn{2}{c}{\textbf{LCTC 11$\times$11+3$\times$3}} \\
    & { PSNR } & { SSIM } & { PSNR } & { SSIM } & { PSNR } & { SSIM } \\        
    \midrule

    Gaussian Noise & 4.8501 & 0.0104 & 8.7346 & 0.1014 &  13.6475 & 0.1523  \\
    Shot Noise & 4.8616 & 0.0127 & 8.9524 & 0.0984  &  13.2464 & 0.1564   \\
    Impulse Noise & 5.0154 & 0.0214 & 9.2451 & 0.1065  &  14.8425 & 0.187    \\
    Brightness & 32.3199 & 0.9576 &  30.676 & 0.9394  &   30.4098 & 0.9361  \\
    Contrast & 26.5941 & 0.7759 &  25.9743 & 0.7561  &  25.8733 & 0.7525   \\
    Elastic Transform & 17.944 & 0.6392 &  19.7686 & 0.703  &  19.7672 & 0.702 \\
    Pixelate & 4.4977 & 0.246 &  4.4999 & 0.246  &  4.4958 & 0.246  \\
    JPEG Compression & 25.2767 & 0.8095 &  25.1014 & 0.8032  &  25.3788 & 0.8104 \\
    Speckle Noise & 4.8287 & 0.0158 &  9.2336 & 0.1044   &  14.6622 & 0.2473  \\
    Saturate & 32.1969 & 0.958 &  30.5904 & 0.9399   &  30.3005 & 0.9365  \\
    \hline
    \textbf{Mean} & 15.8385 & 0.4447 &  17.2776 & 0.4798  &  \textbf{19.262} & \textbf{0.5127} \\
        
    \bottomrule
    \end{tabular}    
    }
    \label{tab:ood_nafnet}
    
\end{table}
Since LCTC leads to improved sampling that provides stability to feature maps learned by the network (not merely defense), inspired by observations from \cite{grabinski2022frequencylowcut}, we hypothesize that the trends on adversarial attacks should translate to Real-World noise. 
We show this in \Cref{tab:ood_nafnet} by applying 2D common corruptions (CC) (severity=3) from \cite{commoncorruptions} on images from the GoPro dataset and using NAFNet models from \Cref{fig:reconstruction_pgd_attack_main}. 
Since the task is deblurring, we consider all common corruptions but additional blurring and weather corruptions, as these would have to be captured before blurring.

\section{Additional Results Disparity Estimation}
\label{sec:appendix:disparity}
Following we report additional results for Disparity Estimation using STTR-light.
In \Cref{tbl:appendix:exp:disparity:sttr} we report the performance of STTR-light architecture on clean test images and under PGD attack.
Whereas in \Cref{fig:depth_estimation}, we present a visual comparison of depth estimation predictions by a vanilla STTR-light as proposed by \cite{sttr} and our proposed modification of \textcolor{BrickRed}{increasing the kernel size of the transposed convolution operation} in the ``feature extractor'' module of the architecture from 3$\times$3 to Large Context Transposed Convolutions with kernel sizes 7$\times$7+3$\times$3 and 11$\times$11+3$\times$3.

\subsection{Disparity Estimation Discussion}
\begin{table*}[ht]
\caption{Comparison of performance of STTR-light architecture with different sized kernels in transposed convolution for upscaling the feature maps in the feature extractor.}
\label{tbl:appendix:exp:disparity:sttr}
\centering
\scalebox{.75}{

\begin{tabular}{@{}p{5cm}cc|cc|cc|cc@{}}
\toprule
 \multirow{3}{5cm}{\textbf{Transposed Convolution Kernels}} & \multicolumn{2}{c}{\textbf{Test Accuracy}}  &  \multicolumn{6}{c}{PGD Attack}   \\
  & & & \multicolumn{2}{c}{3 Iterations} & \multicolumn{2}{c}{5 Iterations} & \multicolumn{2}{c}{10 Iterations} \\
     & epe\textcolor{green}{$\downarrow$} & 3px error\textcolor{green}{$\downarrow$} & epe\textcolor{green}{$\downarrow$} & 3px error\textcolor{green}{$\downarrow$}  & epe\textcolor{green}{$\downarrow$} & 3px error\textcolor{green}{$\downarrow$}  & epe\textcolor{green}{$\downarrow$} & 3px error\textcolor{green}{$\downarrow$} \\
\midrule
STTR-light \cite{sttr} reported   &   0.5  &    1.54 &  \\
\midrule
   3$\times$3 \cite{sttr} reproduced  & 0.4927 & 1.54 & 4.05 & 18.46 & 4.07 & 18.59 & 4.08 & 18.6   \\
   LCTC: 7$\times$7 (Ours)   &   0.487 &  1.52  & 4.26 & 19.09 & 4.289 & 19.21 & 4.294 & 19.23 \\
   LCTC: 7$\times$7 + 3$\times$3 (Ours)  &   \textbf{0.4788} & \textbf{1.50} & 4.02 & 18.3 & 4.0474 & 18.43 & 4.05 & 18.45   \\
   LCTC: 9$\times$9 (Ours) & 0.4983 & \textbf{1.50} & 4.36 & \textbf{18.02} & 4.386 & \textbf{18.14} & 4.39 & \textbf{18.16} \\
   LCTC: 11$\times$11 +3$\times$3 (Ours)  &   0.5124 & 1.57 & \textbf{ 4.004} & 18.29 & \textbf{4.028} & 18.42 & \textbf{4.032} & 18.44 \\

\bottomrule

\end{tabular}
}
\end{table*}

\begin{figure*}    
    \centering
    \includegraphics[width=1.0\textwidth]{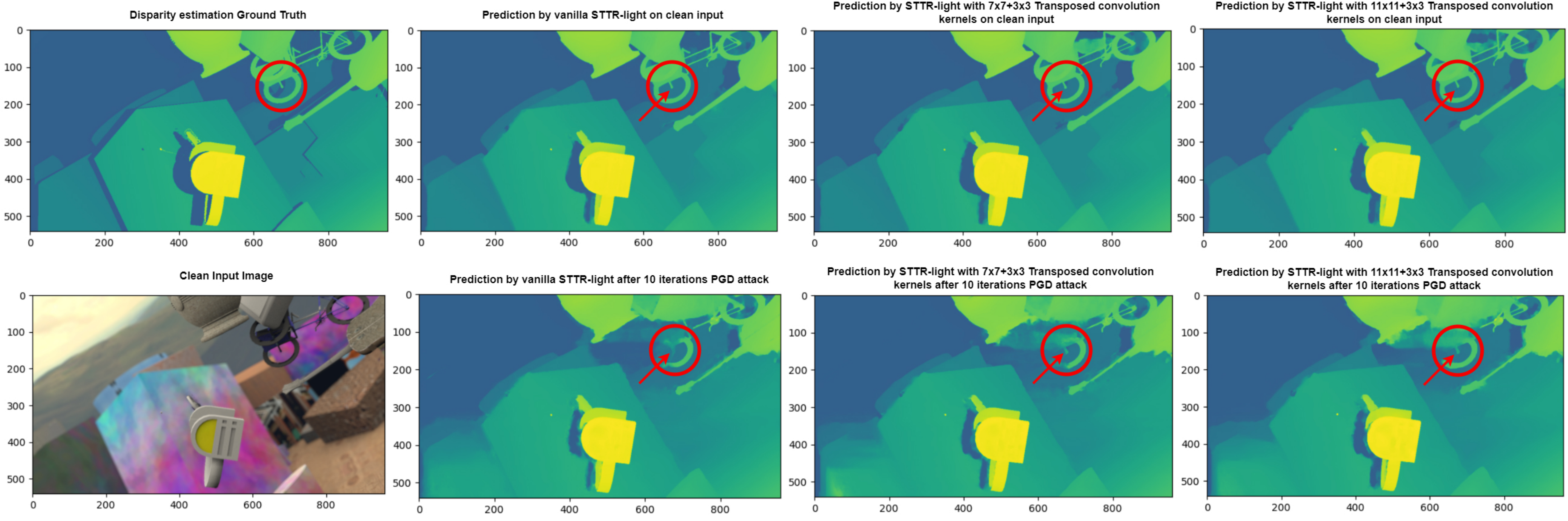}
    \caption{Visual comparison of Disparity Estimation predictions by a vanilla STTR-light as proposed by \cite{sttr} and our proposed modification of \textcolor{BrickRed}{increasing the kernel size of the transposed convolution operation} in the ``feature extractor'' module of the architecture from 3$\times$3 to LCTC with 7$\times$7+3$\times$3 and 11$\times$11+3$\times$3 sized kernels. As shown by the region in the \textcolor{red}{red circle}, both vanilla architecture and the architecture with our proposed change perform well compared to the ground truth on clean images. However, under 10 iteration PGD adversarial attack, we observe small protrusion's depth(shown by the \textcolor{red}{red arrow}) is incorrectly estimated by the vanilla architecture, however, the architectures with LCTC preserve the prediction of the disparity.}
    \label{fig:depth_estimation}
\end{figure*}

\label{subsubsec:appendix:exp:disparity}
In \Cref{fig:depth_estimation} as shown by the region in the \textcolor{red}{red circle}, both vanilla architecture and the architecture with our proposed change perform well compared to the ground truth on clean images. 
However, under a 10 iteration PGD adversarial attack, we observe small protrusion's depth(shown by the \textcolor{red}{red arrow}) is incorrectly estimated by the vanilla architecture.
The architecture with 7$\times$7+3$\times$3 and 11$\times$11+3$\times$3 transposed convolution kernels preserves the prediction of the disparity.

Additionally from \Cref{tbl:appendix:exp:disparity:sttr}, we observe the significance of the parallel 3$\times$3 small kernel with the large 7$\times$7 and 11$\times$11 kernels. 
The stability of the performance of the large kernels without the small parallel kernel compared to the baseline is better. However, the stability of performance when only using larger kernels compared to larger kernels with small parallel kernels is marginally worse.


\section{Nomenclature: What are \textit{Large Context Transposed Convolutions}?}
\label{sec:appendix:nomenclature}

In \Cref{sec:method} we introduce the term \textbf{``Large Context Transposed Convolutions (LCTC)''}. 
In this work, we use this to describe the Transposed Convolution layers in the decoder with large kernel sizes and thus a large spatial context.
However, terms like ``large'' are subjective, this in the following we discuss our interpretation of a ``large'' kernel size.

Most previous works use kernel sizes of 2$\times$2 or 3$\times$3 for any convolution operation, be it for downsampling\cite{resnet,vgg} or be it for upsampling\cite{unet}.
\cite{convnext} introduced performing downsampling using convolution operations with a large kernel size which in their case was 7$\times$7.
This ``larger'' kernel size for downsampling was further extended by other works like \cite{replk,SegNeXt} to 31$\times$31 and even up to 51$\times$51.

In \Cref{sec:fft}, we show how increasing context during upsampling can reduce spectral artifacts from a theoretical perspective. 
Theoretically, we would want an infinite-sized kernel when performing upsampling.
However, this is not practical, thus we used Transposed Convolution with kernel sizes sufficiently large to give a good trade-off between theorized context and practical trainability and compute requirements. 

Thus, inspired by encoding literature\cite{convnext, replk, SegNeXt} we use kernel sizes for upsampling that are larger than those used by previous works.
Given that previous works used kernel sizes like 2$\times$2 or 3$\times$3, anything bigger than this already provides more spatial context.
Thus, even a kernel size of 5$\times$5 would be an interesting exploration and thus we explore this as well in \cref{tbl:appendix:ablation:backbone} and \cref{tbl:appendix:ablation:segpgd_backbone}.

However, given the theoretically ideal kernel size is infinity, a kernel size of 5$\times$5 does not provide enough spatial context and thus we start calling transposed convolution operations as Large Context Transposed Convolution only when their kernel sizes are 7$\times$7 or larger.

\section{Additional visualizations of Upsampling Artifacts and their Frequency Spectra}
\label{sec:appendix:teaser_extension}
\begin{figure}[t]
    \centering
\scalebox{0.54}{
\scriptsize
   \begin{tabular}{@{}c@{\hspace{0.15cm}}c@{}c@{\hspace{0.05cm}}c@{\hspace{0.15cm}}c@{\hspace{0.05cm}}c@{}}

   \multicolumn{3}{c}{} & \textbf{Clean - within domain} & \textbf{Attacked} & \textbf{2D Frequency Spectra} \\

    \rotatebox{90}{{\phantom{sub}Baseline} \cite{chen2022simple}} &  \rotatebox{90}{{\phantom{suB}Pixel Shuffle}} & &
\includegraphics[width=0.325\textwidth]{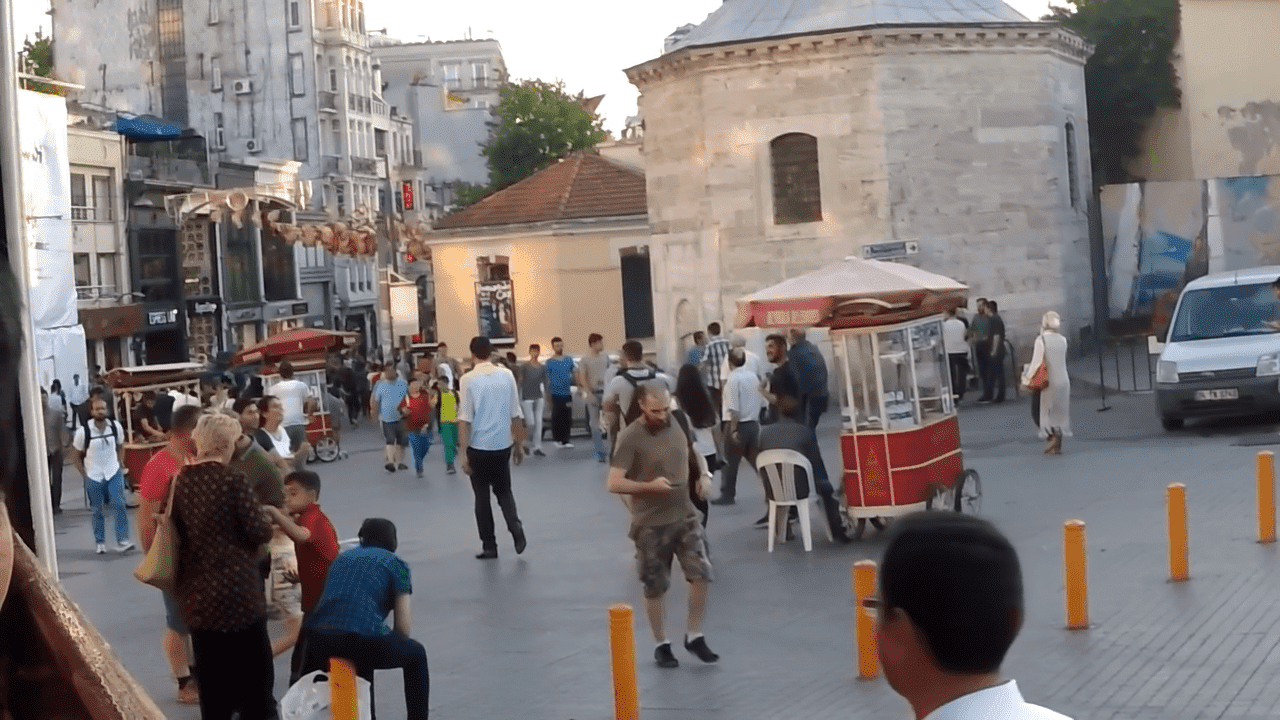} &
  \includegraphics[width=0.325\textwidth]{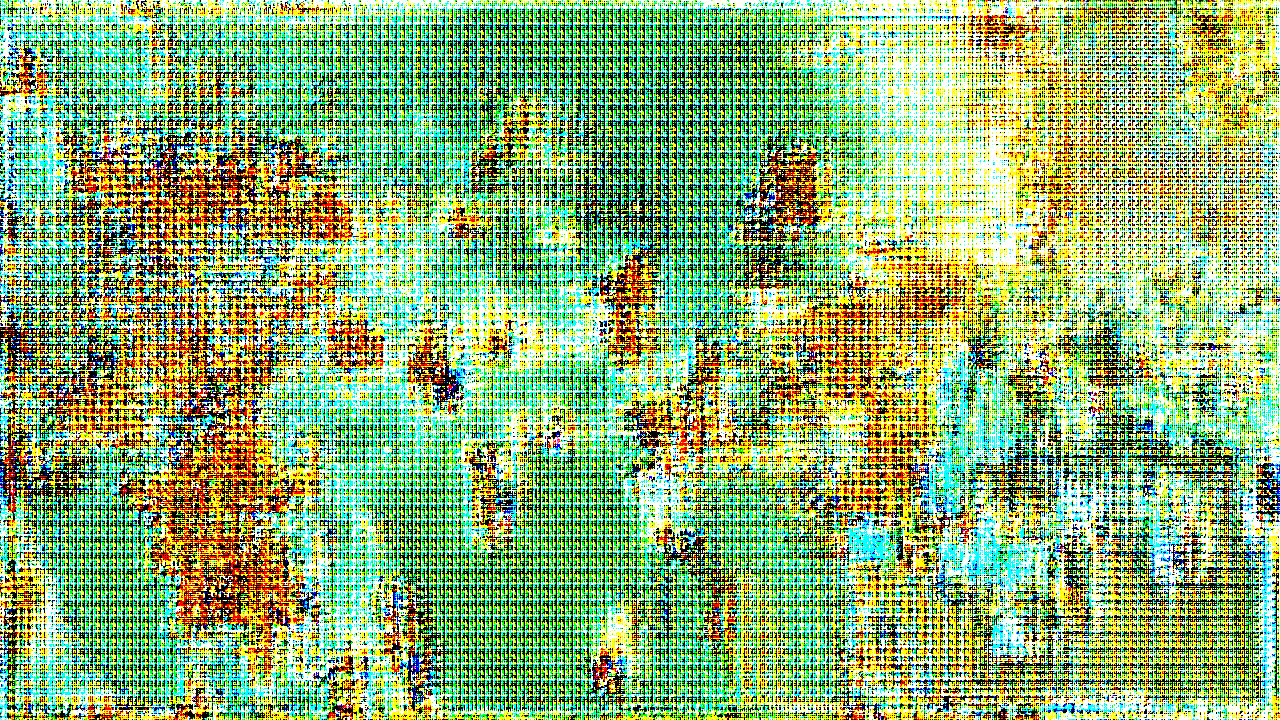} 
& \includegraphics[width=0.325\textwidth, height=2.24cm]{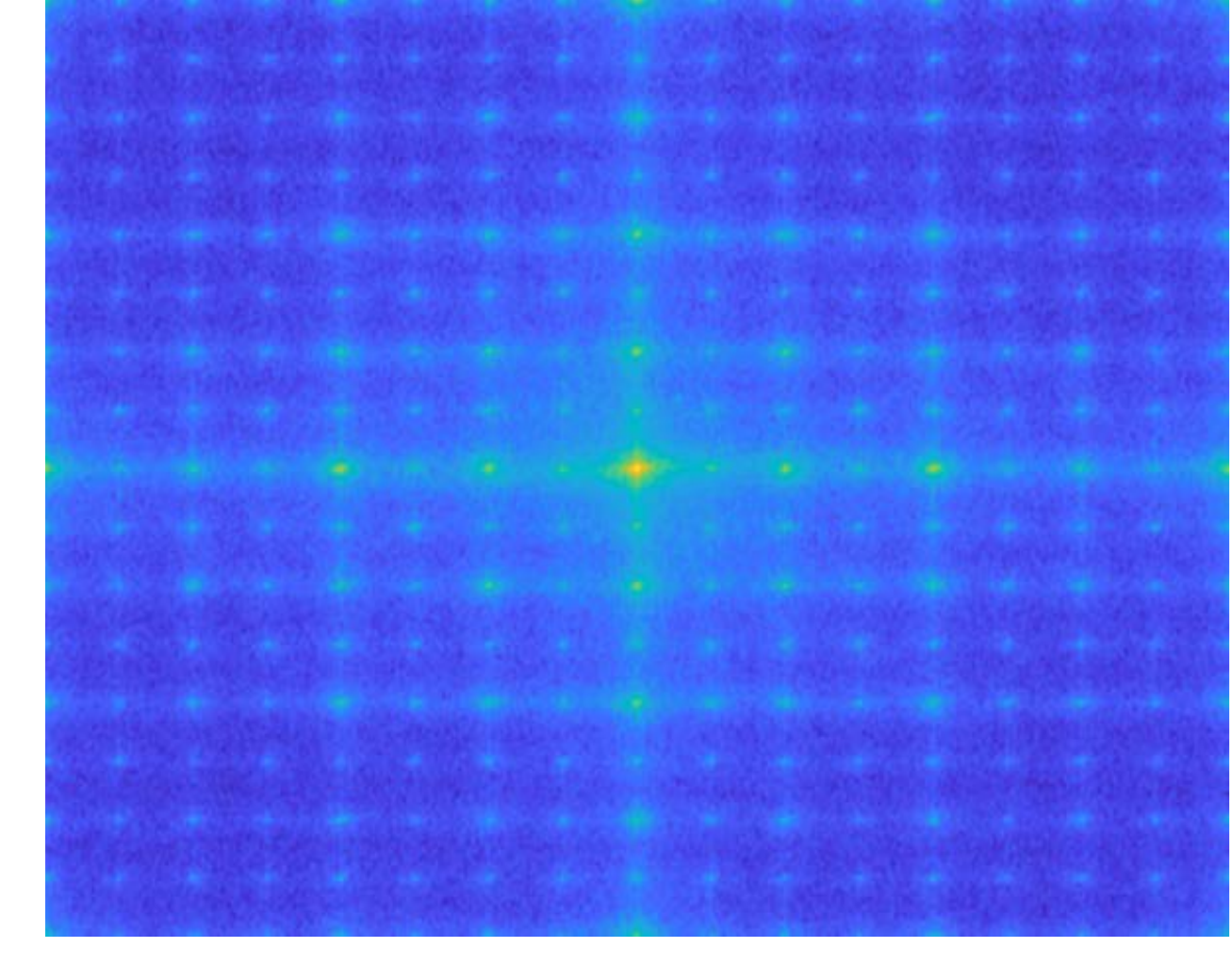} 
  \\
  
  \rotatebox{90}{{\phantom{su}}} &  \rotatebox{90}{{\phantom{suB}Transp.~conv.}} & &
\includegraphics[width=0.325\textwidth]{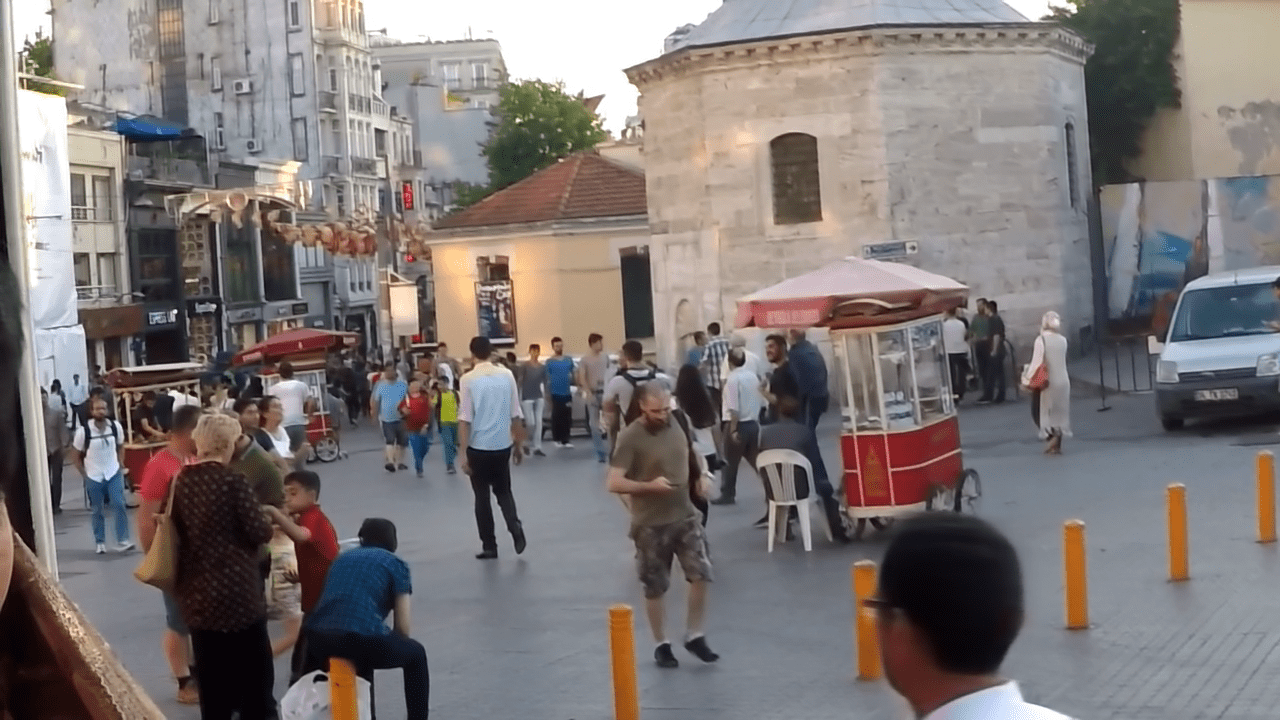} &
  \includegraphics[width=0.325\textwidth]{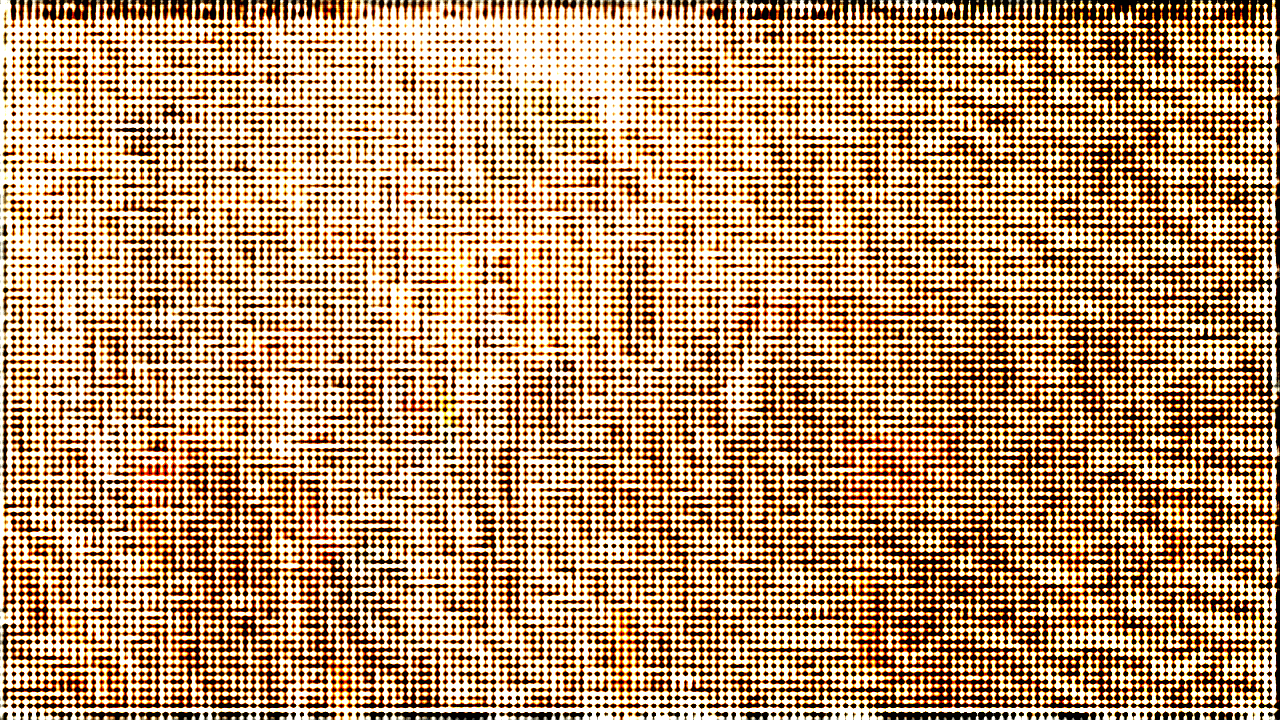} 
& \includegraphics[width=0.325\textwidth, height=2.24cm]{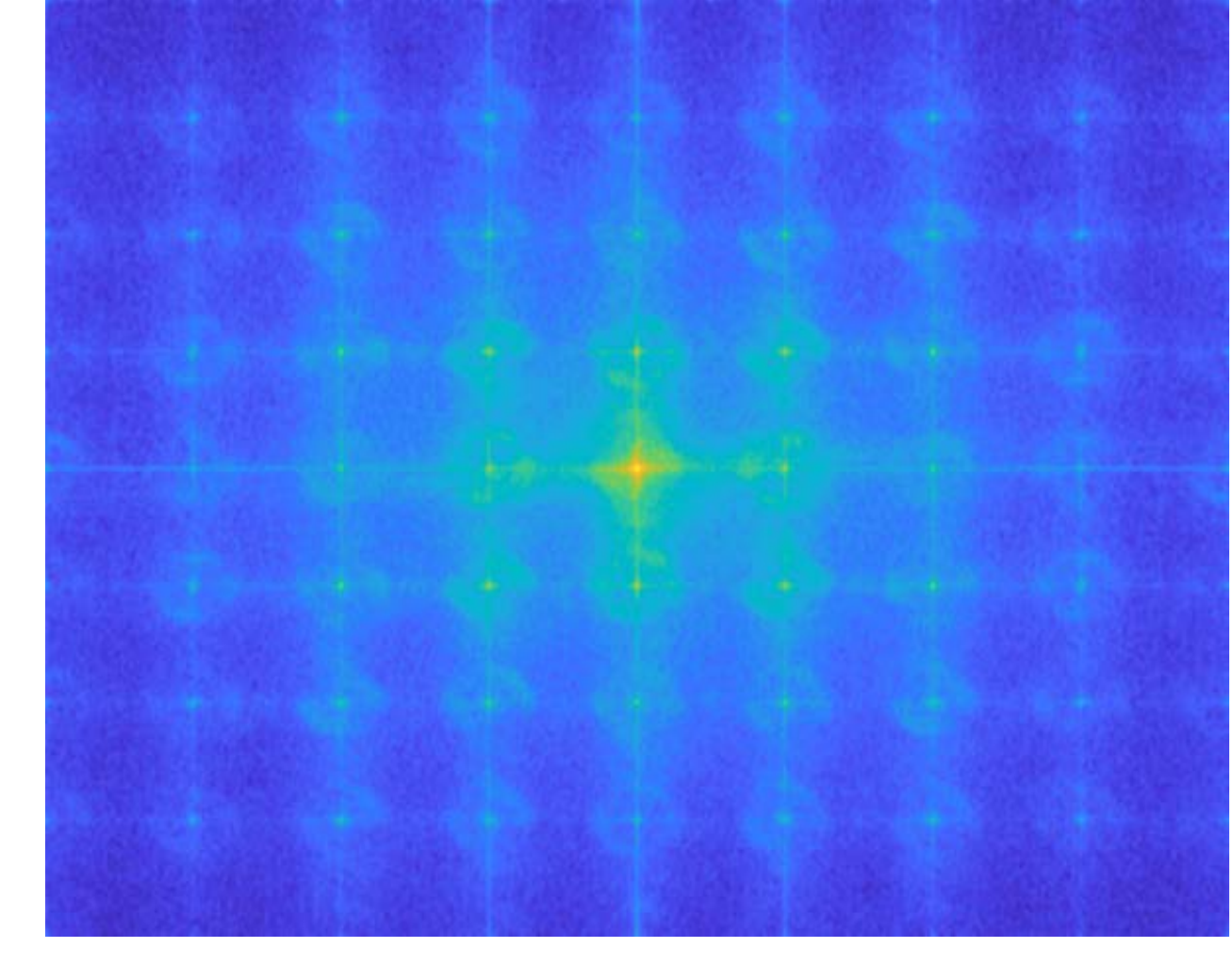} 
  \\
  
  \rotatebox{90}{{{\phantom{s}} \textbf{Large Context}}} & \rotatebox{90}{{\textbf{Transp.~conv. (Ours)}}}
  &  
  &
\includegraphics[width=0.325\textwidth]{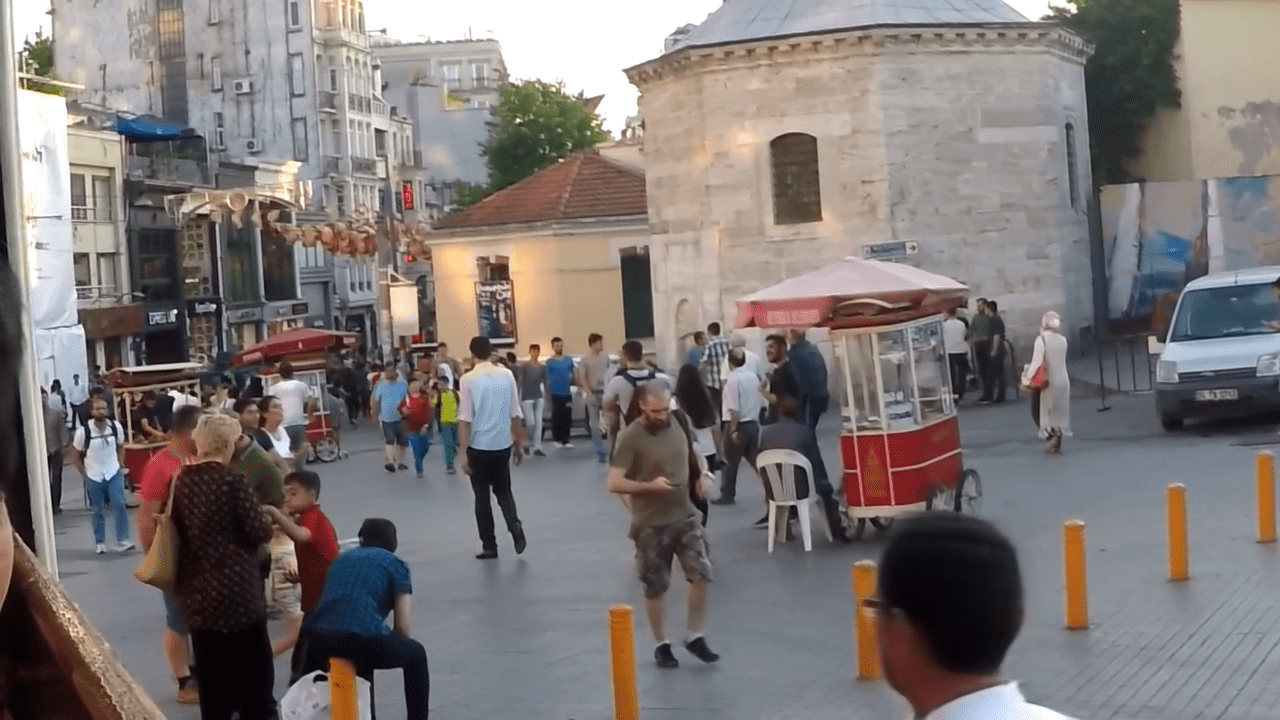} &
  \includegraphics[width=0.325\textwidth]{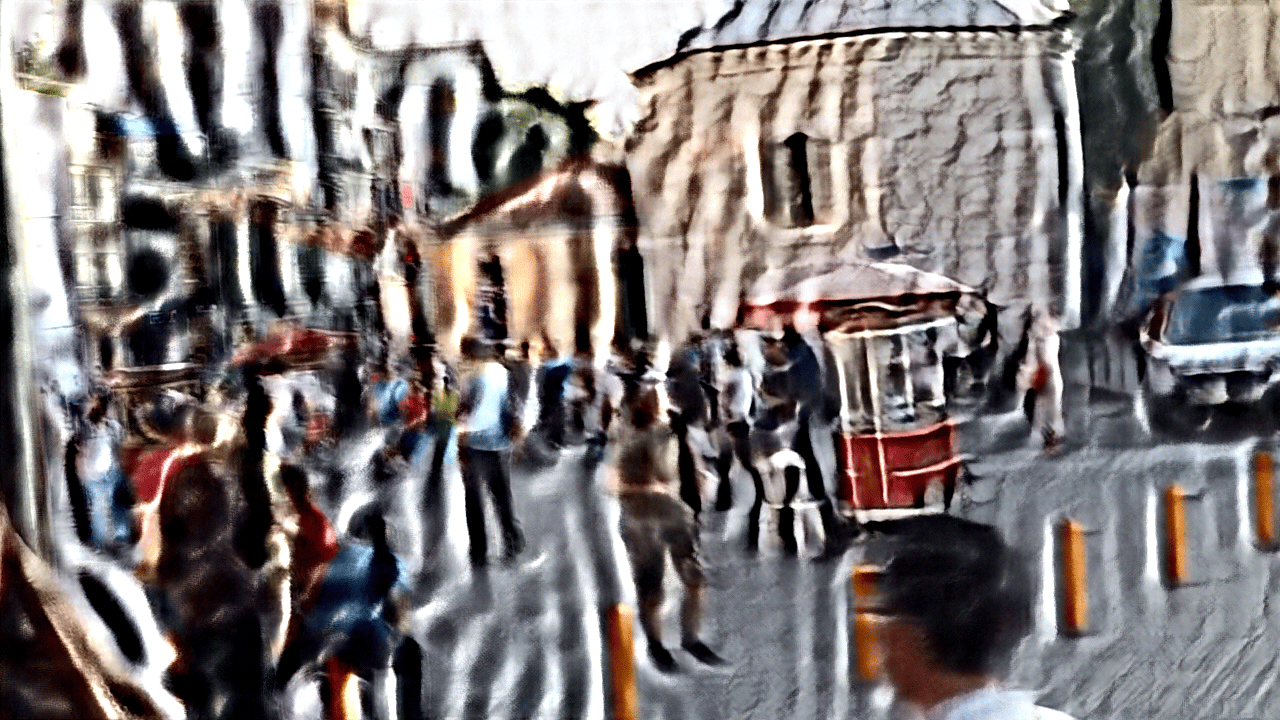} 
& \includegraphics[width=0.325\textwidth, height=2.24cm]{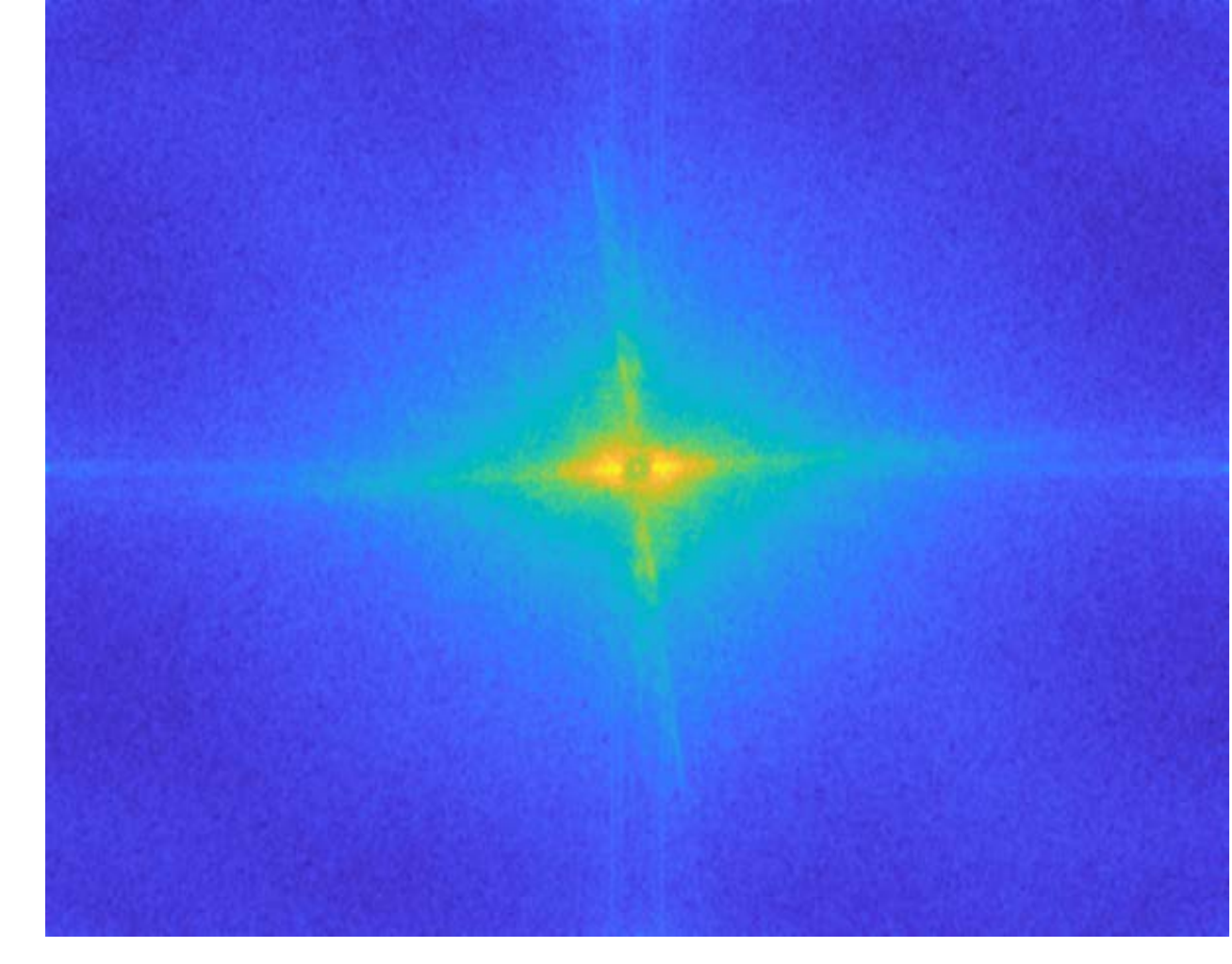} 
  \\

  \multicolumn{6}{c}{Example Image 1} \\

      \rotatebox{90}{{\phantom{sub}Baseline} \cite{chen2022simple}} &  \rotatebox{90}{{\phantom{suB}Pixel Shuffle}} & &
\includegraphics[width=0.325\textwidth]{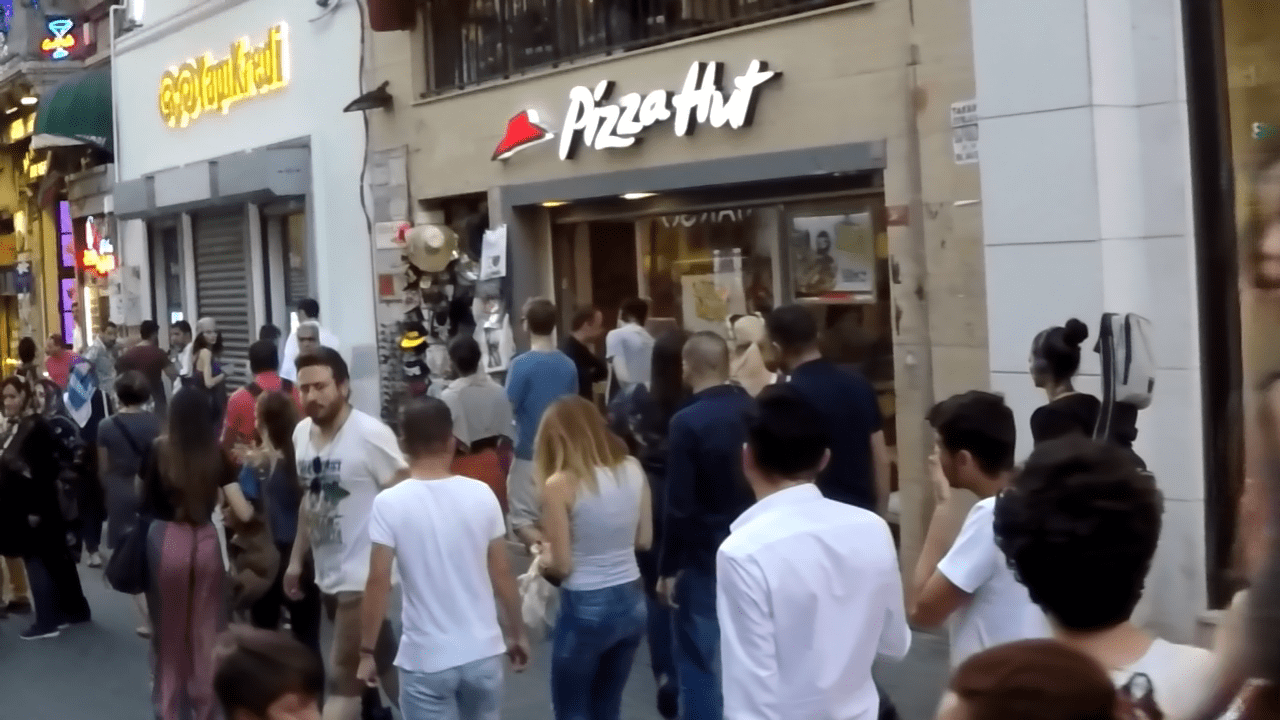} &
  \includegraphics[width=0.325\textwidth]{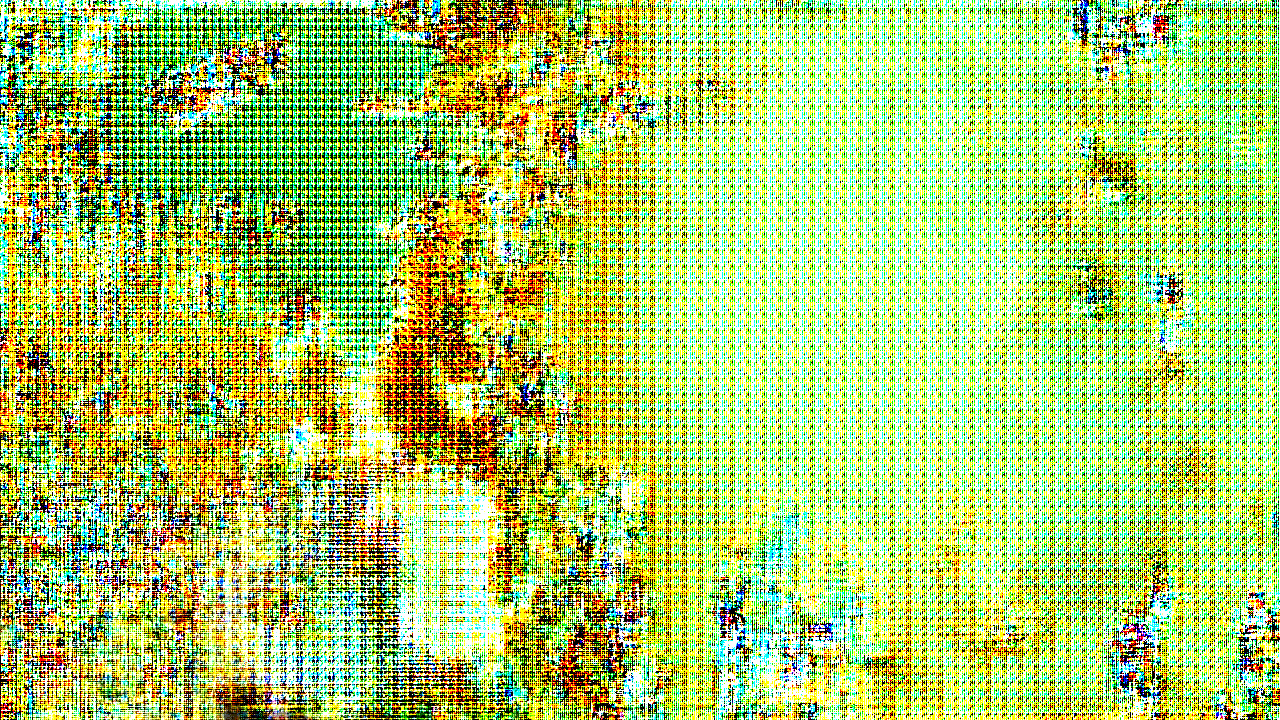} 
& \includegraphics[width=0.325\textwidth, height=2.24cm]{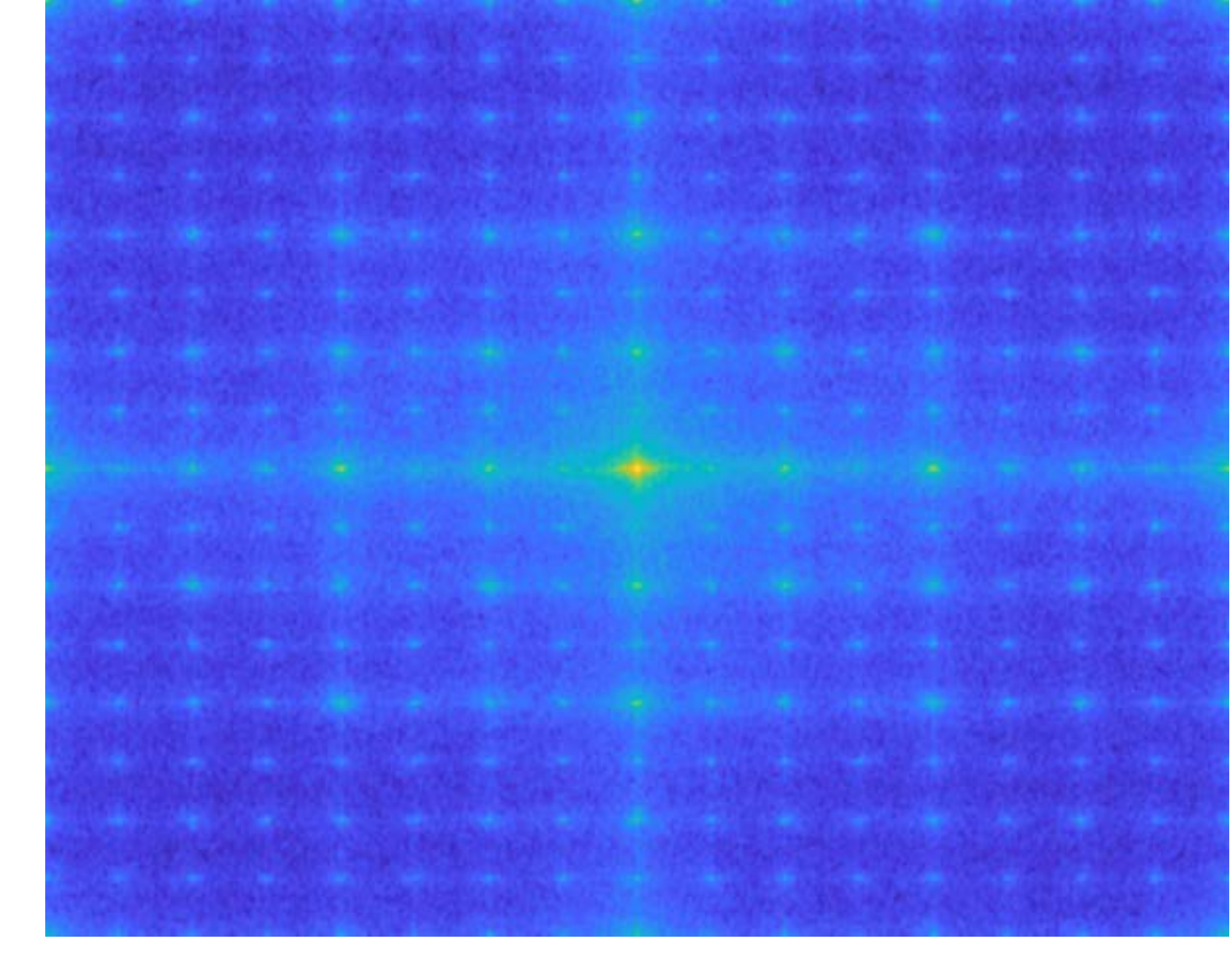} 
  \\
  
  \rotatebox{90}{{\phantom{su}}} &  \rotatebox{90}{{\phantom{suB}Transp.~conv.}} & &
\includegraphics[width=0.325\textwidth]{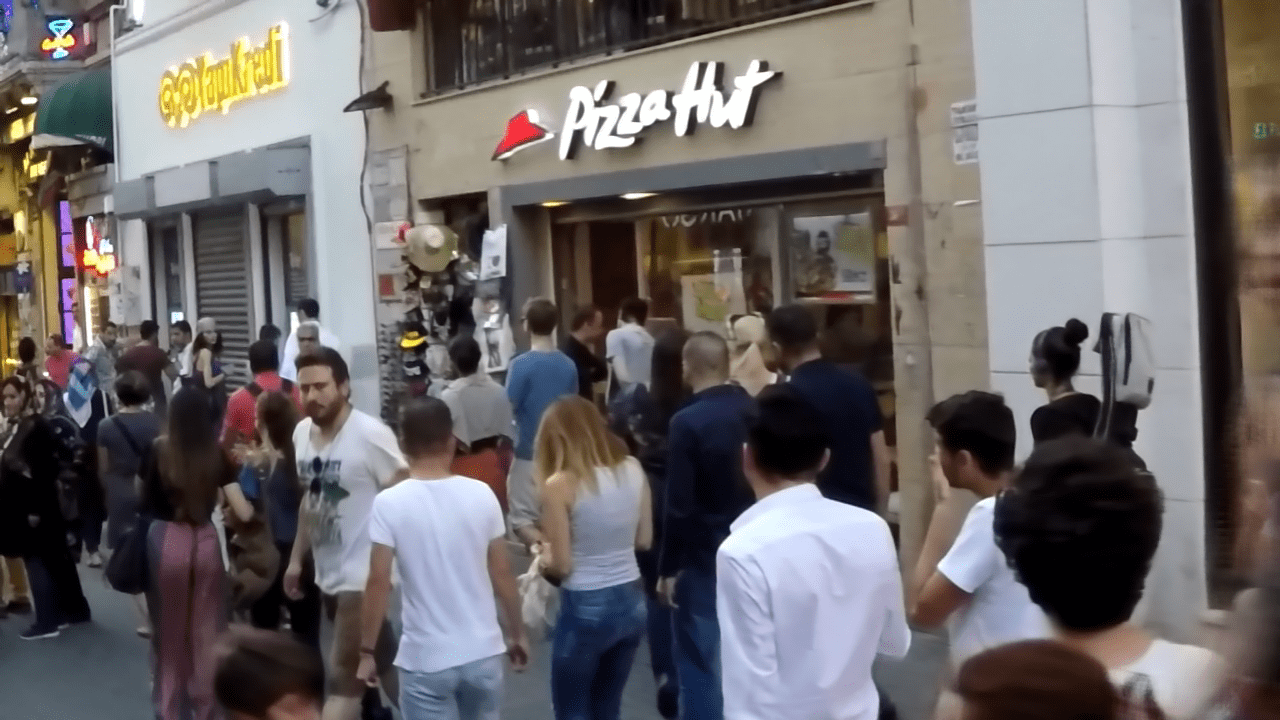} &
  \includegraphics[width=0.325\textwidth]{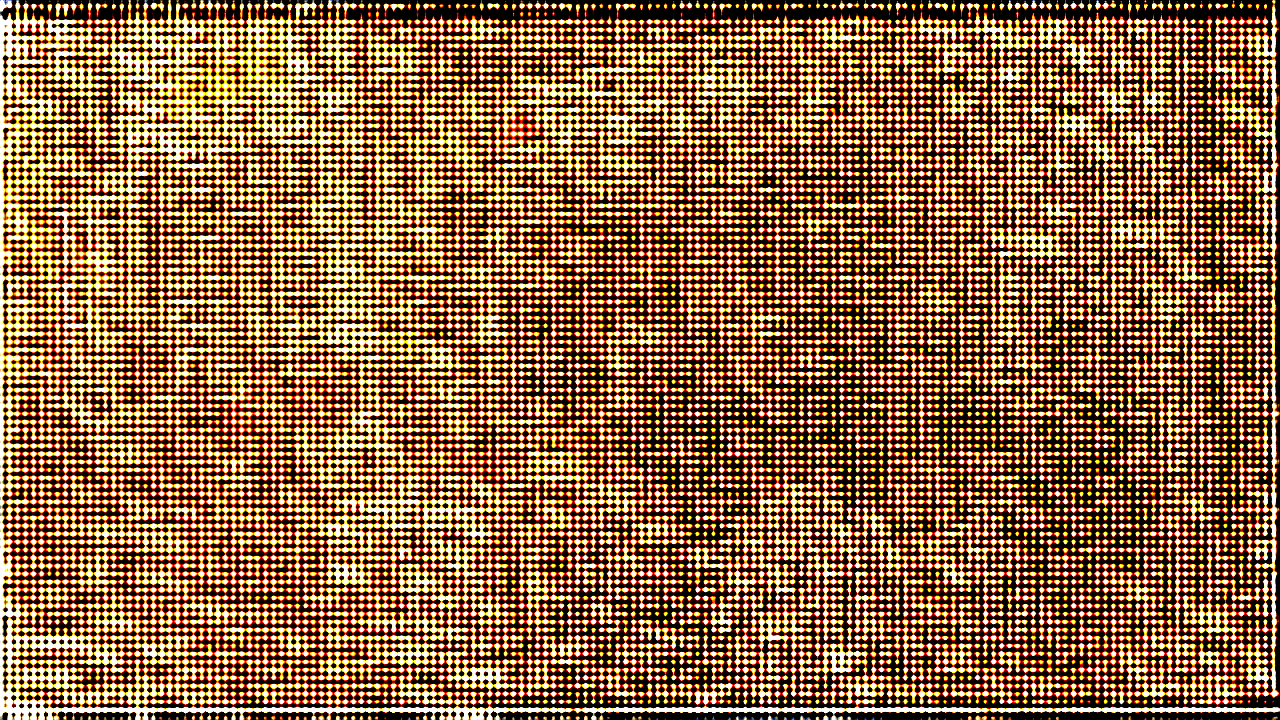} 
& \includegraphics[width=0.325\textwidth, height=2.24cm]{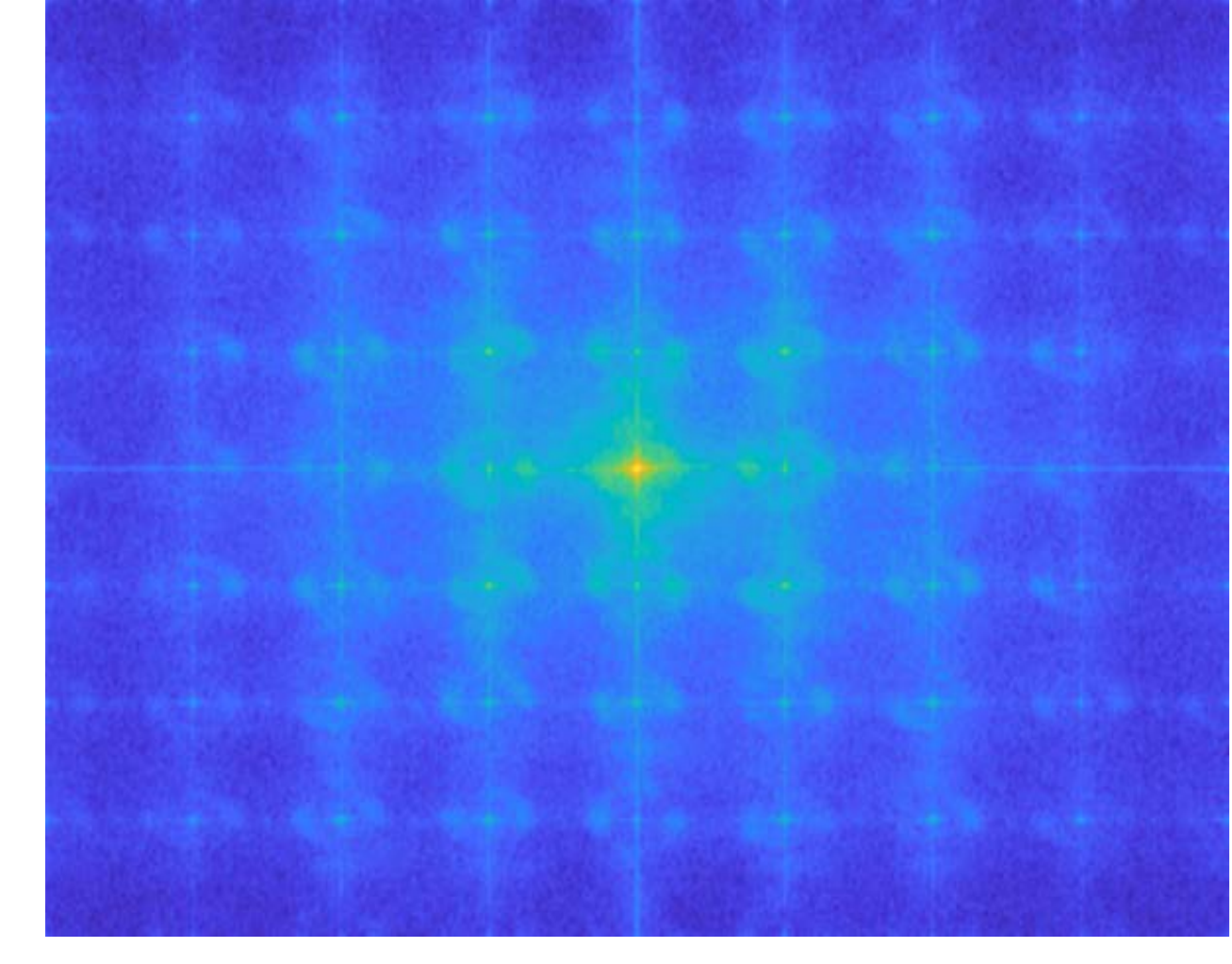} 
  \\
  
  \rotatebox{90}{{{\phantom{s}} \textbf{Large Context}}} & \rotatebox{90}{{\textbf{Transp.~conv. (Ours)}}}
  &  
  &
\includegraphics[width=0.325\textwidth]{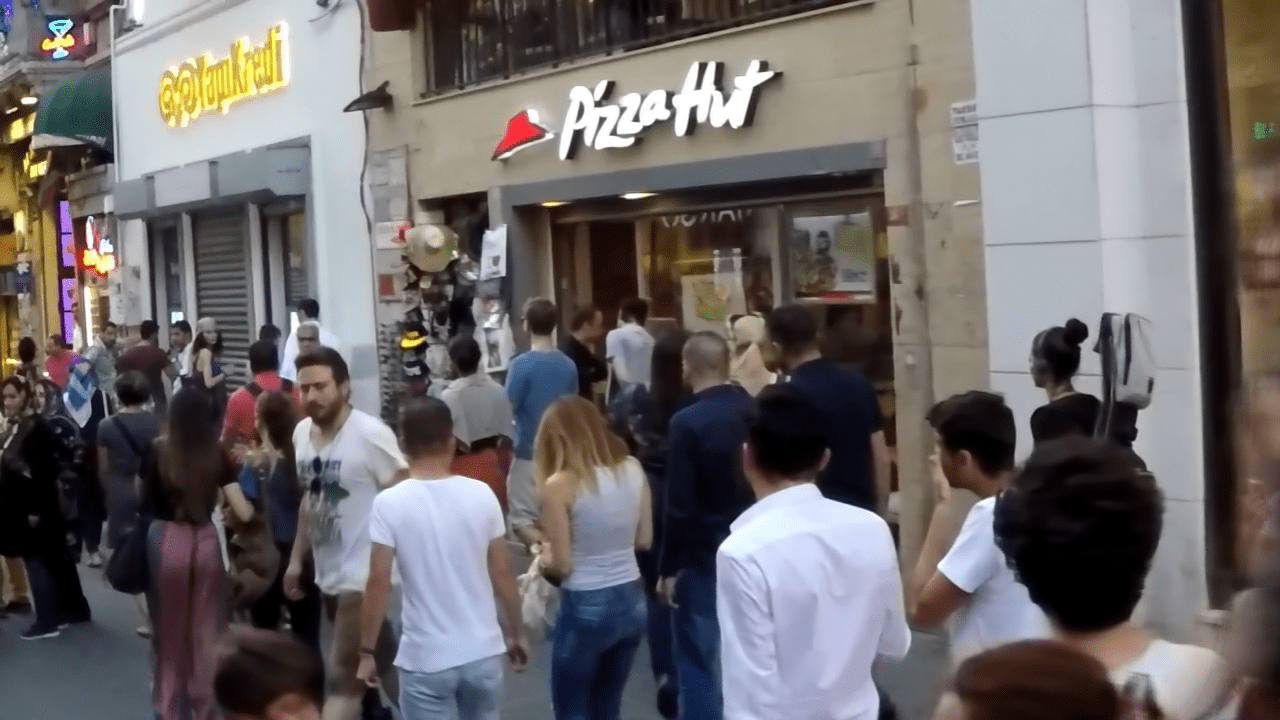} &
  \includegraphics[width=0.325\textwidth]{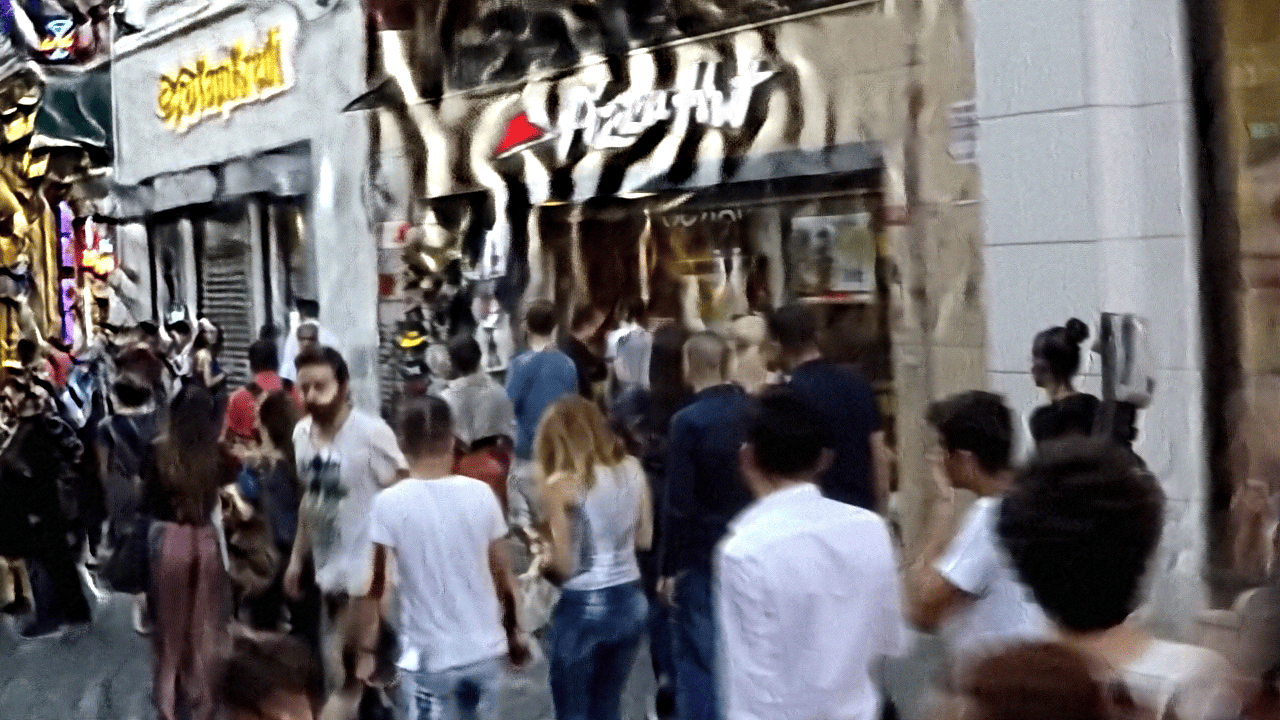} 
& \includegraphics[width=0.325\textwidth, height=2.24cm]{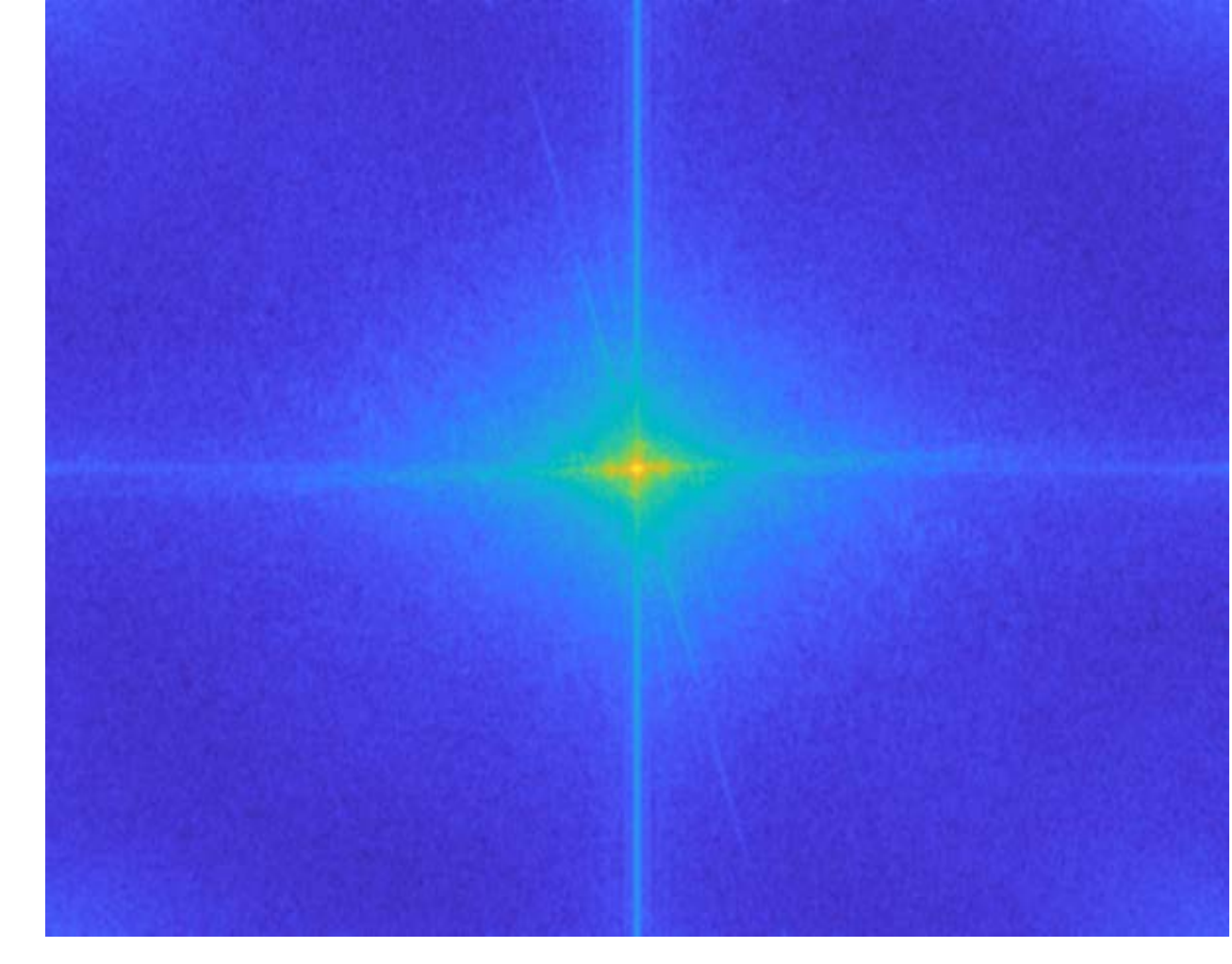} 
  \\

  \multicolumn{6}{c}{Example Image 2} \\

    \rotatebox{90}{{\phantom{sub}Baseline} \cite{chen2022simple}} &  \rotatebox{90}{{\phantom{suB}Pixel Shuffle}} & &
\includegraphics[width=0.325\textwidth]{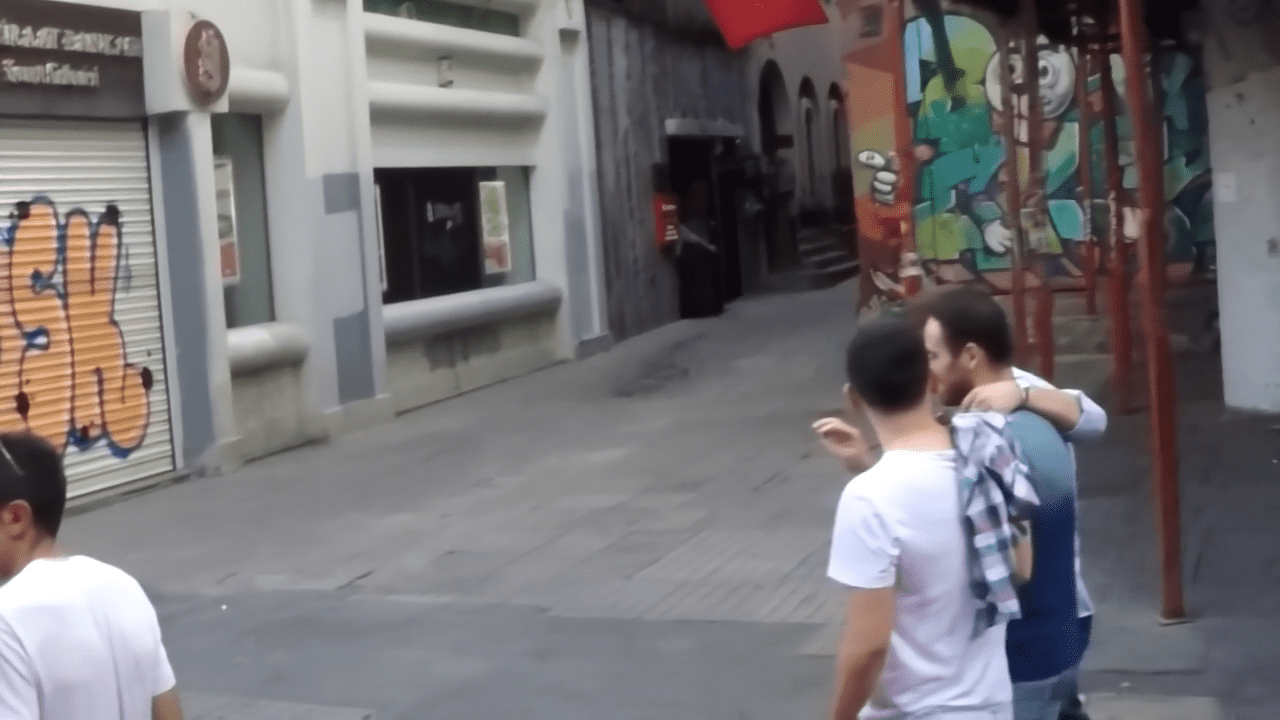} &
  \includegraphics[width=0.325\textwidth]{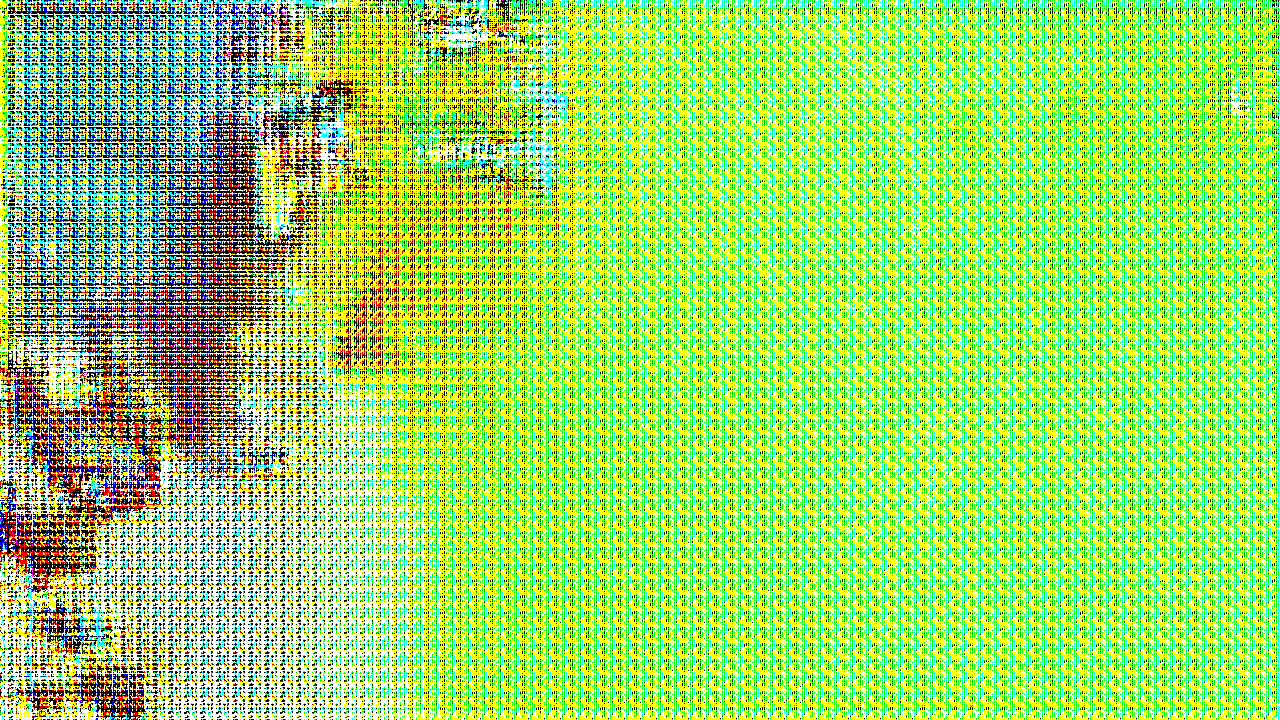} 
& \includegraphics[width=0.325\textwidth, height=2.24cm]{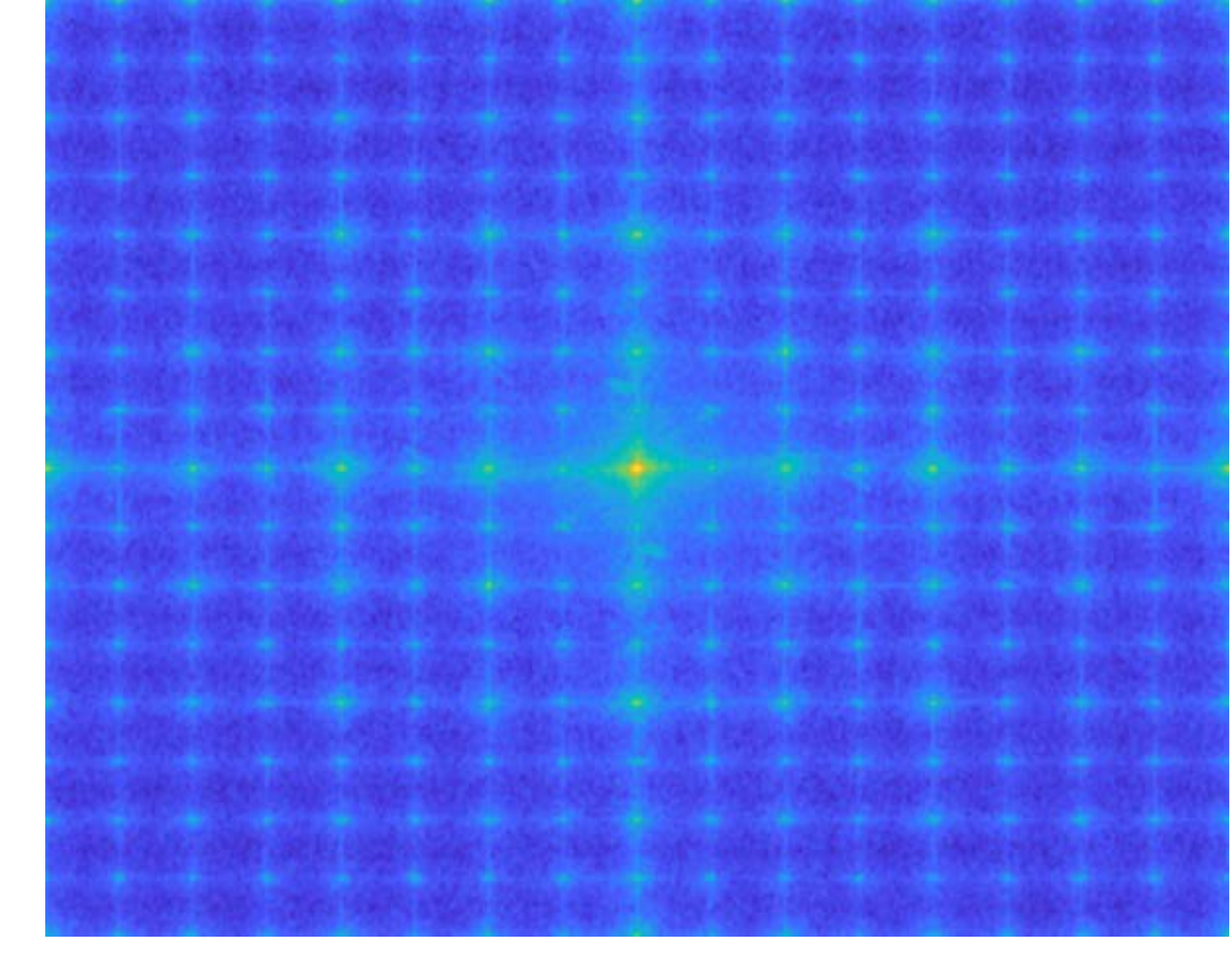} 
  \\
  
  \rotatebox{90}{{\phantom{su}}} &  \rotatebox{90}{{\phantom{suB}Transp.~conv.}} & &
\includegraphics[width=0.325\textwidth]{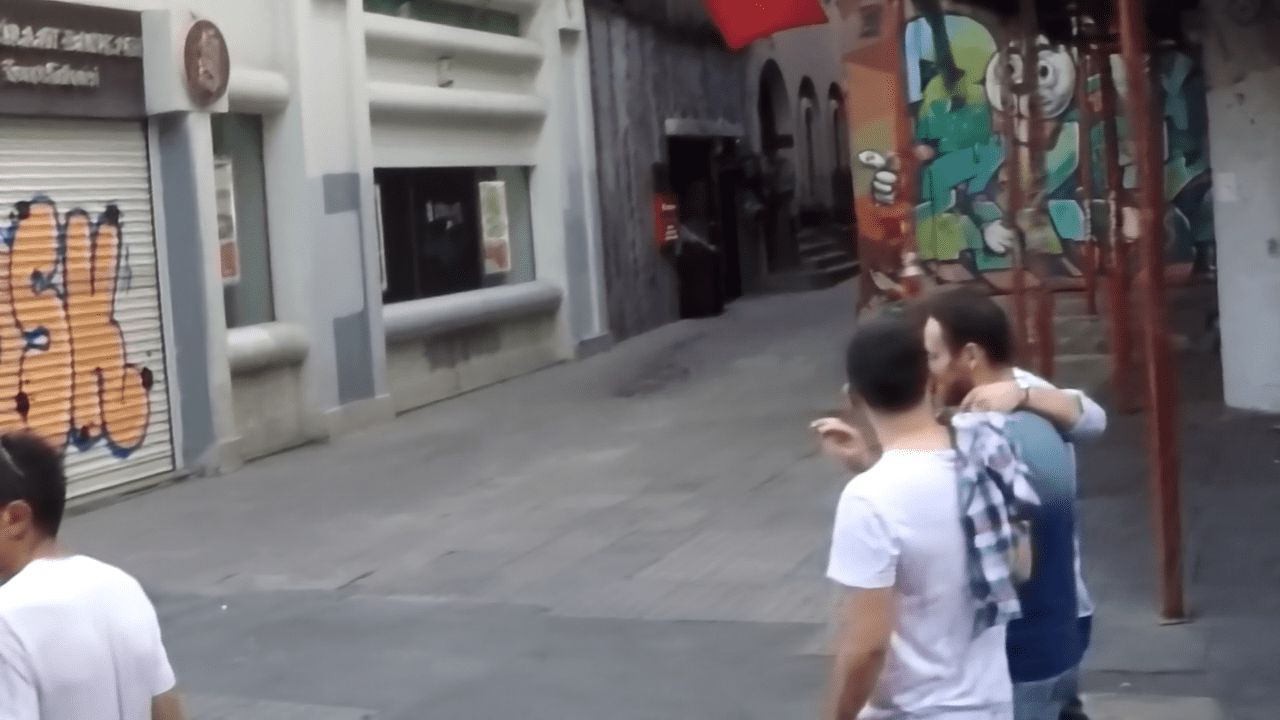} &
  \includegraphics[width=0.325\textwidth]{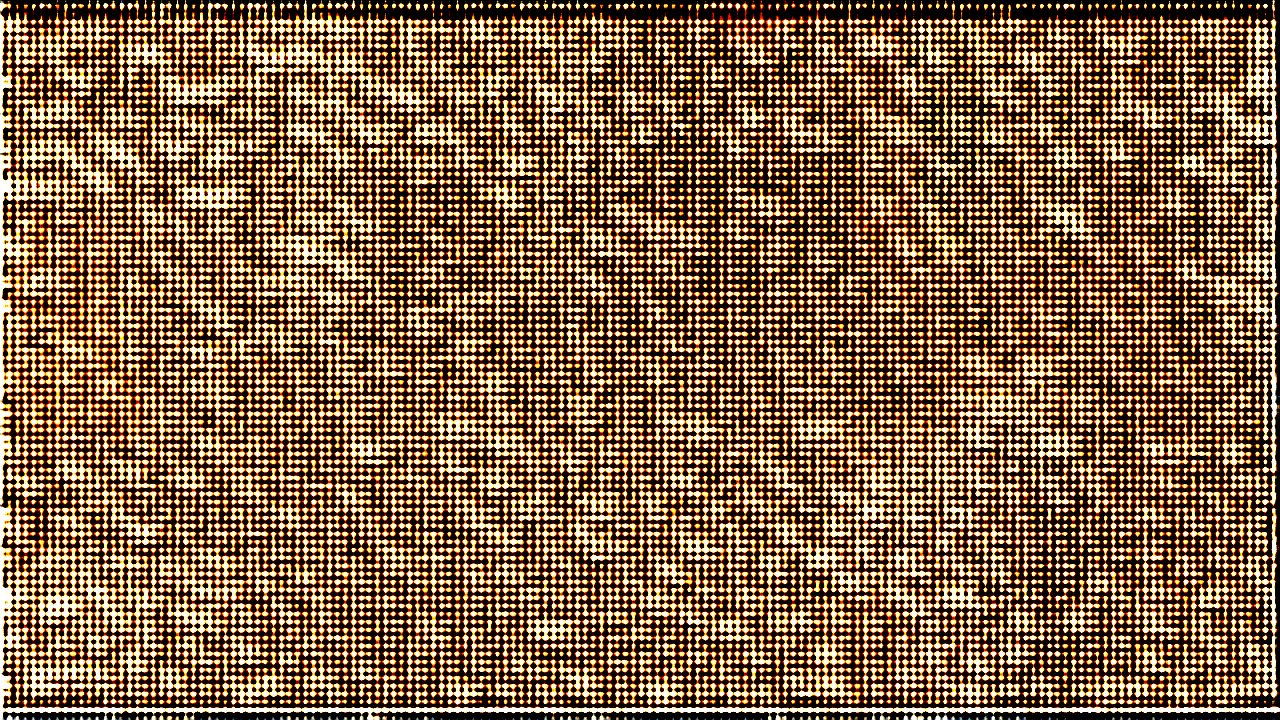} 
& \includegraphics[width=0.325\textwidth, height=2.24cm]{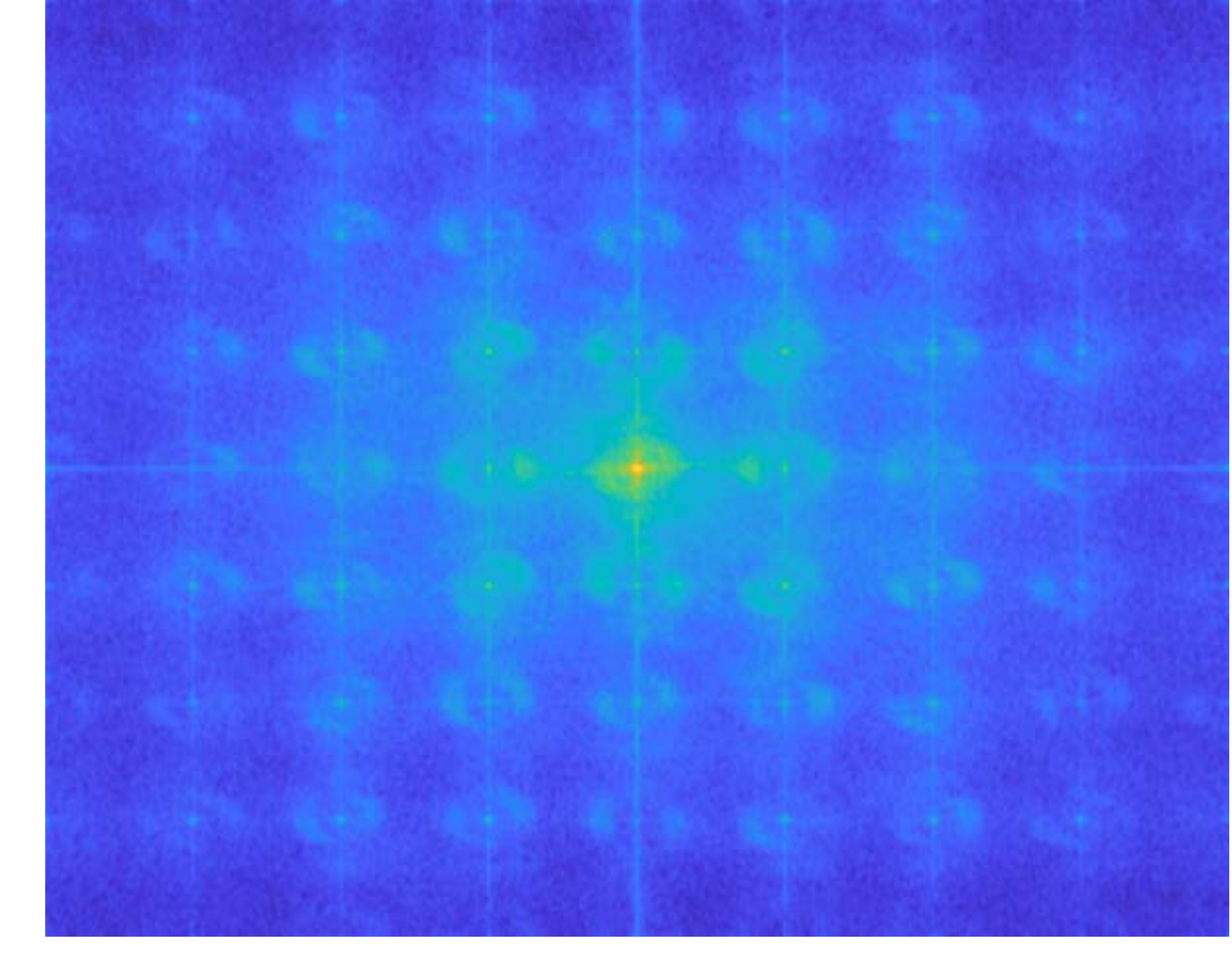} 
  \\
  
  \rotatebox{90}{{{\phantom{s}} \textbf{Large Context}}} & \rotatebox{90}{{\textbf{Transp.~conv. (Ours)}}}
  &  
  &
\includegraphics[width=0.325\textwidth]{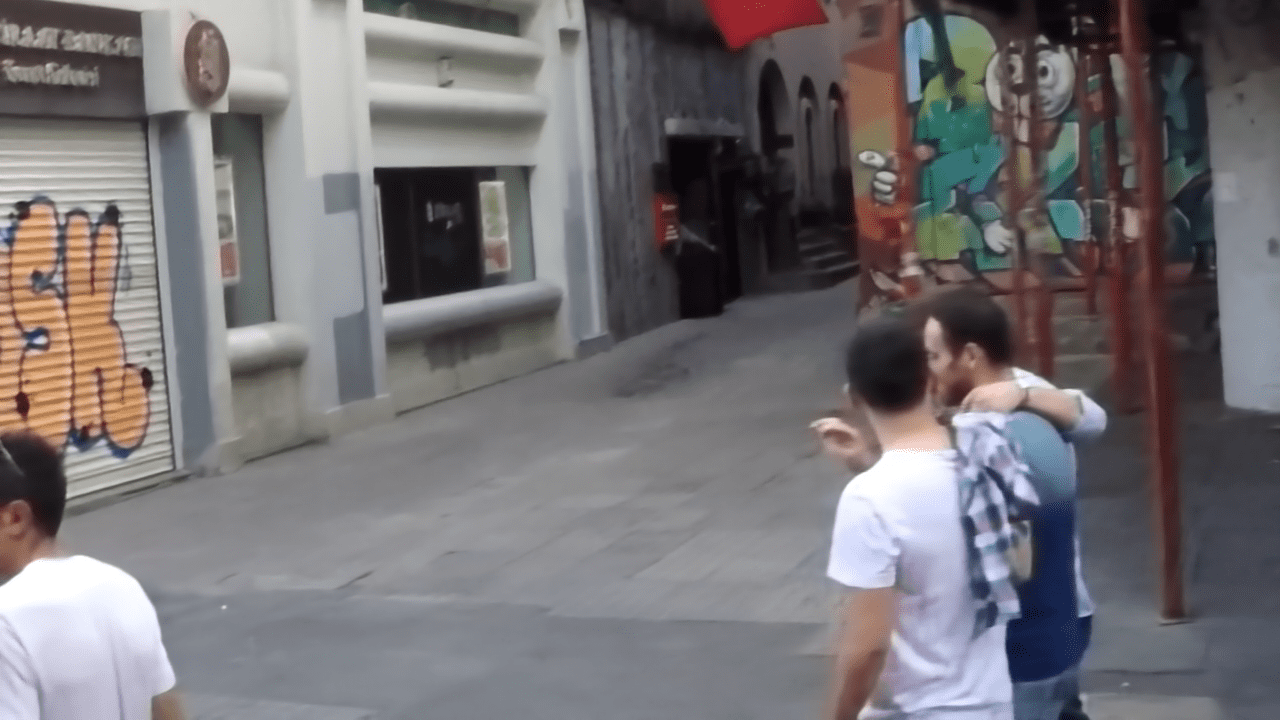} &
  \includegraphics[width=0.325\textwidth]{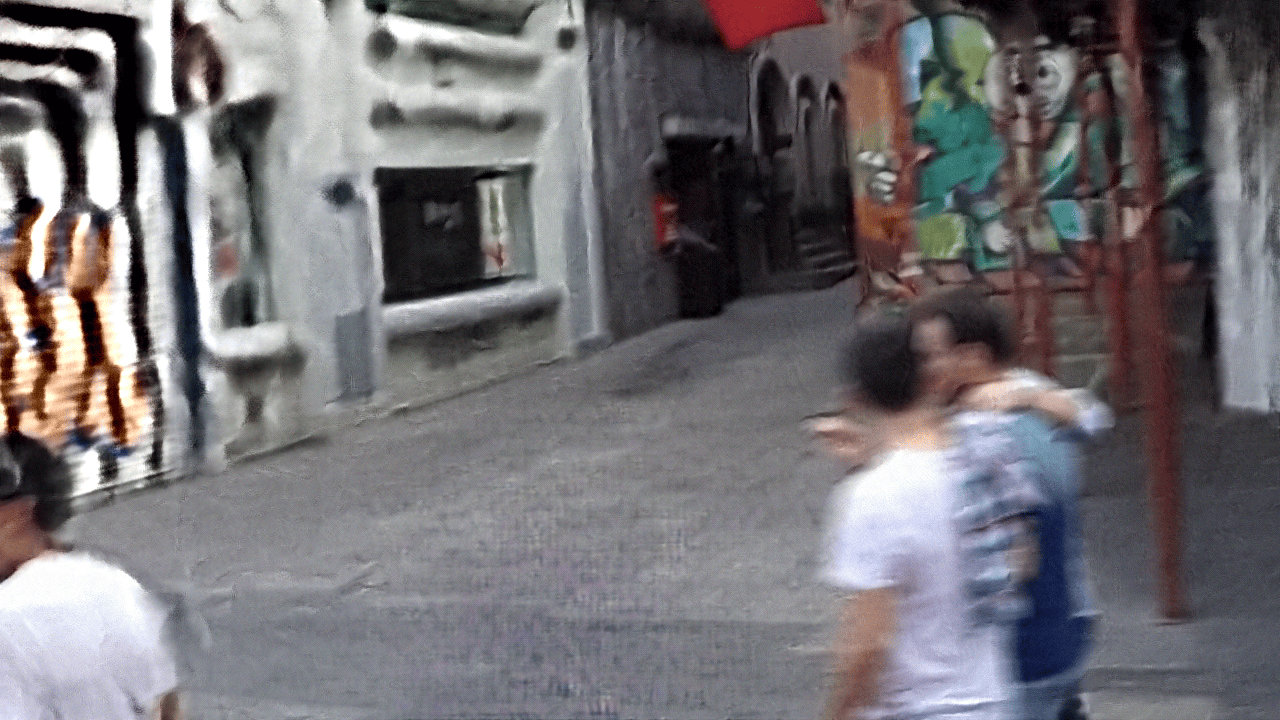} 
& \includegraphics[width=0.325\textwidth, height=2.24cm]{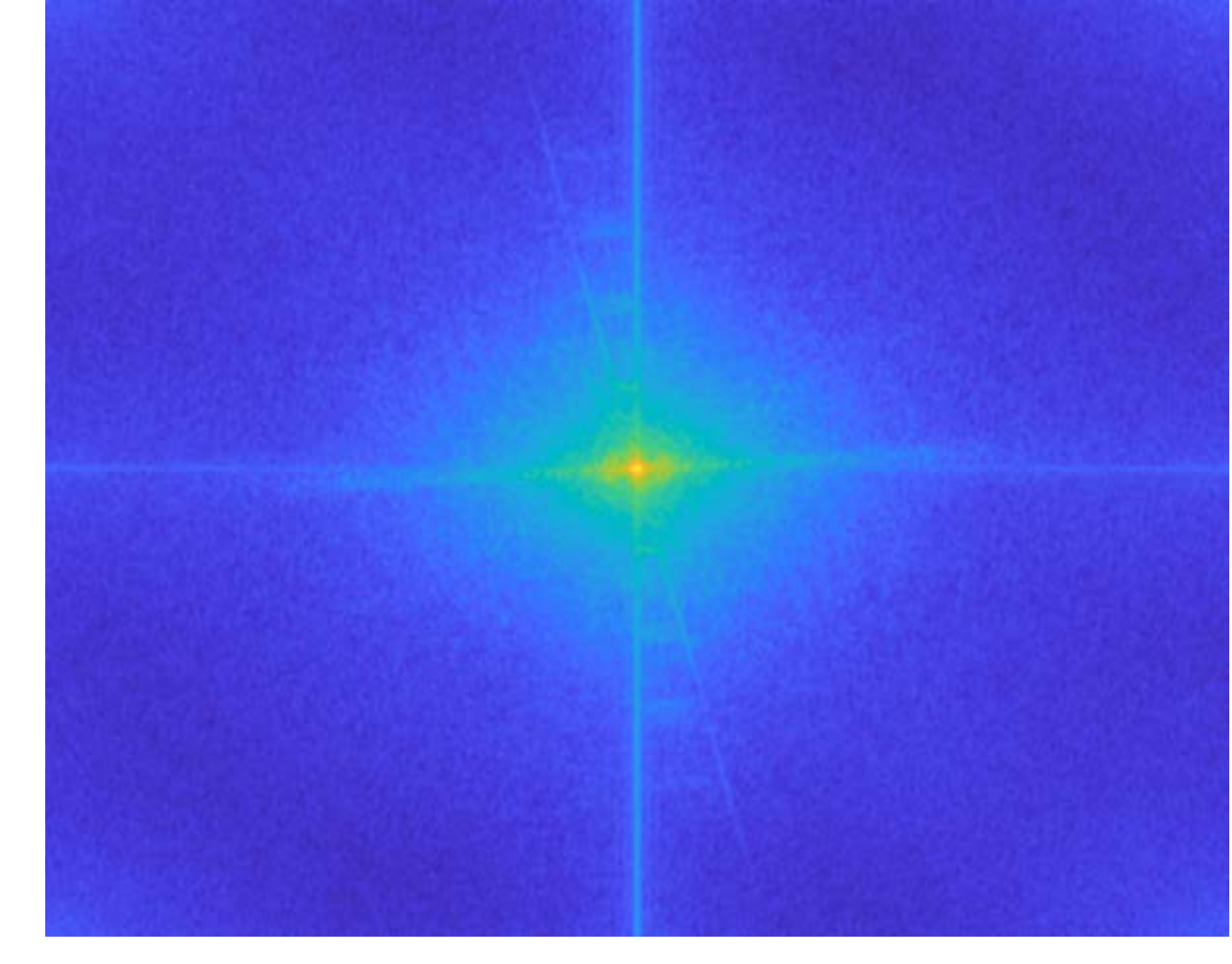} 
  \\

  \multicolumn{6}{c}{Example Image 3} \\
  
    \rotatebox{90}{{\phantom{sub}Baseline} \cite{chen2022simple}} &  \rotatebox{90}{{\phantom{suB}Pixel Shuffle}} & &
\includegraphics[width=0.325\textwidth]{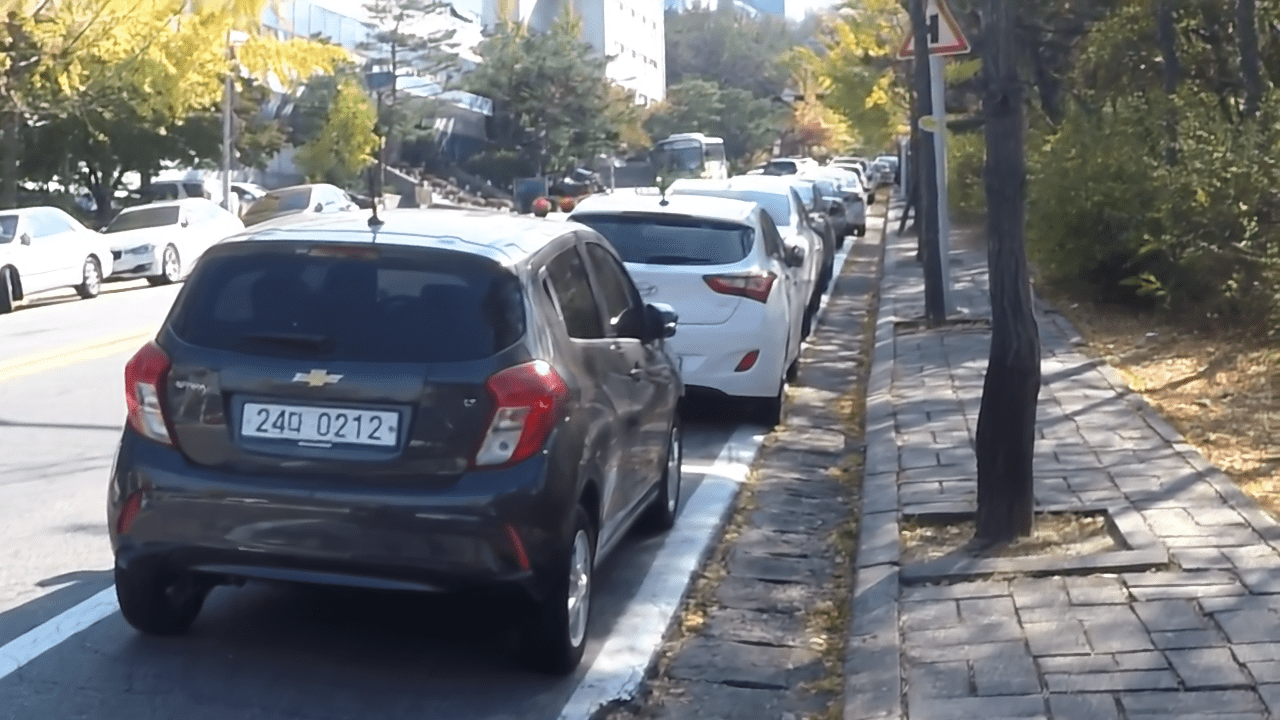} &
  \includegraphics[width=0.325\textwidth]{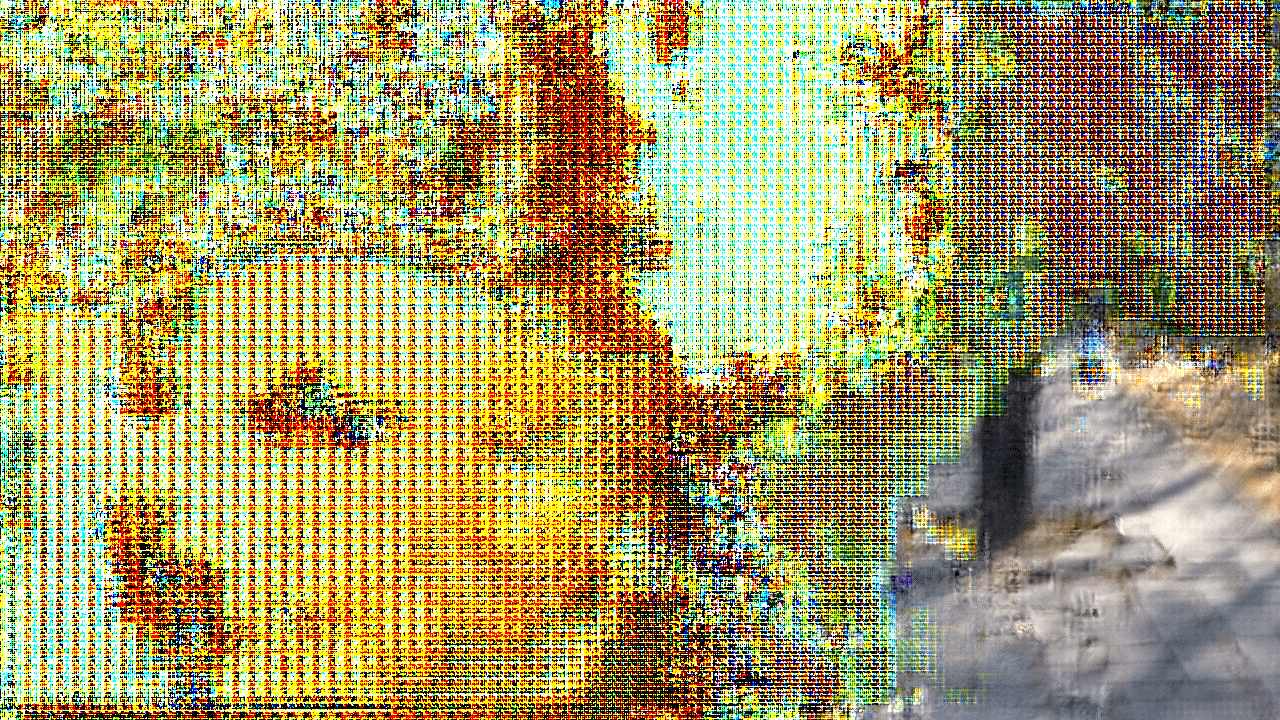} 
& \includegraphics[width=0.325\textwidth, height=2.24cm]{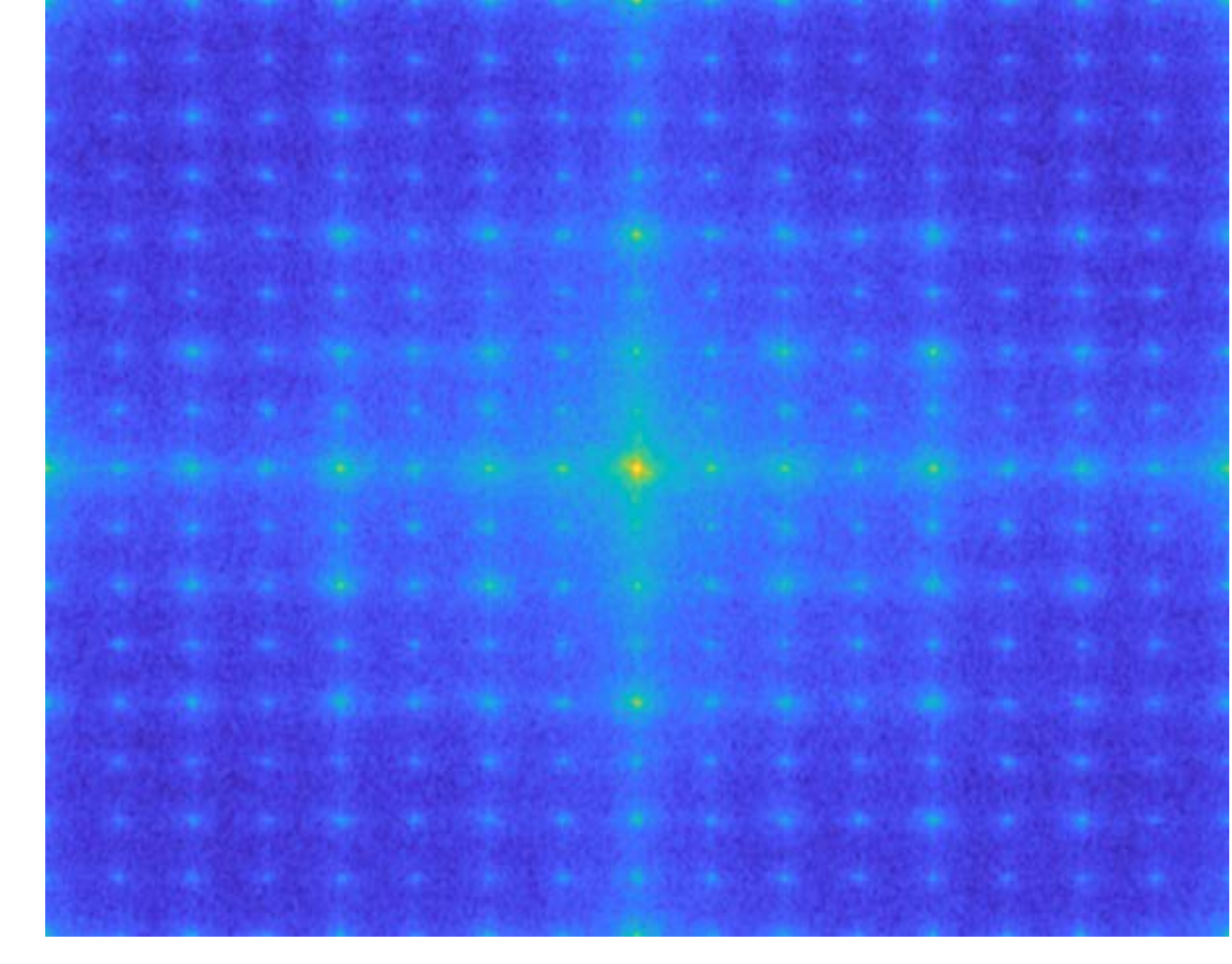} 
  \\
  
  \rotatebox{90}{{\phantom{su}}} &  \rotatebox{90}{{\phantom{suB}Transp.~conv.}} & &
\includegraphics[width=0.325\textwidth]{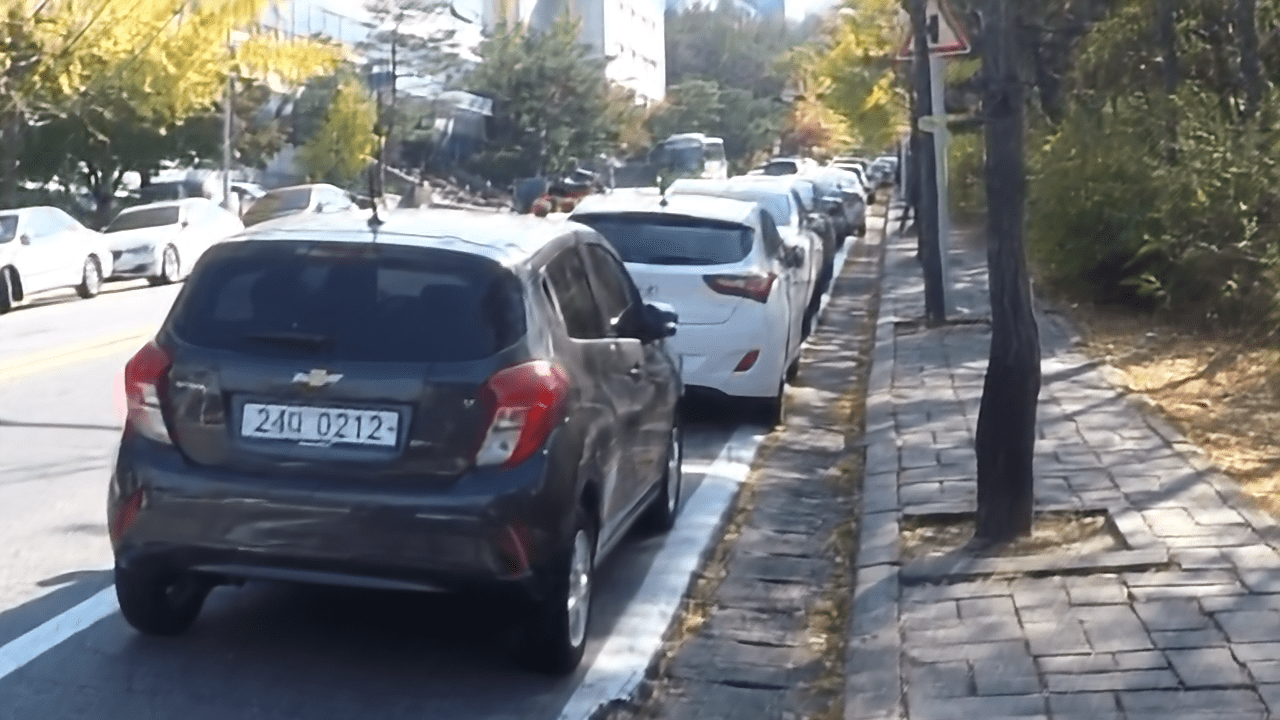} &
  \includegraphics[width=0.325\textwidth]{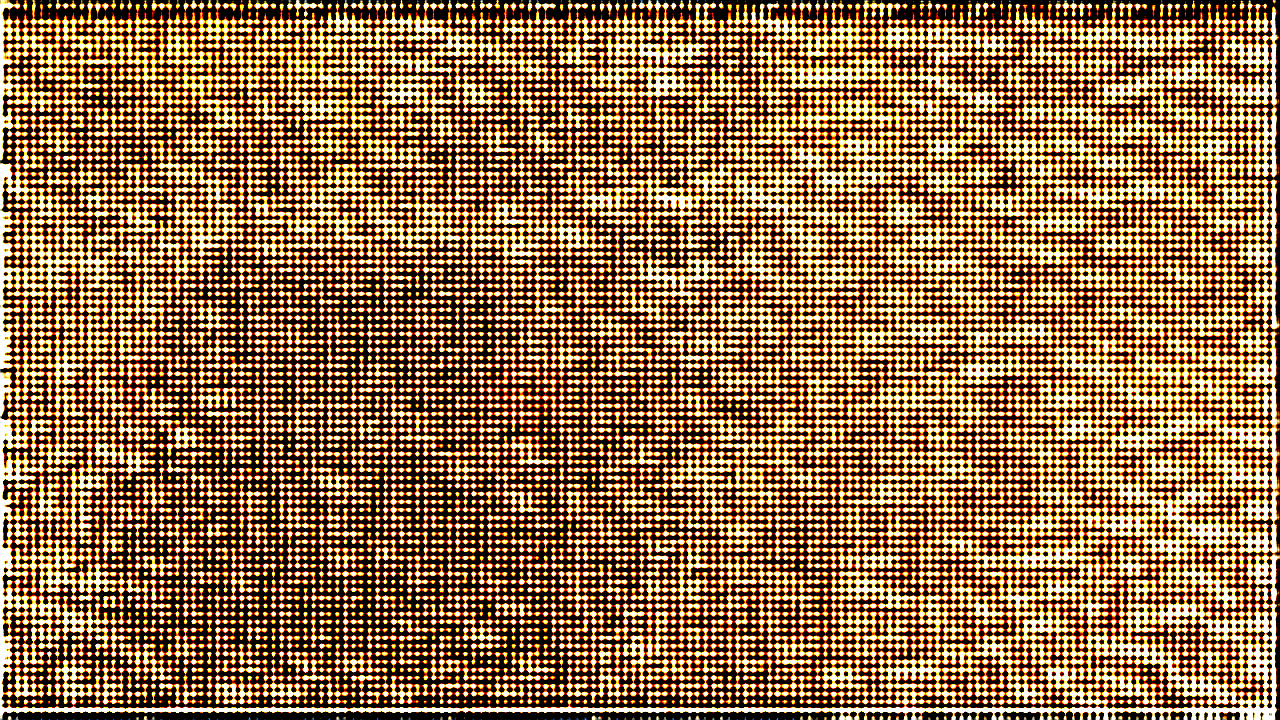} 
& \includegraphics[width=0.325\textwidth, height=2.24cm]{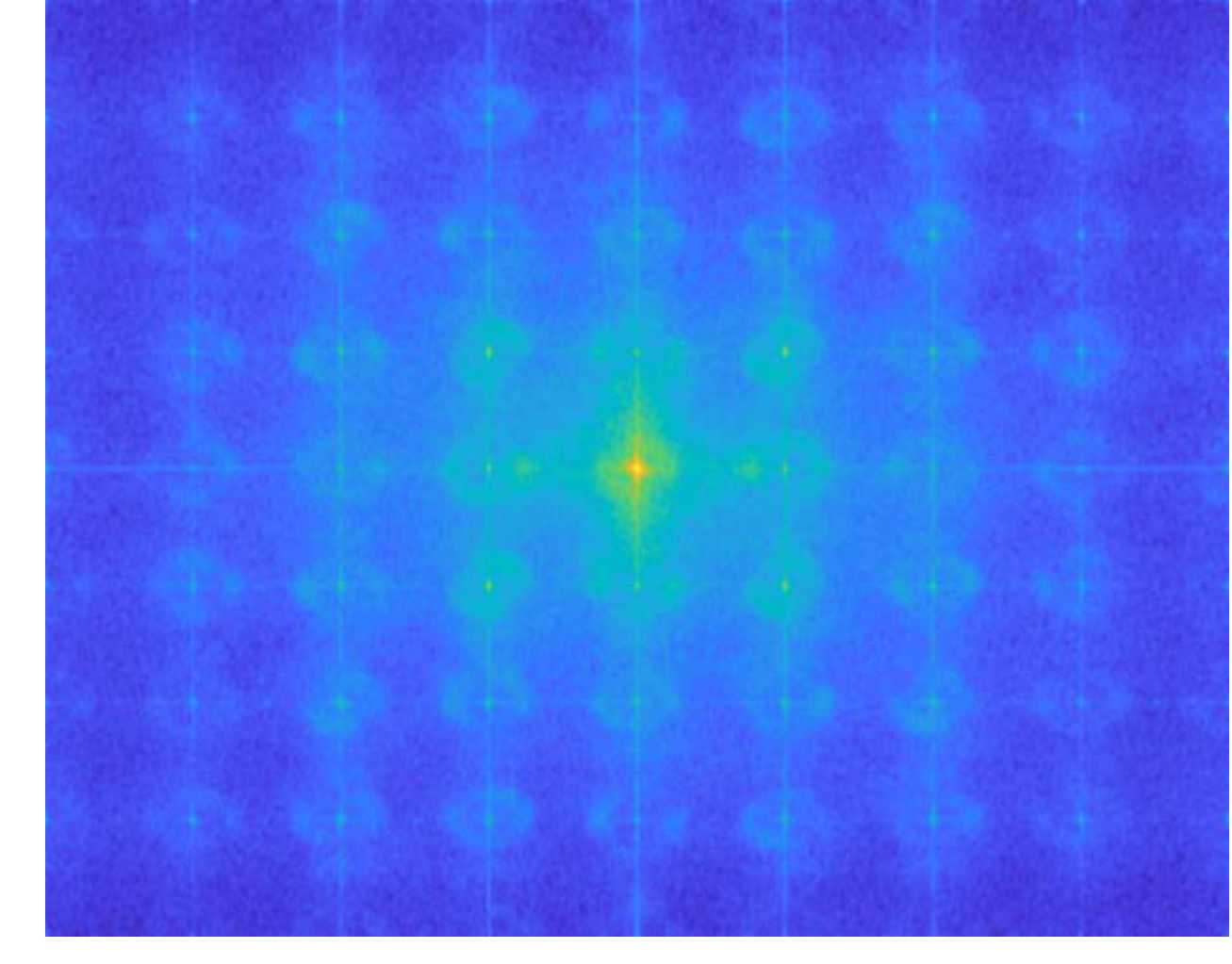} 
  \\
  
  \rotatebox{90}{{{\phantom{s}} \textbf{Large Context}}} & \rotatebox{90}{{\textbf{Transp.~conv. (Ours)}}}
  &  
  &
\includegraphics[width=0.325\textwidth]{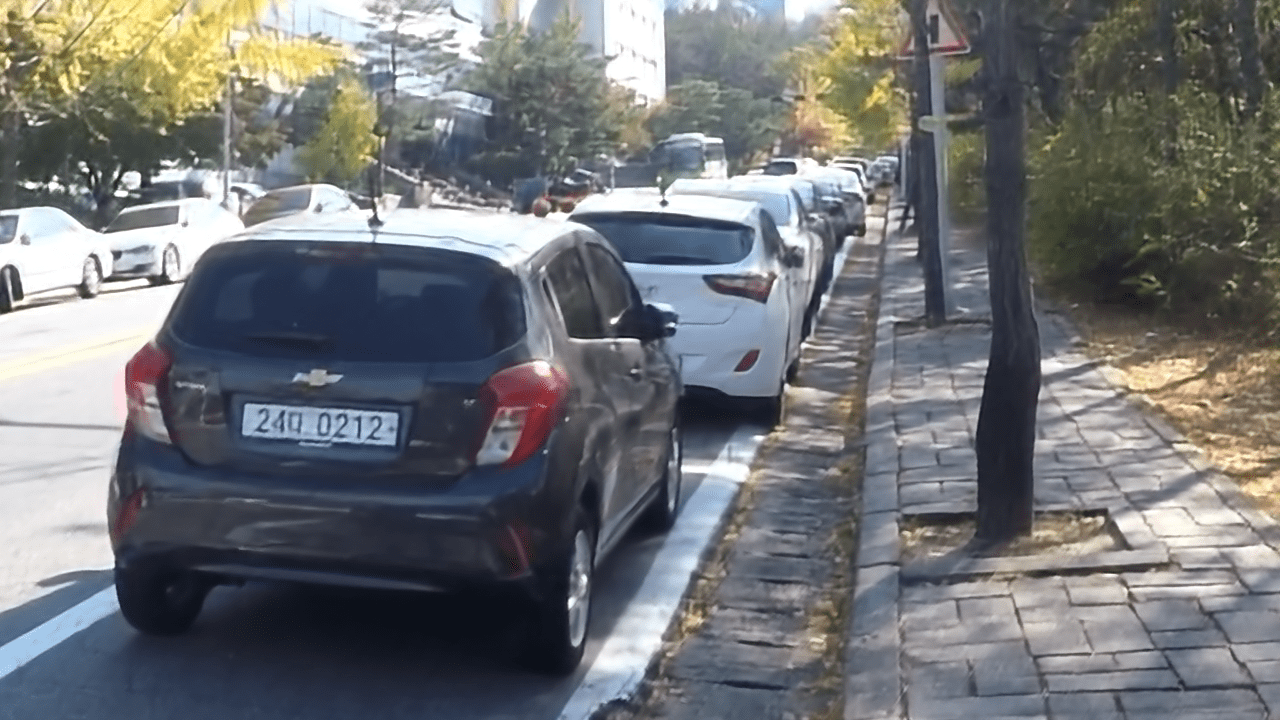} &
  \includegraphics[width=0.325\textwidth]{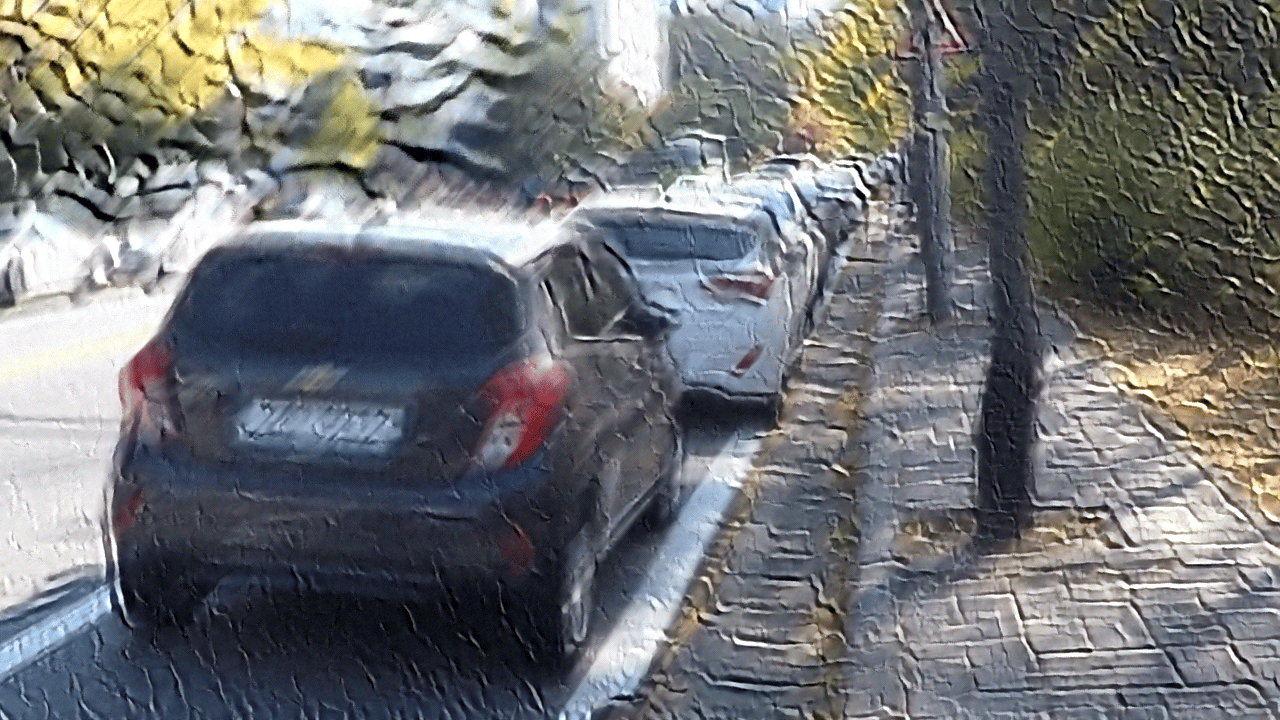} 
& \includegraphics[width=0.325\textwidth, height=2.24cm]{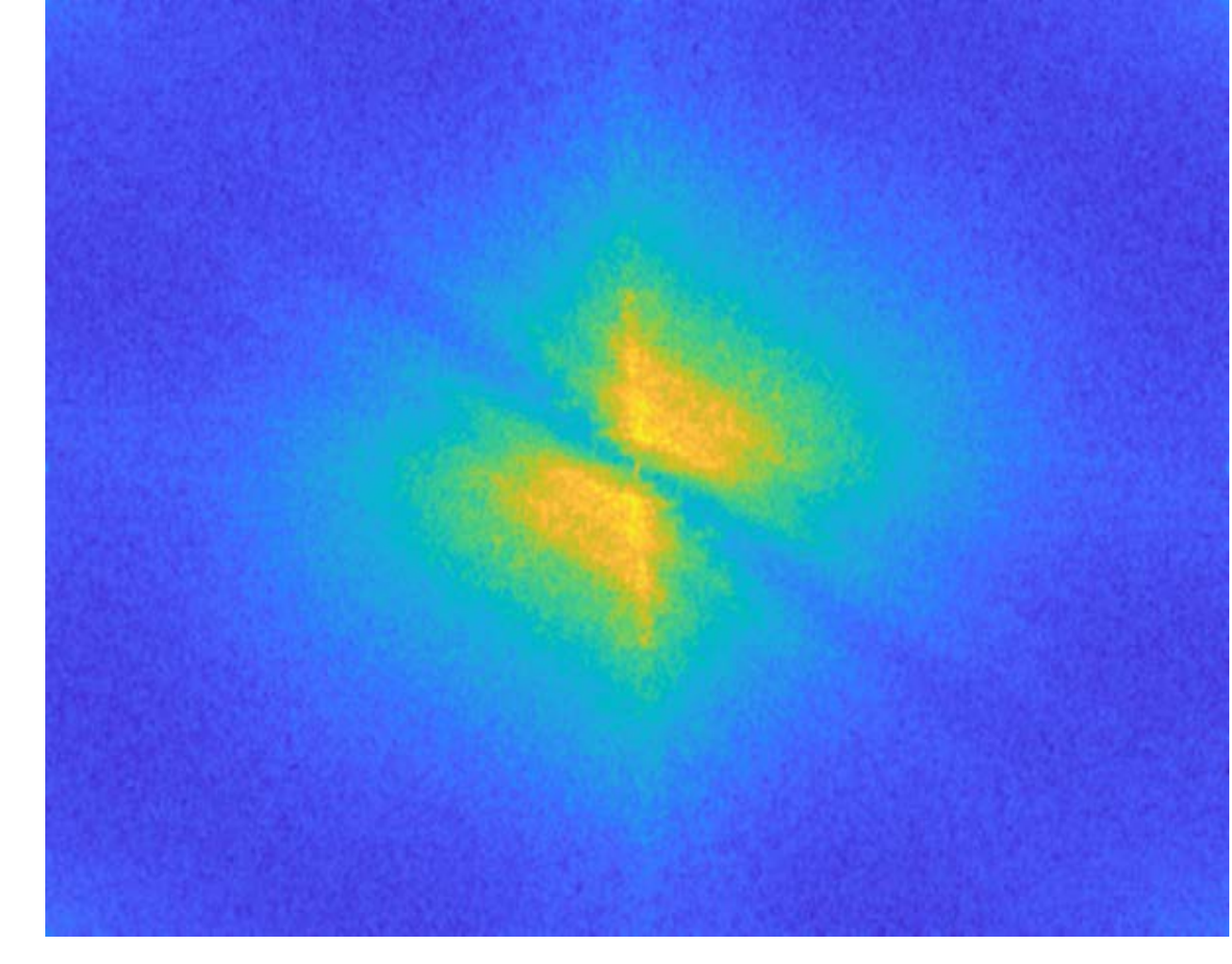} 
  \\

  \multicolumn{6}{c}{Example Image 4} \\

\end{tabular}
}
   \caption{This is extension to \cref{fig:teaser}, here we observe the same artifacts both in the spatial and frequency domain as that observed in \cref{fig:teaser}. Here we perform Image restoration using NAFNet~\cite{chen2022simple} variants on GoPro~\cite{gopro}. Normal Transposed Convolution uses 3$\times$3 sized kernels. Large Context Transposed Convolution uses kernels of size 7$\times$7+3$\times$3 for upsampling. LCTC significantly increases the model's stability during upsampling, observable in the restored image under attack and the frequency spectrum. The procedure for obtaining the 2D Frequency Spectra has been explained in \cref{subsubsec:exp:setup:frequency}.} 
    \label{fig:teaser_extension_appendix}
    \vspace{-1em}
\end{figure}

Following, we extend the example from \Cref{fig:teaser} to \Cref{fig:teaser_extension_appendix} showing similar upsampling artifacts but on different input images to demonstrate that our findings are not limited to one example.

\section{Limitations}
\label{sec:appendix:limitations}
Current metrics for measuring performance do not completely account for spectral artifacts.
Spectral artifacts begin affecting these metrics only when they become pronounced such as under adversarial attacks, and here Large Context Transposed Convolutions consistently perform better across tasks and architectures.
Ideally, we would want infinitely large kernels, however, with increasing kernel size and task complexity, training extremely large kernels can be challenging. Thus, in this work, while having ablated over kernels as large as 31$\times$31, we propose using kernels only as large as 7$\times$7 to 11$\times$11 for good practical trade-offs. Further improvements might be possible when jointly optimizing the encoder \emph{and} decoder of architectures. 

In this work, we are focused on the reduction of spectral artifacts in upsampled images and features introduced due to the theoretical limitations of upsampling operations.
However, there might exist other factors that contribute to the introduction and existence of spectral artifacts such as spatial bias.
This might also present an interesting avenue to explore.

\end{document}